\documentclass[11pt]{article}
\usepackage[bookmarksnumbered = true, colorlinks = true, linktocpage = true, bookmarksopen = true,linkcolor = black, urlcolor = black]{hyperref}
\usepackage[intlimits]{amsmath}
\usepackage{subcaption}
\usepackage{amssymb}
\usepackage{comment}
\usepackage{amsthm}
\usepackage{rotating}
\usepackage{mathtools}
\usepackage[]{algorithm}
\usepackage{algorithmic}
\usepackage{tikz}
\usepackage{authblk}
\usepackage{booktabs}
\usetikzlibrary{fit,shapes.misc}
\usepackage[utf8]{inputenc}
\usepackage[margin =1in]{geometry}
\usepackage{url}
\usepackage[bottom,flushmargin,hang,multiple]{footmisc}
\usepackage{setspace}
\newtheorem{theorem}{Theorem}[section]
\newtheorem{lemma}[theorem]{Lemma}

\newtheorem{assumption}[theorem]{Assumption}

\newtheorem{remark}[theorem]{Remark}

\newtheorem{definition}[theorem]{Definition}

\usepackage{color, colortbl}
\usepackage{xcolor}
\definecolor{Lightgray}{rgb}{0.75,0.75,0.75}
\usepackage[titletoc]{appendix}
\usepackage{indentfirst}
\usepackage[numbers,sort]{natbib}
\usepackage[normalem]{ulem}
\renewenvironment{proof}[1][Proof]{\noindent\textbf{#1.} }{\hfill\qed\vspace{\baselineskip}}

\begin{document}

\title{Stabilizing Temporal Difference Learning \\ via Implicit Stochastic Recursion}

\author[1]{Hwanwoo Kim\thanks{hwanwoo.kim@duke.edu}}
\author[2]{Panos Toulis\thanks{panos.toulis@chicagobooth.edu}}
\author[1]{Eric Laber\thanks{eric.laber@duke.edu}}

\affil[1]{Department of Statistical Science, Duke University}
\affil[2]{Booth School of Business,
University of Chicago}

\date{}

\maketitle

\begin{abstract}
\noindent Temporal difference (TD) learning is a foundational algorithm in reinforcement learning (RL). For nearly forty years, TD learning has served as a workhorse for applied RL as well as a building block for more complex and specialized algorithms. However, despite its widespread use, TD procedures are generally sensitive to step size specification. A poor choice of step size can dramatically increase variance and slow convergence in both on-policy and off-policy evaluation tasks. In practice, researchers often use trial and error to identify stable step sizes, but these approaches tend to be ad hoc and inefficient. As an alternative, we propose implicit TD algorithms that reformulate TD updates into fixed-point equations. Such updates are more stable and less sensitive to step size without sacrificing computational efficiency. Moreover, we derive almost sure convergence guarantees and finite-time error bounds (with a projection step) for the proposed implicit TD algorithms. Our results show that implicit TD algorithms are applicable to a much broader range of step sizes, and thus provide a robust and versatile framework for policy evaluation and value approximation in modern RL tasks. We demonstrate these benefits empirically through extensive numerical experiments spanning both on-policy and off-policy tasks.
\end{abstract}

\section{Introduction}
\noindent Temporal difference (TD) learning, originally introduced by \citet{sutton1988learning}, is a cornerstone of reinforcement learning (RL). Combining the strengths of Monte Carlo methods and dynamic programming, TD learning enables incremental updates using temporally correlated data, making it both simple and efficient for policy evaluation. This foundational algorithm underpins many modern RL techniques and has been applied successfully in a wide range of domains, including robotics, finance, and neuro-imaging, where accurate value prediction is critical for evaluation and control \citep{van200030,o2003temporal,modayil2014multi}. In real-world scenarios, Markov decision processes (MDPs) often operate in large state spaces, making exact value estimation computationally infeasible.  A common approach to address this issue is to apply TD learning with linear function approximation. This approach makes TD learning a practical and scalable solution even for high-dimensional problems \citep{tsitsiklis1996analysis, bertsekas1996neuro}.

Since the seminal work by \citet{tsitsiklis1996analysis} on the almost sure convergence of TD($\lambda$) algorithm with linear function approximation, numerous theoretical analyses have been conducted under a wide range of assumptions and settings \citep{dalal2018finite, bhandari2018finite, 
srikant2019finite, khamaru2021temporal, li2023accelerated, duan2023finite, patil2023finite, li2024high,mitra2024simple, samsonov2024improved, lee2026single}. 
For instance, \citet{dalal2018finite} conducted a finite-time error analysis under the assumption of i.i.d. streaming data.  \citet{bhandari2018finite} extended this work to Markovian data by incorporating a projection step to analyze the mean path of TD iterates. \citet{srikant2019finite} further 
derived finite-time error bounds for TD methods with Markovian data without requiring a  
projection step; their approach relied on refinements of stochastic approximation methods via Lyapunov function-based stability analysis. \citet{patil2023finite} obtained finite-time high-probability bounds for tail averaged TD as well as regularized TD iterates under i.i.d. setting. \citet{samsonov2024improved} established finite-time high-probability bounds for Polyak-averaged TD iterates under the i.i.d. setting and extended to Markovian data setting by incorporating a data-dropping approach together with tail averaging. \citet{mitra2024simple} established finite-time mean-squared error (MSE) bounds of TD iterates under a Markovian data assumption through an elegant induction argument.
More recently, \citet{li2026towards} obtained a finite-time MSE bound for regularized TD(0), while \citet{lee2026single} established a high-probability MSE bound for vanilla TD(0) with step size-dependent averaging, both in the Markovian setting. Orthogonal to convergence analysis and finite-time error bounds, recent works have established instance-dependent minimax lower bounds for policy evaluation \citep{khamaru2021temporal, li2023accelerated} as well as a multi-step lookahead extension of TD \citep{duan2023finite}. \

\indent In off‐policy learning problems, the data are generated under one (behavior) policy but evaluation or improvement is desired under a different (target) policy. Off-policy TD methods must correct for this distributional mismatch, typically via importance sampling ratios or projected Bellman gradients. To that end, numerous algorithms have been proposed \citep{sutton2008convergent, sutton2009fast, maei2011gradient, sutton2016emphatic, ghiassian2020gradient} with provable convergence guarantees under linear approximation and mild mixing conditions on the underlying Markov process. Among the long list of off-policy evaluation algorithms, the temporal difference learning with gradient correction (TDC) algorithm has demonstrated superior empirical performance~\citep{maei2011gradient,dann2014policy,ghiassian2020gradient}. Rigorous theoretical studies on TDC include finite-time error bounds under i.i.d.\ data \citep{dalal2018twofinite, li2024high}, as well as under Markovian data, both with a projection step \citep{xu2019two} and without a projection step \citep{gupta2019finite, doan2021finite, kaledin2020finite, haque2023tight}.

\indent While TD algorithms are celebrated for their efficiency, they remain sensitive to 
the choice of step size in both on-policy and off-policy regimes. 
From a practitioner's perspective, larger step sizes can accelerate the pace of learning but at increased risk of numerical instability or divergence  
\citep{dabney2012adaptive, tamar2014implicit, dalal2018finite}; 
conversely, smaller step sizes
guarantee stability but 
at the cost of slow progress. Adaptive step size mechanisms, such as those proposed by \citet{dabney2012adaptive}, 
which adjust rates based on temporal error signals, or the state‐dependent rules of \citet{hutter2007temporal}, offer 
some relief, but typically rely on heuristics, incur extra computational burden, and lack comprehensive theoretical 
support. From a theoretician's perspective,
existing finite-time error bounds for both TD and TDC impose restrictive conditions on the choice of step size 
\citep{bhandari2018finite,srikant2019finite,xu2019two, mitra2024simple, lee2025finite}, again highlighting the issue of step size calibration. 
Thus, there remains a pressing demand for numerically stable and computationally efficient adaptive schemes with provable 
convergence guarantees.\

\indent Implicit stochastic recursions, as exemplified by implicit stochastic gradient descent  
\citep[SGD;][]{toulis2014statistical, toulis2016towards, toulis2017asymptotic, chee2023plus}, 
provide a promising framework for improving numerical stability in TD learning. Implicit SGD reformulates the standard gradient-based update as a fixed-point equation. This introduces a natural stabilizing effect, reducing sensitivity to step size and preventing divergence. Unlike explicit update methods, which directly apply gradient steps, implicit SGD imposes data-adaptive stabilization in gradient updates to control large deviations, ensuring robustness while maintaining computational simplicity. Utilizing this key idea behind implicit SGD, 
we provide a principled approach to resolve step size sensitivities for both on-policy and off-policy TD algorithms.

\subsection{Contributions}
\noindent We extend and formalize the idea of implicit recursions in TD learning, which was introduced for TD($\lambda$) in an unpublished manuscript by \citet{tamar2014implicit}. We further propose implicit TDC, thereby establishing a comprehensive framework for implicit TD update rules. In implicit TD learning, the standard TD recursion is reformulated into a fixed-point equation, which brings the stabilizing effects of implicit updates and thereby reduces sensitivity to the choice of step size. 

Our work substantially extends \citet{tamar2014implicit}, which offers only a preliminary stability analysis of implicit TD($\lambda$) under a restrictive zero-reward assumption. In contrast, we provide rigorous theoretical justification for the improved numerical stability of implicit TD algorithms without relying on such an unrealistic assumption. This analysis includes almost sure convergence guarantees for implicit TD(0) and TD($\lambda$) algorithms, as well as finite-time MSE bounds for the projected variants. Moreover, we establish finite-time MSE bounds of the implicit version of the projected TDC. 

More specifically, we show that MSE bounds hold for implicit TD(0) and TD($\lambda$) with a step size of the form $\alpha_n = \alpha_1/n^s$, $\alpha_1 > 0$, in either of two regimes: (1) $s = 1$ with $\alpha_1$ sufficiently large, under suitable conditions on the feature norm or the environment; (2) $s \in (1/2, 1)$, independently of $\alpha_1$, the feature norm, or the environment.
This stands in stark contrast to existing bounds for standard TD algorithms, which often require small step sizes to ensure numerical stability as well as theoretical guarantees \citep{bhandari2018finite, srikant2019finite, xu2019two, kaledin2020finite, li2026towards, lee2026single}. A distinct feature of our result is that, for $s \in (1/2, 1)$, none of the choice of step size, the algorithm, or the starting time for the finite-time MSE bound depends on unknown environment-specific quantities. Meanwhile, our bounds require a projection step, a lightweight operation added to ensure all-time bounds. In particular, we show that the convergence rate does not deteriorate with the choice of projection radius (although the multiplicative constants may depend on it), and thus our results continue to hold even with a vacuously large projection radius. In a similar vein, the MSE bound for projected implicit TDC holds for an arbitrary initial step size and iteration index, independent of the environment, a result that is unprecedented in the existing literature on finite-time error bounds for TDC. \citep{gupta2019finite, xu2019two, doan2022nonlinear, kaledin2020finite, haque2023tight, chandak2025finite}.

We demonstrate that the proposed implicit TD algorithms offer substantial improvements in stability and robustness while retaining the computational efficiency of standard TD methods. A summary comparing our results to existing work is provided in Table \ref{tab:stepsize_comparison} and Table \ref{tab:stepsize_comparison_TDC}. Our contributions in this paper can be summarized as follows:
\begin{itemize}
    \item establishing connections between implicit and standard TD methods, showing implicit updates incur virtually no additional computational cost (Lemma \ref{LEMMA:IMP_EXP_RELATION} and Lemma \ref{LEMMA:IMP_TDC_EXP_RELATION}); 
    \item almost sure convergence for implicit TD(0) and TD($\lambda$) algorithms (Theorem \ref{THM:ASYM_IMP_TD}); 
    \item finite-time MSE bounds for projected implicit TD(0) and TD($\lambda$) (Theorem \ref{THM:FIN_PROJ_TD0} and Theorem \ref{THM:FIN_PROJ_TDL}); 
    \item introduction of the implicit TDC algorithm;
    \item a finite-time MSE bound for the projected implicit TDC algorithm (Theorem \ref{thm:tdc_decr});
    \item substantially relaxed initial step size requirements for finite-time MSE bounds;
    \item empirical demonstration of superior numerical stability and learning efficiency of the proposed implicit TD(0) and TD($\lambda$) in synthetic random walk, Markov reward process environments, as well as continuous domain control problems; 
    \item demonstration of implicit TDC's substantially improved numerical stability and value function approximation over TDC in the celebrated Baird's counterexample \citep{baird1995residual}. 
\end{itemize}

\noindent In Section \ref{SEC:BACKGROUND}, 
we provide the mathematical framework for TD algorithms with linear 
function approximation and demonstrate their instability
with respect to the choice of 
step size. In Section \ref{SEC:IMPLICIT_TD}, we formulate implicit TD(0), TD($\lambda$), 
and TDC algorithms. In Section 
\ref{SEC:THEORY}, we present theoretical justifications for the proposed implicit TD(0), TD($\lambda$), and TDC algorithms. We present both almost sure convergence results without projection and 
finite-time MSE bounds with a projection step. In Section \ref{SEC:NUMERICS}, we demonstrate the superior numerical stability of implicit TD algorithms over standard TD algorithms in a range of environments. Finally, in Section \ref{SEC:CONCLUSION}, we provide a summary and concluding remarks. 

\begin{table}[t]
\centering
\begin{tabular}{|c|c|c|c|c|c|c|}
\hline
\textbf{Paper} & $\alpha_n$ $\perp$ Env & Alg $\perp$ Env &  $n \perp$ Env & \sout{Proj} &\sout{Avg} & $\tilde{\mathcal{O}}(1/n^s)$ \\
\hline
\citet{bhandari2018finite} & $\times$ & \checkmark & \checkmark & $\times$ & \checkmark & \checkmark \\
\hline
\citet{srikant2019finite} & $\times$ & \checkmark & $\times$ & \checkmark & \checkmark & \checkmark\\
\hline
\citet{samsonov2024improved} & \checkmark & $\times$ & $\times$ & \checkmark & $\times$ & \checkmark \\
\hline
\citet{mitra2024simple} & $\times$ & \checkmark & $\times$  & \checkmark  & \checkmark & \checkmark \\
\hline
\citet{li2026towards} & \checkmark & $\times$ & $\times$ & \checkmark  & \checkmark & $\times$ \\
\hline
\citet{lee2026single} & $\times$ & \checkmark & \checkmark & \checkmark  & $\times$ & \checkmark \\
\hline
\textbf{This paper} & \textbf{\checkmark} & \checkmark & \checkmark & $\times$ & \checkmark & \checkmark \\
\hline
\end{tabular}
\caption[Summary of step size, algorithmic, timing, projection, averaging requirements, and convergence rate for finite-time TD(0) MSE bounds under Markovian data.]{
Summary of step size, algorithmic, timing, projection, averaging requirements, and convergence rate for finite-time linear TD(0) MSE bounds under Markovian data.
`` $\alpha_n \perp$ Env'' is checked if the step size requires no tuning to environment-dependent quantities (e.g., minimum eigenvalue of feature covariance or mixing time). ``Alg $\perp$ Env'' is checked if the algorithm itself, beyond the step size, requires no environment-dependent modification. ``$n \perp$ Env'' is checked if the bound holds without requiring the iteration index to exceed an environment-dependent threshold. ``\sout{Proj}'' is checked if a projection step is not required. ``\sout{Avg}'' is checked if the bound is established for the last iterate without requiring any form of averaging. ``$\tilde{\mathcal{O}}(1/n^s)$'' is marked if the convergence rate is $\mathcal{O}(1/n^s)$ up to polylogarithmic factors with a step size of the form $\alpha_n = \alpha_1/n^s, ~\alpha_1 > 0$.
}
\label{tab:stepsize_comparison}
\end{table}

\begin{table}[t]
\centering
\begin{tabular}{|c|c|c|c|c|}
\hline
\textbf{Paper} & $\alpha_n\perp$ Env & $n \perp$ Env & \sout{Proj} &  Rate \\
\hline
\citet{xu2019two} & $\times$ & \checkmark & $\times$ & $\mathcal{O}(1/n^{2/3})$ \\
\hline
\citet{gupta2019finite} & $\times$ &  $\times$  & \checkmark & $\mathcal{O}(1/n^{2/3})$ \\
\hline
\citet{kaledin2020finite} & $\times$ & \checkmark & \checkmark &  $\mathcal{O}(1/n)$\\
\hline
\citet{doan2021finite} &$\times$ &  $\times$  & \checkmark & $\mathcal{O}(1/n^{2/3})$ \\
\hline
\citet{haque2023tight} & $\times$ & \checkmark & \checkmark &  $\mathcal{O}(1/n)$ \\
\hline
\citet{chandak2025finite} & $\times$ & $\times$ & \checkmark &  $\mathcal{O}(1/n^{2/3})$ \\
\hline
\textbf{This paper} & \textbf{\checkmark}  & \checkmark & $\times$ &  $\mathcal{O}(1/n^{2/3})$ \\
\hline
\end{tabular}
\caption[Summary of step size, algorithmic, timing, projection, averaging requirements, and convergence rate for finite-time TDC MSE bounds under Markovian data.]{
Summary of step size, algorithmic, timing, projection, averaging requirements, and convergence rate for finite-time linear TDC MSE bounds under Markovian data.
``$\alpha_n \perp$ Env'' is checked if the step size requires no tuning to environment-dependent quantities (e.g., minimum eigenvalue of feature covariance or mixing time). ``$n \perp$ Env'' is checked if the bound holds without requiring the iteration index to exceed an environment-dependent threshold. ``\sout{Proj}'' is checked if a projection step is not required. ''Rate'' indicates the asymptotic convergence rate. 
}
\label{tab:stepsize_comparison_TDC}
\end{table}

\section{Background}\label{SEC:BACKGROUND}
\subsection{Value function}
\noindent We consider a discrete-time Markov decision process with finite state space $\mathcal{X}\subset \mathbb{R}^k$, finite action space $\mathcal{A}$, transition kernel $P(\boldsymbol{x}' \mid a, \boldsymbol{x})$ for $\boldsymbol{x}, \boldsymbol{x}' \in \mathcal{X}$, $a \in \mathcal{A}$, discount factor $\gamma \in (0,1)$, and bounded reward function $r: \mathcal{X}\times \mathcal{A} \to [-r_{\max}, r_{\max}]$ for some $r_{\max} \in \mathbb{R}_{>0}$. In addition, we assume there is a fixed target policy $\pi_\star:\mathcal{X} \to \mathcal{A}$.\footnote{For clarity of presentation, we focus on deterministic policies in the on-policy evaluation setting, equivalent to considering a Markov reward process. In the off-policy evaluation setting, we consider the more general case of stochastic policies.}
Let $\boldsymbol{x}_n$ be the state, $a_n = \pi_{\star}(\boldsymbol{x}_n)$ the action, and $r_n:= r(\boldsymbol{x}_n, a_n)$ the reward at time $n \in \mathbb{N}$. 
The primary object of interest is the value function, given by
$$
V(\boldsymbol{x}) = \mathbb{E}_{\pi_\star}\left(
\sum_{n=1}^\infty \gamma^{n-1} r_n \mid \boldsymbol{x}_1 = \boldsymbol{x}\right)
$$
where the expectation is over the sequence of states 
generated according to the time-homogeneous transition kernel 
$P_{\pi_\star}(\boldsymbol{x}' \mid \boldsymbol{x})= P(\boldsymbol{x}' \mid \pi_{\star}(\boldsymbol{x}), \boldsymbol{x})$. 
The goal of on-policy evaluation is to approximate the value function from observed data $(\boldsymbol{x}_n)_{n \in \mathbb{N}}$ transitioning according to the probability kernel $P_{\pi_\star}$ induced by the target policy.

In many applications, however, collecting data under the target policy can be impossible, expensive, or unethical. 
In such cases, off-policy evaluation is used to estimate the value function 
using data generated under a behavior policy $\pi_b: \mathcal{X} \to \Delta(\mathcal{A})$ instead of the target policy. For both policy regimes, we will assume the Markov chain $(\boldsymbol{x}_n)_{n \in \mathbb{N}}$ admits a unique steady-state distribution $\mu_\pi$, corresponding to the policy $\pi$ (either $\pi_\star$ or $\pi_b$) that governs the observed data dynamics.

When the state-space is high-dimensional, 
it is generally infeasible to compute the value function exactly. In such cases, one must
posit additional structure on the value function. We consider linear function approximation, in which it is assumed that, for some weight vector $\boldsymbol{w}_{\star} \in \mathbb{R}^d$, the value function satisfies 
$$ 
V(\boldsymbol{x})\approx V_{\boldsymbol{w}_{\star}}(\boldsymbol{x}) = \phi(\boldsymbol{x})^{\top} \boldsymbol{w}_{\star} 
$$ 
for some pre-determined feature mapping $\phi:\mathcal{X} \to \mathbb{R}^d$. The problem of estimating $V$ thus reduces to estimating the weight vector $\boldsymbol{w}_{\star}$. Define 
$\boldsymbol{\Phi} = [
    \phi(\boldsymbol{x})^{\top}
]_{\boldsymbol{x} \in {\mathcal{X}}}, 
$
and $V_{\boldsymbol{w}_{\star}} = \boldsymbol{\Phi} \boldsymbol{w}_{\star}$. Throughout, we assume rows of $\boldsymbol{\Phi}$ are linearly independent. Such an assumption is natural, as otherwise, we can attain the same quality of approximation after removing some of the features. 

\subsection{Temporal difference learning}\label{SUBSEC:TD_BACKGROUND}
\noindent TD(0) and TD($\lambda$) algorithms \citep{sutton1988learning, sutton1998reinforcement} 
constitute a widely used class of stochastic approximation methods 
for estimating the value 
function $V$ from accumulating data. 
Under the linear approximation, these algorithms 
provide a recursive estimator of $\boldsymbol{w}_{\star}$. At time $n \in \mathbb{N}$ with a target policy $\pi_\star$, recall that $a_n = \pi_{\star}(\boldsymbol{x}_n), r_n = r(\boldsymbol{x}_n, a_n), \boldsymbol{\phi}_{n} = \phi(\boldsymbol{x}_n)$, and $\boldsymbol{\phi}_{n+1}= \phi(\boldsymbol{x}_{n+1})$. The TD(0) update rule is given by
\begin{align}
    \boldsymbol{w}_{n+1} &= \boldsymbol{w}_n + \alpha_n \delta_n \boldsymbol{\phi}_n, \label{TD0_UPDATE}\\ \delta_n &:= r_n  + \gamma \boldsymbol{\phi}_{n+1}^{\top} \boldsymbol{w}_n - \boldsymbol{\phi}_n^{\top} \boldsymbol{w}_n, \nonumber
\end{align}
where $\alpha_n$ is the step size for the $n^{\text{th}}$ iteration, and $\delta_n$ is referred to as the TD error. The update rule for the TD($\lambda$) algorithm, parametrized by $\lambda \in [0,1]$, is given by 
\begin{align}
    \boldsymbol{w}_{n+1} &= \boldsymbol{w}_{n} + \alpha_n \delta_n \boldsymbol{e}_{n}, \label{TD_LAMB_UPDATE}\\ \delta_n &:= r_n  + \gamma \boldsymbol{\phi}_{n+1}^{\top} \boldsymbol{w}_{n} + (\lambda\gamma)\boldsymbol{e}_{n-1}^{\top} \boldsymbol{w}_{n} - \boldsymbol{e}_{n}^{\top} \boldsymbol{w}_{n}, \nonumber \\
    \boldsymbol{e}_{n} &:= \boldsymbol{\phi}_{n} + (\lambda\gamma)\boldsymbol{e}_{n-1}, ~ \boldsymbol{e}_{0} = 0, \nonumber
\end{align}
where $\boldsymbol{e}_{n}$ is known as the eligibility trace, which contains information on 
all previously visited states. Note that the TD($\lambda$) algorithm subsumes TD(0) and the Monte Carlo evaluation (TD(1)) as special cases. 
In prior numerical experiments, TD($\lambda$) with $\lambda \in (0,1)$ has shown superior performance over TD(0) and the Monte Carlo in approximating the value function \citep{sutton1998reinforcement}. 

While TD($\lambda$) with $\lambda \in [0,1)$ is effective in many on-policy evaluation tasks, it can become numerically unstable in off-policy settings \citep{baird1995residual, sutton1998reinforcement}. In the off-policy evaluation setting, one needs to account for the action $a_n$, which is instead sampled from the so-called behavior policy $\pi_b$, i.e., $a_n \sim \pi_b(\cdot \mid \boldsymbol{x}_{n})$. Several extensions have been proposed to accommodate TD in such settings \citep{sutton2008convergent, sutton2009fast, sutton2016emphatic}. Here, we focus on the TDC algorithm with linear function approximation \citep{sutton2009fast}. To this end, let us define 
$\rho_n := \pi_\star(a_n \mid \boldsymbol{x}_{n})/\pi_b(a_n\mid \boldsymbol{x}_{n})$ which denotes the ratio of target and behavior policies, also known as an importance weight. With $\boldsymbol{\phi}_{n} = \phi(\boldsymbol{x}_{n}), r_n = r(\boldsymbol{x}_{n}, a_n), \boldsymbol{\phi}_{n+1}= \phi(\boldsymbol{x}_{n+1})$ and $\delta_n = r_n  + \gamma \boldsymbol{\phi}_{n+1}^{\top} \boldsymbol{w}_{n} - \boldsymbol{\phi}_{n}^{\top} \boldsymbol{w}_{n}$, the TDC update is given by
\begin{align}
    \boldsymbol{w}_{n+1} &= \boldsymbol{w}_{n} + \alpha_n \rho_n \delta_n \boldsymbol{\phi}_{n} - \alpha_n \rho_n \gamma \boldsymbol{\phi}_{n+1}\boldsymbol{\phi}_{n}^{\top} \boldsymbol{u}_{n}, \label{TDC_main_update} \\
    \boldsymbol{u}_{n+1} &= \boldsymbol{u}_{n} + \beta_n \rho_n \delta_n \boldsymbol{\phi}_{n} - \beta_n \rho_n \boldsymbol{\phi}_{n} \boldsymbol{\phi}_{n}^{\top} \boldsymbol{u}_{n}, \label{TDC_aux_update} 
\end{align}
where $(\alpha_n)_{n \in \mathbb{N}}$ and $(\beta_n)_{n 
\in \mathbb{N}}$ are non-negative step size sequences. The primary iterates $(\boldsymbol{w}_{n})_{n \in \mathbb{N}}$ parameterize the value function, while the auxiliary iterates 
$(\boldsymbol{u}_{n})_{n \in \mathbb{N}}$ serve to modify the direction of TD updates. The term $\alpha_n \rho_n \gamma 
\boldsymbol{\phi}_{n+1}\boldsymbol{\phi}_{n}^{\top} \boldsymbol{u}_{n}$ in the primary iterate update is known as the gradient correction term. It is common to assume $\alpha_n \ll \beta_n$, and 
hence, the auxiliary iterates update on a faster time-scale than that of the target parameter iterates.

To facilitate the analysis of TD algorithms, incorporating a projection step that ensures the iterates $(\boldsymbol{w}_{n})_{n \in \mathbb{N}}$ fall within an $\ell_2$-ball of radius $R \in \mathbb{R}_{>0}$ is a common technique in the stochastic optimization and reinforcement learning literature \citep{nemirovski2009robust, bubeck2015convex, bhandari2018finite, xu2019two, zou2019finite}. In addition to the recursive updates in \eqref{TD0_UPDATE} and \eqref{TD_LAMB_UPDATE}, the projected TD update includes the following step:
\begin{align*}
\Pi_{R}(\boldsymbol{w}) = \underset{\boldsymbol{w}': \|\boldsymbol{w}'\| \le R}{\operatorname{argmin}} \|\boldsymbol{w}-\boldsymbol{w}'\| = \begin{cases}
    \frac{R}{\|\boldsymbol{w}\| }\boldsymbol{w} &~~\text{if}~~ \|\boldsymbol{w}\| > R \\
    \boldsymbol{w} &~~\text{otherwise},
\end{cases}   
\end{align*}
where the projection radius $R$ is chosen to be sufficiently large to guarantee $\|\boldsymbol{w}_{\star}\| \le R$.
Such a projection step not only serves as a way to facilitate finite-time MSE analysis for TD(0) and TD($\lambda$) that holds for all time $n \in \mathbb{N}$, but also prevents potential divergent behavior with no extra computational burden \citep{bhandari2018finite}. In the same spirit, the projected TDC algorithm, incorporating the following
projection steps to \eqref{TDC_main_update} and \eqref{TDC_aux_update} 
\begin{align*}
\Pi_{R_{\boldsymbol{w}}}(\boldsymbol{w}) = \begin{cases}
    \frac{R_{\boldsymbol{w}}}{\|\boldsymbol{w}\|} \boldsymbol{w}  &~~\text{if}~~ \|\boldsymbol{w}\| > R_{\boldsymbol{w}} \\
    \boldsymbol{w} &~~\text{otherwise}
    \end{cases}  
\quad \text{and} \quad 
\Pi_{R_{\boldsymbol{u}}}(\boldsymbol{u}) = \begin{cases}
    \frac{R_{\boldsymbol{u}}}{\|\boldsymbol{u}\|} \boldsymbol{u} &~~\text{if}~~ \|\boldsymbol{u}\| > R_{\boldsymbol{u}} \\
    \boldsymbol{u} &~~\text{otherwise}
\end{cases}  
\end{align*}
was studied in depth, 
and finite-time MSE bounds of projected TDC were established \citep{xu2019two}.

\subsection{Stochastic approximation}\label{SUBSEC:STOC_APPROX}
\noindent The aforementioned TD algorithms fall into a broader class of linear stochastic approximation methods \citep{robbins1951stochastic,benveniste2012adaptive,lakshminarayanan2018linear, srikant2019finite}, 
characterized by updates of the form 
$$
\boldsymbol{w}_{n+1} = \boldsymbol{w}_{n} + \alpha_n(\boldsymbol{b}_{n} + \boldsymbol{A}_{n} \boldsymbol{w}_{n}), \quad n \in \mathbb{N},
$$
where $\alpha_n$ is the step size for the $n^{\text{th}}$ iteration, and $(\boldsymbol{b}_{n}, \boldsymbol{A}_{n})$ are random quantities. Under suitable technical assumptions on $\alpha_n, \boldsymbol{b}_{n}$ and $\boldsymbol{A}_{n}$, various types of convergence of the stochastic approximation algorithms have been established \citep{robbins1951stochastic, borkar2008stochastic, ljung2012stochastic, benveniste2012adaptive}. Of particular relevance to our setting are convergence results when randomness in $(\boldsymbol{b}_{n}, \boldsymbol{A}_{n})$ is induced by the underlying time-homogeneous Markov chain $(\boldsymbol{x}_{n})_{n\in \mathbb{N}}$, which is assumed to mix at a geometric rate.  Let $\mathbb{E}_{\mu_\pi}$ denote expectation with respect to the
steady-state distribution of $(\boldsymbol{x}_{n})_{n \in \mathbb{N}}$. Define $\boldsymbol{b} = \mathbb{E}_{\mu_\pi}(\boldsymbol{b}_{n})$ and $\boldsymbol{A} = \mathbb{E}_{\mu_\pi}(\boldsymbol{A}_{n})$. The so-called Robbins-Monro condition on the step size, i.e., 
$\sum_{n=1}^\infty \alpha_n = \infty$ and $\sum_{n=1}^\infty 
\alpha^2_n < \infty$, combined with suitable assumptions on $\boldsymbol{A}$ and $\boldsymbol{b}$ guarantees convergence of iterates $\boldsymbol{w}_{n}$ to $\boldsymbol{w}_{\star}$, where $\boldsymbol{w}_{\star}$ is a solution of the equation $\boldsymbol{A}\boldsymbol{w} + \boldsymbol{b} = 0$ \citep[e.g., see][]{bertsekas1996neuro, tsitsiklis1996analysis, benveniste2012adaptive}. 

Rewriting the TD(0) and TD($\lambda$) updates as above, it can be shown that both algorithms fall into the class of 
linear stochastic approximation algorithms. A range of approaches, including Lyapunov-function-based analysis \citep{tsitsiklis1996analysis, bertsekas1996neuro, benveniste2012adaptive, srikant2019finite}, the ODE method \citep{borkar2000ode, borkar2025ode, liu2025ode}, information-theoretic inequalities \citep{bhandari2018finite}, and mathematical induction \citep{mitra2024simple}, have established asymptotic convergence and finite-time MSE bounds of TD(0) and TD($\lambda$) iterates, respectively, to the solution $\boldsymbol{w}_{\star}$ satisfying 
\begin{align}
      \mathbb{E}_{\mu_{\pi}}(\boldsymbol{\phi}_{n} \boldsymbol{\phi}_{n}^{\top} - \gamma \boldsymbol{\phi}_{n} \boldsymbol{\phi}_{n+1}^{\top})\boldsymbol{w}_{\star} &= \mathbb{E}_{\mu_{\pi}}(r_n \boldsymbol{\phi}_{n})\label{MEAN_PATH_TD0}\\
    \mathbb{E}_{\mu_{\pi}}( \boldsymbol{e}_{-\infty:n} \boldsymbol{\phi}_{n}^{\top} - \gamma \boldsymbol{e}_{-\infty:n} \boldsymbol{\phi}_{n+1}^{\top})\boldsymbol{w}_{\star} &= \mathbb{E}_{\mu_{\pi}}(r_n  \boldsymbol{e}_{-\infty:n})\label{MEAN_PATH_TDL}
\end{align}
where $\boldsymbol{e}_{-\infty:n} = \sum_{k = -\infty}^n (\lambda\gamma)^{n-k} \boldsymbol{\phi}_k$ is the steady-state eligibility trace.

The previously discussed linear stochastic approximation framework naturally extends to two-timescale linear stochastic approximation, characterized by coupled iterative updates operating at distinct timescales. A subclass of these algorithms is of the form:
\begin{align*}
\boldsymbol{w}_{n+1} &= \boldsymbol{w}_{n} + \alpha_n\left(\boldsymbol{b}_{n} + \boldsymbol{A}_{n} \boldsymbol{w}_{n} + \boldsymbol{B}_{n} \boldsymbol{u}_{n}\right), \quad n \in \mathbb{N}\\
\boldsymbol{u}_{n+1} &= \boldsymbol{u}_{n} + \beta_n\left(\boldsymbol{b}_{n} + \boldsymbol{A}_{n} \boldsymbol{w}_{n}  +\boldsymbol{C}_{n} \boldsymbol{u}_{n}\right), \quad n \in \mathbb{N}
\end{align*}
where $\left(\boldsymbol{A}_{n}, \boldsymbol{B}_{n}, \boldsymbol{C}_{n}, \boldsymbol{b}_{n}\right)$ are random quantities driven by the underlying Markov processes, and the sequences $(\alpha_n)_{n \in \mathbb{N}}$ and $(\beta_n)_{n \in \mathbb{N}}$ are positive step sizes satisfying:
$\alpha_n \ll \beta_n$ and ${\alpha_n}/{\beta_n} \rightarrow 0.$
This separation of scales ensures that the auxiliary iterates $\boldsymbol{u}_{n}$ evolve faster than the 
primary iterates $\boldsymbol{w}_{n}$. Under additional assumptions and mixing conditions on the underlying Markov chain $(\boldsymbol{x}_{n})_{n \in \mathbb{N}}$, these iterates converge to a solution of coupled equilibrium equations involving expectations under the steady-state distribution \citep{borkar1997stochastic,konda2004convergence, borkar2008stochastic,karmakar2018two}.\ 

In fact, the TDC algorithm precisely fits this two-timescale framework. The primary parameter $\boldsymbol{w}_{n}$ and the auxiliary parameter $\boldsymbol{u}_{n}$ are updated concurrently but at differing step sizes $\alpha_n$ and $\beta_n$, respectively, with $\alpha_n\ll\beta_n$. The auxiliary iterates $\boldsymbol{u}_{n}$ are introduced to estimate a correction term that aligns the update direction of the primary iterates $\boldsymbol{w}_{n}$ with the gradient of a suitable objective function. This adjustment allows the TDC algorithm to become gradient-based, differentiating it from the standard TD method, which does not correspond to gradient-based updates. Convergence analyses of TDC explicitly leverage the two-timescale stochastic approximation theory, demonstrating that the iterates asymptotically approach the equilibrium solutions of a coupled linear system \citep{sutton2008convergent, sutton2009fast}. The two-timescale structure captures the interplay between primary and auxiliary iterates inherent to the TDC algorithm, providing rigorous convergence results and finite-time MSE bounds \citep{dalal2018twofinite, gupta2019finite, xu2019two, ma2020variance, kaledin2020finite,doan2021finite, haque2023tight} within a unified theoretical framework.

\subsection{Numerical instability}\label{SUBSEC:INSTAB}
\noindent Despite the widespread use of TD algorithms, their sensitivity to 
step size selection presents a persistent practical challenge. 
While larger step sizes can lead to faster progress, the updates may become unstable and cause divergence \citep{dabney2012adaptive, tamar2014implicit, dalal2018finite}. 
Conversely, using smaller step sizes improves numerical stability but can significantly slow the learning process. The primary issue stems from the recursive nature of TD methods, where updates are based on estimates that rely on prior updates, causing errors to propagate and potentially compound over time. Analogous to standard TD algorithms, it has been extensively documented that the choice of step size sequences for the TDC algorithm also requires careful calibration and restricts the usage of large step sizes, which may be
inefficient \citep{dann2014policy, white2016investigating, ghiassian2020gradient}. \

\begin{figure}[!h]
    \centering
    
    \includegraphics[height=.25\textwidth]{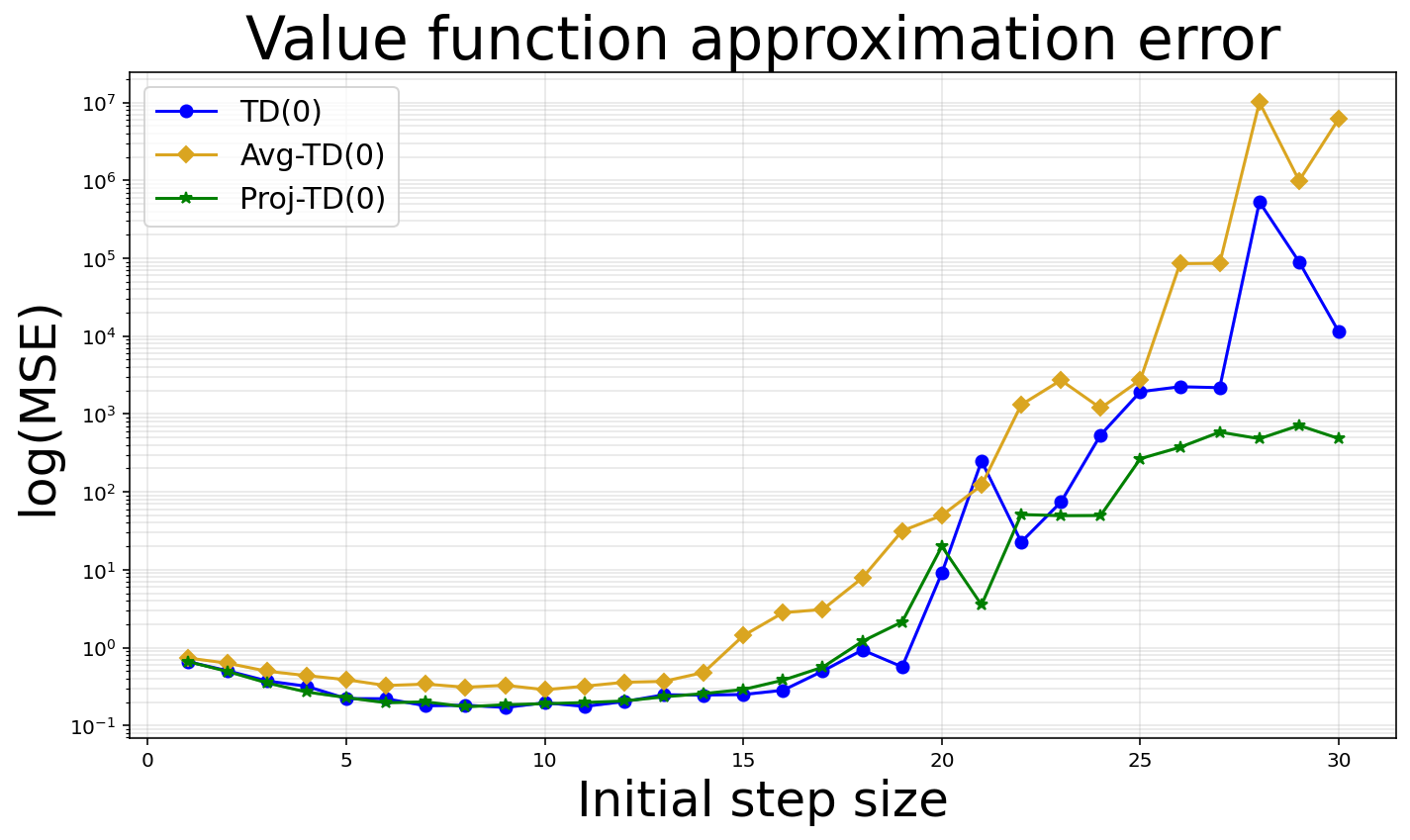}
    \includegraphics[height=.25\textwidth]{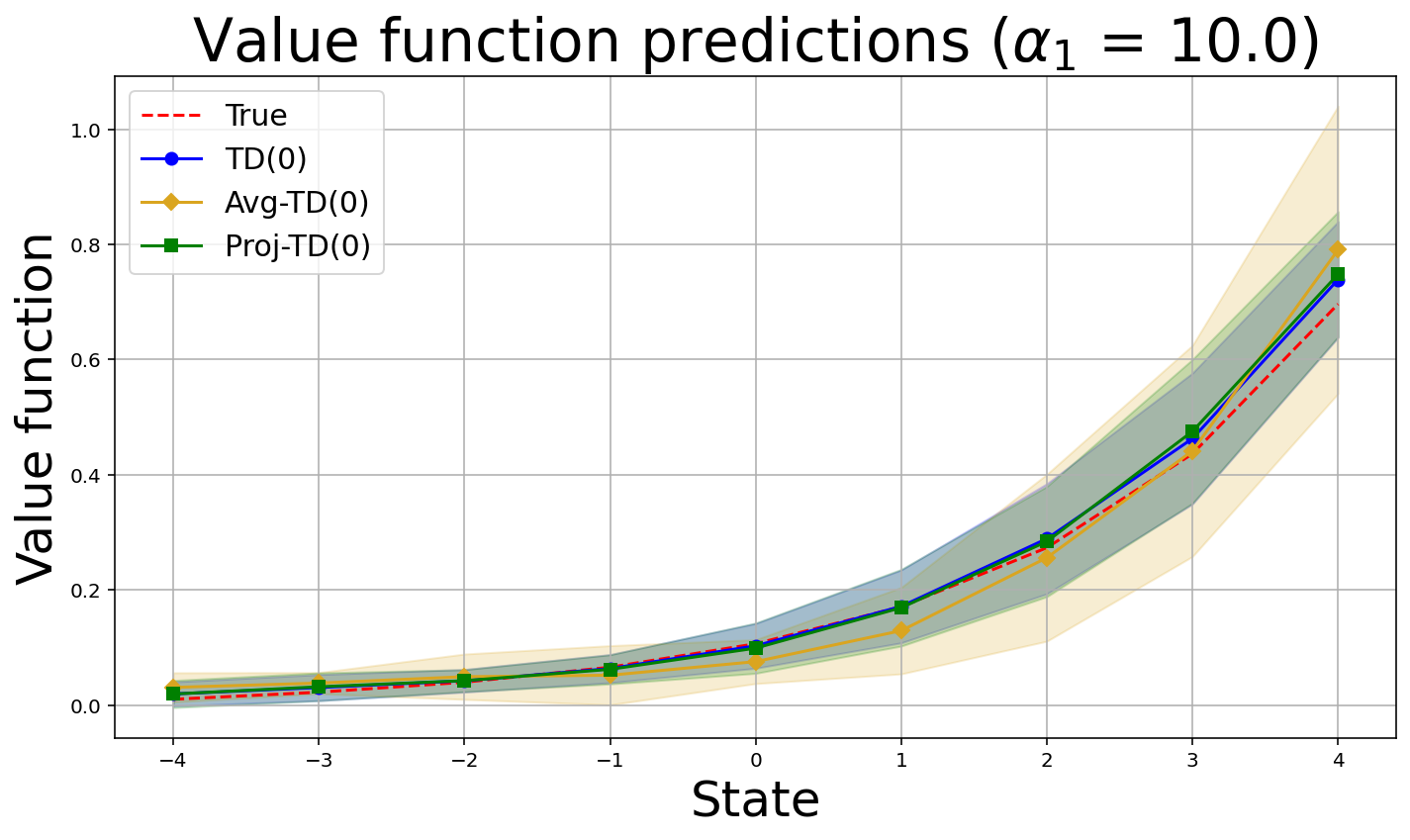}
    \includegraphics[height=.25\textwidth]{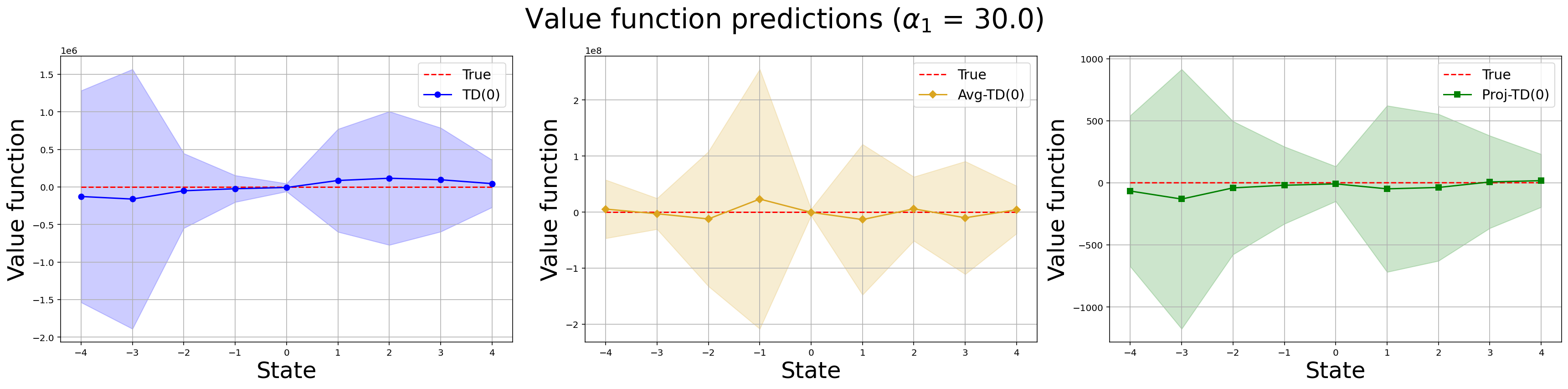}

    \caption{Value function approximation in the random walk environment. \textbf{Top left:} Log-MSE between estimated and true value functions versus initial step size $\alpha_1$; small step sizes are inefficient while large ones cause error to grow sharply. \textbf{Top right ($\boldsymbol{\alpha_1 = 10}$):} Value functions estimated by standard TD(0) (blue), averaged TD(0) (yellow), and projected TD(0) (green) closely match the true value function (red dashed), and the pointwise one standard deviation bands (shaded regions) remain narrow, indicating stable performance. \textbf{Bottom left, bottom middle, and bottom right ($\boldsymbol{\alpha_1 = 30}$):} Value function estimates from standard TD(0), averaged TD(0), and projected TD(0), respectively, diverge substantially from the true value function, and the pointwise one standard deviation bands inflate to extreme magnitudes, reflecting a loss of numerical stability under a large initial step size.}\label{fig:VALUE_FUNC_TD0_INSTAB}
\end{figure}

\begin{figure}[!h]
    \centering
    \includegraphics[height=.295\textwidth]{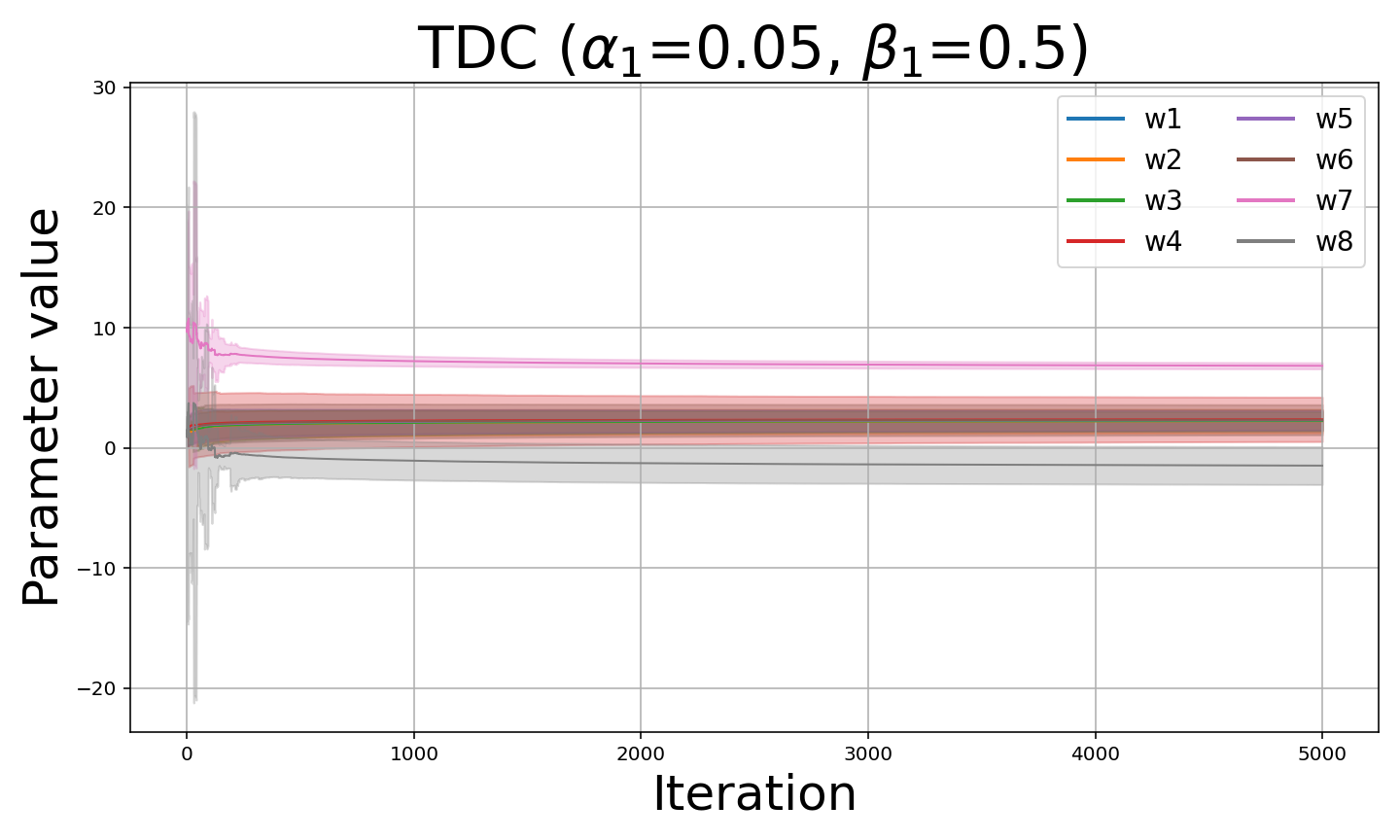}
    \includegraphics[height=.295\textwidth]{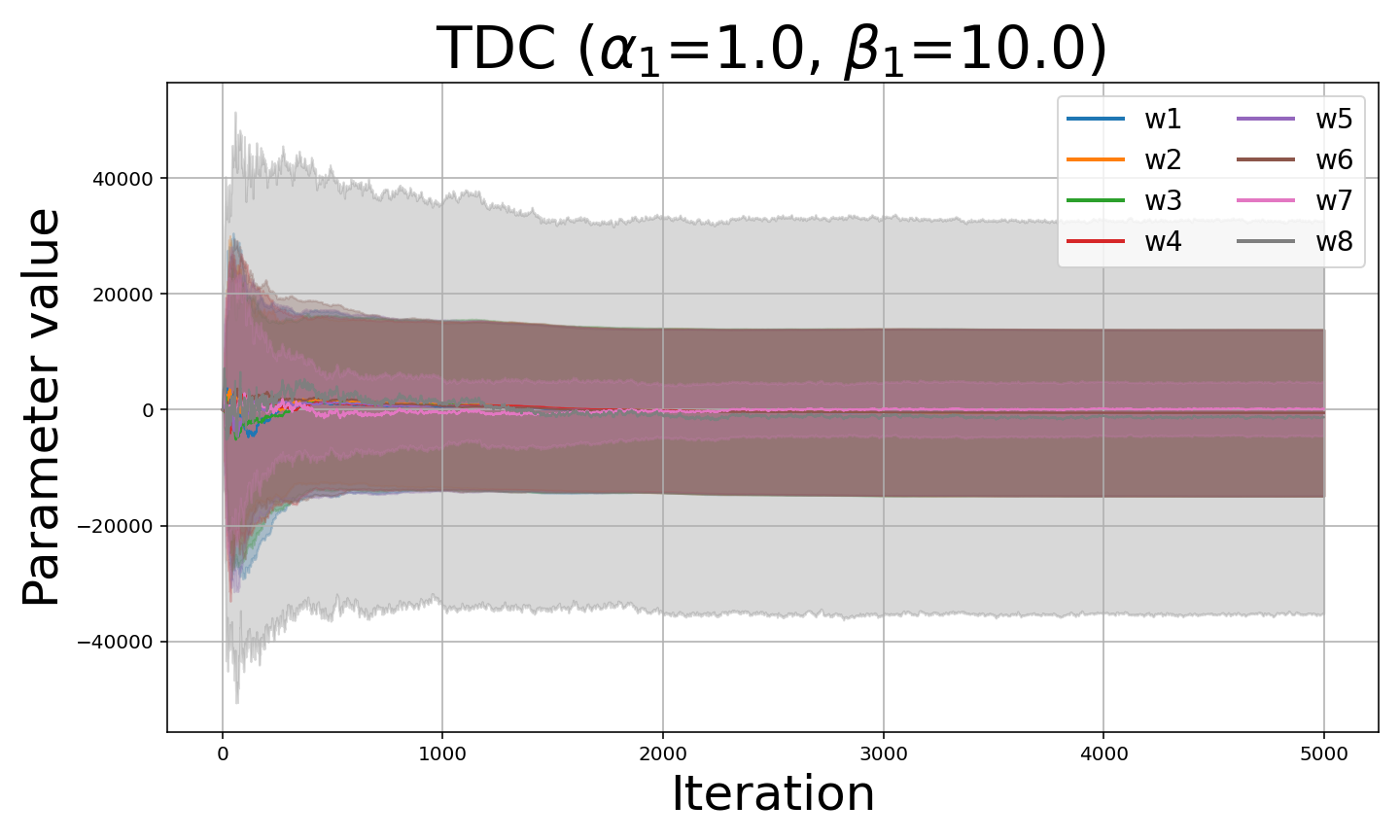}
    \caption{Trajectories of the estimated weight parameters for standard TDC on Baird's counterexample. \textbf{Left ($\boldsymbol{\alpha_1 = 0.05, \beta_1 = 0.5}$):} Standard TDC exhibits stable performance. \textbf{Right ($\boldsymbol{\alpha_1 = 1.0, \beta_1 = 10.0}$):} Standard TDC exhibits large run-to-run variance. The TDC estimates show pronounced oscillations, and the pointwise one-standard deviation bands are substantially wider, reflecting TDC's sensitivity to step size selection.}\label{fig:BAIRD_TDC_INSTAB}
\end{figure}

To demonstrate the numerical instability caused by an inappropriate choice of step size, we consider value function approximation on a simple 11-state random walk and on the celebrated Baird's counterexample \citep{baird1995residual}. For each environment, we ran one hundred independent experiments using a step size sequence of the form $\alpha_n = \alpha_1/n^{0.7}$ for the random walk example, and $\alpha_n = \alpha_1/n^{0.8}$ with $\beta_n = \beta_1/n^{0.6}$ for Baird's counterexample, where $\alpha_1 > 0$ and $\beta_1 > 0$. Detailed descriptions of both environments are given in Section~\ref{SEC:NUMERICS}. In panels of Figure \ref{fig:VALUE_FUNC_TD0_INSTAB}, the average performance of each method and the true value function are shown as lines, while the shaded bands indicate one standard deviation above and below the mean. In the random walk environment, the goal is to approximate the true value function, shown as the red dashed line, using cosine and sine basis functions. The top left panel of Figure \ref{fig:VALUE_FUNC_TD0_INSTAB} plots the value function approximation error (log-MSE) against the initial step size for standard TD(0), averaged TD(0), and projected TD(0): small initial step sizes incur slow learning progress, reflected in the higher MSE at the left end of the step size range, while step sizes beyond a moderate range cause the error to grow sharply due to numerical instability. The remaining four panels of Figure \ref{fig:VALUE_FUNC_TD0_INSTAB} show the corresponding value function predictions. As the top right panel shows, with a moderate initial step size ($\alpha_1 = 10.0$), all three versions of TD(0), standard TD(0), averaged TD(0), and projected TD(0), shown in blue, yellow, and green, respectively, closely approximate the true value function, with tight standard deviation bands reflecting stable estimates. With a larger initial step size ($\alpha_1 = 30.0$), however, all three methods become highly unstable: as shown in the bottom three panels, their estimates deviate drastically from the true value function, and the standard deviation bands widen by several orders of magnitude, illustrating the sensitivity of these methods to the choice of initial step size. For Baird's example, Figure~\ref{fig:BAIRD_TDC_INSTAB} shows a substantial deterioration in the TDC algorithm's weight estimates as the initial step sizes change from $(\alpha_1, \beta_1) = (0.05, 0.5)$ to $(\alpha_1, \beta_1) = (1.0, 10.0)$, highlighting the sensitivity of standard TDC to step size selection. This observation is consistent with the step size sensitivity reported by \citet{white2016investigating}.

To address the numerical instability of TD algorithms, such as those shown in Figures \ref{fig:VALUE_FUNC_TD0_INSTAB}, a variety of strategies have been proposed in the literature. \citet{hutter2007temporal} introduced adaptive step size schedules based on discounted state visitation counters. While this method demonstrated improved stability in several settings, it can diverge in continuous domains. \citet{mahmood2012tuning} proposed an alternative adaptive step size scheme that requires tuning meta-parameters governing the decay rate. However, to avoid divergent behavior, they recommended using the step size schedule with sufficiently small initial values. This approach thus suffers from the same instability issues as standard TD methods, as it remains sensitive to the choice of step size. The Alpha-Bound algorithm \citep{dabney2012adaptive} introduced an adaptive bound on the effective step size and demonstrated improved stability over prior approaches. Nonetheless, it requires storing vector-valued quantities across all past TD iterations and can still exhibit high variance and divergence when initialized with a large step size. Importantly, none of the aforementioned step size adaptation methods provide theoretical guarantees or error bounds on the convergence of TD iterates. For a comprehensive discussion of other existing deterministic and stochastic step size strategies, we refer the reader to a review by \citet{george2006adaptive}.

\section{Implicit temporal difference learning}\label{SEC:IMPLICIT_TD}
\noindent In this section, we introduce implicit TD algorithms, which are designed  
to alleviate the numerical instability discussed in 
Section \ref{SUBSEC:INSTAB}. The key idea behind implicit updates lies in rewriting the recursive update as a fixed-point equation, where the future iterate appears on both the left- and right-hand sides of the update rule. To provide a concrete example, consider the standard stochastic gradient descent (SGD) algorithm \citep{bottou2010large, bottou2012stochastic} applied to an objective function $f$,
\begin{equation}\label{EQN:SGD}
\boldsymbol{w}_{n+1} = \boldsymbol{w}_{n} + \alpha_n \nabla f(\boldsymbol{w}_{n} ; \xi_n), \quad n\ge 1,
\end{equation}
where $\alpha_n$ is a non-negative step size and $\xi_n$ represents the random component involved in computing the $n^{\text{th}}$ stochastic gradient. An implicit version of the aforementioned SGD algorithm, given by
\begin{equation}\label{EQN:IMP_SGD}
\boldsymbol{w}^{\text{im}}_{n+1} = \boldsymbol{w}^{\text{im}}_n + \alpha_n \nabla f(\boldsymbol{w}^{\text{im}}_{n+1} ; \xi_n), \quad n\ge 1,
\end{equation}
was proposed and analyzed in \citet{toulis2014statistical, toulis2016towards, toulis2017asymptotic}, 
and \citet{chee2023plus}. Note that the 
highlighted term in~\eqref{EQN:IMP_SGD} indicates that the gradient is evaluated at the future iterate, resulting in a fixed-point equation. For a range of objective functions $f$, this equation admits a simple update rule \citep{toulis2014statistical}. Moreover, such an implicit update was shown to provide significant stability improvements over the standard SGD algorithm given in \eqref{EQN:SGD}. Motivated by this central idea of implicit SGD, our goal in the following two subsections is to develop implicit variants of TD algorithms.

\subsection{Implicit TD($0$)/TD($\lambda$) algorithms}
\noindent In this subsection, inspired by the principles of implicit SGD, we reformulate the TD update rules as fixed-point equations. Recall that $a_n = \pi_{\star}(\boldsymbol{x}_{n})$, $r_n = r(\boldsymbol{x}_{n}, a_n), \boldsymbol{\phi}_{n} = \phi(\boldsymbol{x}_{n})$, and $\boldsymbol{\phi}_{n+1}= \phi(\boldsymbol{x}_{n+1})$. Based on this formulation, we propose the following implicit TD(0) algorithm
\begin{align}
    \boldsymbol{w}^{\text{im}}_{n+1} &= \boldsymbol{w}^{\text{im}}_n + \alpha_n \delta^{\text{im}}_n \boldsymbol{\phi}_{n}, \label{IMP_TD0_ORIG}\\
    \delta^{\text{im}}_n &:= r_n + \gamma \boldsymbol{\phi}_{n+1}^{\top} \boldsymbol{w}^{\text{im}}_n - \boldsymbol{\phi}_{n}^{\top} \boldsymbol{w}^{\text{im}}_{n+1}, \nonumber 
\end{align}
and the implicit TD($\lambda$) algorithm \citep{tamar2014implicit}
\begin{align}
\boldsymbol{w}^{\text{im}}_{n+1} &= \boldsymbol{w}^{\text{im}}_n + \alpha_n \delta^{\text{im}}_n \boldsymbol{e}_{n},  \label{IMP_TDL_ORIG}\\
\delta^{\text{im}}_n &:= r_n + \gamma \boldsymbol{\phi}_{n+1}^{\top} \boldsymbol{w}^{\text{im}}_n + \lambda\gamma \boldsymbol{e}_{n-1}^{\top}\boldsymbol{w}^{\text{im}}_{n}- \boldsymbol{e}_{n}^{\top}\boldsymbol{w}^{\text{im}}_{n+1}, \nonumber \\
\boldsymbol{e}_{n} &:= \boldsymbol{\phi}_{n} + (\lambda\gamma)\boldsymbol{e}_{n-1}, ~ \boldsymbol{e}_{0} = 0, \nonumber
\end{align}
where $(\alpha_n)_{n\in \mathbb{N}}$ is a sequence of non-negative step sizes.
Combining the future iterate value $\boldsymbol{w}^{\text{im}}_{n+1}$ from both sides, 
\eqref{IMP_TD0_ORIG} can be rewritten as 
\begin{align*}
    \left(\mathbf{I} + \alpha_n \boldsymbol{\phi}_{n}\boldsymbol{\phi}_{n}^{\top} \right)\boldsymbol{w}^{\text{im}}_{n+1} = \boldsymbol{w}^{\text{im}}_n + \alpha_n (r_n + \gamma \boldsymbol{\phi}_{n+1}^{\top} \boldsymbol{w}^{\text{im}}_n ) \boldsymbol{\phi}_{n}.
\end{align*}
Analogously, equation~\eqref{IMP_TDL_ORIG} can be reexpressed as
\begin{align*}
\left(\mathbf{I} + \alpha_n \boldsymbol{e}_{n} \boldsymbol{e}_{n}^{\top} \right)\boldsymbol{w}^{\text{im}}_{n+1} = \boldsymbol{w}^{\text{im}}_n + \alpha_n (r_n + \gamma \boldsymbol{\phi}_{n+1}^{\top} \boldsymbol{w}^{\text{im}}_n + \lambda\gamma \boldsymbol{e}_{n-1}^{\top}\boldsymbol{w}^{\text{im}}_{n}) \boldsymbol{e}_{n}.     
\end{align*}
Using the Sherman-Morrison-Woodbury formula, we have
\begin{align*}
    \left(\mathbf{I} + \alpha_n \boldsymbol{\phi}_{n}\boldsymbol{\phi}_{n}^{\top} \right)^{-1} = \mathbf{I} - \frac{\alpha_n}{1 + \alpha_n ||\boldsymbol{\phi}_{n}||^2} \boldsymbol{\phi}_{n}\boldsymbol{\phi}_{n}^{\top} ~~\text{and}~~
    \left(\mathbf{I} + \alpha_n \boldsymbol{e}_{n} \boldsymbol{e}_{n}^{\top} \right)^{-1} =
    \mathbf{I} - \frac{\alpha_n}{1 + \alpha_n ||\boldsymbol{e}_{n}||^2} \boldsymbol{e}_{n} \boldsymbol{e}_{n}^{\top}.
\end{align*}
These expressions provide insight into why implicit TD algorithms are more stable than standard TD.
First, note that the norms of the update matrices shown above are both less than or equal to one, resulting in a stable update from 
$\boldsymbol{w}^{\text{im}}_n $ to $\boldsymbol{w}^{\text{im}}_{n+1}$. In each iteration, implicit algorithms utilize both feature and eligibility trace
information to impose adaptive shrinkage on the running iterates.  
In Algorithm \ref{ALG:Imp_TD} shown below, we present a concrete description of the implicit TD algorithms, with and without the projection step.

\begin{algorithm}[!h]
\begin{algorithmic}
    \caption{Implicit TD(0)/TD($\lambda)$}\label{ALG:Imp_TD}
        \STATE \textbf{Input:} initial guess $\boldsymbol{w}^{\text{im}}_1$, initial state $\boldsymbol{x}_1$, step size $(\alpha_n)_{n \in \mathbb{N}}$, trace parameter $\lambda$ (for TD($\lambda$)), projection radius $R > 0$ (for projected version)
        \STATE \textbf{For} $n \in \mathbb{N}$, 
        \textbf{do}:
        \vspace{-2.5mm}
        \begin{enumerate}
            \item Obtain values of the reward $r_n$ and next state $\boldsymbol{x}_{n+1}$.
            \vspace{-2.5mm}
            \item Compute the temporal difference error:
            \vspace{-2.5mm}
            \begin{equation*}
                \delta_{n} = r_n + \gamma \boldsymbol{\phi}_{n+1}^{\top} \boldsymbol{w}^{\text{im}}_{n} - \boldsymbol{\phi}_{n}^{\top} \boldsymbol{w}^{\text{im}}_{n}
                \vspace{-3mm}
            \end{equation*}
            \item For TD(0), update:
            \vspace{-2.5mm}
            \begin{align*}
                \boldsymbol{w}^{\text{im}}_{n+1} = \boldsymbol{w}^{\text{im}}_{n} + \frac{\alpha_n}{1+\alpha_n\|\boldsymbol{\phi}_{n}\|^2} \delta_{n} \boldsymbol{\phi}_{n} 
                \vspace{-2.5mm}
            \end{align*}\vspace{-2.5mm}For TD($\lambda$), update:
            \begin{align*}
            \boldsymbol{w}^{\text{im}}_{n+1} &= \boldsymbol{w}^{\text{im}}_{n} + \frac{\alpha_n}{1+\alpha_n\|\boldsymbol{e}_{n}\|^2} \delta_{n} \boldsymbol{e}_{n}, \\ 
            \boldsymbol{e}_{n} &= \boldsymbol{\phi}_{n} + (\lambda\gamma)\boldsymbol{e}_{n-1}, ~\text{with} ~ \boldsymbol{e}_{0} = 0
            \end{align*}\vspace{-7.5mm}
            \item (For projected implicit TD) If $\|\boldsymbol{w}^{\text{im}}_{n+1}\| > R$:
            \vspace{-2.5mm}
            \begin{equation*}
                \boldsymbol{w}^{\text{im}}_{n+1} = \frac{R}{\|\boldsymbol{w}^{\text{im}}_{n+1}\|} \boldsymbol{w}^{\text{im}}_{n+1}
                \vspace{-2.5mm}
            \end{equation*}
            \vspace{-2.5mm}
        \end{enumerate}
\end{algorithmic}
\end{algorithm}

We begin our analysis of implicit TD algorithms by 
establishing a connection to standard TD updates. This result 
is presented in Lemma \ref{LEMMA:IMP_EXP_RELATION} that follows.

\begin{lemma}\label{LEMMA:IMP_EXP_RELATION}
    An implicit update of TD($0$) given in \eqref{IMP_TD0_ORIG}
    can be written as
\begin{align} \label{IMP_TD0_RANDOM_STEP}
    \boldsymbol{w}^{\text{im}}_{n+1} = \boldsymbol{w}^{\text{im}}_n + \tilde\alpha_n \left(r_n + \gamma \boldsymbol{\phi}_{n+1}^{\top} \boldsymbol{w}^{\text{im}}_n - \boldsymbol{\phi}_{n}^{\top} \boldsymbol{w}^{\text{im}}_{n} \right)\boldsymbol{\phi}_{n},
\end{align}
where $\tilde\alpha_n = \frac{\alpha_n}{1+\alpha_n \|\boldsymbol{\phi}_{n}\|^2}$. Similarly, the implicit TD($\lambda$) given in \eqref{IMP_TDL_ORIG} can be expressed as
\begin{align}\label{IMP_TDL_RANDOM_STEP}
\boldsymbol{w}^{\text{im}}_{n+1} = \boldsymbol{w}^{\text{im}}_n + \tilde\alpha_n \left(r_n + \gamma \boldsymbol{\phi}_{n+1}^{\top} \boldsymbol{w}^{\text{im}}_n - \boldsymbol{\phi}_{n}^{\top} \boldsymbol{w}^{\text{im}}_{n} \right)\boldsymbol{e}_{n},
\end{align}
where $\tilde\alpha_n = \frac{\alpha_n}{1+\alpha_n \|\boldsymbol{e}_{n}\|^2}$.
\end{lemma}

From Lemma \ref{LEMMA:IMP_EXP_RELATION}, we see that the implicit TD(0) and TD($\lambda$) algorithms move along the same direction as the standard TD(0) and TD($\lambda$) algorithms, respectively. Unlike the standard TD algorithms, implicit TD algorithms provide an additional source of shrinkage in the running iterates through adaptive step sizes $(\tilde\alpha_n)_{n\in\mathbb{N}}$, which scale inversely with the norm of the feature or eligibility trace. Lemma \ref{LEMMA:IMP_EXP_RELATION} further shows that implicit updates can be computed without much additional cost, since the implicit TD(0) and TD($\lambda$) algorithms amount to using random step sizes $(\tilde\alpha_n)_{n\in\mathbb{N}}$. Combined with the projection step discussed in Section \ref{SUBSEC:TD_BACKGROUND}, this leads to projected implicit TD algorithms, which further enhance numerical stability.

Implicit updates thus adapt to the magnitude of the feature vector or eligibility trace at each step. When $\|\boldsymbol{\phi}_{n}\|$ or $\|\boldsymbol{e}_{n}\|$ is large, indicating a high-signal or potentially destabilizing transition, the effective step size is automatically reduced to prevent numerical blowup. When $\|\boldsymbol{\phi}_n\|$ or $\|\boldsymbol{e}_n\|$ is small to moderate, the step size stays close to $\alpha_n$, allowing the algorithm to fully exploit the available information. This built-in adaptivity lets the step size stay large when it is safe to do so, and shrink when caution is required. By adjusting updates in a state- and feature-dependent manner, implicit TD algorithms achieve considerably greater numerical stability and learning efficiency than the standard algorithms.

\subsection{Implicit TDC algorithm}
\noindent In the same spirit as the implicit TD(0) and TD($\lambda$) algorithms, here we introduce an implicit version of the TDC algorithm, which we refer to as the implicit TDC algorithm. 
Recall that $a_n \sim \pi_{b}(\cdot \mid \boldsymbol{x}_{n}), r_n = r(\boldsymbol{x}_{n}, a_n)$, $\boldsymbol{\phi}_{n} = \phi(\boldsymbol{x}_{n}), \boldsymbol{\phi}_{n+1}= \phi(\boldsymbol{x}_{n+1}),$ and $\rho_n = \pi_\star(a_n \mid \boldsymbol{x}_{n})/\pi_b(a_n \mid \boldsymbol{x}_{n})$. We propose the following implicit version of the aforementioned TDC algorithm:
\begin{align}
    \boldsymbol{w}^{\text{im}}_{n+1} &= \boldsymbol{w}^{\text{im}}_n + \alpha_n \rho_n \left(r_n \boldsymbol{\phi}_{n} + \gamma \boldsymbol{\phi}_{n} \boldsymbol{\phi}_{n+1}^{\top} \boldsymbol{w}^{\text{im}}_n - \gamma \boldsymbol{\phi}_{n+1}\boldsymbol{\phi}_{n}^{\top} \boldsymbol{u}^{\text{im}}_{n}  \right) - \alpha_n \rho_n \boldsymbol{\phi}_{n} \boldsymbol{\phi}_{n}^{\top} \boldsymbol{w}^{\text{im}}_{n+1}, \label{IMP_WEIGHT_UPDATE}  \\
    \boldsymbol{u}^{\text{im}}_{n+1} &= \boldsymbol{u}^{\text{im}}_{n} + \beta_n \rho_n \left(r_n \boldsymbol{\phi}_{n} + \gamma \boldsymbol{\phi}_{n} \boldsymbol{\phi}_{n+1}^{\top}\boldsymbol{w}^{\text{im}}_n - \boldsymbol{\phi}_{n} \boldsymbol{\phi}_{n}^{\top} \boldsymbol{w}^{\text{im}}_n \right) - \beta_n \rho_n \boldsymbol{\phi}_{n} \boldsymbol{\phi}_{n}^{\top} \boldsymbol{u}^{\text{im}}_{n+1}.\label{IMP_AUX_UPDATE} 
\end{align}
In these expressions, the sequences $(\alpha_n)_{n \in \mathbb{N}}$ and $(\beta_n)_{n \in \mathbb{N}}$ are non-negative step sizes. To gain insight into the numerical stability of the implicit TDC update, observe that the implicit TDC update for the primary parameter $w^{\text{im}}$ can be rewritten as follows
\begin{align*}
    &\left(\mathbf{I} + \alpha_n \rho_n \boldsymbol{\phi}_{n} \boldsymbol{\phi}_{n}^{\top} \right)\boldsymbol{w}^{\text{im}}_{n+1} = \boldsymbol{w}^{\text{im}}_n + \alpha_n \rho_n \left(r_n \boldsymbol{\phi}_{n} + \gamma \boldsymbol{\phi}_{n} \boldsymbol{\phi}_{n+1}^{\top} \boldsymbol{w}^{\text{im}}_n - \gamma \boldsymbol{\phi}_{n+1}\boldsymbol{\phi}_{n}^{\top} \boldsymbol{u}^{\text{im}}_{n}  \right) \nonumber \\
    &\Leftrightarrow \boldsymbol{w}^{\text{im}}_{n+1} = \left(\mathbf{I} + \alpha_n \rho_n \boldsymbol{\phi}_{n} \boldsymbol{\phi}_{n}^{\top} \right)^{-1}\left\{\boldsymbol{w}^{\text{im}}_n + \alpha_n \rho_n \left(r_n \boldsymbol{\phi}_{n} + \gamma \boldsymbol{\phi}_{n} \boldsymbol{\phi}_{n+1}^{\top} \boldsymbol{w}^{\text{im}}_n - \gamma \boldsymbol{\phi}_{n+1}\boldsymbol{\phi}_{n}^{\top} \boldsymbol{u}^{\text{im}}_{n}  \right)\right\} \nonumber \\
    &\Leftrightarrow \boldsymbol{w}^{\text{im}}_{n+1} = \left(\mathbf{I} - \alpha'_n \rho_n \boldsymbol{\phi}_{n} \boldsymbol{\phi}_{n}^{\top} \right)\left\{\boldsymbol{w}^{\text{im}}_n + \alpha_n \rho_n \left(r_n \boldsymbol{\phi}_{n} + \gamma \boldsymbol{\phi}_{n} \boldsymbol{\phi}_{n+1}^{\top} \boldsymbol{w}^{\text{im}}_n - \gamma \boldsymbol{\phi}_{n+1}\boldsymbol{\phi}_{n}^{\top} \boldsymbol{u}^{\text{im}}_{n}  \right)\right\} 
\end{align*}
where $\alpha'_n = \frac{\alpha_n}{1+\alpha_n \rho_n \|\boldsymbol{\phi}_{n}\|^2}$. Similarly, the implicit TDC update for the auxiliary parameter $u^{\text{im}}$ can be rewritten as follows
\begin{align*}
   &\left(\mathbf{I} + \beta_n \rho_n \boldsymbol{\phi}_{n}\boldsymbol{\phi}_{n}^{\top} \right) \boldsymbol{u}^{\text{im}}_{n+1} = \boldsymbol{u}^{\text{im}}_{n} + \beta_n \rho_n \delta_n^{\text{im}}\boldsymbol{\phi}_{n} \nonumber \\
    &\Leftrightarrow \boldsymbol{u}^{\text{im}}_{n+1} = \left(\mathbf{I} + \beta_n \rho_n \boldsymbol{\phi}_{n}\boldsymbol{\phi}_{n}^{\top} \right)^{-1}\left(\boldsymbol{u}^{\text{im}}_{n} + \beta_n \rho_n \delta_n^{\text{im}}\boldsymbol{\phi}_{n} \right)\nonumber \\
    &\Leftrightarrow \boldsymbol{u}^{\text{im}}_{n+1} = \left(\mathbf{I} - \beta'_n \rho_n \boldsymbol{\phi}_{n}\boldsymbol{\phi}_{n}^{\top} \right)\left(\boldsymbol{u}^{\text{im}}_{n} + \beta_n \rho_n \delta_n^{\text{im}}\boldsymbol{\phi}_{n} \right) 
\end{align*}
where $\delta^{\text{im}}_{n} = r_n + \gamma \boldsymbol{\phi}_{n+1}^{\top} \boldsymbol{w}^{\text{im}}_{n} - \boldsymbol{\phi}_{n}^{\top} \boldsymbol{w}^{\text{im}}_{n}$ and $\beta'_n = \frac{\beta_n}{1+\beta_n \rho_n \|\boldsymbol{\phi}_{n}\|^2}$.
It turns out that the implicit TDC admits a succinct update expression, requiring the same order of computational cost as the standard TDC algorithm. A complete characterization of the implicit TDC update is in  
Lemma \ref{LEMMA:IMP_TDC_EXP_RELATION}.

\begin{lemma}\label{LEMMA:IMP_TDC_EXP_RELATION}
Implicit TDC algorithm given in \eqref{IMP_WEIGHT_UPDATE}
and \eqref{IMP_AUX_UPDATE} can be written as
\begin{align} 
    \boldsymbol{w}^{\text{im}}_{n+1} &= \boldsymbol{w}^{\text{im}}_n +  \alpha'_n\rho_n \delta^{\text{im}}_n \boldsymbol{\phi}_{n} - \alpha_n \rho_n \gamma \left(\boldsymbol{\phi}_{n}^{\top} \boldsymbol{u}^{\text{im}}_{n}\right) \left\{\boldsymbol{\phi}_{n+1}- \alpha'_n \rho_n \left(\boldsymbol{\phi}_{n}^{\top} \boldsymbol{\phi}_{n+1}\right) \boldsymbol{\phi}_{n} \right\}, \label{IMP_TDC_WEIGHT}\\
    \boldsymbol{u}^{\text{im}}_{n+1} &= \boldsymbol{u}^{\text{im}}_{n} + \beta'_n \rho_n \delta_n^{\text{im}}\boldsymbol{\phi}_{n} - \beta'_n \rho_n \boldsymbol{\phi}_{n} \boldsymbol{\phi}_{n}^{\top} \boldsymbol{u}_{n}^{\text{im}}, \label{IMP_TDC_AUX}
\end{align}
where $\alpha'_n = \frac{\alpha_n}{1+\alpha_n\rho_n \|\boldsymbol{\phi}_{n}\|^2}$, $\beta'_n = \frac{\beta_n}{1+\beta_n\rho_n \|\boldsymbol{\phi}_{n}\|^2}$ and $\delta_n^{\text{im}}=r_n + \gamma \boldsymbol{\phi}_{n+1}^{\top} \boldsymbol{w}^{\text{im}}_n -\boldsymbol{\phi}_{n}^{\top} \boldsymbol{w}_{n}^{\text{im}}$.
\end{lemma}

Compared to the standard TDC updates given in \eqref{TDC_main_update} and \eqref{TDC_aux_update}, Lemma \ref{LEMMA:IMP_TDC_EXP_RELATION} reveals that the implicit TDC algorithm closely resembles the standard TDC algorithm, but with adjusted step sizes and a modified correction term. Specifically, in the primary parameter update of the implicit TDC algorithm, $\alpha'_n$ serves as a data-adaptive version of the step size $\alpha_n$. Similarly, $\beta'_n$ replaces the original step size $\beta_n$ in the auxiliary parameter update. Regarding the gradient correction term, the implicit TDC algorithm adjusts the TD update in the direction of
$$
-\alpha_n \rho_n \gamma \left(\boldsymbol{\phi}_{n}^{\top} \boldsymbol{u}^{\text{im}}_{n}\right) \left\{\boldsymbol{\phi}_{n+1}- \alpha'_n \rho_n \left(\boldsymbol{\phi}_{n}^{\top} \boldsymbol{\phi}_{n+1}\right) \boldsymbol{\phi}_{n} \right\},
$$
in contrast to the standard TDC algorithm’s correction term
$$
-\alpha_n \rho_n \gamma (\boldsymbol{\phi}_{n}^{\top} \boldsymbol{u}_{n})\boldsymbol{\phi}_{n+1}.
$$ 
Roughly speaking, the standard TDC algorithm leverages the full information contained in $\boldsymbol{\phi}_{n+1}$, whereas the implicit TDC algorithm effectively filters out the component of $\boldsymbol{\phi}_{n+1}$ that is aligned with $\boldsymbol{\phi}_{n}$. This reduces the correlation between consecutive gradient correction terms, which can enhance numerical stability. In Algorithm \ref{ALG:Imp_TDC} shown below, we present a concrete description of the implicit TDC algorithm that was introduced in this section.

\begin{algorithm}[!ht]
\begin{algorithmic}
    \caption{Implicit TDC}\label{ALG:Imp_TDC}
        \STATE \textbf{Input:} initial guess $\boldsymbol{w}^{\text{im}}_1$, $\boldsymbol{u}^{\text{im}}_1$, initial state $\boldsymbol{x}_1$, step size $(\alpha_n)_{n \in \mathbb{N}}$, step size $(\beta_n)_{n \in \mathbb{N}}$, projection radius $R_{\boldsymbol{w}}, R_{\boldsymbol{u}} \in \mathbb{R}_{>0}$ (for projected version)
        \STATE \textbf{For} $n \in \mathbb{N} $, 
        \textbf{do}:
        \vspace{-2.5mm}
        \begin{enumerate}
            \item Obtain values of the reward $r_n$ and next state $\boldsymbol{x}_{n+1}$
            \vspace{-2.5mm}
            \item Compute the temporal difference error:
            \vspace{-2.5mm}
            \begin{equation*}
                \delta^{\text{im}}_{n} = r_n + \gamma \boldsymbol{\phi}_{n+1}^{\top} \boldsymbol{w}^{\text{im}}_{n} - \boldsymbol{\phi}_{n}^{\top} \boldsymbol{w}^{\text{im}}_{n}
                \vspace{-3mm}
            \end{equation*}
            \item Update:
            \vspace{-2.5mm}
            \begin{align*}
             \boldsymbol{w}^{\text{im}}_{n+1} &= \boldsymbol{w}^{\text{im}}_n + \alpha'_n \rho_n \delta^{\text{im}}_n \boldsymbol{\phi}_{n} - \alpha_n \rho_n \gamma \left(\boldsymbol{\phi}_{n}^{\top} \boldsymbol{u}^{\text{im}}_{n}\right) \left\{\boldsymbol{\phi}_{n+1}- \alpha'_n \rho_n \left(\boldsymbol{\phi}_{n}^{\top} \boldsymbol{\phi}_{n+1}\right) \boldsymbol{\phi}_{n} \right\} \\
            \boldsymbol{u}^{\text{im}}_{n+1} &= \boldsymbol{u}^{\text{im}}_{n} + \beta'_n\rho_n \delta_n^{\text{im}}\boldsymbol{\phi}_{n} - \beta'_n \rho_n\boldsymbol{\phi}_{n} \boldsymbol{\phi}_{n}^{\top} \boldsymbol{u}_{n}^{\text{im}}
            \end{align*}
            with $\alpha'_n = \frac{\alpha_n}{1+\alpha_n\rho_n \|\boldsymbol{\phi}_{n}\|^2}$ and $\beta'_n = \frac{\beta_n}{1+\beta_n\rho_n \|\boldsymbol{\phi}_{n}\|^2}$
            \item (For projected implicit TDC) \vspace{2.5mm}\\
            If $\|\boldsymbol{w}^{\text{im}}_{n+1}\| > R_{\boldsymbol{w}}$:
            \vspace{-2.5mm}
            \begin{equation*}
                \boldsymbol{w}^{\text{im}}_{n+1} = \frac{R_{\boldsymbol{w}}}{\|\boldsymbol{w}^{\text{im}}_{n+1}\|} \boldsymbol{w}^{\text{im}}_{n+1}
                \vspace{-2.5mm}
            \end{equation*}
            If $\|\boldsymbol{u}^{\text{im}}_{n+1}\| > R_{\boldsymbol{u}}$:
            \vspace{-2.5mm}
            \begin{equation*}
                \boldsymbol{u}^{\text{im}}_{n+1} = \frac{R_{\boldsymbol{u}}}{\|\boldsymbol{u}^{\text{im}}_{n+1}\|} \boldsymbol{u}^{\text{im}}_{n+1}
                \vspace{-2.5mm}
            \end{equation*}
            \vspace{-2.5mm}
        \end{enumerate}
\end{algorithmic}
\end{algorithm}

\section{Theoretical analysis}\label{SEC:THEORY}
\noindent In this section, we provide a theoretical analysis of proposed implicit TD algorithms. We begin by listing out assumptions and definitions that will be used throughout this section. Unless explicitly noted otherwise, $\|\cdot\|$ denotes the Euclidean norm for vectors and the corresponding induced norm for matrices. The first assumption we introduce imposes restrictions on the data generating process.

\begin{assumption}[Aperiodicity and irreducibility of Markov chain]\label{ASSUMP:IRR_APE_MARKOV} The Markov chain $(\boldsymbol{x}_{n})_{n \in \mathbb{N}}$ is aperiodic and irreducible with a unique steady-state distribution $\mu_\pi$ with $\mu_{\pi}(\boldsymbol{x}) > 0$ for all $\boldsymbol{x} \in \mathcal{X}$. In the on-policy evaluation setting, we assume that $\mu_\pi = \mu_{\pi_\star}$ for some target policy $\pi_\star$. In the off-policy evaluation setting, we assume $\mu_\pi = \mu_{\pi_b}$, where $\pi_b$ is the behavior policy used to generate the data. 
\end{assumption}

\noindent Note that Assumption \ref{ASSUMP:IRR_APE_MARKOV}, together with the finiteness of the state space, implies that the Markov chain $(\boldsymbol{x}_{n})_{n \in \mathbb{N}}$ mixes at a uniform geometric rate \citep{levin2017markov}, i.e., $(\boldsymbol{x}_{n})_{n \in \mathbb{N}}$ is uniformly ergodic. That is, there exist constants $m > 0$ and $\rho \in (0, 1)$ such that
\begin{align}\label{GEOM_MIX}
\sup_{\boldsymbol{x} \in \mathcal{X}} d_{\text{TV}}\left\{\mathbb{P}(\boldsymbol{x}_{n} \in \cdot \mid \boldsymbol{x}_1 = \boldsymbol{x}), \mu_\pi \right\} \le m \rho^n \quad \forall n \in \mathbb{N},
\end{align}
where $d_{\text{TV}}(P, Q)$ denotes the total-variation distance between probability measures $P$ and $Q$. Here, the initial distribution of $\boldsymbol{x}_1$ is the steady-state distribution $\mu_\pi$, i.e., $(\boldsymbol{x}_1, \boldsymbol{x}_2, \ldots)$ is a stationary sequence. We next list out some assumptions on the environment and feature mapping used for approximating the value function.

\begin{assumption}[Bounded reward]\label{ASSUMP:BDD_RWD}
There exists $r_{\max} > 0$ such that $|r_n| \le r_{\max}, \forall n \in \mathbb{N}$.
\end{assumption}

\begin{assumption}[Bounded features]\label{ASSUMP:NORM_FEAT}
There exists $\phi_{\max} > 0$ such that $\|\boldsymbol{\phi}_{n}\| \le \phi_{\max}, \forall n \in \mathbb{N}$.
\end{assumption}

\begin{assumption}[Full rank]\label{ASSUMP:FULL_RANK}
Define $\Phi = [
    \phi(\boldsymbol{x})^{\top}
]_{\boldsymbol{x} \in {\mathcal{X}}}$ as 
the full state matrix where the $k^{\text{th}}$ row corresponds to $\phi$ evaluated at the $k^{\text{th}}$ state in $\mathcal{X}$. We assume that $\Phi$ is full rank.
\end{assumption}

\noindent Assumptions \ref{ASSUMP:IRR_APE_MARKOV}, \ref{ASSUMP:BDD_RWD}, \ref{ASSUMP:NORM_FEAT} and \ref{ASSUMP:FULL_RANK} are widely accepted in the literature \citep{tsitsiklis1996analysis, bertsekas1996neuro, bhandari2018finite, srikant2019finite, samsonov2024improved,li2026towards,lee2026single}. They are considered to be mild as they encompass many real-world RL environments. In particular, Assumption \ref{ASSUMP:FULL_RANK} can be satisfied by removing redundant features. In our theory, the combined role of Assumption \ref{ASSUMP:IRR_APE_MARKOV} and Assumption~\ref{ASSUMP:FULL_RANK} is to preclude 
irregularities in the long-term behavior of the TD algorithm since, under these assumptions, the steady-state feature covariance matrix,
$$
    \boldsymbol{\Gamma} = \boldsymbol{\Phi}^{\top} \mathbf{D} \Phi = \sum_{\boldsymbol{x} \in \mathcal{X}} \mu_\pi(\boldsymbol{x}) \phi(\boldsymbol{x})\phi(\boldsymbol{x})^{\top},
$$
    is positive definite, where we set $\mathbf{D} := \text{diag}\{\mu_\pi(\boldsymbol{x})\}_{\boldsymbol{x} \in \mathcal{X}}$.
We will denote the minimum eigenvalue of $\boldsymbol{\Gamma}$ as $\lambda_{\text{min}}$. Moreover, thanks to Assumption \ref{ASSUMP:NORM_FEAT}, we have that $\lambda_{\text{min}} \in (0, \phi_{\max}^2]$. Lastly, for the statement of the finite-time MSE bounds, we introduce the mixing time of the Markov chain $(\boldsymbol{x}_{n})_{n \in \mathbb{N}}$ which appears in the bounds we establish.

 \begin{definition}[Mixing time]\label{DEF:MC_MIX_TIME}
Given a threshold $\epsilon > 0$, constants $\rho \in (0,1)$ and $m \in (0, \infty)$, the mixing time of the uniformly ergodic Markov chain $(\boldsymbol{x}_{n})_{n \in \mathbb{N}}$ is defined as
$$
 \tau_\epsilon = \min \{n \in \mathbb{N} \mid m \rho^n \le \epsilon\}.
$$ 
\end{definition}

\noindent For the TD($\lambda$) algorithm, a modified definition of mixing time, which reflects the geometric weighting of the eligibility trace will be used in the finite-time MSE bound expression. A formal definition is given below. 

\begin{definition}[$\lambda$-mixing time]\label{DEF:DOUBLE_MIX}
Given a trace-decay parameter $\lambda \in (0, 1)$, a discount factor $\gamma \in (0, 1)$, and a threshold $\epsilon > 0$, the $\lambda$-mixing time of the uniformly ergodic Markov chain $(\boldsymbol{x}_{n})_{n\in \mathbb{N}}$ is defined as
\begin{align*}
\tau_{\lambda, \epsilon} = \max\{\tau_{\epsilon}, \tau_{\epsilon}^{\lambda}\} \quad \text{where}\quad  \tau^{\lambda}_{\epsilon} := \min\left\{n\in \mathbb{N} \mid (\lambda\gamma)^n \le \epsilon\right\}.
\end{align*}
\end{definition}

\noindent To understand how these quantities behave as $\epsilon$ decreases, consider the case where $\epsilon =\mathcal{O}(1/t^s)$ for some $s > 0$. Under this condition, it can be shown that both $\tau_{\epsilon}$ and $\tau_{\lambda, \epsilon}$ grow at a rate of $\mathcal{O}(\log t)$.


\subsection{Almost sure convergence of implicit TD}
\label{SUBSEC:ASYMPTOTIC}
\noindent Under the aforementioned assumptions, we can establish the almost sure convergence of the implicit TD(0) and TD($\lambda$) algorithms.

\begin{theorem}[Almost sure convergence of implicit TD] \label{THM:ASYM_IMP_TD}
Under Assumptions \ref{ASSUMP:IRR_APE_MARKOV}-\ref{ASSUMP:FULL_RANK} with a step size sequence $\alpha_n = \alpha_1n^{-s}, ~\alpha_1 \in \mathbb{R}_{>0}, ~s \in (0.5, 1]$, implicit TD($0$) and TD($\lambda$) iterates 
converge almost surely to $\boldsymbol{w}_{\star}$, respectively satisfying \eqref{MEAN_PATH_TD0} and \eqref{MEAN_PATH_TDL}.
\end{theorem}

\noindent The main challenge in proving almost sure convergence of the implicit algorithms is that, unlike standard TD algorithms, where the deterministic step sizes satisfy the Robbins–Monro condition, i.e., $\sum_{n=1}^\infty \alpha_n = \infty$ and $\sum_{n=1}^\infty \alpha_n^2 < \infty$, the effective step sizes $(\tilde\alpha_n)_{n \in \mathbb{N}}$ for implicit algorithms are random. This rules out the direct application of numerous existing convergence results for stochastic approximation algorithms, including the well-known ODE method \citep{borkar2000ode, kushner2003stochastic, borkar2008stochastic, meyn2022control, borkar2025ode, liu2025ode} as well as almost-supermartingale based approaches \citep{robbins1971convergence, karandikar2024convergence, qian2024almost, liu2025extensions}. Instead, our approach relies on the Lyapunov-function-based approach of \citep{benveniste2012adaptive} (Theorem 17, p. 239), which can incorporate the difference between the standard TD update and the implicit TD update into a second-order term of the form $\mathcal{O}(\alpha_n^2)$ to yield the almost sure convergence of the implicit TD iterates.

\subsection{Finite-time error analysis of implicit TD with projection}
\noindent While the almost sure convergence established in the previous subsection provides an asymptotic justification for the use of implicit TD algorithms, their advantage over standard TD methods is not fully captured by Theorem~\ref{THM:ASYM_IMP_TD}. To further justify the superior numerical stability of implicit TD algorithms, we establish finite-time MSE bounds for implicit algorithms. We extend the information-theoretic approach of \citet{bhandari2018finite}, which relies on a projection step to ensure the deviation of the TD update from the mean-path TD update remains uniformly bounded, to the setting of random step sizes. Thanks to this projection, the deviation between the standard TD update and the implicit TD update is uniformly bounded and can be absorbed into a bias term that vanishes as the step size approaches zero. Such a projection step is easy to incorporate into the algorithm, and the projection radius $R \in \mathbb{R}_{>0}$ can be set arbitrarily large to include the optimal weight without sacrificing convergence. The strength of incorporating projection lies in that it allows one to establish finite-time MSE bounds holding for all $n \in \mathbb{N}$, which provides insight into potential numerical instability that may arise in the early phase.

We highlight that our setting operates under a Markovian data framework, where dependence between observations is explicitly accounted for. This distinction is practically important: recent works by \citet{huo2023bias} and \citet{lauand2023curse} have shown that, with Polyak–Ruppert averaging, a constant step size $\alpha$ in the Markovian setting leads to an irreducible asymptotic bias of order $\mathcal{O}(\alpha)$ in the MSE. In contrast, several existing error bounds derived under i.i.d. assumptions report asymptotically vanishing bias \citep{lakshminarayanan2018linear,patil2023finite, li2024high,samsonov2024improved}, suggesting that the gap between the Markovian and i.i.d.\ settings deserves careful attention.

The MSE bound established below holds in the Markovian setting for all $n \in \mathbb{N}$, and, to the best of our knowledge, is the first all time MSE bound obtained with a data-adaptive step size for TD algorithms. Furthermore, the step size we use does not depend on any environment specific quantities. We introduce the following function, which captures the effect of the initialization point:
$$
\psi(n) := \begin{cases}
\dfrac{n^{1-s}-1}{1-s} \quad &\text{for } s \in (0.5, 1),\\[4pt]
\log n \quad &\text{for } s =1.
\end{cases}
$$
We first present an MSE bound between the optimal weight $\boldsymbol{w}_{\star}$ and the implicit TD(0) iterates.

\begin{theorem}[Finite-time analysis of implicit TD(0) with projection]\label{THM:FIN_PROJ_TD0}
Suppose Assumptions \ref{ASSUMP:IRR_APE_MARKOV}--\ref{ASSUMP:FULL_RANK} hold with a decreasing step size $\alpha_n = \frac{\alpha_1}{n^s}, ~\alpha_1 \in \mathbb{R}_{>0}, ~s \in (0.5, 1]$. If $s \in (0.5, 1)$, or $s = 1$ and $\eta \alpha_1 > 1$ where
$$
\eta = \frac{2(1-\gamma)\lambda_{\min}}{1+\alpha_1\{\phi_{\max}^2 - 2(1-\gamma)\lambda_{\min}\}\mathbb{I}\{\phi_{\max}^2 > 2(1-\gamma)\lambda_{\min}\}},
$$
the projected implicit TD($0$) iterates with $R \ge \|\boldsymbol{w}_{\star}\|$ satisfy
\begin{align*}
\mathbb{E}_{\mu_{\pi_\star}}\left\|\boldsymbol{w}_{\star}-\boldsymbol{w}^{\text{im}}_{n+1}\right\|^2 \lesssim \exp\left\{-\eta\alpha_1
\psi(n)\right\} \left\|\boldsymbol{w}_{\star}-\boldsymbol{w}^{\text{im}}_{1}\right\|^2 + \mathcal{O}\left(\tau_{\alpha_n} \alpha_{(n-\tau_{\alpha_n})\vee 1} \right), \quad \forall n \in \mathbb{N}.
\end{align*}
\end{theorem}

\begin{remark}
Under the Markovian data setting, the above finite-time bound holds regardless of the initial step size when $s \in (0.5, 1)$. In comparison, one line of existing finite-time bounds holds only for a sufficiently large time index \citep{srikant2019finite, mitra2024simple}, which fails to capture the potential numerical instability of TD algorithms in the early phase. Another line of work relies on step sizes that require knowledge of the environment or of quantities dependent on the stationary distribution \citep{bhandari2018finite, mitra2024simple, lee2026single}. Other approaches modify TD algorithms with a data-dropping procedure that similarly requires environment knowledge \citep{samsonov2024improved}, or incorporate environment-dependent regularization \citep{li2026towards}. All of these are difficult to deploy in practice. Furthermore, such dependence of the step size or algorithm on environment specific quantities highlights the sensitivity of the standard TD(0) algorithm to the choice of step size. In contrast, for the implicit TD algorithm with $s \in (0.5, 1)$, the above finite-time MSE bound holds for any initial step size. 

\end{remark}

\begin{remark}
For $s \in (0.5, 1)$, the bound above holds for all step sizes regardless of the size of the maximum feature norm $\phi_{\max}$, which further shows that the implicit algorithm is robust to the choice of feature basis. In contrast, existing works \citep{lakshminarayanan2018linear, patil2023finite, lee2025finite, lee2026single} choose a step size inversely proportional to the maximum feature norm to guarantee finite-time MSE bounds, or explicitly assume feature normalization \citep{bhandari2018finite, mitra2024simple, samsonov2024improved}, both of which further complicate step size calibration. 
\end{remark}

\begin{remark}
For $s = 1$, if the maximum feature norm satisfies $\phi_{\max}^2 \le 2(1-\gamma)\lambda_{\min}$, where $\lambda_{\min} \in (0, \phi_{\max}^2]$, the above bound motivates the use of a larger initial step size satisfying $\eta\alpha_1 > 1$ without suffering numerical instability; in fact, it precludes the use of small step sizes. Meanwhile, when $\phi_{\max}^2 > 2(1-\gamma)\lambda_{\min}$, for larger initial step sizes to satisfy $\eta\alpha_1 > 1$, one needs $\phi_{\max}^2 < 4(1-\gamma)\lambda_{\min}$. This requirement reflects the difficulty of the problem: the easier the problem (i.e., the larger $\lambda_{\min}$), the safer it is to deploy the algorithm with a feature basis of larger scale.
\end{remark}

\begin{remark}
Our analysis indicates that using a large initial step size for implicit TD(0) is both feasible and desirable: doing so reduces the initialization effect much more rapidly in the early phase, while the asymptotic convergence rate remains governed by the $\mathcal{O}(\tau_{\alpha_n}\alpha_{(n-\tau_{\alpha_n})\vee1})$ term, which yields a rate of $\tilde{\mathcal{O}}(n^{-s})$ where $\tilde{\mathcal{O}}$ denotes $\mathcal{O}$ up to polylogarithmic order terms. For $s = 1$, this matches the rates established in \citep{bhandari2018finite, mitra2024simple, samsonov2024improved, lee2026single}. Implicit TD(0) therefore retains superior numerical stability without sacrificing standard TD's asymptotic convergence rate. This flexibility stands in stark contrast to standard TD(0), whose finite-time bounds often impose restrictive conditions on the step size that preclude the use of large values \citep{bhandari2018finite, patil2023finite, mitra2024simple, lee2025finite, li2026towards, lee2026single}.
\end{remark}

\noindent Next, we provide a finite-time MSE bound for the projected implicit TD($\lambda$) algorithm.

\begin{theorem}[Finite-time analysis of implicit TD($\lambda$) with projection]\label{THM:FIN_PROJ_TDL}
Suppose Assumptions \ref{ASSUMP:IRR_APE_MARKOV}--\ref{ASSUMP:FULL_RANK} hold with a decreasing step size $\alpha_n = \frac{\alpha_1}{n^s}, ~\alpha_1 \in \mathbb{R}_{>0}, ~s \in (0.5, 1]$. If $s \in (0.5, 1)$, or $s = 1$ and $\eta_\lambda \alpha_1 > 1$ where
$$
\eta_{\lambda} = \frac{2(1-\lambda\gamma)(1-\gamma)\lambda_{\min}}{1+\alpha_1\{e_{\max}^2 - 2(1-\lambda\gamma)(1-\gamma)\lambda_{\min}\}\mathbb{I}\{e_{\max}^2 > 2(1-\lambda\gamma)(1-\gamma)\lambda_{\min}\}}, \quad e_{\max} = \frac{\phi_{\max}}{1-\lambda\gamma} 
$$
the projected implicit TD($\lambda$) iterates with $R \ge \|\boldsymbol{w}_{\star}\|$ satisfy
\begin{align*}
\mathbb{E}_{\mu_{\pi_\star}}\left\|\boldsymbol{w}_{\star}-\boldsymbol{w}^{\text{im}}_{n+1}\right\|^2 \lesssim \exp\left\{-\eta_{\lambda}\alpha_1
\psi(n)\right\} \left\|\boldsymbol{w}_{\star}-\boldsymbol{w}^{\text{im}}_{1}\right\|^2 + \mathcal{O}(\tau_{\lambda,\alpha_n} \alpha_{(n-2\tau_{\lambda,\alpha_n})\vee 1}), \quad \forall n \in \mathbb{N}.
\end{align*}
\end{theorem}

\begin{remark}
When $\lambda = 0$, the above result reduces to the case of TD(0). Implicit TD($\lambda$) possesses the same benefits as implicit TD(0) relative to standard TD(0). For $s \in (0.5, 1)$, it admits any initial step size with any feature basis, while for $s = 1$, our analysis indicates that the initial step size $\alpha_1$ should be chosen large enough, with the feature basis satisfying $e_{\max}^2 < 4(1-\lambda\gamma)(1-\gamma)\lambda_{\min}$, where $\lambda_{\min} \in (0, \phi_{\max}^2]$. The asymptotic convergence rate of implicit TD($\lambda$) is $\tilde{\mathcal{O}}(n^{-s})$ for $s \in (0.5, 1]$, which coincides with that of standard TD($\lambda$).
\end{remark}

\begin{remark}
We note that finite-time MSE bounds for implicit TD methods without a projection step can be established either using step sizes that depend on quantities associated with the stationary distribution of the underlying Markov chain \citep{lee2026single}, or by deriving bounds that hold only for sufficiently large time indices \citep{srikant2019finite, mitra2024simple}, an approach that overlooks the numerical instability that TD methods can exhibit in its early phase. 
\end{remark}

\begin{remark}
The finite-time bounds obtained in this section serve as a natural extension of the finite-time bounds for implicit SGD established in \citep{toulis2016towards, toulis2017asymptotic} to the Markovian data setting. The implicit framework thus naturally brings superior numerical stability over standard algorithms while retaining the same asymptotic convergence rate.
\end{remark}

\subsection{Finite-time analysis of implicit TDC with projection}
\noindent In this subsection, we establish finite-time MSE bounds for the projected implicit TDC algorithm to demonstrate its superior numerical stability with respect to the choice of step size. Recall that in the off-policy evaluation setting, the goal is to approximate the value function under the target policy with data generated from the behavior policy. To this end, we restrict the class of behavior policies to a reasonable set of policies to ensure optimal value function approximation is identifiable. Widely accepted assumptions for such a guarantee are listed below \citep{wang2017finite,xu2019two}.

\begin{assumption}[Importance weights]\label{ASSUMP:importance_weight}
Given $\rho(\boldsymbol{x},a) := \pi_\star(a\mid \boldsymbol{x}) / \pi_b(a\mid \boldsymbol{x})$,
\begin{align*}
\rho_{\max} &:= \max_{\boldsymbol{x} \in \mathcal{X}, a \in \mathcal{A}}\pi_\star(a\mid \boldsymbol{x}) / \pi_b(a\mid \boldsymbol{x}) < \infty.
\end{align*}
\end{assumption}

\noindent Assumption \ref{ASSUMP:importance_weight} is a mild condition to guarantee that the support of the behavior policy contains the support of the target policy. Furthermore, with a slight abuse of notation, in the context of off-policy evaluation\footnote{In the context of TD(0), the definitions on page 6 imply $\boldsymbol{A}:= \mathbb{E}_{\mu_{\pi_\star}}[\phi(\boldsymbol{x})\left\{\gamma \phi\left(\boldsymbol{x}'\right)-\phi(\boldsymbol{x})\right\}^{\top}]$ and $\boldsymbol{b}:=\mathbb{E}_{\mu_{\pi_\star}}\{r(\boldsymbol{x},a)\phi(\boldsymbol{x})\}$.}, let us define 
\begin{align*}
&\boldsymbol{A}:=\mathbb{E}_{\mu_{\pi_{b}}}\left[\rho(\boldsymbol{x},a) \phi(\boldsymbol{x})\left\{\gamma \phi\left(\boldsymbol{x}'\right)-\phi(\boldsymbol{x})\right\}^{\top}\right], \quad \boldsymbol{C}:=-\mathbb{E}_{\mu_{\pi_{b}}}\left\{\rho(\boldsymbol{x},a)\phi(\boldsymbol{x}) \phi(\boldsymbol{x})^{\top}\right\},
\end{align*}
as well as $\boldsymbol{b} :=\mathbb{E}_{\mu_{\pi_{b}}}\left\{\rho(\boldsymbol{x},a)r(\boldsymbol{x},a) \phi(\boldsymbol{x})\right\}$. Here, the expectation is taken with respect to the stationary distribution of the transition tuple $(\boldsymbol{x},a,\boldsymbol{x}')$, where $\boldsymbol{x}\sim\mu_{\pi_b}, a\sim\pi_b(\cdot\mid \boldsymbol{x})$, and $\boldsymbol{x}'\sim P(\cdot\mid \boldsymbol{x},a)$. It can be shown that an optimal linear function approximation with respect to the mean-square
projected Bellman error is obtained when $\boldsymbol{w}_{\star} = -\boldsymbol{A}^{-1}\boldsymbol{b}$ \citep{sutton2009fast, xu2019two}. We make the following assumption on matrices $\boldsymbol{A}$ and $\boldsymbol{C}$ to guarantee the existence of a unique optimal linear function approximation representation of the target value function.
\begin{assumption}[Problem solvability]\label{ASSUMP:nonsingularity}
The matrices $\boldsymbol{A}$ and $\boldsymbol{C}$ are nonsingular. We denote the minimum absolute eigenvalue of the matrix $\boldsymbol{C}$ to be $\lambda_c > 0$. Furthermore, there exist $\lambda_u$ and $\lambda_w$ such that
$\lambda_{\max }(2 \boldsymbol{C}) \leq \lambda_{u}<0$ and $ \lambda_{\max }\left(2\boldsymbol{A}^{\top} \boldsymbol{C}^{-1} \boldsymbol{A}\right) \leq \lambda_{w}<0$.
\end{assumption}

\noindent 
Recall that in the implicit TDC algorithm, there are two sequences of iterates: $\boldsymbol{w}^{\text{im}}_{n}$, which parameterizes the value function of interest, and $\boldsymbol{u}^{\text{im}}_{n}$, which serves as an auxiliary variable to compute the gradient correction term for the primary iterate. To facilitate the error analysis, we introduce the tracking error vector $\boldsymbol{v}_n := \boldsymbol{u}^{\text{im}}_{n} - \boldsymbol{u}^{\star}_n$, where $\boldsymbol{u}^{\star}_n := -\boldsymbol{C}^{-1}(\boldsymbol{b} + \boldsymbol{A} \boldsymbol{w}^{\text{im}}_{n})$ denotes the stationary point of the ODE: $\boldsymbol{u'} = \boldsymbol{b} + \boldsymbol{A} \boldsymbol{w}^{\text{im}}_n + \boldsymbol{C} \boldsymbol{u}$. In short, for a fixed value of $\boldsymbol{w}^{\text{im}}_{n}$, $\boldsymbol{u}^{\star}_n$ represents the point to which the auxiliary iterates $\boldsymbol{u}^{\text{im}}_{n}$ would converge. The tracking error $\boldsymbol{v}_n$ thus quantifies the deviation of the auxiliary iterates from their instantaneous stationary point, providing a handle to assess how much of the overall error in the primary iterate $\boldsymbol{w}^{\text{im}}_{n}$ can be attributed to imperfect tracking by the auxiliary sequence.

\begin{theorem}[Finite-time analysis for implicit TDC with projection]\label{thm:tdc_decr}
Given Assumptions \ref{ASSUMP:IRR_APE_MARKOV}, \ref{ASSUMP:BDD_RWD}, \ref{ASSUMP:NORM_FEAT}, \ref{ASSUMP:importance_weight} and \ref{ASSUMP:nonsingularity}, suppose  $\alpha_n = \frac{\alpha_1}{n^\sigma},~\beta_n  = \frac{\beta_1}{n^\nu},$
with \(0<\nu<\sigma<1\) and $\alpha_1, \beta_1 > 0$. Then for any $\epsilon\in(0,\sigma-\nu]$, $\epsilon'\in (0,0.5]$, the projected implicit TDC with $R_{\boldsymbol{w}} \ge \|\boldsymbol{w}_{\star}\|$, $R_{\boldsymbol{u}} \ge 2\|\boldsymbol{C}^{-1}\| \|\boldsymbol{A}\|R_{\boldsymbol{w}}$ and 
$$
\eta_w = \frac{\left|\lambda_{w}\right|}{1 + \mathbb{I}\{\rho_{\max}\phi_{\max}^2 > |\lambda_w|\}(\rho_{\max}\phi_{\max}^2 - |\lambda_w|)\alpha_1}
$$
yields
\begin{align*}
\mathbb{E}_{\mu_{\pi_b}}\bigl\|\boldsymbol{w}^{\text{im}}_n - \boldsymbol{w}_{\star}\bigr\|^2
&\lesssim
\exp\left[-\eta_w \alpha_1\left\{(1+n)^{1-\sigma}-1\right\}\right]\left\|\boldsymbol{w}_{\star}-\boldsymbol{w}^{\text{im}}_1\right\|^2 \\&\quad + \mathcal{O}\left(\frac{\log n}{n^{\sigma}}\right) + \mathcal{O}\left(\frac{\log n}{n^{\nu}}+h(\sigma, \nu)\right)^{1-\epsilon'}
\\
\mathbb{E}_{\mu_{\pi_b}}\bigl\|\boldsymbol{v}_n\bigr\|^2
& = \mathcal{O}\left(\frac{\log n}{n^{\nu}}\right) + \mathcal{O}\left(h(\sigma, \nu)\right), \quad 
h(\sigma,\nu)
=
\begin{cases}
\frac{1}{n^{\nu}}, & \sigma>1.5\,\nu,\\
\frac{1}{n^{2(\sigma-\nu)-\epsilon}}, & \nu<\sigma\le1.5\,\nu
\end{cases}
\end{align*}
for all $n \in \mathbb{N}.$
\end{theorem}

\begin{remark}
The implication of Theorem~\ref{thm:tdc_decr} is that implicit TDC offers greater flexibility than standard TDC in choosing a step size schedule while retaining the asymptotic convergence rate established in \citep{xu2019two}. In particular, the bound holds for any $\alpha_1 > 0$ and $\beta_1 > 0$. In comparison, existing bounds often require knowledge of the environment and its stationary distribution \citep{xu2019two, gupta2019finite, kaledin2020finite, doan2021finite, haque2023tight, chandak2025finite}, which greatly restricts the applicability of the theory in practice. For instance, previous results established by \citet{xu2019two}, imposed restrictive conditions on initial step sizes, i.e., $\alpha_1 < 1/|\lambda_w|$ and $\beta_1 < 1/|\lambda_u|$. Unlike such bounds, the above bounds hold regardless of the initial step size one uses, and to our knowledge, they are the first all time MSE bounds for TDC established under a data-adaptive step size that does not require knowledge of the environment.
\end{remark}

\begin{remark}
Analogous to implicit TD, finite-time MSE bounds for implicit TDC without a projection step can be established either using step sizes that depend on quantities associated with the stationary distribution of the underlying Markov chain \citep{kaledin2020finite, doan2021finite, haque2023tight}, or by deriving bounds that hold only for sufficiently large time indices \citep{gupta2019finite, doan2021finite}, an approach that overlooks the numerical instability that TDC can exhibit in its early phase. Furthermore, we do not claim that our established convergence rate is the tightest possible; it may potentially be improved to match the asymptotic convergence rate of TDC established in \citep{kaledin2020finite, haque2023tight}, and we leave this as a direction for future work.
\end{remark}

\begin{remark}
The implicit construction developed for TDC may also be extended to other gradient-based temporal difference methods, such as GTD and GTD2. \citep{sutton2008convergent,sutton2009fast}. In particular, TDC and GTD2 are closely related two-timescale methods that introduce an auxiliary parameter to obtain gradient-based updates for minimizing the mean squared projected Bellman error (MSPBE), but they differ in the form of the primary-parameter update. Thus, although TDC and GTD2 target the same MSPBE objective, their distinct stochastic update structures naturally lead to different implicit counterparts. In contrast, GTD is based on the norm of the expected update (NEU) objective and may therefore give rise to a fundamentally different implicit formulation. Developing implicit versions of other off-policy evaluation algorithms, and investigating whether they inherit the improved step size robustness of implicit TDC, is an interesting direction for future work.    
\end{remark}

\section{Numerical experiments}
\label{SEC:NUMERICS}
\subsection{Random walk with absorbing states: sensitivity to step size and feature}\label{SUBSEC:RW_ABS}
\begin{figure}[!h]
    \centering

    \includegraphics[height=0.25\textwidth]{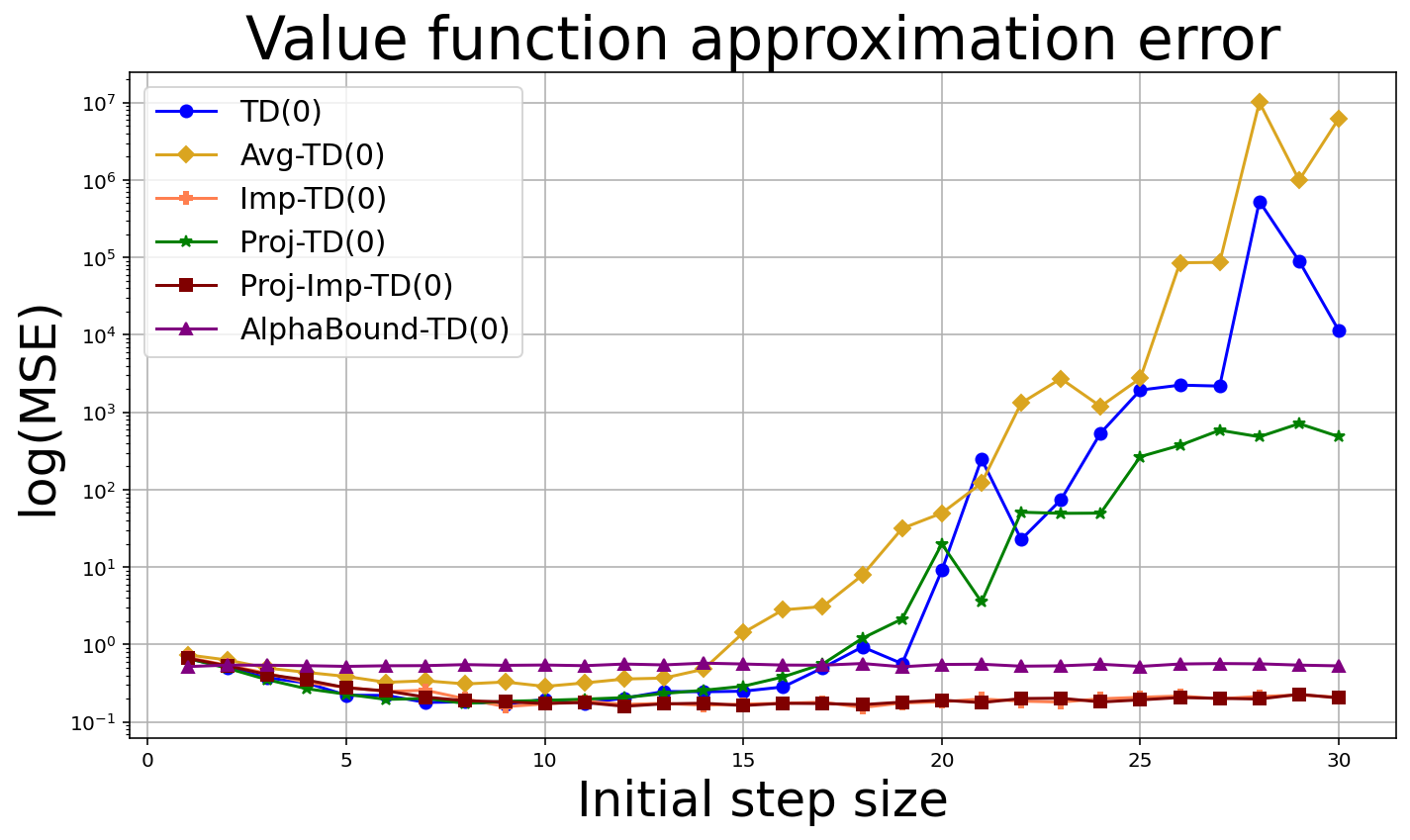} \includegraphics[height=0.25\textwidth]{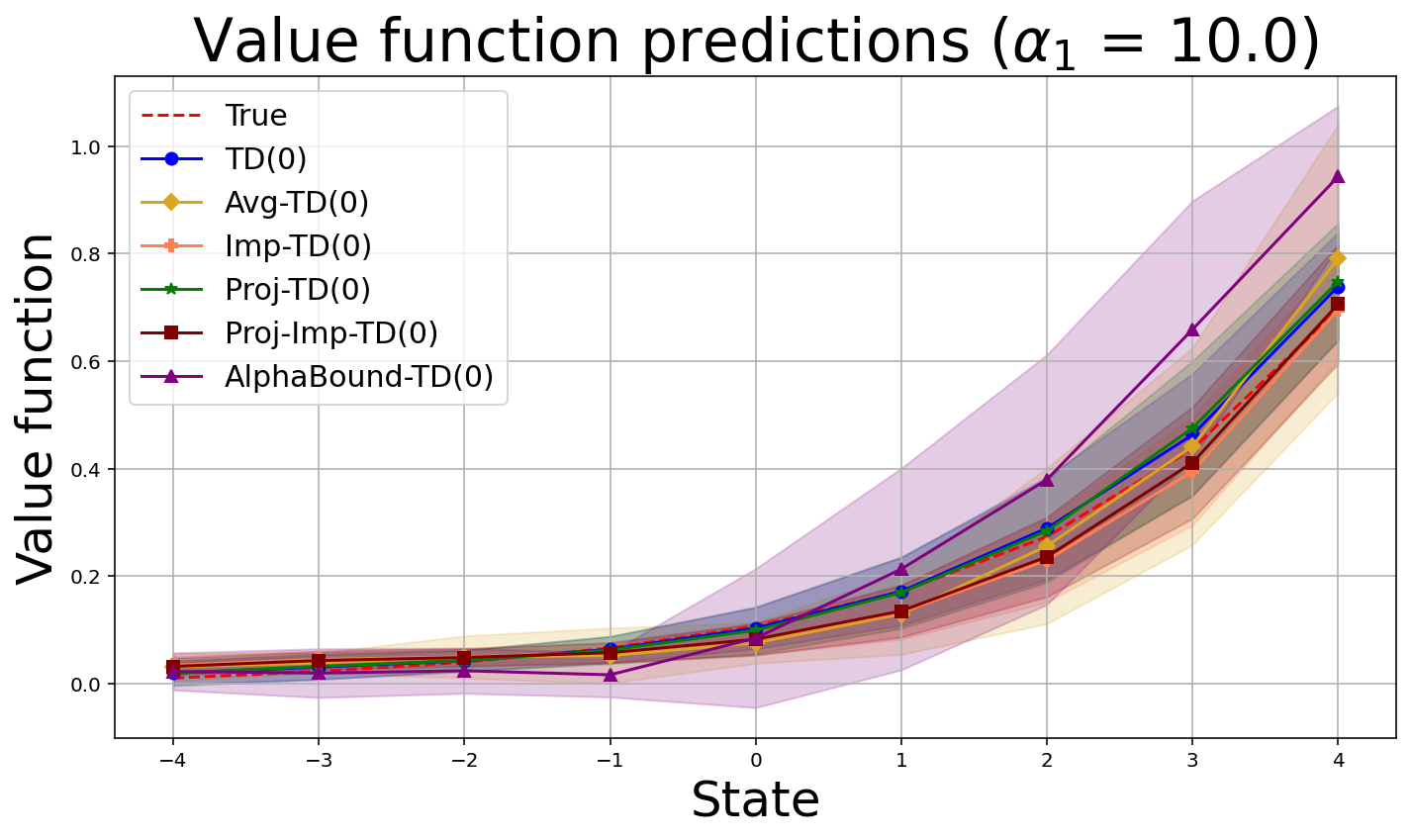}

    \includegraphics[height=0.25\textwidth]{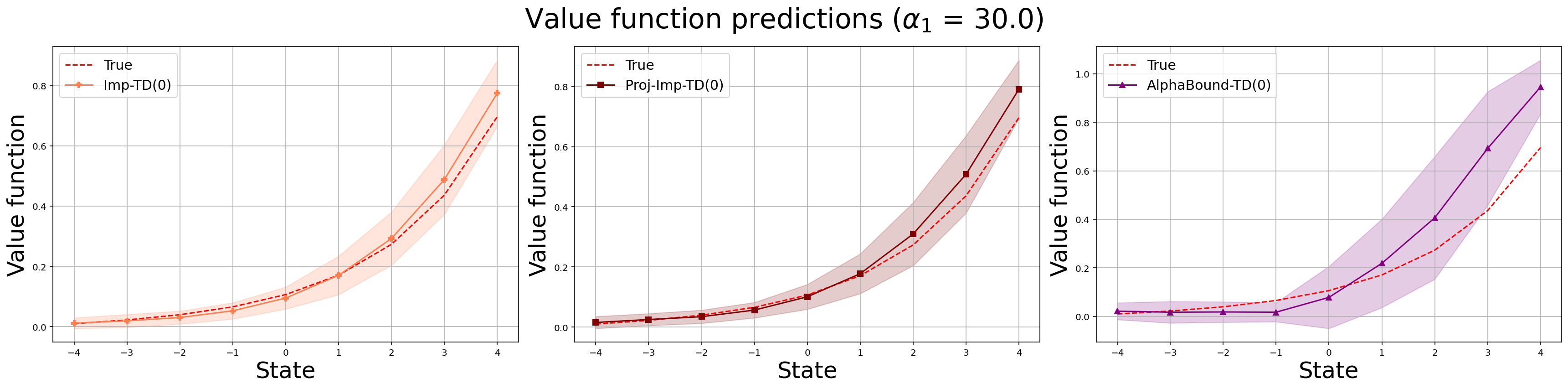}
    \caption{Value function approximation in the random walk environment. \textbf{Top left:} Value function approximation error (log-MSE) versus initial step size over the interval $[10.0, 30.0]$. While the log-MSE for standard/averaged/projected TD(0) grow sharply as the initial step size increases, implicit TD(0) and projected implicit TD(0) remain nearly flat across the entire range, reflecting their robustness to large step sizes. AlphaBound-TD(0) \citep{dabney2012adaptive} is also flat with a higher log-MSE by construction, since it does not depend on an initial step size parameter; \textbf{Top right:} Value function approximation with $\alpha_1 = 10.0$. All methods approximate the true value function with tight standard deviation bands, while AlphaBound-TD(0) is noticeably inferior, with a wider standard deviation and a less accurate fit. \textbf{Bottom left:} Value function approximation with $\alpha_1 = 30.0$ using implicit TD(0). Implicit TD(0) continues to track the true value function reasonably well, with a moderate standard deviation band. \textbf{Bottom middle:} Value function approximation with $\alpha_1 = 30.0$ using projected implicit TD(0). Projected implicit TD(0) also closely tracks the true value function with a tight standard deviation. \textbf{Bottom right:} Value function approximation using AlphaBound-TD(0). AlphaBound-TD(0) tracks the true value function less accurately than implicit TD(0) or projected implicit TD(0), with a wider standard deviation.}\label{fig:VALUE_FUNC_TD0}
\end{figure}

\noindent In this subsection, we consider a one-dimensional random walk environment with 11 integer-valued states arranged on a real line, with zero at the center. The two endpoints, the leftmost and rightmost states, are absorbing and thus omitted from the value function approximation. The reward is zero for all states except for the rightmost state, where the reward is one. A total number of 100 independent experiments were run with a discount factor $\gamma = 0.9$ and a projection radius $R = 10^6$. We employ TD(0), averaged TD(0), implicit TD(0), projected TD(0), and projected implicit TD(0), and show their average performance, along with one standard deviation bands shown as shaded regions, in Figure \ref{fig:VALUE_FUNC_TD0}. We consider a range of initial step sizes between 10.0 and 30.0, with a step size sequence of the form $\alpha_n = \alpha_1/n^{0.7}$.

Based on the top left plot in Figure \ref{fig:VALUE_FUNC_TD0}, we observe that as the initial step size increases, the average value approximation error (measured in log-MSE) increases sharply for standard TD(0), averaged TD(0), and projected TD(0), while implicit TD(0) and projected implicit TD(0) show little to no increase, remaining nearly flat across the entire range. We also compare against the adaptive step size proposed by \citet{dabney2012adaptive}, denoted AlphaBound TD(0). With an initial step size $\alpha_1 = 10.0$, all methods provided accurate value function approximation with tight standard deviation bands, with the exception of AlphaBound-TD(0), which is noticeably inferior, exhibiting a wider standard deviation and a less accurate fit, as shown in the top right plot in Figure \ref{fig:VALUE_FUNC_TD0}. For a large initial step size ($\alpha_1 = 30.0$), however, standard TD(0), averaged TD(0), and projected TD(0) all suffered from numerical instability, yielding poor value function approximation results, as shown in Figure \ref{fig:VALUE_FUNC_TD0_INSTAB}. Unlike these methods, the implicit procedures remained numerically stable even under this large initial step size with lower log-MSE values: implicit TD(0) alone is shown in the bottom left plot of Figure \ref{fig:VALUE_FUNC_TD0}, and projected implicit TD(0) alone is shown in the bottom middle plot of the same figure.

Since AlphaBound TD(0) requires no initial step size parameter, its log-MSE curve is flat, and value approximations are identical across the top right and bottom right plots, regardless of the nominal $\alpha_1$ value used elsewhere. This stability comes at a cost, however, as it consistently yields inferior value function predictions compared to the other methods, including implicit TD(0) and projected implicit TD(0), which achieve the same robustness to step size while still approximating the true value function accurately. We also employed standard and implicit TD(0.5) algorithms and observed qualitatively identical results; see Section \ref{APPEND_SEC_LAMBDA} in the appendix for further details. Together, these results confirm our theoretical findings in Theorems \ref{THM:FIN_PROJ_TD0} and \ref{THM:FIN_PROJ_TDL}.

 \begin{figure}[!h]
 \centering
    \includegraphics[height=.295\textwidth]{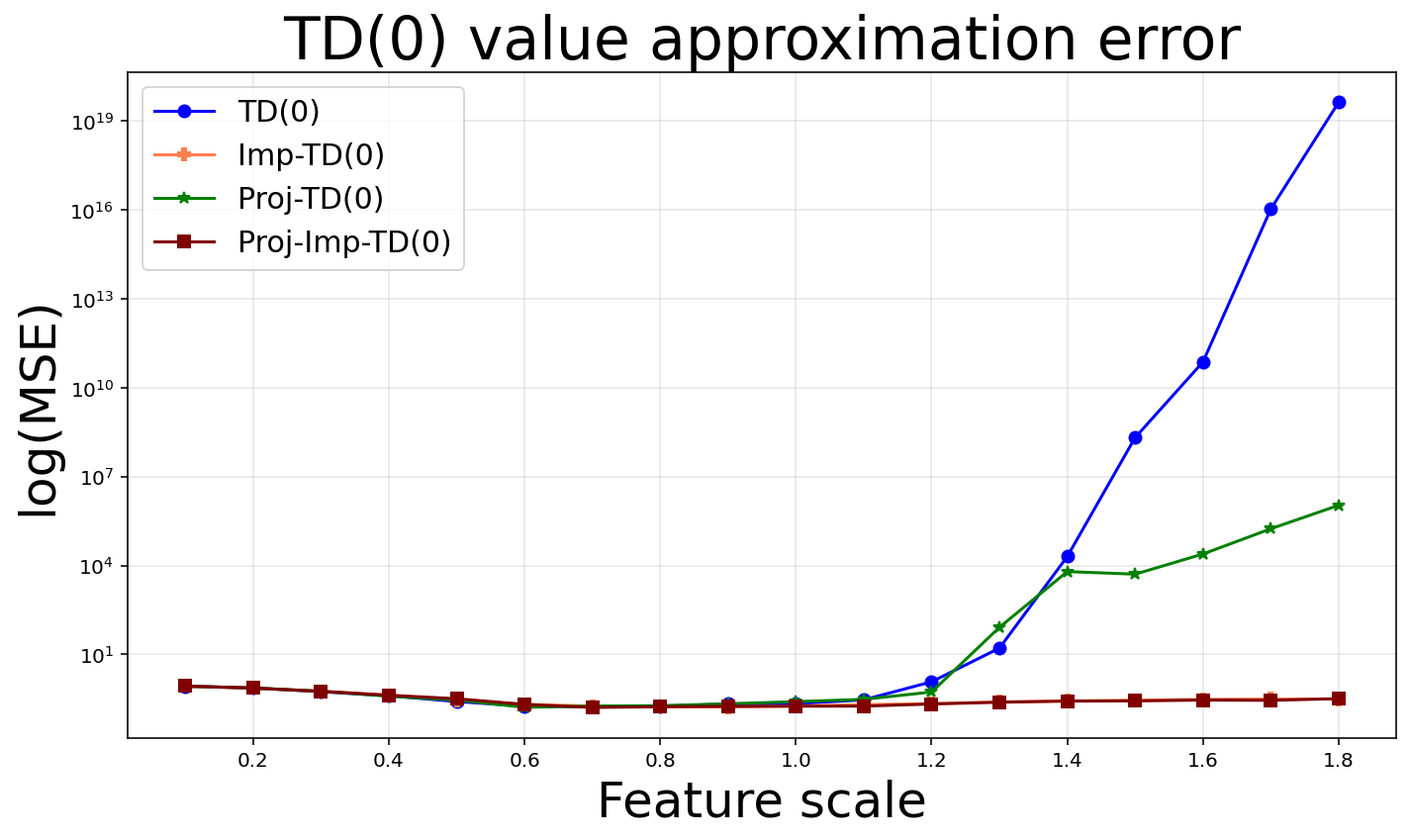}
    \includegraphics[height=.295\textwidth]{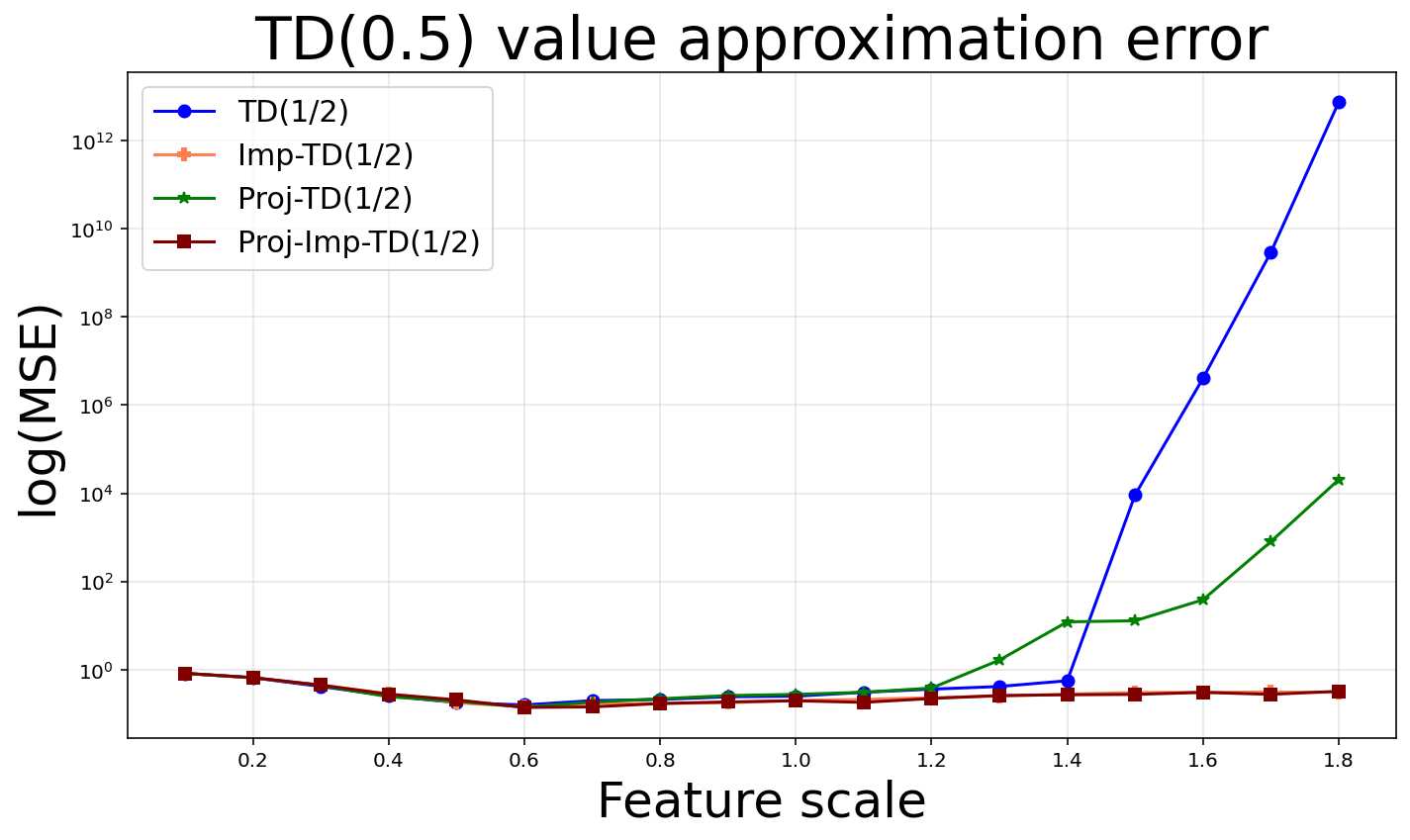}
    \caption{Value function approximation error versus feature scale in the 11-state random walk environment considered in Subsection \ref{SUBSEC:RW_ABS}. In both panels, standard TD and projected TD exhibit rapidly increasing approximation error as the feature scale grows, while implicit TD and projected implicit TD remain stable across the entire range.}
    \label{fig:FEATURE_NORM}
\end{figure}

We next examine the sensitivity of these algorithms to the norm of the feature vectors, with the step size fixed at $\alpha_n = 1/\sqrt{n}$. Figure \ref{fig:FEATURE_NORM} shows the value approximation error as the feature norm scale increases from 0.1 to 1.8, for both TD(0) (left) and TD(0.5) (right). For features with small norm values, all four algorithms perform comparably and achieve low approximation error. However, as the feature scale grows beyond approximately 1.2, standard TD and projected TD exhibit a sharp increase in error, with the log-MSE for standard TD growing especially fast, exceeding $10^{19}$ for TD(0) and $10^{12}$ for TD(0.5) at a feature scale of 1.8. Projected TD grows more moderately but still diverges substantially from its performance at smaller feature scales. In contrast, implicit TD and projected implicit TD remain remarkably stable throughout the entire range, showing little to no increase in approximation error even at the largest feature scale considered. This suggests that the benefits of implicit updates are not limited to robustness against large step sizes: they also provide robustness to the scale of the features, a related but distinct source of instability in temporal difference learning. Larger feature norms amplify the update magnitude in each TD step, similar to how a larger step size does. Because implicit updates are data-adaptive, effectively scaling the step size according to the squared feature norm, they mitigate the instability induced by feature scaling as well.

\subsection{Reward process with 100-dimensional states}
\noindent Motivated by \citet{zhang2021finite}, we construct a synthetic Markov reward process with 100 states whose transition probability matrix is generated at random. For each state, we sample 99 independent $\mathrm{Uniform}(0,1)$ samples, sort them, and take successive differences—treating 0 and 1 as boundary points—to form a valid transition probability distribution. Repeating this procedure for all 100 states and stacking the resulting distributions row-wise yields the full transition probability matrix $\boldsymbol{P}$. Rewards are assigned by drawing one $\mathrm{Uniform}(0,1)$ sample per state and collecting these into the reward vector $\boldsymbol{r}$. We set the discount factor to $\gamma = 0.9$.

\begin{figure}[!h]
 \centering
    \includegraphics[height=.295\textwidth]{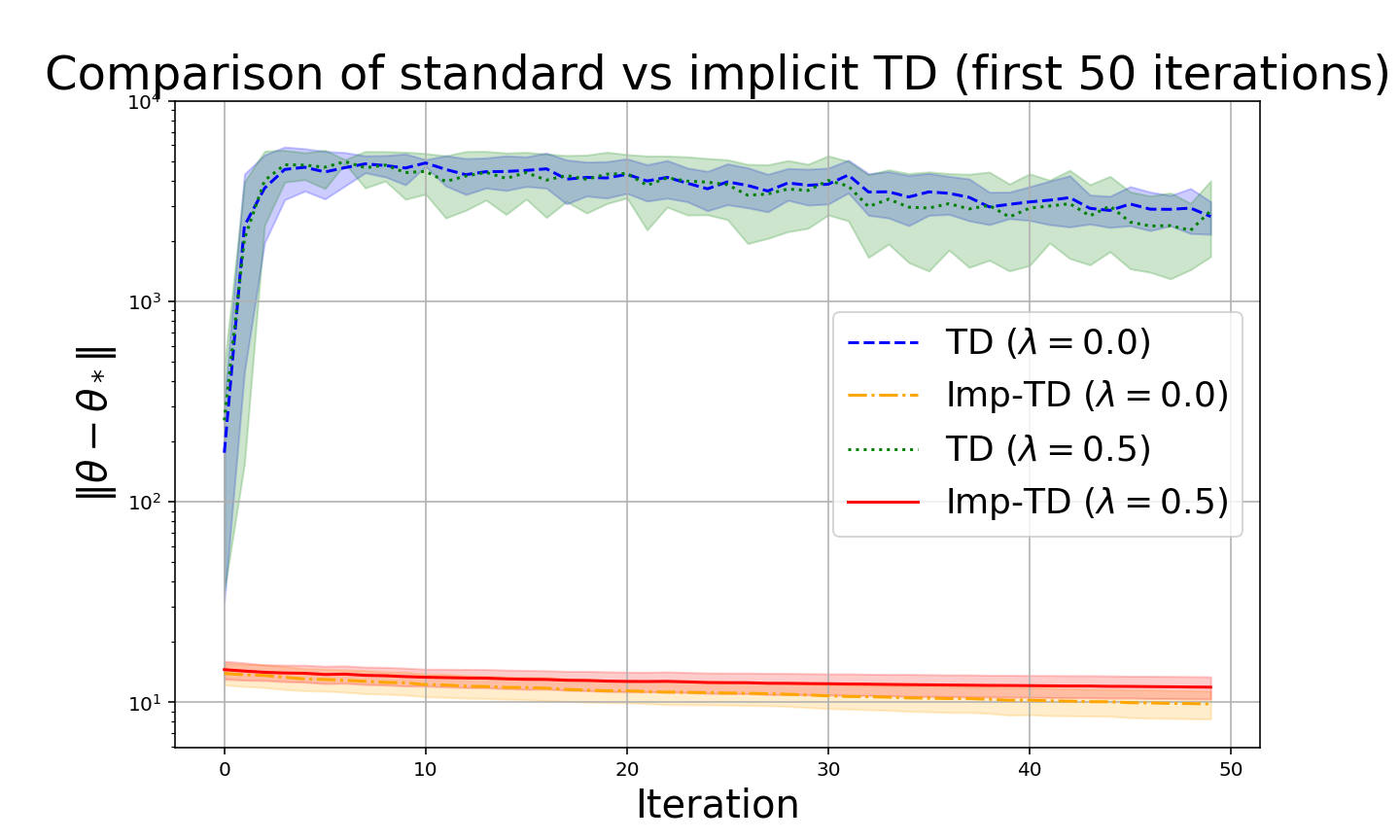}
    \includegraphics[height=.295\textwidth]{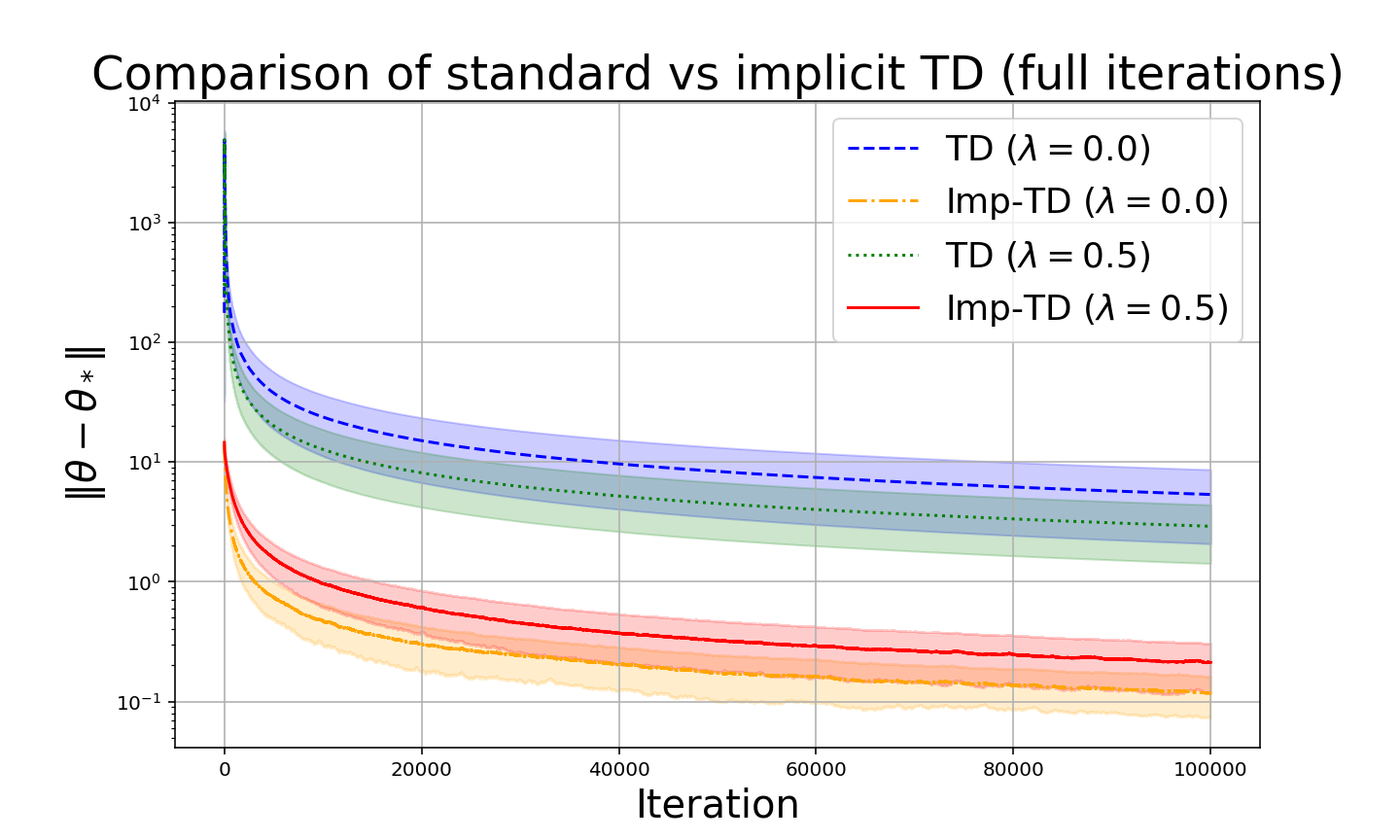}
    \caption{Parameter estimation error in a synthetic 100-state Markov reward process, comparing standard TD and implicit TD for $\lambda = 0$ and $\lambda = 0.5$. Step size was set to $\alpha_n = 300/n$. \textbf{Left:} Over the first 50 iterations, standard TD exhibits pronounced error amplification before slowly decaying, whereas implicit TD yields an immediate error reduction. \textbf{Right:} Over $10^5$ iterations, implicit TD consistently converges toward the optimal parameter with superior accuracy, whereas standard TD remains hindered by its initial error amplification.}
    \label{fig:MDP100}
\end{figure}

\begin{figure}[h!]
  \centering
  \begin{subfigure}[c]{0.425\textwidth}
    \centering
    \resizebox{\textwidth}{0.08\textheight}{%
      \begin{tabular}{@{}l c c@{}}
        \toprule
        \textbf{Method}    & $\boldsymbol{\lambda}$ & \textbf{Mean $\pm$ Std} \\
        \midrule
        Standard TD        & 0.0    & 5.356 $\pm$ 3.279   \\
        Implicit TD        & 0.0    & 0.117 $\pm$ 0.044   \\
        Standard TD        & 0.5    & 2.906 $\pm$ 1.484   \\
        Implicit TD        & 0.5    & 0.212 $\pm$ 0.094   \\
        \bottomrule
      \end{tabular}%
    }
    \caption{Final average parameter estimation error}
    \label{tab:MRP}
  \end{subfigure}\hfill
    \begin{subfigure}[c]{0.48\textwidth}
    \centering
    \includegraphics[height=0.65\textwidth]{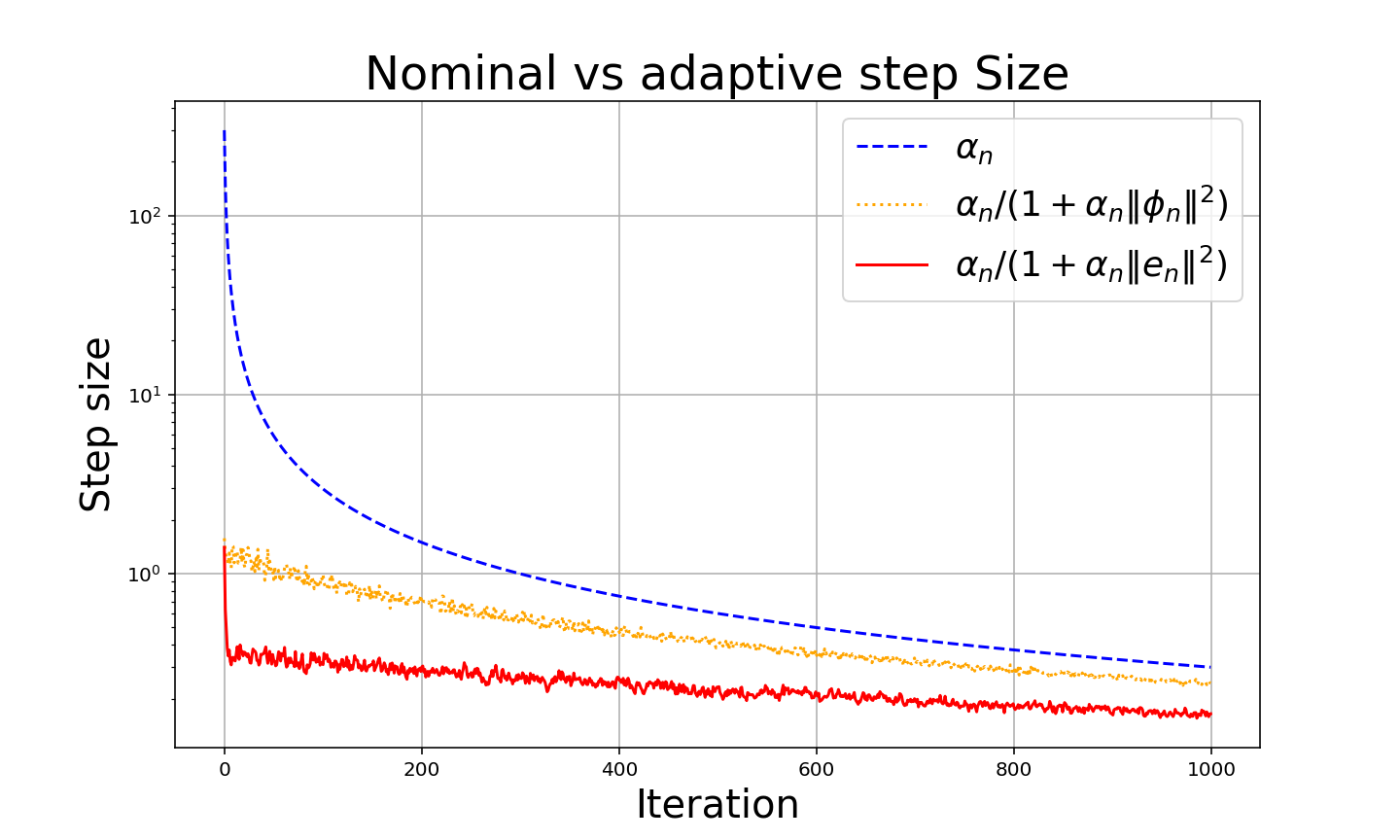}
    \caption{Nominal step size vs effective step size}
    \label{fig:step_size}
  \end{subfigure}
  \caption{\textbf{Left:} Average final parameter estimation errors and standard deviation for standard and implicit TD algorithms in a synthetic 100-state Markov reward process. Implicit TD yields substantially improved parameter estimation error when used with a large initial step size. \textbf{Right:} Nominal step size $\alpha_n = 300/n$ versus effective step size trajectories $\alpha_n/(1+\alpha_n \|\boldsymbol{\phi}_{n}
\|)^2$ for TD(0) and $\alpha_n/(1+\alpha_n \|\boldsymbol{e}_{n}\|)^2$ for TD(0.5). Although the effective step sizes in the implicit algorithms exhibit non-monotonicity, they eventually converge to zero.}
\end{figure}

In this example, the true value function can be analytically computed and is given by $\boldsymbol{V}_*=(\mathbf{I}-\gamma \boldsymbol{P})^{-1}\boldsymbol{r}$. Our job is to approximate the true value function via $\boldsymbol{\Phi} \boldsymbol{w}$, where each row of $\boldsymbol{\Phi}\in\mathbb{R}^{100\times20}$ represents a normalized random binary feature. The oracle parameter $\boldsymbol{w}_{\star}$ was obtained by solving $\min_{\boldsymbol{w}} \|\Phi \boldsymbol{w} - \boldsymbol{V}_*\|$. Both standard and implicit TD algorithms were run for $10^5$ iterations with $\lambda\in\{0,0.5\}$ under the decaying step size schedule $\alpha_n = 300/n.$ We set a projection radius $R = 5000$ and conducted 20 independent experiments. Figure \ref{fig:MDP100} depicts the mean estimation error, with shaded bands indicating one standard deviation. Figure \ref{fig:MDP100} and Figure \ref{tab:MRP} present parameter estimation results for standard versus implicit TD(0) and TD($\lambda$) algorithms.

In Figure~\ref{tab:MRP}, we see that the 
mean estimation error for standard TD(0) is $5.356$ (std $3.279$), while 
for implicit TD(0) it is $0.117$ (std $0.044$), a reduction of roughly 98\%. Figure~\ref{fig:MDP100} (left) further shows that, within the first 50 iterations, the standard TD(0) trajectory deviates from the true parameter, whereas the implicit TD(0) algorithm immediately reduces the estimation error. After $10^5$ iterations (Figure~\ref{fig:MDP100}, right), standard TD(0) plateaus at a high error, but implicit TD(0) has already reached near-zero error. This comparison extends to TD(0.5) variants as well. From Figure~\ref{tab:MRP}, we see that standard TD(0.5) achieves mean error $2.906$ (std $1.484$), while implicit TD(0.5) attains mean $0.212$ (std $0.094$). Although standard TD(0.5) roughly halved the estimation error relative to TD(0), the implicit TD(0.5) algorithm nonetheless outperformed the standard method by an order of magnitude and exhibited smaller variance across independent runs. Implicit TD methods consistently improved numerical stability, allowing the use of large step sizes for fast early learning, and produced both lower bias and lower variance in the final parameter estimates.

In Figure \ref{fig:step_size}, 
we provide a plot of decreasing step size $\alpha_n = 300/n$ versus effective step sizes for implicit TD(0): $\alpha_n/(1+\alpha_n\|\boldsymbol{\phi}_{n}\|^2)$ and implicit TD(0.5): $\alpha_n/(1+\alpha_n\|\boldsymbol{e}_{n}\|^2)$. 
The figure shows that all three step size schedules decrease to zero. In the meantime, the effective step sizes for the implicit algorithms are not necessarily monotonic, as they depend on the random quantities $\boldsymbol{\phi}_{n}$ and $\boldsymbol{e}_{n}$. Such an adaptive step size prevents numerical instability by appropriately scaling down drastic temporal difference updates. 

\subsection{Policy evaluation for continuous domain control}
\begin{figure}[!h]
\centering
\includegraphics[height=.3125\textwidth]{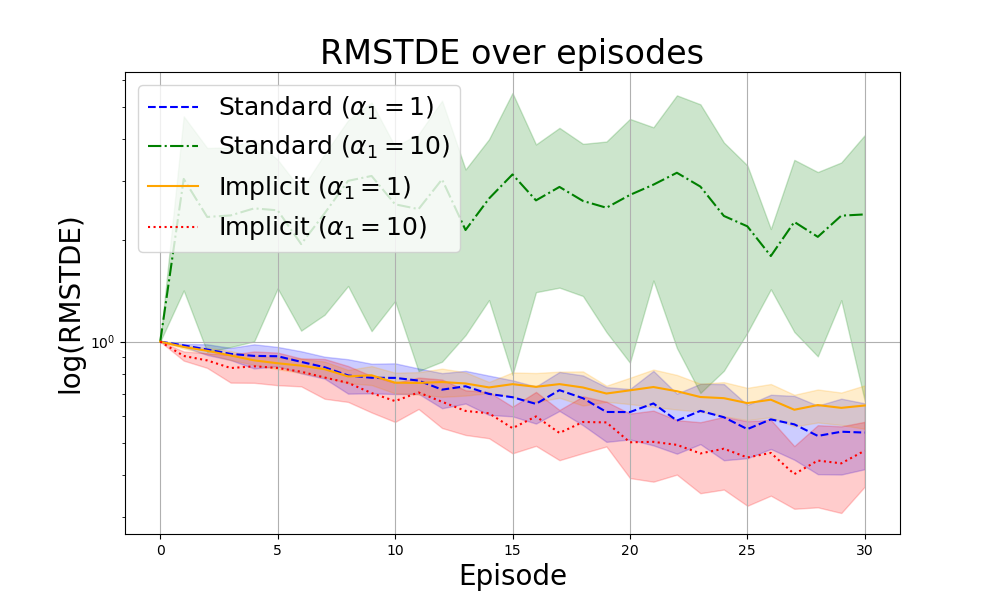}
\includegraphics[height=.295\textwidth]{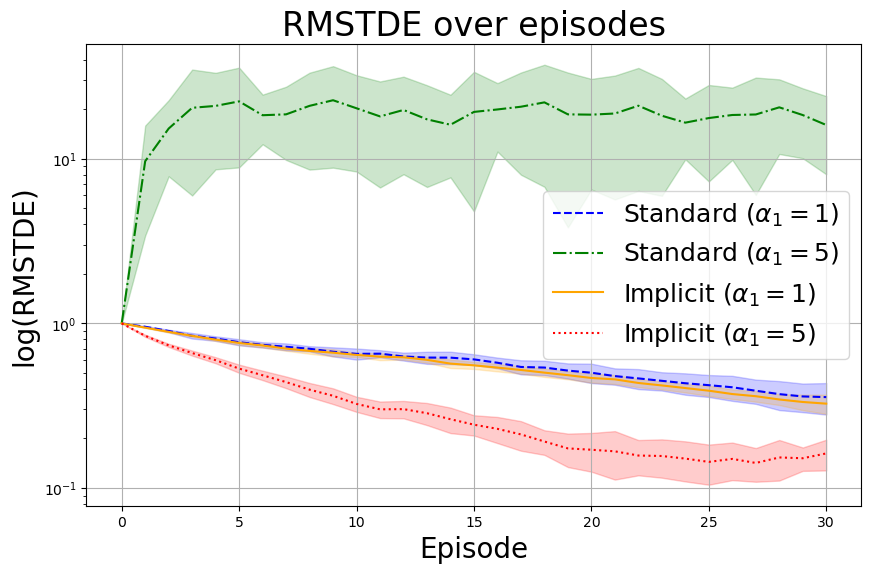}
\caption{Average root mean squared temporal difference error (RMSTDE) versus episode index for standard and implicit TD(0). Shaded bands indicate pointwise one standard deviation over 20 independent runs. \textbf{Left:}  In the Acrobot environment with a large initial step size, implicit TD(0) delivers accelerated RMSTDE decay and reduced variance, whereas standard TD(0) exhibits poor convergence and amplified variance. \textbf{Right:} In the Mountain Car environment, implicit TD(0) showcases rapid RMSTDE reduction under a large initial step size, reflecting superior numerical stability relative to standard TD(0).}
\label{fig:control}
\end{figure}

\noindent In this subsection, we test the robustness of implicit updates in classical control tasks. We considered both Acrobot and Mountain Car environments available through \texttt{Gymnasium} library in Python \citep{towers2024gymnasium}. The Acrobot environment consists of a two link pendulum system with two joints, where only the joint between the two links is actuated. The episode begins with both links hanging downward, and the objective is to swing the end of the lower link upward to reach a specified target height. The agent receives a reward of $-1$ at each time step until the goal is achieved, which ends the episode with a reward of $0$. In the Mountain Car environment, a car is positioned between two hills, where the goal is to reach the top of the right hill. The car’s engine is underpowered, so the agent must build momentum by driving back and forth. The state includes position and velocity; actions apply force left, right, or none. Each step incurs a reward of $-1$, and the episode ends upon reaching the goal.\\
\indent We applied State-Action-Reward-State-Action algorithm \citep{rummery1994line}, based on both standard TD(0) and implicit TD(0) updates, to approximate the state-action value function in the Acrobot and Mountain Car environments. In each case, the state-action value function was approximated by radial basis features \(\boldsymbol{\phi}_{n}\in\mathbb{R}^{100}\), and we measured performance by the empirical root mean squared temporal difference error (RMSTDE) computed over 1000 input values. We used a decaying step size schedule
$\alpha_n = \alpha_1/n,~\alpha_1\in\{1.0,\,10.0\} ~\text{for Acrobot}$ and $\alpha_1\in\{1.0,\,5.0\} ~\text{for Mountain Car}$. A total of 20 independent experiments were conducted. Figure \ref{fig:control} presents the mean RMSTDE for both environments, with shaded regions showing one standard deviation bands.
\\
\indent The results for the Acrobot environment are shown in Figure \ref{fig:control} (left) and Table \ref{tab:rmstde_side_by_side} (left). 
With an initial step size of $\alpha_1=1.0$, standard TD(0) attained a mean RMSTDE of 0.546 (std 0.167), whereas implicit TD(0) yielded a higher mean RMSTDE of 0.655 but with markedly lower variability (std 0.062).
When \(\alpha_1\) was increased to 10.0, standard TD(0) performed much worse than implicit TD(0), achieving a mean RMSTDE of 2.585 (std 2.308) against a mean RMSTDE of 0.428 (std 0.099). This demonstrates that the implicit procedure remains stable and even benefits from larger step sizes, while the standard TD procedure suffers from a large initial step size and greater run-to-run variance. \\
\indent The results for the Mountain Car environment are shown in Figure \ref{fig:control} (right) and Table \ref{tab:rmstde_side_by_side} (right). 
In this environment, the advantage of implicit updates under an aggressive step size is more evident. With \(\alpha_1=1.0\), both methods performed similarly (standard: mean RMSTDE of \(0.379\) with std\ \(0.088\); implicit: mean RMSTDE of \(0.324\) with std\ \(0.043\)).  But with \(\alpha_1=5.0\), the standard update drastically deteriorated (mean RMSTDE of \(19.827\) with std \ \(10.395\)), whereas implicit version obtained an improved error (mean RMSTDE of \(0.162\) with std \ \(0.042\)). These results demonstrate that implicit algorithms retain the ease of implementation of standard methods while substantially enhancing numerical stability even in continuous domain control problems.  

\begin{table}[h!]
  \centering
  \vspace{1em}
  \begin{minipage}{0.48\textwidth}\label{tab:rmsbe_acrobot}
    \centering
    \begin{tabular}{@{}l c c@{}}
      \specialrule{1pt}{0pt}{4pt}
      \multicolumn{3}{c}{\textbf{Acrobot (RMSTDE)}} \\
      \midrule
      \textbf{Method}    & $\alpha_1$ & \textbf{Mean $\pm$ Std} \\
      \midrule
      Standard & 1.0 & 0.546 $\pm$ 0.167 \\
      Standard & 10.0 & 2.585 $\pm$ 2.308 \\
      Implicit & 1.0 & 0.655 $\pm$ 0.062 \\
      Implicit & 10.0 & 0.428 $\pm$ 0.099 \\
      \bottomrule
    \end{tabular}
  \end{minipage}\hfill
  \begin{minipage}{0.48\textwidth}\label{tab:rmsbe_mountain}
    \centering
    \begin{tabular}{@{}l c c@{}}
      \specialrule{1pt}{0pt}{4pt}
      \multicolumn{3}{c}{\textbf{Mountain Car (RMSTDE)}} \\
      \midrule
      \textbf{Method} & $\alpha_1$ & \textbf{Mean $\pm$ Std} \\
      \midrule
      Standard & 1.0 & 0.379 $\pm$ 0.088 \\
      Standard & 5.0 & 19.827 $\pm$ 10.395 \\
      Implicit & 1.0 & 0.324 $\pm$ 0.043 \\
      Implicit & 5.0 & 0.162 $\pm$ 0.042 \\
      \bottomrule
    \end{tabular}
  \end{minipage}
  \caption{Final root mean squared temporal difference error (RMSTDE) for standard and implicit TD(0) on the Acrobot (left) and Mountain Car (right) environments. In the Acrobot environment, implicit TD(0) matches standard TD(0) at $\alpha_1=1$, while at $\alpha_1=10$ it substantially reduces both error magnitude and variance compared to standard TD(0). In the Mountain Car environment, implicit TD(0) offers modest improvement at $\alpha_1=1$ and, at $\alpha_1=5$, prevents the severe RMSTDE explosion exhibited by standard TD(0), thereby demonstrating superior numerical stability.
}\label{tab:rmstde_side_by_side}
\end{table}

\subsection{Baird's counterexample}
\noindent In this last subsection, we consider a celebrated off-policy evaluation problem, known as Baird's counterexample. This is a classical benchmark problem in reinforcement learning, specifically constructed to expose instability and convergence issues in off-policy TD algorithms when combined with linear function approximation. Originally introduced by \citet{baird1995residual}, this example is notable for its simplicity yet significant theoretical and practical implications. The environment consists of seven states, with one center state connected directly to six peripheral states. The behavior policy used in Baird’s example uniformly selects one of the six peripheral states with equal probability (1/6), while the target policy deterministically transitions to the center state, creating a distinct discrepancy between the two policies. This deliberate mismatch poses substantial difficulties for algorithms relying on off-policy updates. Linear function approximation is employed in this counterexample, characterized by eight distinct features designed to create an inherently challenging setting for standard TD methods. Every peripheral state has its own unique feature, and there is one extra feature that remains active in every state, including the center. This particular choice of feature representation leads to nontrivial correlation among the features, further complicating the convergence and stability of existing TD-based algorithms.\

\begin{figure}[!h]
\centering
\includegraphics[height=.2925\textwidth]{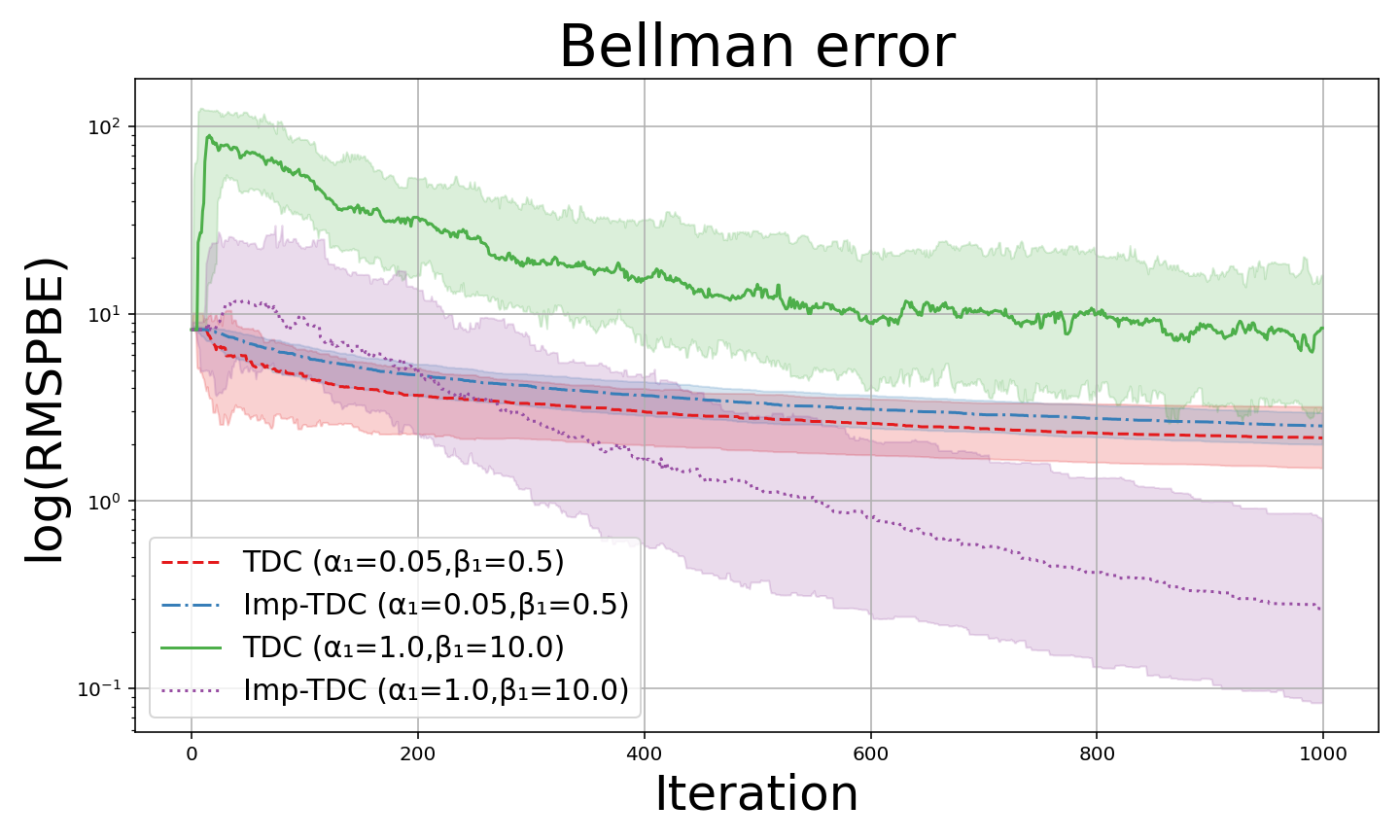}.
\includegraphics[height=.2925\textwidth]{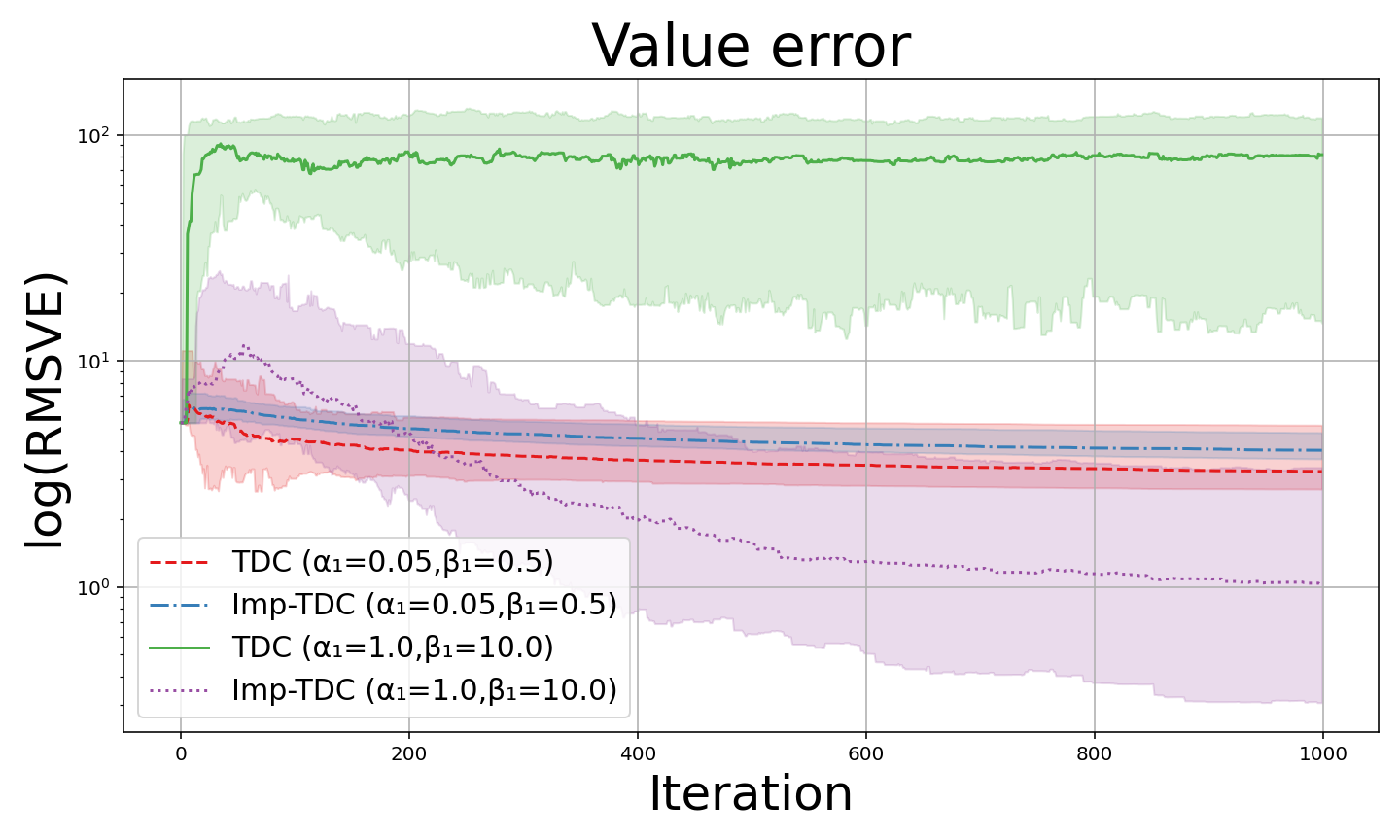}
\caption{Projected Bellman and value error of the standard TDC and implicit TDC on Baird’s counterexample. \textbf{Left (Bellman error):} Implicit TDC significantly reduces projected Bellman error with a large initial step size, reaching far lower errors than standard TDC. \textbf{Right (value error):} Implicit TDC maintains consistent value error reduction, while standard TDC exhibits elevated error under a large initial step size.}
\label{fig:Baird}
\end{figure}

We performed 100 independent experiments and report the mean along with the 10th and 90th percentiles of the outcomes. The results depicted in Figure \ref{fig:Baird} and Table \ref{tab:combined_error} demonstrate the critical role that step size selection plays in the performance of TDC and implicit TDC on Baird's counterexample. Under decreasing step sizes, where $\alpha_n = \alpha_1/n^{0.8}$ and $\beta_n = \beta_1/n^{0.6}$, the same pattern persists: implicit TDC remains numerically robust while standard TDC does not. When the initial step sizes are large (e.g., $\alpha_1 = 1.0, \beta_1 = 10.0$), standard TDC suffers severe numerical instability, with error amplifying to an RMSPBE of $8.390$ (10th/90th percentile: $3.123$, $16.069$) and an RMSVE of $81.682$ (10th/90th percentile: $14.668$, $119.105$). In stark contrast, implicit TDC achieves substantially lower error under the same large initial step sizes, with an RMSPBE of $0.256$ (10th/90th percentile: $0.084$, $0.763$) and an RMSVE of $1.001$ (10th/90th percentile: $0.307$, $3.371$). Crucially, implicit updates remain stable even under large initial step sizes, avoiding the numerical instability seen in standard TDC. Furthermore, Figure \ref{fig:Baird} illustrates the clear advantage of a large initial step size, which facilitates more efficient learning within the first 1000 iterations.

\begin{figure}[!h]
    \centering
    \includegraphics[height=.295\textwidth]{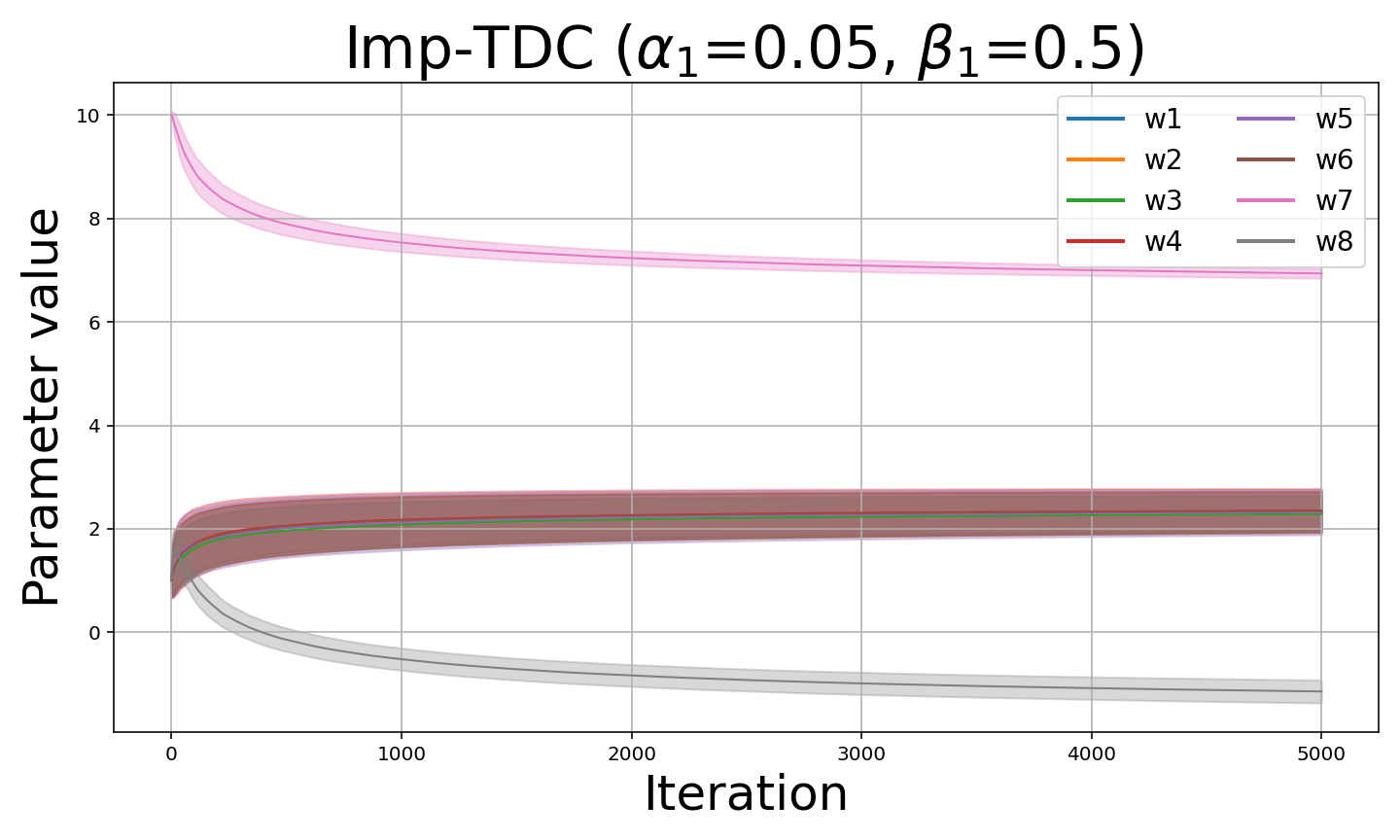}
    \includegraphics[height=.295\textwidth]{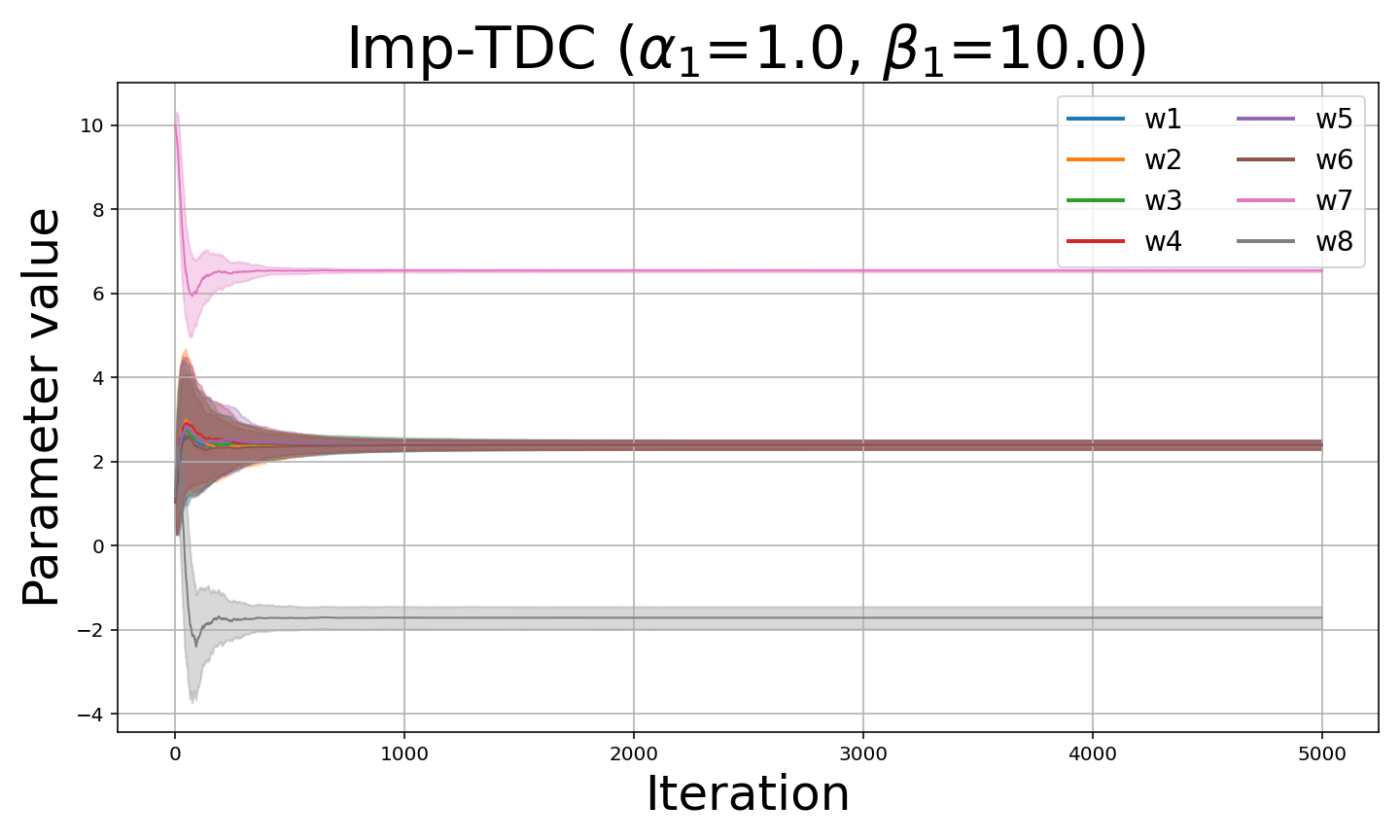}
    \caption{Trajectories of the estimated weight parameters for implicit TDC on Baird's counterexample. \textbf{Left (small initial step size):} implicit TDC performs comparably to standard TDC. \textbf{Right (large initial step size):} implicit TDC converges quickly with substantially lower run-to-run variance.}\label{fig:BAIRD_TRAJECT}
\end{figure}

Building on Figure \ref{fig:BAIRD_TDC_INSTAB}, Figure \ref{fig:BAIRD_TRAJECT} further demonstrates the improved numerical stability of implicit TDC by presenting the trajectories of the parameter updates. Recall that in Figure \ref{fig:BAIRD_TDC_INSTAB}, the standard TDC trajectories oscillate significantly under larger initial step sizes; in contrast, the implicit TDC trajectories in Figure \ref{fig:BAIRD_TRAJECT} quickly stabilize and remain bounded. These empirical results validate our theoretical findings in Theorem \ref{thm:tdc_decr}, underline the critical importance of selecting appropriate step sizes in off-policy algorithms, and demonstrate the robustness and efficiency of the implicit TDC approach on off-policy evaluation problems such as Baird's counterexample. 

\begin{table}[!h]
  \centering
  \begin{tabular}{l l c c c c c}
    \toprule
    \textbf{Metric} & \textbf{Method} & $\alpha_1$ & $\beta_1$ & \textbf{10th percentile} & \textbf{Mean} & \textbf{90th percentile} \\
    \midrule
    RMSPBE & TDC & 0.05 & 0.50 & 1.503 & 2.173 & 3.195 \\
           & Imp-TDC & 0.05 & 0.50 & 2.010 & 2.518 & 2.966 \\
           & TDC & 1.00 & 10.00 & 3.123 & 8.390 & 16.069 \\
           & Imp-TDC & 1.00 & 10.00 & 0.084 & 0.256 & 0.763 \\
    \midrule
    RMSVE  & TDC & 0.05 & 0.50 & 2.704 & 3.237 & 5.187 \\
           & Imp-TDC & 0.05 & 0.50 & 3.683 & 4.105 & 4.811 \\
           & TDC & 1.00 & 10.00 & 14.668 & 81.682 & 119.105 \\
           & Imp-TDC & 1.00 & 10.00 & 0.307 & 1.001 & 3.371 \\
    \bottomrule
  \end{tabular}
  \caption{Final root mean squared projected Bellman error (RMSPBE) and root mean squared value error (RMSVE) for standard and implicit TDC on Baird's counterexample, reported as the 10th percentile, mean, and 90th percentile across 100 independent runs. Implicit TDC is comparable to standard TDC under small step size configurations; however, with large initial step sizes ($\alpha_1 = 1.0$, $\beta_1 = 10.0$), it attains substantially smaller error on both metrics, whereas standard TDC exhibits severe amplification.}
  \label{tab:combined_error}
\end{table}

\section{Conclusion}
\label{SEC:CONCLUSION}
\noindent This paper introduces implicit versions of TD algorithms, ranging from TD(0) and TD($\lambda)$ for on-policy evaluation to TDC for off-policy evaluation. Combined with a feature approximation framework, we extend the classical TD algorithm to address the critical challenge of step size sensitivity. By reformulating TD updates as fixed-point equations, we show that implicit TD enhances robustness in algorithm convergence. Our theoretical contributions include results on almost sure convergence and finite-time MSE bounds of the implicit TD algorithms. The proposed algorithms are computationally efficient and scalable, making them well-suited for high-dimensional state spaces, as illustrated in several empirical applications. Looking ahead, we believe that an interesting area for future research is the application of implicit TD algorithms to policy learning and multi-agent RL. Furthermore, implicit algorithms could be beneficial in actor-critic algorithms, where sensitivity to step size specification remains an issue.

\clearpage 

\appendix
\section{Proofs of preliminary results}
\noindent We list and establish foundational lemmas on eligibility trace and implicit update, which will be heavily used in establishing asymptotic convergence as well as finite-time MSE bounds. Unless explicitly stated, $\|\cdot\|$ implies the Euclidean norm for vectors and their induced norm for matrices.

\begin{lemma}\label{APPEND_LEMMA_ELIGB_BOUND}
Given a trace-decay parameter $\lambda \in (0,1)$ and a discount factor $\gamma \in (0, 1), \|\boldsymbol{e}_{n}\| \le e_{\max}:=\frac{\phi_{\max}}{1-\lambda \gamma}$, for all $n \in \mathbb{N}$.
\end{lemma}

\begin{proof}
Recall that
$\boldsymbol{e}_{n} = \sum_{i=1}^n (\lambda\gamma)^{n-i}\boldsymbol{\phi}_{i}$. Using the triangle inequality, we have
$$ 
\|\boldsymbol{e}_{n}\| \le \phi_{\max} \sum_{i=1}^n (\lambda\gamma)^{n-i} \le \phi_{\max}\sum_{i=0}^\infty (\lambda\gamma)^{i} = \frac{\phi_{\max}}{1-\lambda\gamma} = : e_{\max}.
$$
\end{proof}

\noindent We now prove Lemma \ref{LEMMA:IMP_EXP_RELATION}, establishing a connection between standard and implicit TD algorithms. \\

\begin{proof}[Proof of Lemma \ref{LEMMA:IMP_EXP_RELATION}]
Rearranging terms for the implicit TD(0) update, we have
\begin{align*}
    \left(\boldsymbol{I} + \alpha_n \boldsymbol{\phi}_{n}\boldsymbol{\phi}_{n}^{\top} \right)\boldsymbol{w}^{\text{im}}_{n+1} = \boldsymbol{w}^{\text{im}}_n + \alpha_n (r_n + \gamma \boldsymbol{\phi}_{n+1}^{\top} \boldsymbol{w}^{\text{im}}_n ) \boldsymbol{\phi}_{n} 
\end{align*}
Multiplying both sides by the inverse of $\left(\boldsymbol{I} + \alpha_n \boldsymbol{\phi}_{n}\boldsymbol{\phi}_{n}^{\top} \right)$, we get
\begin{align*}
\boldsymbol{w}^{\text{im}}_{n+1} &= \left(\boldsymbol{I} + \alpha_n \boldsymbol{\phi}_{n}\boldsymbol{\phi}_{n}^{\top} \right)^{-1} \left\{\boldsymbol{w}^{\text{im}}_n + \alpha_n (r_n + \gamma \boldsymbol{\phi}_{n+1}^{\top} \boldsymbol{w}^{\text{im}}_n ) \boldsymbol{\phi}_{n}\right\} \\
&= \left(\boldsymbol{I} - \frac{\alpha_n}{1 + \alpha_n ||\boldsymbol{\phi}_{n}||^2} \boldsymbol{\phi}_{n}\boldsymbol{\phi}_{n}^{\top}\right)\left\{\boldsymbol{w}^{\text{im}}_n + \alpha_n (r_n + \gamma \boldsymbol{\phi}_{n+1}^{\top} \boldsymbol{w}^{\text{im}}_n ) \boldsymbol{\phi}_{n}\right\}.
\end{align*}
where the second equality follows from the Woodbury identity. Expanding terms out, we have
\begin{align*}
\boldsymbol{w}^{\text{im}}_{n+1} &= \boldsymbol{w}^{\text{im}}_n + \alpha_n r_n \boldsymbol{\phi}_{n} + \alpha_n \gamma \boldsymbol{\phi}_{n+1}^{\top} \boldsymbol{w}^{\text{im}}_n  \boldsymbol{\phi}_{n} - \frac{\alpha_n}{1+\alpha_n \|\boldsymbol{\phi}_{n}\|^2}\boldsymbol{\phi}_{n}^{\top} \boldsymbol{w}^{\text{im}}_n \boldsymbol{\phi}_{n} - \frac{\alpha_n^2 r_n \|\boldsymbol{\phi}_{n}\|^2}{1+\alpha_n \| \boldsymbol{\phi}_{n}\|^2}\boldsymbol{\phi}_{n} - \frac{\alpha_n^2 \gamma \|\boldsymbol{\phi}_{n}\|^2 \boldsymbol{\phi}_{n+1}^{\top} \boldsymbol{w}^{\text{im}}_{n}}{1+\alpha_n \|\boldsymbol{\phi}_{n}\|^2}\boldsymbol{\phi}_{n} \\
&= \boldsymbol{w}^{\text{im}}_n + \alpha_n r_n \left(1- \frac{\alpha_n \|\boldsymbol{\phi}_{n}\|^2}{1+\alpha_n \|\boldsymbol{\phi}_{n}\|^2}\right)\boldsymbol{\phi}_{n} + \alpha_n \gamma \boldsymbol{\phi}_{n+1}^{\top} \boldsymbol{w}^{\text{im}}_n \left(1 - \frac{\alpha_n \|\boldsymbol{\phi}_{n}\|^2}{1 + \alpha_n \|\boldsymbol{\phi}_{n}\|^2} \right)  \boldsymbol{\phi}_{n} - \frac{\alpha_n}{1+\alpha_n \|\boldsymbol{\phi}_{n}\|^2}\boldsymbol{\phi}_{n}^{\top} \boldsymbol{w}^{\text{im}}_n \boldsymbol{\phi}_{n} \\
&= \boldsymbol{w}^{\text{im}}_n + \frac{\alpha_n}{1+\alpha_n \|\boldsymbol{\phi}_{n}\|^2}\left(r_n + \gamma \boldsymbol{\phi}_{n+1}^{\top} \boldsymbol{w}^{\text{im}}_n - \boldsymbol{\phi}_{n}^{\top} \boldsymbol{w}^{\text{im}}_n \right)\boldsymbol{\phi}_{n},
\end{align*}
Analogously, for the implicit TD($\lambda$) algorithm, we have 
\begin{align*}
\left(\boldsymbol{I} + \alpha_n \boldsymbol{e}_{n} \boldsymbol{e}_{n}^{\top} \right)\boldsymbol{w}^{\text{im}}_{n+1} \nonumber = \boldsymbol{w}^{\text{im}}_n + \alpha_n (r_n + \gamma \boldsymbol{\phi}_{n+1}^{\top} \boldsymbol{w}^{\text{im}}_n + \lambda\gamma \boldsymbol{e}_{n-1}^{\top}\boldsymbol{w}^{\text{im}}_{n}) \boldsymbol{e}_{n}.  
\end{align*}
Multiplying by inverse of $\left(\boldsymbol{I} + \alpha_n \boldsymbol{e}_{n} \boldsymbol{e}_{n}^{\top} \right)$, we get
\begin{align*}
\boldsymbol{w}^{\text{im}}_{n+1} &= \left(\boldsymbol{I} + \alpha_n \boldsymbol{e}_{n} \boldsymbol{e}_{n}^{\top} \right)^{-1} \left\{\boldsymbol{w}^{\text{im}}_n + \alpha_n (r_n + \gamma \boldsymbol{\phi}_{n+1}^{\top} \boldsymbol{w}^{\text{im}}_n + \lambda\gamma \boldsymbol{e}_{n-1}^{\top}\boldsymbol{w}^{\text{im}}_{n}) \boldsymbol{e}_{n}\right\}.
\end{align*}
Using the Woodbury identity, we get
\begin{align*}
\boldsymbol{w}^{\text{im}}_{n+1} &= \left(\boldsymbol{I} - \frac{\alpha_n}{1 + \alpha_n ||\boldsymbol{e}_{n}||^2} \boldsymbol{e}_{n} \boldsymbol{e}_{n}^{\top}\right)\left\{\boldsymbol{w}^{\text{im}}_n + \alpha_n (r_n + \gamma \boldsymbol{\phi}_{n+1}^{\top} \boldsymbol{w}^{\text{im}}_n + \lambda\gamma \boldsymbol{e}_{n-1}^{\top}\boldsymbol{w}^{\text{im}}_{n}) \boldsymbol{e}_{n}\right\}. 
\end{align*}
Expanding terms and collecting terms, we have
\begin{align*}
\boldsymbol{w}^{\text{im}}_{n+1} &= \boldsymbol{w}^{\text{im}}_n + \alpha_n r_n \boldsymbol{e}_{n} + \alpha_n \gamma \boldsymbol{\phi}_{n+1}^{\top} \boldsymbol{w}^{\text{im}}_n \boldsymbol{e}_{n} + \alpha_n \lambda\gamma \boldsymbol{e}_{n-1}^{\top} \boldsymbol{w}^{\text{im}}_n \boldsymbol{e}_{n} \\
&\quad - \frac{\alpha_n}{1+\alpha_n \|\boldsymbol{e}_{n}\|^2}\boldsymbol{e}_{n}^{\top} \boldsymbol{w}^{\text{im}}_n \boldsymbol{e}_{n} - \frac{\alpha_n^2 r_n \|\boldsymbol{e}_{n}\|^2}{1+\alpha_n \| \boldsymbol{e}_{n}\|^2}\boldsymbol{e}_{n} - \frac{\alpha_n^2 \gamma \|\boldsymbol{e}_{n}\|^2 \boldsymbol{\phi}_{n+1}^{\top} \boldsymbol{w}^{\text{im}}_{n}}{1+\alpha_n \|\boldsymbol{e}_{n}\|^2}\boldsymbol{e}_{n} - \frac{\alpha_n^2  \lambda \gamma\|\boldsymbol{e}_{n}\|^2 \boldsymbol{e}_{n-1}^{\top} \boldsymbol{w}^{\text{im}}_{n}}{1+\alpha_n \|\boldsymbol{e}_{n}\|^2}\boldsymbol{e}_{n} \\
&= \boldsymbol{w}^{\text{im}}_n + \left(\alpha_n r_n \boldsymbol{e}_{n} - \frac{\alpha_n^2 r_n \|\boldsymbol{e}_{n}\|^2}{1+\alpha_n \| \boldsymbol{e}_{n}\|^2}\boldsymbol{e}_{n}\right) + \left(\alpha_n \gamma \boldsymbol{\phi}_{n+1}^{\top} \boldsymbol{w}^{\text{im}}_n \boldsymbol{e}_{n} -\frac{\alpha_n^2 \gamma \|\boldsymbol{e}_{n}\|^2 \boldsymbol{\phi}_{n+1}^{\top} \boldsymbol{w}^{\text{im}}_{n}}{1+\alpha_n \|\boldsymbol{e}_{n}\|^2}\boldsymbol{e}_{n}\right) \\ &\quad + \left(\alpha_n \lambda\gamma \boldsymbol{e}_{n-1}^{\top} \boldsymbol{w}^{\text{im}}_n \boldsymbol{e}_{n} - \frac{\alpha_n^2  \lambda \gamma\|\boldsymbol{e}_{n}\|^2 \boldsymbol{e}_{n-1}^{\top} \boldsymbol{w}^{\text{im}}_{n}}{1+\alpha_n \|\boldsymbol{e}_{n}\|^2}\boldsymbol{e}_{n}\right)
- \frac{\alpha_n}{1+\alpha_n \|\boldsymbol{e}_{n}\|^2}\boldsymbol{e}_{n}^{\top} \boldsymbol{w}^{\text{im}}_n \boldsymbol{e}_{n} \\
&= \boldsymbol{w}^{\text{im}}_n + \alpha_n r_n \left(1- \frac{\alpha_n \|\boldsymbol{e}_{n}\|^2}{1+\alpha_n \|\boldsymbol{e}_{n}\|^2}\right)\boldsymbol{e}_{n} + \alpha_n \gamma \boldsymbol{\phi}_{n+1}^{\top} \boldsymbol{w}^{\text{im}}_n \left(1 - \frac{\alpha_n \|\boldsymbol{e}_{n}\|^2}{1 + \alpha_n \|\boldsymbol{e}_{n}\|^2} \right) \boldsymbol{e}_{n} \\ &\quad + \alpha_n \lambda\gamma \boldsymbol{e}_{n-1}^{\top} \boldsymbol{w}^{\text{im}}_n \left(1 - \frac{\alpha_n \|\boldsymbol{e}_{n}\|^2}{1 + \alpha_n \|\boldsymbol{e}_{n}\|^2} \right)\boldsymbol{e}_{n} - \frac{\alpha_n}{1+\alpha_n \|\boldsymbol{e}_{n}\|^2}\boldsymbol{e}_{n}^{\top} \boldsymbol{w}^{\text{im}}_n \boldsymbol{e}_{n} \\
&= \boldsymbol{w}^{\text{im}}_n + \frac{\alpha_n}{1+\alpha_n \|\boldsymbol{e}_{n}\|^2}\left(r_n + \gamma \boldsymbol{\phi}_{n+1}^{\top} \boldsymbol{w}^{\text{im}}_n + \lambda\gamma \boldsymbol{e}_{n-1}^{\top} \boldsymbol{w}^{\text{im}}_n - \boldsymbol{e}_{n}^{\top} \boldsymbol{w}^{\text{im}}_n \right)\boldsymbol{e}_{n}.
\end{align*}
\end{proof}

\noindent We next prove Lemma \ref{LEMMA:IMP_TDC_EXP_RELATION} to compare standard and implicit TDC algorithms. \vspace{2mm}  

\begin{proof}[Proof of Lemma \ref{LEMMA:IMP_TDC_EXP_RELATION}]
The implicit TDC update for $w^{\text{im}}$ can be rewritten as
\begin{align}
    &\left(\boldsymbol{I} + \alpha_n \rho_n \boldsymbol{\phi}_{n} \boldsymbol{\phi}_{n}^{\top} \right)\boldsymbol{w}^{\text{im}}_{n+1} = \boldsymbol{w}^{\text{im}}_n + \alpha_n \rho_n \left(r_n \boldsymbol{\phi}_{n} + \gamma \boldsymbol{\phi}_{n} \boldsymbol{\phi}_{n+1}^{\top} \boldsymbol{w}^{\text{im}}_n - \gamma \boldsymbol{\phi}_{n+1}\boldsymbol{\phi}_{n}^{\top} \boldsymbol{u}^{\text{im}}_{n}  \right) \nonumber \\
    &\Leftrightarrow \boldsymbol{w}^{\text{im}}_{n+1} = \left(\boldsymbol{I} + \alpha_n \rho_n \boldsymbol{\phi}_{n} \boldsymbol{\phi}_{n}^{\top} \right)^{-1}\left\{\boldsymbol{w}^{\text{im}}_n + \alpha_n \rho_n \left(r_n \boldsymbol{\phi}_{n} + \gamma \boldsymbol{\phi}_{n} \boldsymbol{\phi}_{n+1}^{\top} \boldsymbol{w}^{\text{im}}_n - \gamma \boldsymbol{\phi}_{n+1}\boldsymbol{\phi}_{n}^{\top} \boldsymbol{u}^{\text{im}}_{n}  \right)\right\} \nonumber \\
    &\Leftrightarrow \boldsymbol{w}^{\text{im}}_{n+1} = \left(\boldsymbol{I} - \alpha'_n \rho_n \boldsymbol{\phi}_{n} \boldsymbol{\phi}_{n}^{\top} \right)\left\{\boldsymbol{w}^{\text{im}}_n + \alpha_n \rho_n \left(r_n \boldsymbol{\phi}_{n} + \gamma \boldsymbol{\phi}_{n} \boldsymbol{\phi}_{n+1}^{\top} \boldsymbol{w}^{\text{im}}_n - \gamma \boldsymbol{\phi}_{n+1}\boldsymbol{\phi}_{n}^{\top} \boldsymbol{u}^{\text{im}}_{n}  \right)\right\} \label{imp_w_updt_sherman}
\end{align}
for $\alpha'_n = \frac{\alpha_n}{1+\alpha_n \rho_n \|\boldsymbol{\phi}_{n}\|^2}$.
Expanding the right hand side of \eqref{imp_w_updt_sherman}, we have
\begin{align*}
    \boldsymbol{w}^{\text{im}}_{n+1} &= \boldsymbol{w}^{\text{im}}_n + \alpha_n \rho_n r_n \boldsymbol{\phi}_{n} + \alpha_n \rho_n \gamma \left(\boldsymbol{\phi}_{n+1}^{\top} \boldsymbol{w}^{\text{im}}_{n}\right)\boldsymbol{\phi}_{n} - \alpha_n \rho_n \gamma \left(\boldsymbol{\phi}_{n}^{\top} \boldsymbol{u}^{\text{im}}_{n}\right) \boldsymbol{\phi}_{n+1}\\ 
    & \quad - \alpha'_n \rho_n \alpha_n \rho_n r_n \|\boldsymbol{\phi}_{n}\|^2  \boldsymbol{\phi}_{n} - \alpha'_n \rho_n \alpha_n \rho_n \gamma \|\boldsymbol{\phi}_{n}\|^2 \left(\boldsymbol{\phi}_{n+1}^{\top} \boldsymbol{w}^{\text{im}}_{n}\right)  \boldsymbol{\phi}_{n} + \alpha'_n \rho_n \alpha_n \rho_n \gamma \left(\boldsymbol{\phi}_{n}^{\top} \boldsymbol{\phi}_{n+1}\right) \left(\boldsymbol{\phi}_{n}^{\top} \boldsymbol{u}^{\text{im}}_{n}\right)  \boldsymbol{\phi}_{n} \\
    & \quad  -\alpha'_n \rho_n 
    \left(\boldsymbol{\phi}_{n}^{\top} \boldsymbol{w}_{n}^{\text{im}}\right)\boldsymbol{\phi}_{n} \\&= \boldsymbol{w}^{\text{im}}_n + \alpha_n \rho_n r_n \left(1-\alpha'_n \rho_n\|\boldsymbol{\phi}_{n}\|^2\right)\boldsymbol{\phi}_{n} + \alpha_n \rho_n \gamma \left(1-\alpha'_n \rho_n\|\boldsymbol{\phi}_{n}\|^2\right)\left(\boldsymbol{\phi}_{n+1}^{\top} \boldsymbol{w}^{\text{im}}_{n}\right)\boldsymbol{\phi}_{n} -\alpha'_n \rho_n \left(\boldsymbol{\phi}_{n}^{\top} \boldsymbol{w}_{n}^{\text{im}}\right)\boldsymbol{\phi}_{n} \\
    &\quad - \alpha_n \rho_n \gamma \left(\boldsymbol{\phi}_{n}^{\top} \boldsymbol{u}^{\text{im}}_{n}\right) \boldsymbol{\phi}_{n+1}+ \alpha'_n \rho_n \alpha_n \rho_n \gamma \left(\boldsymbol{\phi}_{n}^{\top} \boldsymbol{\phi}_{n+1}\right) \left(\boldsymbol{\phi}_{n}^{\top} \boldsymbol{u}^{\text{im}}_{n}\right)  \boldsymbol{\phi}_{n} \\
    & = \boldsymbol{w}^{\text{im}}_n + \alpha'_n \rho_n \delta_n^{\text{im}} \boldsymbol{\phi}_{n} - \alpha_n \rho_n \gamma \left(\boldsymbol{\phi}_{n}^{\top} \boldsymbol{u}^{\text{im}}_{n}\right) \left\{\boldsymbol{\phi}_{n+1}- \alpha'_n \rho_n \left(\boldsymbol{\phi}_{n}^{\top} \boldsymbol{\phi}_{n+1}\right) \boldsymbol{\phi}_{n} \right\}
\end{align*}
where $\delta_n^{\text{im}}\boldsymbol{\phi}_{n}:= r_n \boldsymbol{\phi}_{n} + \gamma \boldsymbol{\phi}_{n} \boldsymbol{\phi}_{n+1}^{\top}\boldsymbol{w}^{\text{im}}_n - \boldsymbol{\phi}_{n} \boldsymbol{\phi}_{n}^{\top} \boldsymbol{w}^{\text{im}}_{n}$. An implicit TDC update for auxiliary parameter $u^{\text{im}}$ can be analogously derived. Consider
\begin{align}
   \left(\boldsymbol{I} + \beta_n \rho_n \boldsymbol{\phi}_{n}\boldsymbol{\phi}_{n}^{\top} \right) \boldsymbol{u}^{\text{im}}_{n+1} = \boldsymbol{u}^{\text{im}}_{n} + \beta_n \rho_n \delta_n^{\text{im}}\boldsymbol{\phi}_{n} 
    &\Leftrightarrow \boldsymbol{u}^{\text{im}}_{n+1} = \left(\boldsymbol{I} + \beta_n \rho_n \boldsymbol{\phi}_{n}\boldsymbol{\phi}_{n}^{\top} \right)^{-1}\left(\boldsymbol{u}^{\text{im}}_{n} + \beta_n \rho_n \delta_n^{\text{im}}\boldsymbol{\phi}_{n} \right)\nonumber \\
    &\Leftrightarrow \boldsymbol{u}^{\text{im}}_{n+1} = \left(\boldsymbol{I} - \beta'_n \rho_n\boldsymbol{\phi}_{n}\boldsymbol{\phi}_{n}^{\top} \right)\left(\boldsymbol{u}^{\text{im}}_{n} + \beta_n \rho_n \delta_n^{\text{im}}\boldsymbol{\phi}_{n} \right) \label{imp_u_updt_sherman}
\end{align}
for $\beta'_n = \frac{\beta_n}{1 + \beta_n\rho_n \|\boldsymbol{\phi}_{n}\|^2}$.  Re-expressing the right hand side of \eqref{imp_u_updt_sherman}, we obtain
\begin{align*}
    \boldsymbol{u}^{\text{im}}_{n+1} &= \boldsymbol{u}^{\text{im}}_{n} + \beta_n \rho_n \delta_n^{\text{im}}\boldsymbol{\phi}_{n} - \beta'_n \rho_n \boldsymbol{\phi}_{n}^{\top} \boldsymbol{u}_{n}^{\text{im}} \boldsymbol{\phi}_{n} - \beta'_n \rho_n \beta_n \rho_n \|\boldsymbol{\phi}_{n}\|^2 \delta_n^{\text{im}} \boldsymbol{\phi}_{n} \\
    &= \boldsymbol{u}^{\text{im}}_{n} - \beta'_n \rho_n \boldsymbol{\phi}_{n}^{\top} \boldsymbol{u}_{n}^{\text{im}} \boldsymbol{\phi}_{n} + \beta_n \rho_n \delta_n^{\text{im}}\boldsymbol{\phi}_{n} \left(1- \beta'_n \rho_n \|\boldsymbol{\phi}_{n}\|^2 \right) \\
    &= \boldsymbol{u}^{\text{im}}_{n} + \beta'_n \rho_n \delta_n^{\text{im}}\boldsymbol{\phi}_{n} - \beta'_n \rho_n \boldsymbol{\phi}_{n} \boldsymbol{\phi}_{n}^{\top} \boldsymbol{u}_{n}^{\text{im}}.
\end{align*}
\end{proof}

\subsection{Auxiliary lemmas for finite-time bounds}
\begin{lemma}\label{APPEND_LEMMA:IMP_STEP}
Given a positive step size sequence $(\alpha_n)_{n \in \mathbb{N}}$, define the sequence $(\tilde{\alpha}_n)_{n \in \mathbb{N}}$ as
$$
\tilde{\alpha}_n = \begin{cases}
    \frac{\alpha_n}{1+\alpha_n\|\boldsymbol{\phi}_{n}\|^2} & \text{for TD}(0) \\
    \frac{\alpha_n}{1+\alpha_n\|\boldsymbol{e}_{n}\|^2} & \text{for TD}(\lambda).
\end{cases}
$$
Then the following bounds hold almost surely
\begin{align*}
    \frac{\alpha_n}{1+\alpha_n \phi_{\max}^2} &\le \tilde{\alpha}_n \le \alpha_n \qquad \text{for TD}(0) \\
    \frac{\alpha_n}{1+\alpha_n e_{\max}^2} &\le \tilde{\alpha}_n \le \alpha_n \qquad \text{for TD}(\lambda).
\end{align*}
\end{lemma}

\begin{proof}
    Since $1+\alpha_n\|\boldsymbol{\phi}_{n}\|^2 \ge 1$, we have $\tilde \alpha_n \le \alpha_n$ for TD(0). Analogously, $1+\alpha_n\|\boldsymbol{e}_{n}\|^2 \ge 1$ implies $\tilde \alpha_n \le \alpha_n$ for TD($\lambda$). Lower bounds follow from $\frac{1}{1+\alpha_n\|\boldsymbol{\phi}_{n}\|^2} \ge \frac{1}{1+\alpha_n\phi_{\max}^2}$ and $\frac{1}{1+\alpha_n\|\boldsymbol{e}_{n}\|^2} \ge \frac{1}{1+\alpha_n e_{\max}^2}$. 
\end{proof}

\begin{lemma}\label{harmonic_sum_bound} Let $\alpha_n = \alpha_1 n^{-s}$ with $\alpha_1 > 0$, $\gamma > 0$, and $1 \leq \ell \leq n$. Then
$$
e^{-\gamma\alpha_1\sum_{k=\ell+1}^n k^{-s}} \lesssim \begin{cases} e^{-\frac{\gamma\alpha_1}{1-s}\left\{(n+1)^{1-s}-(\ell+1)^{1-s}\right\}} & s \in (0.5, 1) \\[6pt] \left(\frac{\ell+1}{n+1}\right)^{\gamma\alpha_1} & s = 1 \end{cases}
$$
\end{lemma}
\begin{proof}
Suppose $s \in (0.5,1)$. Since $x^{-s}$ is decreasing, the integral comparison gives for each $k \geq 1$:
$k^{-s} \geq \int_k^{k+1} x^{-s}dx$. Summing from $k = \ell+1$ to $n$:
$$
\sum_{k=\ell+1}^n k^{-s} \geq \int_{\ell+1}^{n+1} x^{-s}dx = \frac{(n+1)^{1-s} - (\ell+1)^{1-s}}{1-s}.
$$
Therefore
$$
e^{-\gamma\alpha_1\sum_{k=\ell+1}^n k^{-s}} \leq \exp\left(-\frac{\gamma\alpha_1}{1-s}\left[(n+1)^{1-s} - (\ell+1)^{1-s}\right]\right).
$$
Next, consider the case when $s = 1$. Since $x^{-1}$ is decreasing, again the same integral comparison gives:
$$
\sum_{k=\ell+1}^n k^{-1} \geq \int_{\ell+1}^{n+1} x^{-1}dx = \log(n+1) - \log(\ell+1) = \log\frac{n+1}{\ell+1}.
$$
Therefore
$$
e^{-\gamma\alpha_1\sum_{k=\ell+1}^n k^{-1}} \leq \exp\left(-\gamma\alpha_1\log\frac{n+1}{\ell+1}\right) = \left(\frac{\ell+1}{n+1}\right)^{\gamma\alpha_1} \lesssim \left(\frac{\ell}{n}\right)^{\gamma\alpha_1}.
$$
\end{proof}

\begin{lemma}\label{LEMMA:prelim_TD_exp_bound}
For $n \in \mathbb{N}$, let $\alpha_n = \frac{\alpha_1}{n^s}$, $\gamma > 0$ 
and $\ell \in \{1, \cdots, n\}$. If $s\in (0.5,1)$ or $s =1$ and 
$\gamma \alpha_1 > 1$,
\begin{enumerate}
    \item $\sum_{i=1}^{\ell} e^{-\gamma \sum_{k=i+1}^{n} \alpha_{k}} 
    \alpha_{i} \lesssim \begin{cases} e^{-\frac{\gamma\alpha_1}{1-s}\left\{(n+1)^{1-s}-(\ell+1)^{1-s}\right\}} & s \in (0.5, 1) \\[6pt] \left(\frac{\ell+1}{n+1}\right)^{\gamma\alpha_1} & s = 1 \end{cases} $ 
    \item $\sum_{i=\tau_{\alpha_n}+1}^{n} \left(e^{-\gamma 
    \sum_{k=i+1}^{n} \alpha_{k}} \alpha_{i-\tau_{\alpha_n}} \alpha_{i} 
    \right) \lesssim \alpha_{(n-\tau_{\alpha_n})\vee 1}$
\end{enumerate}
with the convention that a sum is zero when its lower limit exceeds its upper limit.
\end{lemma}

\begin{proof}
Let $T_n = \sum_{i=1}^{n-1} \alpha_i$ and use the convention 
$\sum_{k=n+1}^n \alpha_k = 0$. When $s=1$ and $\gamma\alpha_1 > 1$, 
fix any $c > \frac{\gamma\alpha_1}{\gamma\alpha_1-1}$, so that 
$\frac{(c-1)\gamma\alpha_1}{c} > 1$; when $s \in (0.5,1)$ set $c=2$.

\medskip
\noindent\textbf{Proof of statement 1.}
\begin{align}
\sum_{i=1}^{\ell} e^{-\gamma \sum_{k=i+1}^{n} \alpha_{k}} \alpha_{i} 
& \leq \max_{i \geq 1}\left(e^{\gamma \alpha_{i}}\right) 
\sum_{i=1}^{\ell} e^{-\gamma \sum_{k=i}^{n} \alpha_{k}} \alpha_{i} 
= e^{\gamma \alpha_1} \sum_{i=1}^{\ell} 
e^{-\gamma\left(T_{n+1}-T_{i}\right)} \alpha_{i} \nonumber \\
& \leq e^{\gamma \alpha_1} \int_{0}^{T_{\ell+1}} 
e^{-\gamma\left(T_{n+1}-z\right)} dz 
\leq \frac{e^{\gamma \alpha_1}}{\gamma} 
e^{-\gamma\left(T_{n+1}-T_{\ell+1}\right)} \nonumber \\
&= \frac{e^{\gamma \alpha_1}}{\gamma} 
e^{-\gamma \alpha_1 \sum_{k=\ell+1}^{n} k^{-s}}
\lesssim \begin{cases} e^{-\frac{\gamma\alpha_1}{1-s}\left\{(n+1)^{1-s}-(\ell+1)^{1-s}\right\}} & s \in (0.5, 1) \\[6pt] \left(\frac{\ell+1}{n+1}\right)^{\gamma\alpha_1} & s = 1 \end{cases} \label{alpha_bound_cbc}
\end{align}
where the last step uses Lemma \ref{harmonic_sum_bound}. 

\medskip
\noindent\textbf{Proof of statement 2.}
The bound trivially holds for $\tau_{\alpha_n} \ge n$. Therefore, we consider the case $\tau_{\alpha_n} < n$. First notice that
\begin{align}
\sum_{i=\tau_{\alpha_n}+1}^{n} e^{-\gamma \sum_{k=i+1}^{n} \alpha_{k}} 
\alpha_{i} & \leq \max_{i \geq 1}\left(e^{\gamma \alpha_{i}}\right) 
\sum_{i=\tau_{\alpha_n}+1}^{n} e^{-\gamma \sum_{k=i}^{n} \alpha_{k}} 
\alpha_{i} = e^{\gamma \alpha_1} \sum_{i=\tau_{\alpha_n}+1}^{n} 
e^{-\gamma\left(T_{n+1}-T_{i}\right)} \alpha_{i} \nonumber \\
& \leq e^{\gamma \alpha_1}\int_{T_{\tau_{\alpha_n}+1}}^{T_{n+1}} 
e^{-\gamma\left(T_{n+1}-z\right)} dz 
= \frac{e^{\gamma \alpha_1}}{\gamma} \left\lbrace 1 - 
e^{-\gamma\left(T_{n+1}-T_{\tau_{\alpha_n}+1}\right)}\right\rbrace 
\leq \frac{e^{\gamma \alpha_1}}{\gamma}. 
\label{prelim_second_pre}
\end{align}
Then, we have
\begin{align}
\sum_{i=\tau_{\alpha_n}+1}^{n} e^{-\gamma \sum_{k=i+1}^{n} \alpha_{k}} 
\alpha_{i-\tau_{\alpha_n}} \alpha_{i} 
& \leq \max_{i \in\left[\tau_{\alpha_n}+1, n\right]}
\left\{e^{-\frac{(c-1)\gamma}{c} \sum_{k=i+1}^{n} \alpha_{k}} 
\alpha_{i-\tau_{\alpha_n}}\right\} 
\sum_{i=\tau_{\alpha_n}+1}^{n} e^{-\frac{\gamma}{c} 
\sum_{k=i+1}^{n} \alpha_{k}} \alpha_{i} \nonumber \\
& \lesssim \max_{i \in\left[\tau_{\alpha_n}+1, n\right]}
\left\{e^{-\frac{(c-1)\gamma}{c} \sum_{k=i+1}^{n} \alpha_{k}} 
\alpha_{i-\tau_{\alpha_n}}\right\} 
\label{TD_aux_bound3}
\end{align}
where the second inequality follows from \eqref{prelim_second_pre} 
(with $\gamma$ replaced by $\gamma/c$). To bound \eqref{TD_aux_bound3}, 
note that $\left\{e^{-\frac{(c-1)\gamma}{c} \sum_{k=i+1}^{n} \alpha_{k}} 
\alpha_{i-\tau_{\alpha_n}}\right\}_{i\in \mathbb{N}}$ is eventually 
increasing, i.e., there exists $i_{\alpha} \in \mathbb{N}$ such that
\begin{align*}
\max_{i \in\left[\tau_{\alpha_n}+1, n\right]}
\left\{e^{-\frac{(c-1)\gamma}{c} \sum_{k=i+1}^{n} \alpha_{k}} 
\alpha_{i-\tau_{\alpha_n}}\right\} = \alpha_{n-\tau_{\alpha_n}} 
\quad \text{if} \quad \tau_{\alpha_n}+1 \ge i_{\alpha}.
\end{align*}
On the other hand, if $\tau_{\alpha_n}+1 < i_{\alpha}$, we get
\begin{align*}
&\max_{i \in\left[\tau_{\alpha_n}+1, n\right]}
\left\{e^{-\frac{(c-1)\gamma}{c} \sum_{k=i+1}^{n} \alpha_{k}} 
\alpha_{i-\tau_{\alpha_n}}\right\} \\
&\quad\leq \max_{i \in\left[\tau_{\alpha_n}+1, i_{\alpha}\right]}
\left\{e^{-\frac{(c-1)\gamma}{c} \sum_{k=i+1}^{n} \alpha_{k}} 
\alpha_{i-\tau_{\alpha_n}}\right\} 
+ \max_{i \in\left[i_{\alpha}+1, n\right]}
\left\{e^{-\frac{(c-1)\gamma}{c} \sum_{k=i+1}^{n} \alpha_{k}} 
\alpha_{i-\tau_{\alpha_n}}\right\} \\
&\quad\leq e^{-\frac{(c-1)\gamma}{c} \sum_{k=1}^{n} \alpha_{k}} 
\max_{i \in\left[\tau_{\alpha_n}+1, i_{\alpha}\right]}
\left\{e^{\frac{(c-1)\gamma}{c} \sum_{k=1}^{i} \alpha_{k}} 
\alpha_{i-\tau_{\alpha_n}}\right\} + \alpha_{n-\tau_{\alpha_n}} \\
&\quad\lesssim e^{-\frac{(c-1)\gamma\alpha_1}{c} \sum_{k=1}^n k^{-s}} 
+ \alpha_{n-\tau_{\alpha_n}} \\
&\quad\leq e^{-\frac{(c-1)\gamma\alpha_1}{c}\phi(n)} 
+ \alpha_{n-\tau_{\alpha_n}}
\end{align*}
where
$$
\phi(n) := \begin{cases}
\frac{n^{1-s}-1}{1-s} & \text{for } s \in (0.5, 1),\\[6pt]
\log n & \text{for } s = 1.
\end{cases}
$$
For $s \in (0.5,1)$, we have 
$e^{-\frac{(c-1)\gamma\alpha_1}{c}\phi(n)} = e^{-\frac{\gamma\alpha_1}{2}\left(\frac{n^{1-s}-1}{1-s}\right)} \lesssim \alpha_{n-\tau_{\alpha_n}}$. For $s=1$, we have 
$e^{-\frac{(c-1)\gamma\alpha_1}{c}\log n} = n^{-\frac{(c-1)\gamma\alpha_1}{c}} 
\lesssim \frac{1}{n} \lesssim \alpha_{n-\tau_{\alpha_n}}$ since 
$\frac{(c-1)\gamma\alpha_1}{c} > 1$ by the choice of $c$. 
Combining both cases,
\begin{align*}
\sum_{i=\tau_{\alpha_n}+1}^{n} e^{-\gamma \sum_{k=i+1}^{n} \alpha_{k}} 
\alpha_{i-\tau_{\alpha_n}} \alpha_{i} \lesssim \alpha_{n-\tau_{\alpha_n}}.
\end{align*}
\end{proof}

\section{Theoretical analysis for implicit TD(0) and TD($\lambda$)}
\noindent To simplify our presentation, for the TD(0) algorithm, let us introduce
\begin{align*}
    S_{\boldsymbol{z}_n}(\boldsymbol{w}) &:= r_n \boldsymbol{\phi}_{n} + \gamma \boldsymbol{\phi}_{n} \boldsymbol{\phi}_{n+1}^{\top} \boldsymbol{w} - \boldsymbol{\phi}_{n} \boldsymbol{\phi}_{n}^{\top} \boldsymbol{w} = \boldsymbol{b}_{n} + \boldsymbol{A}_n \boldsymbol{w}, \\
    S(\boldsymbol{w})  &:= \mathbb{E}_\mu\left(r_n \boldsymbol{\phi}_{n}\right) + \mathbb{E}_\mu(\gamma \boldsymbol{\phi}_{n} \boldsymbol{\phi}_{n+1}^{\top})\boldsymbol{w} - \mathbb{E}_\mu(\boldsymbol{\phi}_{n} \boldsymbol{\phi}_{n}^{\top}) \boldsymbol{w} = \boldsymbol{b} + \boldsymbol{A}\boldsymbol{w},
\end{align*}
where $\boldsymbol{A}_n = \gamma \boldsymbol{\phi}_{n} \boldsymbol{\phi}_{n+1}^{\top}-\boldsymbol{\phi}_{n} \boldsymbol{\phi}_{n}^{\top}, ~\boldsymbol{A} = \mathbb{E}_\mu\left(\boldsymbol{A}_n\right)$, $\boldsymbol{b}_{n} = r_n \boldsymbol{\phi}_{n}, \boldsymbol{b} = \mathbb{E}_\mu\left(\boldsymbol{b}_{n}\right)$. Here $\mathbb{E}_\mu$ is the expectation with respect to the steady-state distribution of the Markov process $(\boldsymbol{z}_n)_{n \in \mathbb{N}}$ where $\boldsymbol{z}_n = (\boldsymbol{x}_{n}, \boldsymbol{x}_{n+1}) \in \mathcal{Z}:= \mathcal{X}\times \mathcal{X} \subset \mathbb{R}^{2k}$. Note that transition probabilities of $(\boldsymbol{z}_n)_{n\in \mathbb{N}}$ coincide with that of $(\boldsymbol{x}_{n})_{n \in \mathbb{N}}$. 
Similarly, for the TD($\lambda$) algorithm, let us define
\begin{align*}
    S_{\boldsymbol{z}_n}(\boldsymbol{w}) &:= r_n \boldsymbol{e}_{n} + \gamma \boldsymbol{e}_{n} \boldsymbol{\phi}_{n+1}^{\top} \boldsymbol{w} - \boldsymbol{e}_{n} \boldsymbol{\phi}_{n}^{\top} \boldsymbol{w} = \boldsymbol{b}_{n} + \boldsymbol{A}_n \boldsymbol{w}, \\
    S(\boldsymbol{w}) &:= \mathbb{E}_\mu\left(r_n \boldsymbol{e}_{-\infty:n}\right) + \mathbb{E}_\mu(\gamma \boldsymbol{e}_{-\infty:n} \boldsymbol{\phi}_{n+1}^{\top})\boldsymbol{w} - \mathbb{E}_\mu(\boldsymbol{e}_{-\infty:n} \boldsymbol{\phi}_{n}^{\top}) \boldsymbol{w} = \boldsymbol{b} +  \boldsymbol{A} \boldsymbol{w}, 
\end{align*}
where $\boldsymbol{e}_{-\infty:n}:= \sum_{k=0}^\infty (\lambda\gamma)^k \boldsymbol{\phi}_{n-k}$ represents the steady-state eligibility trace and
$\boldsymbol{A}_n = \gamma \boldsymbol{e}_{n} \boldsymbol{\phi}_{n+1}^{\top} - \boldsymbol{e}_{n} \boldsymbol{\phi}_{n}^{\top},  ~\boldsymbol{A} = \mathbb{E}_\mu(\gamma \boldsymbol{e}_{-\infty:n} \boldsymbol{\phi}_{n+1}^{\top}) - \mathbb{E}_\mu\left(\boldsymbol{e}_{-\infty:n} \boldsymbol{\phi}_{n}^{\top}\right) = \lim_{n \to \infty}\mathbb{E}_\mu(\boldsymbol{A}_n)$, $\boldsymbol{b}_{n} = r_n \boldsymbol{e}_{n}$
and $\boldsymbol{b} = \mathbb{E}_\mu(r_n \boldsymbol{e}_{-\infty:n}) = \lim_{n\to\infty}\mathbb{E}_\mu(\boldsymbol{b}_{n})$. Note that $\mathbb{E}_\mu$ is the expectation with respect to the steady-state distribution of the Markov process $(\boldsymbol{z}_n)_{n \in \mathbb{N}}$ where $\boldsymbol{z}_n = (\boldsymbol{x}_{n}, \boldsymbol{x}_{n+1}, \boldsymbol{e}_{n}) \in \mathcal{Z}:=\mathcal{X}\times \mathcal{X}\times \mathcal{E} \subset \mathbb{R}^{2k+d}$ with $\mathcal{E}: = \{\boldsymbol{e} \in \mathbb{R}^d: \|\boldsymbol{e}\| \le e_{\max}\}$. Again, transition probabilities of $(\boldsymbol{z}_n)_{n\in \mathbb{N}}$ coincide with that of $(\boldsymbol{x}_{n})_{n \in \mathbb{N}}$. \citet{tsitsiklis1996analysis} has shown that the limit point of TD algorithms, denoted by $\boldsymbol{w}_{\star}$ solves the equation $S(\boldsymbol{w}) = 0$. 

\subsection{Almost sure convergence of implicit TD(0)/TD($\lambda$)}
\noindent We begin by stating the key theorem, along with the conditions we use to establish almost sure convergence of the implicit TD algorithms. For any function $f(\boldsymbol{z},\boldsymbol{w})$ on $\mathcal{Z} \times \mathbb{R}^d$, we denote by $f_{\boldsymbol{w}}$ the mapping $\boldsymbol{z} \mapsto f(\boldsymbol{z},\boldsymbol{w})$, and by $\Pi f_{\boldsymbol{w}}$ the function
$$
\Pi f_{\boldsymbol{w}}:\boldsymbol{z} \mapsto \int f(\boldsymbol{y},\boldsymbol{w})\, \Pi(\boldsymbol{z}, d\boldsymbol{y}) = \mathbb{E}\{f_{\boldsymbol{w}}(\boldsymbol{y}) \mid \boldsymbol{z} \},
$$
where $\Pi$ is a probability kernel. Notice that 
$$
\Pi^2 f_{\boldsymbol{w}}(\boldsymbol{z}) = \int \Pi f_{\boldsymbol{w}}(\boldsymbol{y}) \Pi(\boldsymbol{z}, d\boldsymbol{y}) = \int \int f(\boldsymbol{u}, \boldsymbol{w}) \Pi(\boldsymbol{y}, d\boldsymbol{u})\Pi(\boldsymbol{z}, d\boldsymbol{y}) = \mathbb{E}\{f_{\boldsymbol{w}}(\boldsymbol{u}) \mid \boldsymbol{z}\}.
$$
In general, we have
$$
\Pi^k f_{\boldsymbol{w}}(\boldsymbol{z}_1) = \mathbb{E}\{f_{\boldsymbol{w}}(\boldsymbol{z}_{k+1}) \mid \boldsymbol{z}_1\}.
$$
\begin{assumption}\label{ASSUMP_B1}
$(\alpha_n)_{n \in \mathbb N}$ is a step size sequence of the form $\alpha_n = \frac{\alpha_1}{n^s}$ where $s \in (0.5, 1]$.
\end{assumption}

\begin{assumption}\label{ASSUMP_B2}
$\exists$ a probability kernel $\Pi$ associated with the Markov chain $(\boldsymbol{z}_n)_{n \in \mathbb{N}}$ such that, for any Borel set $A \subset \mathcal{Z}$,
$
P(\boldsymbol{z}_{n+1} \in \mathcal{A} \mid \boldsymbol{z}_n) = \Pi(\boldsymbol{z}_n, \mathcal{A}).
$
\end{assumption}

\noindent The above two assumptions are easily satisfied in many circumstances and therefore we assume them throughout in this subsection. We now consider an iterative algorithm of the form
\begin{equation}
    \boldsymbol{w}_{n+1} = \boldsymbol{w}_{n} + \alpha_n S_{\boldsymbol{z}_n}(\boldsymbol{w}_{n}) + \alpha_n^2 R_{n}(\boldsymbol{w}_{n}, \boldsymbol{z}_n) \label{w_iterate}
\end{equation}
with the following conditions
\begin{align}
    & S_{\boldsymbol{z}}(\boldsymbol{w}) ~\text{is locally Lipschitz on } \mathbb{R}^d \label{cond_Lipshitz}\\
    &\|S_{\boldsymbol{z}}(\boldsymbol{w})\| \lesssim (1 + \|\boldsymbol{w}\|)(1 + \|\boldsymbol{z}\|^{q_1}) \label{cond_S}\\
    &\|R_{n}(\boldsymbol{w}, \boldsymbol{z})\| \lesssim (1 + \|\boldsymbol{w}\|)(1 + \|\boldsymbol{z}\|^{q_2}), \quad \forall n \in \mathbb{N} \label{cond_R}\\
    &\mathbb{E}(1 + \|\boldsymbol{z}_{n+1}\|^{q_3} \mid \boldsymbol{z}_1 = \boldsymbol{z}, \boldsymbol{w}_1 = \boldsymbol{w}) \lesssim (1 + \|\boldsymbol{z}\|^{q_3})\label{cond_z} \\
& \exists~\nu_{\boldsymbol{w}} ~\text{such that}~ (I - \Pi)\nu_{\boldsymbol{w}} = S_{\boldsymbol{z}}(\boldsymbol{w}) - S(\boldsymbol{w}) \label{cond_nu_exist}\\
&\|\nu_{\boldsymbol{w}}(\boldsymbol{z})\| \lesssim  (1 + \|\boldsymbol{w}\|)(1 + \|\boldsymbol{z}\|^{q_4}) \label{cond_nu}
\end{align}
for all $\boldsymbol{w} \in \mathbb{R}^d$ and $\boldsymbol{z} \in \mathcal{Z}$ with $q_1, q_2, q_3, q_4 \in \mathbb{R}_{>0}$.
Furthermore, suppose that there exist $q_5 > 0$ such that, for all $\boldsymbol{w}, \boldsymbol{w}'$ with $\|\boldsymbol{w}\| \le R_{\boldsymbol{w}}$ and $\|\boldsymbol{w}'\| \le R_{\boldsymbol{w}}$, where $R_{\boldsymbol{w}}\in \mathbb{R}_{>0}$,
\begin{equation}
\|\Pi \nu_{\boldsymbol{w}}(\boldsymbol{z}) - \Pi \nu_{\boldsymbol{w}'}(\boldsymbol{z})\| \lesssim \|\boldsymbol{w} - \boldsymbol{w}'\| (1 + \|\boldsymbol{z}\|^{q_5}).
\label{cond_pi_nu}
\end{equation}

\begin{theorem}\label{THM:benveniste}
Suppose Assumptions \ref{ASSUMP_B1} - \ref{ASSUMP_B2} and \eqref{cond_Lipshitz}--\eqref{cond_pi_nu} hold. Assume that $S(\boldsymbol{w}_{\star}) = 0$ and that there exists a non-negative function $U$ on $\mathbb{R}^d$ of
class $\mathcal{C}^2$ with bounded second derivatives, such that
\begin{enumerate}
\item[(i)] $\langle \nabla U(\boldsymbol{w}),  S(\boldsymbol{w})\rangle < 0$ for all $w \neq \boldsymbol{w}_{\star}$
\item[(ii)] $U(\boldsymbol{w}) \ge \alpha\|\boldsymbol{w}\|^2$ whenever $\|\boldsymbol{w}\| \ge R_0$ for some $\alpha > 0$ and $R_0 \in \mathbb{R}_{>0}$
\item[(iii)] $U(\boldsymbol{w}) = 0$ if and only if $w = \boldsymbol{w}_{\star}$.
\end{enumerate}
Then the iterates $(\boldsymbol{w}_{n})_{n \in \mathbb{N}}$ of \eqref{w_iterate}  converge almost surely to $\boldsymbol{w}_{\star}$.
\end{theorem}
\begin{proof}
This is a special case of Theorem 17 on page 239 of \citep{benveniste2012adaptive}.
\end{proof}

\begin{proof}[Proof of Theorem \ref{THM:ASYM_IMP_TD}]
Since TD($0$) is subsumed by TD($\lambda$) as a special case, we present the proof for TD($\lambda$).
\noindent For $n \in \mathbb{N}$, let us rewrite the implicit TD($\lambda$) update as
\begin{align*}
 \boldsymbol{w}^{\text{im}}_{n+1} = \boldsymbol{w}^{\text{im}}_n + \alpha_n S_{\boldsymbol{z}_n}(\boldsymbol{w}^{\text{im}}_{n}) +  \alpha_n^2 R_{n}(\boldsymbol{w}_{n}^{\text{im}},\boldsymbol{z}_n) \quad \text{where} \quad 
 R_{n}(\boldsymbol{w}_{n}^{\text{im}}, \boldsymbol{z}_n) := \frac{-\|\boldsymbol{e}_{n}\|^2}{1+\alpha_n \|\boldsymbol{e}_{n}\|^2}S_{\boldsymbol{z}_n}(\boldsymbol{w}^{\text{im}}_{n}).
\end{align*}
We again verify conditions \eqref{cond_Lipshitz}--\eqref{cond_pi_nu} to invoke Theorem \eqref{THM:benveniste}. Condition \eqref{cond_Lipshitz} follows from the linearity of $S_{\boldsymbol{z}}(\boldsymbol{w})$ with respect to $\boldsymbol{w}$. Let $\boldsymbol{w}' \in \mathbb{R}^d$ with $\|\boldsymbol{w}'\| = 1$. Condition \eqref{cond_S} is met since
\begin{align*}
  \|S_{\boldsymbol{z}}(\boldsymbol{w})\| &\le 2e_{\max}\phi_{\max} \|\boldsymbol{w} - \boldsymbol{w}'\| +  \|S_{\boldsymbol{z}}(\boldsymbol{w}')\|\nonumber \\
   &\le 2e_{\max}\phi_{\max} \left(\|\boldsymbol{w}\|+1\right) +  r_{\max}e_{\max} + 2e_{\max}\phi_{\max}  \nonumber \\
   &\le 4e_{\max}\phi_{\max} \left(\|\boldsymbol{w}\|+1\right) +  r_{\max}e_{\max} \left(\|\boldsymbol{w}\|+1\right)  \nonumber \\
    &\le (4e_{\max}\phi_{\max} + r_{\max}e_{\max})\left(\|\boldsymbol{w}\|+1\right).  \nonumber
\end{align*}
Similarly, the condition \eqref{cond_R} is met as we have, for all $n \in \mathbb{N}$,
\begin{align*}
   \|R_{n}(\boldsymbol{w}, \boldsymbol{z})\| &\le (4e_{\max}\phi_{\max} + r_{\max}e_{\max})\left(\|\boldsymbol{w}\|+1\right).
\end{align*}
Since $\mathcal{X}$ is finite and $\|\boldsymbol{e}\| \le e_{\max}$ for any realized eligibility trace $\boldsymbol{e}$, there exists $B_{\boldsymbol{z}} \in \mathbb{R}_{>0}$ such that
$
\sup \|\boldsymbol{z}_n\|^q \le B_{\boldsymbol{z}}^q
$
for all $q > 0$. Therefore, condition \eqref{cond_z} is met because
$$
\mathbb{E}\left(1 + \|\boldsymbol{z}_n\|^q \mid \boldsymbol{z}_1  = \boldsymbol{z}, \boldsymbol{w}_1 = \boldsymbol{w}\right) \le 1 + B_{\boldsymbol{z}}^q \le (1 + B_{\boldsymbol{z}}^q)(1 + \|\boldsymbol{z}\|^q).
$$
for all $q > 0$. We now validate conditions \eqref{cond_nu_exist}, \eqref{cond_nu}, and \eqref{cond_pi_nu}. To this end, let us introduce 
\begin{align*}
    \nu_{\boldsymbol{w}}(\boldsymbol{z}) &:= \sum_{k=0}^\infty \{\Pi^k S_{\boldsymbol{z}}(\boldsymbol{w}) - S(\boldsymbol{w})\} \\&= \sum_{k=0}^\infty \left\{\mathbb{E} \{S_{\boldsymbol{z}_{k+1}}(\boldsymbol{w})|\boldsymbol{z}_1 = \boldsymbol{z}\} - S(\boldsymbol{w})\right\} \\&= \sum_{k=0}^\infty \left\{\mathbb{E}(\boldsymbol{b}_{k+1}+\boldsymbol{A}_{k+1}\boldsymbol{w}|\boldsymbol{z}_1 = \boldsymbol{z}) - (\boldsymbol{b}+\boldsymbol{A}\boldsymbol{w})\right\} \\
     &= \sum_{k=0}^\infty \left\{\mathbb{E}(\boldsymbol{b}_{k+1}|\boldsymbol{z}_1 = \boldsymbol{z}) - \boldsymbol{b}\right\} + \sum_{k=0}^\infty \left\{\mathbb{E}(\boldsymbol{A}_{k+1}|\boldsymbol{z}_1 = \boldsymbol{z}) - \boldsymbol{A}\right\}\boldsymbol{w}.
\end{align*}
Since the series is absolutely convergent by Lemma 6.7 of \citep{bertsekas1996neuro}, the infinite sum is well defined. We first verify that $\nu_{\boldsymbol{w}}$ solves the Poisson equation given in \eqref{cond_nu_exist}. Note that
\begin{align*}
    \Pi\nu_{\boldsymbol{w}}(\boldsymbol{z}) &= \sum_{k=0}^\infty \Pi\left\{\Pi^k S_{\boldsymbol{z}}(\boldsymbol{w}) - S(\boldsymbol{w})\right\} \\
    &= \sum_{k=0}^\infty \left\{\Pi^{k+1} S_{\boldsymbol{z}}(\boldsymbol{w}) - \Pi S(\boldsymbol{w})\right\} \\
    &= \sum_{k=0}^\infty \left\{\Pi^{k+1} S_{\boldsymbol{z}}(\boldsymbol{w}) - S(\boldsymbol{w})\right\} &&\text{($S(\boldsymbol{w})$ is constant)}\\
    &= \sum_{k=1}^\infty \left\{\mathbb{E}(\boldsymbol{b}_{k+1}|\boldsymbol{z}_1 = \boldsymbol{z}) - \boldsymbol{b}\right\} + \sum_{k=1}^\infty \left\{\mathbb{E}(\boldsymbol{A}_{k+1}|\boldsymbol{z}_1 = \boldsymbol{z}) - \boldsymbol{A}\right\}\boldsymbol{w}.
\end{align*}
Therefore, we get condition \eqref{cond_nu_exist} satisfied:
$$
\nu_{\boldsymbol{w}}(\boldsymbol{z}) - \Pi \nu_{\boldsymbol{w}}(\boldsymbol{z}) = \mathbb{E}(\boldsymbol{b}_{1}|\boldsymbol{z}_1 = \boldsymbol{z}) - \boldsymbol{b} + \mathbb{E}(\boldsymbol{A}_{1}|\boldsymbol{z}_1 = \boldsymbol{z})\boldsymbol{w} - \boldsymbol{A}\boldsymbol{w} = S_{\boldsymbol{z}}(\boldsymbol{w}) - S(\boldsymbol{w}).
$$
Next, we show condition \eqref{cond_nu}:
\begin{align*}
    \|\nu_{\boldsymbol{w}}(\boldsymbol{z})\|
     &\le \sum_{k=0}^\infty \left\|\mathbb{E}(\boldsymbol{b}_{k+1}|\boldsymbol{z}_1 = \boldsymbol{z}) - \boldsymbol{b}\right\| +  \|\boldsymbol{w}\| \sum_{k=0}^\infty \left\|\mathbb{E}(\boldsymbol{A}_{k+1}|\boldsymbol{z}_1 = \boldsymbol{z}) - \boldsymbol{A}\right\|\\
     &\le\sum_{k=0}^\infty \tilde m \tilde \rho^k +  \|\boldsymbol{w}\| \sum_{k=0}^\infty \tilde m \tilde \rho^k \quad \text{for} ~~\tilde\rho \in (0,1), ~\tilde m \in \mathbb{R}_{>0} &&\text{(Lemma 6.7 of \citep{bertsekas1996neuro})} \\
     &= \frac{\tilde{m}(1 + \|\boldsymbol{w}\|)}{1-\tilde \rho}.
\end{align*}
Finally, we verify condition \eqref{cond_pi_nu}:
\begin{align*}
    \|\Pi\nu_{\boldsymbol{w}}(\boldsymbol{z}) - \Pi\nu_{\boldsymbol{w}'}(z)\|
     &\le   \|\boldsymbol{w}-\boldsymbol{w}'\| \sum_{k=0}^\infty \left\|\mathbb{E}(\boldsymbol{A}_{k+1}|\boldsymbol{z}_1 = \boldsymbol{z}) - A\right\|\\
     &\le \|\boldsymbol{w}-\boldsymbol{w}'\| \sum_{k=0}^\infty \tilde m \tilde \rho^k \quad \text{for} ~~\tilde\rho \in (0,1), ~\tilde m \in \mathbb{R}_{>0} &&\text{(Lemma 6.7 of \citep{bertsekas1996neuro})} \\
     &= \frac{\tilde{m}}{1-\tilde \rho}\|\boldsymbol{w}-\boldsymbol{w}'\|.
\end{align*}
Finally, consider the function $U(\boldsymbol{w}) := \frac{1}{2}\|\boldsymbol{w}-\boldsymbol{w}_{\star}\|^2$ and note that $S(\boldsymbol{w}) = \boldsymbol{A}(\boldsymbol{w}-\boldsymbol{w}_{\star})$. 
\begin{itemize}
    \item From Lemma 9 of \citep{tsitsiklis1996analysis}, we have the negative-definiteness of $A$. Therefore, we have $\langle \nabla U(\boldsymbol{w}), S(\boldsymbol{w}) \rangle = (\boldsymbol{w}-\boldsymbol{w}_{\star})^{\top} \boldsymbol{A}(\boldsymbol{w}-\boldsymbol{w}_{\star})$ < 0.
    \item From the reverse triangle inequality, we know
    $\|\boldsymbol{w}-\boldsymbol{w}_{\star}\| \ge \|\boldsymbol{w}\|-\|\boldsymbol{w}_{\star}\| \ge \frac{1}{2}\|\boldsymbol{w}\|$ for $w$ such that $\|\boldsymbol{w}\| \ge 2\|\boldsymbol{w}_{\star}\|$. Then, 
    $$
    U(\boldsymbol{w}) = \frac{1}{2}\|\boldsymbol{w}-\boldsymbol{w}_{\star}\|^2 \ge \frac{1}{8}\|\boldsymbol{w}\|^2.
    $$
    \item By construction, $U(\boldsymbol{w}) = 0$ if and only if $\boldsymbol{w} = \boldsymbol{w}_{\star}$.
    \end{itemize}
By Theorem \ref{THM:benveniste}, we get the almost sure convergence of implicit TD($\lambda$) iterates.
\end{proof}

\subsection{Finite-time analysis for implicit TD(0)/TD($\lambda$)}
\noindent In this section, we establish a finite-time MSE bound for the implicit TD algorithms. To this end, we review notations which will be used in this section. 
\begin{itemize}
    \item $\xi_{n}(\boldsymbol{w}):=\left\{S_{\boldsymbol{z}_n}(\boldsymbol{w})-S(\boldsymbol{w})\right\}^{\top}\left(\boldsymbol{w}-\boldsymbol{w}_{\star}\right), ~~ \forall \boldsymbol{w}\in \mathbb{R}^{d}$
    \item $\boldsymbol{\Gamma}:=\sum_{\boldsymbol{x} \in \mathcal{X}} \mu_{\pi}(\boldsymbol{x}) \phi(\boldsymbol{x}) \phi(\boldsymbol{x})^{T} = \boldsymbol{\Phi}^{\top} \boldsymbol{D} \boldsymbol{\Phi}, \quad \boldsymbol{D} :=\operatorname{diag}\left\{\mu_{\pi}(\boldsymbol{x}):\boldsymbol{x} \in \mathcal{X} \right\}$
    \item $\min\{\text{eig}(\boldsymbol{\Gamma})\}=\lambda_{\text{min}}$, $\max\{\text{eig}(\boldsymbol{\Gamma})\}=\lambda_{\text{max}}$  
    \item $V_{\boldsymbol{w}_{\star}}(\boldsymbol{x}):=\phi(\boldsymbol{x})^{\top} \boldsymbol{w}_{\star}$,$~~ \forall \boldsymbol{x} \in \mathcal{X}$
    \item $\left\|V_{\boldsymbol{w}}-V_{\boldsymbol{w}'}\right\|_{\boldsymbol{D}}=\left\|\boldsymbol{w}-\boldsymbol{w}'\right\|_{\boldsymbol{\Gamma}} \text{where} ~~\|\boldsymbol{u}\|_{\boldsymbol{Q}} := \sqrt{\boldsymbol{u}^{\top}\boldsymbol{Q}\boldsymbol{u}}$
\end{itemize}

\noindent We first relate the value function difference to that of the parameter difference.

\begin{lemma}\label{APPEND_LEMMA:BHAND1}
For all $\boldsymbol{w}, \boldsymbol{w}' \in \mathbb{R}^{d}$,
$$
\sqrt{\lambda_{\min}}\left\|\boldsymbol{w}-\boldsymbol{w}'\right\| \le\left\|V_{\boldsymbol{w}}-V_{\boldsymbol{w}'}\right\|_{\boldsymbol{D}} \le \phi_{\max}\left\|\boldsymbol{w}-\boldsymbol{w}'\right\|.
$$
\end{lemma}
\begin{proof}
Note that 
$$
\left\|V_{\boldsymbol{w}}-V_{\boldsymbol{w}'}\right\|_{\boldsymbol{D}}=\sqrt{\sum_{\boldsymbol{x} \in \mathcal{X}} \mu_\pi(\boldsymbol{x})\left(\phi(\boldsymbol{x})^{\top}\left(\boldsymbol{w}-\boldsymbol{w}'\right)\right)^2}=\left\{\left(\boldsymbol{w}-\boldsymbol{w}'\right)^{\top} \boldsymbol{\Gamma} \left(\boldsymbol{w}-\boldsymbol{w}'\right)\right\}^{1/2}.
$$
By the definition of $\boldsymbol{\Gamma} $,
$$
\lambda_{\max }(\boldsymbol{\Gamma})=\lambda_{\text {max}} \left\{\sum_{\boldsymbol{x} \in \mathcal{X}} \mu_{\pi}(\boldsymbol{x}) \phi(\boldsymbol{x}) \phi(\boldsymbol{x})^{\top}\right\} \le \sum_{\boldsymbol{x} \in \mathcal{X}} \mu_{\pi}(\boldsymbol{x}) \lambda_{\max }\left\{\phi(\boldsymbol{x}) \phi(\boldsymbol{x})^{\top}\right\} \le \phi_{\max}^2\sum_{\boldsymbol{x} \in \mathcal{X}} \mu_{\pi}(\boldsymbol{x}) = \phi_{\max}^2.
$$
Therefore, we have 
$$
(\boldsymbol{w}-\boldsymbol{w}')^{\top} \boldsymbol{\Gamma} (\boldsymbol{w}-\boldsymbol{w}') \le \phi_{\max}^2(\boldsymbol{w}-\boldsymbol{w}')^{\top} (\boldsymbol{w}-\boldsymbol{w}').
$$
The lower bound of $\|V_{\boldsymbol{w}}-V_{\boldsymbol{w}'}\|_{\boldsymbol{D}}$ comes from the definition
$\lambda_{\min }=\min_{\boldsymbol{u}} \frac{\boldsymbol{u}^{\top} \boldsymbol{\Gamma} \boldsymbol{u}}{\|\boldsymbol{u}\|^2}$. Plugging in $\boldsymbol{u} = \boldsymbol{w}-\boldsymbol{w}'$, we get the lower bound.
\end{proof}

\subsubsection{Finite-time analysis for implicit TD(0) with projection}
\noindent In this subsection, we present a finite-time MSE bound for projected implicit TD(0). Our approach utilizes results from \citep{bhandari2018finite}, and accounts for the data-adaptive step size used in implicit TD algorithms. To ensure clarity and completeness, we restate theorems from \citep{bhandari2018finite}. An upshot of our result is that an implicit update will yield a finite-time MSE bound independent of the initial step size. We first list results from \citep{bhandari2018finite}, which will be used in establishing finite-time MSE bounds for the projected implicit TD(0) algorithm.

\begin{lemma}[Lemma 3 of \citet{bhandari2018finite}]\label{APPEND_LEMMA:BHAND3}
    For any $\boldsymbol{w} \in \mathbb{R}^{d}$,
\[
\left(\boldsymbol{w}_{\star}-\boldsymbol{w}\right)^{\top} S(\boldsymbol{w}) \ge(1-\gamma)\left\|V_{\boldsymbol{w}_{\star}}-V_{\boldsymbol{w}}\right\|_{\boldsymbol{D}}^2 \ge 0
\]
\end{lemma}

\begin{lemma}[Lemma 6 of \citet{bhandari2018finite}]\label{APPEND_LEMMA:BHAND6}
For all $\boldsymbol{w} \in \{\tilde{\boldsymbol{w}}: \|\tilde{\boldsymbol{w}}\| \le R\}$ and $i \in \mathbb{N}$, 
$$
\left\|S_{\boldsymbol{z}_i}(\boldsymbol{w})\right\| \le G:=r_{\max}\phi_{\max}+2\phi_{\max}^2 R
$$
with probability one.
\end{lemma}

\begin{lemma}[Lemma 9 of \citet{bhandari2018finite}]\label{APPEND_LEMMA:BHAND9} Consider two random variables $\boldsymbol{u}$ and $\tilde{\boldsymbol{u}}$ such that
\[
\boldsymbol{u} \rightarrow \boldsymbol{x}_{n} \rightarrow \boldsymbol{x}_{n+\tau} \rightarrow \tilde{\boldsymbol{u}}
\]
for some fixed $n \in \mathbb{N}$ and $\tau \in \mathbb{N}$. Assume the Markov chain mixes as stated in Expression \ref{GEOM_MIX}. Let $\boldsymbol{u}'$ and $\tilde{\boldsymbol{u}}'$ be independent copies drawn from the marginal distributions of $\boldsymbol{u}$ and $\tilde{\boldsymbol{u}}$. Then, for any bounded function $h$,
\[
\left|\mathbb{E}_{\mu_{\pi}}\left\{h(\boldsymbol{u}, \tilde{\boldsymbol{u}})\right\} - \mathbb{E}\left\{h(\boldsymbol{u}', \tilde{\boldsymbol{u}}')\right\}\right| \le 2 \|h\|_\infty m \rho^\tau
\]
for some $m > 0$, $\rho \in (0,1)$.
\end{lemma}

\begin{lemma}[Lemma 10 of \citet{bhandari2018finite}]\label{APPEND_LEMMA:BHAND10}
     With probability one, for all $\boldsymbol{w}, \boldsymbol{v} \in \{\boldsymbol{w}': \|\boldsymbol{w}'\| \le R\}$ and $i \in \mathbb{N}$,
     \begin{align*}
        \left|\xi_{i}(\boldsymbol{w}) \right| &\le 4GR \\
        \left|\xi_{i}(\boldsymbol{w})-\xi_{i}\left(\boldsymbol{v}\right)\right| &\lesssim \left\|\boldsymbol{w}-\boldsymbol{v}\right\|
     \end{align*}
     where $\xi_i(\boldsymbol{w}) = (S_{\boldsymbol{z}_i}(\boldsymbol{w}) - S(\boldsymbol{w}))^{\top}(\boldsymbol{w}-\boldsymbol{w}_{\star})$.
\end{lemma}

\begin{lemma}\label{APPEND_LEMMA:TD0_FIN_PROJ_ITER} 
For every $n \in \mathbb{N}$,
$$
\left\|\boldsymbol{w}_{\star}-\boldsymbol{w}^{\text{im}}_{n+1}\right\|^2 
\le \|\boldsymbol{w}_{\star} - \boldsymbol{w}^{\text{im}}_{n}\|^2 - \frac{2\alpha_n(1 - \gamma)}{1+\alpha_n\phi_{\max}^2}\left\|V_{\boldsymbol{w}_{\star}}-V_{\boldsymbol{w}^{\text{im}}_{n}}\right\|_{\boldsymbol{D}}^2 + 2\tilde \alpha_n \xi_n(\boldsymbol{w}^{\text{im}}_{n}) + \alpha_n^2 G^2
$$
holds with probability one.
\end{lemma}
\begin{proof}
\begin{align}
\|\boldsymbol{w}_{\star} - \boldsymbol{w}^{\text{im}}_{n+1}\|^2 
&= \|\Pi_{R}(\boldsymbol{w}_{\star}) - \Pi_{R}\{\boldsymbol{w}^{\text{im}}_n + \tilde \alpha_n S_{\boldsymbol{z}_n}(\boldsymbol{w}^{\text{im}}_{n})\}\|^2 \label{LEMMA:3-24-2}\\
&\le \|\boldsymbol{w}_{\star} - \boldsymbol{w}^{\text{im}}_n - \tilde \alpha_n S_{\boldsymbol{z}_n}(\boldsymbol{w}^{\text{im}}_{n})\|^2 \label{LEMMA:3-24-3}\\
&= \|\boldsymbol{w}_{\star} - \boldsymbol{w}^{\text{im}}_{n}\|^2 - 2\tilde \alpha_n S_{\boldsymbol{z}_n}(\boldsymbol{w}^{\text{im}}_{n})^{\top} (\boldsymbol{w}_{\star} - \boldsymbol{w}^{\text{im}}_{n}) +  \|\tilde \alpha_n S_{\boldsymbol{z}_n}(\boldsymbol{w}^{\text{im}}_{n})\|^2 \nonumber \\
&\le \|\boldsymbol{w}_{\star} - \boldsymbol{w}^{\text{im}}_{n}\|^2 - 2\tilde \alpha_n S_{\boldsymbol{z}_n}(\boldsymbol{w}^{\text{im}}_{n})^{\top} (\boldsymbol{w}_{\star} - \boldsymbol{w}^{\text{im}}_{n}) + \alpha_n^2 G^2 \label{LEMMA:3-24-5}\\
&= \|\boldsymbol{w}_{\star} - \boldsymbol{w}^{\text{im}}_{n}\|^2 - 2\tilde \alpha_n S(\boldsymbol{w}^{\text{im}}_{n})^{\top} (\boldsymbol{w}_{\star} - \boldsymbol{w}^{\text{im}}_{n}) + 2\tilde \alpha_n \xi_n(\boldsymbol{w}^{\text{im}}_{n}) + \alpha_n^2 G^2 \nonumber \\
&\le \|\boldsymbol{w}_{\star} - \boldsymbol{w}^{\text{im}}_{n}\|^2 - 2\tilde \alpha_n(1 - \gamma)\left\|V_{\boldsymbol{w}_{\star}}-V_{\boldsymbol{w}^{\text{im}}_{n}}\right\|_{\boldsymbol{D}}^2 + 2\tilde \alpha_n \xi_n(\boldsymbol{w}^{\text{im}}_{n}) + \alpha_n^2 G^2 \label{LEMMA:3-24-7} \\
&\le \|\boldsymbol{w}_{\star} - \boldsymbol{w}^{\text{im}}_{n}\|^2 - \frac{2\alpha_n(1 - \gamma)}{1+\alpha_n \phi^2_{\max}}\left\|V_{\boldsymbol{w}_{\star}}-V_{\boldsymbol{w}^{\text{im}}_{n}}\right\|_{\boldsymbol{D}}^2 + 2\tilde \alpha_n \xi_n(\boldsymbol{w}^{\text{im}}_{n}) + \alpha_n^2 G^2 \label{LEMMA:3-24-8}
\end{align}
where \eqref{LEMMA:3-24-2} is due to the fact that $\boldsymbol{w}_{\star} = \Pi_{R}(\boldsymbol{w}_{\star})$, \eqref{LEMMA:3-24-3} is thanks to non-expansiveness of the projection operator on the convex set, 
\eqref{LEMMA:3-24-5} comes from the fact $\tilde \alpha_n \le \alpha_n$ with Lemma \ref{APPEND_LEMMA:BHAND6} and \eqref{LEMMA:3-24-7} is by Lemma \ref{APPEND_LEMMA:BHAND3}. Finally, the last inequality is a direct consequence of the Lemma \ref{APPEND_LEMMA:IMP_STEP}.
\end{proof}

\begin{lemma}\label{APPEND_LEMMA:RANDOM_STOCHASTIC_BOUND}
Given a non-increasing sequence $(\alpha_{n})_{n \in \mathbb{N}}$, for any fixed $i < n$, we have
\begin{align*}
&\mathbb{E}_{\mu_{\pi_\star}}\left\{\xi_{i}\left(\boldsymbol{w}^{\text{im}}_{i}\right)\right\} \lesssim \tau_{\alpha_n}  \quad \text{for}~i \le \tau_{\alpha_n}\\
&\mathbb{E}_{\mu_{\pi_\star}}\left\{\xi_{i}\left(\boldsymbol{w}^{\text{im}}_{i}\right)\right\} \lesssim  \tau_{\alpha_n} \alpha_{i-\tau_{\alpha_n}} \quad \text{for}~i > \tau_{\alpha_n}.
\end{align*}
\end{lemma}
\begin{proof}
We first establish a bound on $\mathbb{E}\left\{ \xi_{i}\left(\boldsymbol{w}^{\text{im}}_{i}\right)\right\}$. To this end, recall from Lemma \ref{APPEND_LEMMA:BHAND10} that 
\begin{equation}\label{xi_recur_triangle}
\xi_i(\boldsymbol{w}^{\text{im}}_{i}) \lesssim \xi_i(\boldsymbol{w}^{\text{im}}_{i-1}) + \|\boldsymbol{w}^{\text{im}}_{i} - \boldsymbol{w}^{\text{im}}_{i-1}\|. 
\end{equation}
For $i > \tau_{\alpha_n}$, repeated application of \eqref{xi_recur_triangle} gives us
\begin{align*}
     \xi_{i}\left(\boldsymbol{w}^{\text{im}}_{i}\right)
     & \lesssim  \xi_{i}\left(\boldsymbol{w}^{\text{im}}_{i-2}\right) + \left\|\boldsymbol{w}^{\text{im}}_{i-1}-\boldsymbol{w}^{\text{im}}_{i-2}\right\| + \left\|\boldsymbol{w}^{\text{im}}_{i}-\boldsymbol{w}^{\text{im}}_{i-1}\right\| \\
    & \lesssim  \xi_{i}\left(\boldsymbol{w}^{\text{im}}_{i-\tau_{\alpha_n}}\right)+  \sum_{k=i-\tau_{\alpha_n}}^{i-1}\left\|\boldsymbol{w}^{\text{im}}_{k+1}-\boldsymbol{w}^{\text{im}}_{k}\right\|.
\end{align*}
Note that 
$$ 
\left\|\boldsymbol{w}^{\text{im}}_{k+1}-\boldsymbol{w}^{\text{im}}_{k}\right\| = \left\|\Pi_{R}\{\boldsymbol{w}^{\text{im}}_{k}+ \tilde \alpha_k S_{\boldsymbol{z}_k}(\boldsymbol{w}^{\text{im}}_{k})\}-\Pi_{R}(\boldsymbol{w}^{\text{im}}_{k})\right\| \le 
\left\|\boldsymbol{w}^{\text{im}}_{k} + \tilde \alpha_k S_{\boldsymbol{z}_k}(\boldsymbol{w}^{\text{im}}_{k})-\boldsymbol{w}^{\text{im}}_{k}\right\| \le \alpha_k G
$$
where in the first inequality, we have used the non-expansiveness of the projection operator, and
in the second inequality, both Lemma \ref{APPEND_LEMMA:IMP_STEP} and \ref{APPEND_LEMMA:BHAND6} were used. Therefore, we have
\begin{align}
     \xi_{i}\left(\boldsymbol{w}^{\text{im}}_{i}\right) 
     & \lesssim \xi_{i}\left(\boldsymbol{w}^{\text{im}}_{i-\tau_{\alpha_n}}\right) + \sum_{k=i-\tau_{\alpha_n}}^{i-1} {\alpha}_{k} \label{LEMMA3_25_SUM_BOUND}\\
    & \lesssim \xi_{i}\left(\boldsymbol{w}^{\text{im}}_{i-\tau_{\alpha_n}}\right) + \tau_{\alpha_n} \alpha_{i-\tau_{\alpha_n}} \label{LEMMA3_25_NSUM_BOUND}
\end{align}
where \eqref{LEMMA3_25_NSUM_BOUND} follows from non-increasingness of $(\alpha_n)_{n \in \mathbb{N}}$. 

We now establish an upper bound of an expectation of the term in \eqref{LEMMA3_25_NSUM_BOUND}. To this end, recall that $\xi_i(\boldsymbol{w}) = \left\{S_{\boldsymbol{z}_i}(\boldsymbol{w})-S(\boldsymbol{w})\right\}^{\top} (\boldsymbol{w}-\boldsymbol{w}_{\star})$ is as a function of $\boldsymbol{z}_i = \{\boldsymbol{x}_i, \boldsymbol{x}_{i+1}\}$ and $\boldsymbol{w}$. First, notice that $\{\boldsymbol{z}_i\}_{i \in \mathbb{N}}$ is a Markov process with the same transition probability as $\{\boldsymbol{x}_i\}_{i \in \mathbb{N}}$. Second, we can view $\boldsymbol{w}^{\text{im}}_{i-\tau_{\alpha_n}}$ as a function of $\{\boldsymbol{z}_1, \cdots, \boldsymbol{z}_{i-\tau_{\alpha_n}-1}\}$. Now consider $\xi_i(\boldsymbol{w}^{\text{im}}_{i-\tau_{\alpha_n}})$, which is a function of both $\boldsymbol{u} = \{\boldsymbol{z}_1, \cdots, \boldsymbol{z}_{i-\tau_{\alpha_n}-1}\}$ and $\tilde{\boldsymbol{u}} = \boldsymbol{z}_{i}$. Set $h(\boldsymbol{u}, \tilde{\boldsymbol{u}}) = \xi_i (\boldsymbol{w}^{\text{im}}_{i-\tau_{\alpha_n}})$ and note the condition for Lemma \ref{APPEND_LEMMA:BHAND9} is met since $\boldsymbol{u} =\{\boldsymbol{z}_1, \cdots, \boldsymbol{z}_{i-\tau_{\alpha_n}-1}\} \to \boldsymbol{x}_{i-\tau_{\alpha_n}} \to \boldsymbol{x}_i = \tilde{\boldsymbol{u}}$ forms a Markov chain. From Lemma \ref{APPEND_LEMMA:BHAND9}, we get
$$
\mathbb{E}_{\mu_{\pi_\star}} \left\{h(\boldsymbol{u}, \tilde{\boldsymbol{u}}) \right\} - \mathbb{E} \left\{h(\boldsymbol{u}', \tilde{\boldsymbol{u}}') \right\} \le 2\|h\|_\infty m\rho ^{\tau_{\alpha_n}},
$$
where $\boldsymbol{u}'$ 
and $\tilde{\boldsymbol{u}}'$ 
are independent samples drawn from the marginal distributions of $\boldsymbol{u}$ and $\tilde{\boldsymbol{u}}$, respectively. Let us denote the $(i-\tau_{\alpha_n})^{\text{th}}$ implicit TD(0) iterate computed using $\boldsymbol{u}'$ as $\boldsymbol{w}'_{i-\tau_{\alpha_n}}$. Conditioning on $\boldsymbol{u}'$, we get
$$
\mathbb{E}_{\mu_{\pi_\star}} \left\{h(\boldsymbol{u}', \tilde{\boldsymbol{u}}') \right\} = \mathbb{E} \left [\mathbb{E} \left\{ \xi_{i}\left(\boldsymbol{w}'_{i-\tau_{\alpha_n}}\right)\mid \boldsymbol{u}' \right\} \right] = 0,
$$
since
$
\mathbb{E}_{\mu_{\pi_\star}}\left\{\xi_i(\boldsymbol{w})\right\} = 0
$
for any fixed $w$. Combined with Lemma \ref{APPEND_LEMMA:BHAND10}, we have
$$
\mathbb{E}_{\mu_{\pi_\star}} \left\{ \xi_{i}\left(\boldsymbol{w}^{\text{im}}_{i-\tau_{\alpha_n}}\right)\right\} \lesssim m\rho ^{\tau_{\alpha_n}}.
$$
Taking the expectation of \eqref{LEMMA3_25_NSUM_BOUND} with respect to the stationary distribution, we get
\begin{align*}
\mathbb{E}_{\mu_{\pi_\star}}\{\xi_{i}\left(\boldsymbol{w}^{\text{im}}_{i}\right)\} &\lesssim \mathbb{E}_{\mu_{\pi_\star}}\left\{ \xi_{i}\left(\boldsymbol{w}^{\text{im}}_{i-\tau_{\alpha_n}}\right)\right\}+ \tau_{\alpha_n} \alpha_{i-\tau_{\alpha_n}}\\ &\lesssim m\rho ^{\tau_{\alpha_n}} + \tau_{\alpha_n} \alpha_{i-\tau_{\alpha_n}} \nonumber \\&\lesssim \alpha_n + \tau_{\alpha_n} \alpha_{i-\tau_{\alpha_n}}
\\&\le \tau_{\alpha_n} \alpha_{i-\tau_{\alpha_n}}.
\end{align*}
Next we consider when $i \le \tau_{\alpha_{n}}$. Repeated application of \eqref{xi_recur_triangle} gives us
$$
\xi_{i}\left(\boldsymbol{w}^{\text{im}}_{i}\right) 
\lesssim \xi_{i}\left(\boldsymbol{w}^{\text{im}}_{1}\right)+ \tau_{\alpha_{n}}.
$$
Taking the expectation with respect to the steady-state distribution, we get
$$
\mathbb{E}_{\mu_{\pi_\star}}\left\{\xi_{i}\left(\boldsymbol{w}^{\text{im}}_{i}\right)\right\} \lesssim \mathbb{E}_{\mu_{\pi_\star}}\left\{\xi_{i}\left(\boldsymbol{w}^{\text{im}}_{1}\right)\right\} + \tau_{\alpha_{n}}.
$$
Since
$
\mathbb{E}_{\mu_{\pi_\star}}\left\{\xi_i(\boldsymbol{w})\right\} = 0
$
for any fixed $w$, we get
$
\mathbb{E}_{\mu_{\pi_\star}}\left\{\xi_{i}\left(\boldsymbol{w}^{\text{im}}_{i}\right)\right\} \lesssim \tau_{\alpha_n}.
$
\end{proof}

\noindent We now establish a finite-time MSE bound for implicit TD(0).\\
\begin{proof}[Proof of Theorem \ref{THM:FIN_PROJ_TD0}]
Starting from Lemma \ref{APPEND_LEMMA:TD0_FIN_PROJ_ITER}, we have
\begin{align}
\mathbb{E}_{\mu_{\pi_\star}}\left\|\boldsymbol{w}_{\star}-\boldsymbol{w}^{\text{im}}_{n+1}\right\|^2
&\le \mathbb{E}_{\mu_{\pi_\star}}\|\boldsymbol{w}_{\star} - \boldsymbol{w}^{\text{im}}_{n}\|^2 - \frac{2\alpha_n(1 - \gamma)}{1+\alpha_n \phi_{\max}^2}\mathbb{E}_{\mu_{\pi_\star}}\left\|V_{\boldsymbol{w}_{\star}}-V_{\boldsymbol{w}^{\text{im}}_{n}}\right\|_{\boldsymbol{D}}^2 + 2\mathbb{E}_{\mu_{\pi_\star}}\left\{\tilde \alpha_n \xi_n(\boldsymbol{w}^{\text{im}}_{n})\right\} + \alpha_n^2 G^2 \nonumber\\
&\le \mathbb{E}_{\mu_{\pi_\star}}\|\boldsymbol{w}_{\star} - \boldsymbol{w}^{\text{im}}_{n}\|^2 -\frac{2\alpha_n(1-\gamma)\lambda_{\min }}{1+\alpha_n\phi^2_{\max}}\mathbb{E}_{\mu_{\pi_\star}}\left\|\boldsymbol{w}_{\star}-\boldsymbol{w}^{\text{im}}_{n}\right\|^2 + 2\mathbb{E}_{\mu_{\pi_\star}}\left\{\tilde \alpha_n \xi_n(\boldsymbol{w}^{\text{im}}_{n})\right\} + \alpha_n^2 G^2 \nonumber\\
& =\left\{1-\frac{2\alpha_n(1-\gamma)\lambda_{\min }}{1+\alpha_n\phi_{\max}^2}\right\} \mathbb{E}_{\mu_{\pi_\star}}\left\|\boldsymbol{w}_{\star}-\boldsymbol{w}^{\text{im}}_{n}\right\|^2+ 2\mathbb{E}_{\mu_{\pi_\star}}\left\{\tilde \alpha_n \xi_n(\boldsymbol{w}^{\text{im}}_{n})\right\} + \alpha_n^2 G^2 \nonumber
\end{align}
where the second inequality follows from Lemma \ref{APPEND_LEMMA:BHAND1}. Applying Lemma \ref{APPEND_LEMMA:BHAND10}, together with $|\tilde\alpha_n - \alpha_n| \le \alpha_n^2\phi^2_{\max}$, we obtain
\begin{align}
    \mathbb{E}_{\mu_{\pi_\star}}\left\|\boldsymbol{w}_{\star}-\boldsymbol{w}^{\text{im}}_{n+1}\right\|^2 
    &= \left\{1-\frac{2\alpha_n(1-\gamma)\lambda_{\min }}{1+\alpha_n\phi_{\max}^2}\right\} \mathbb{E}_{\mu_{\pi_\star}}\left\|\boldsymbol{w}_{\star}-\boldsymbol{w}^{\text{im}}_{n}\right\|^2 + 2\alpha_n\mathbb{E}_{\mu_{\pi_\star}}\left\{ \xi_n(\boldsymbol{w}^{\text{im}}_{n})\right\} + \mathcal{O}(\alpha_n^2).\label{FIN_TD0_RECUR}
\end{align}

\noindent Note that one can show that
\begin{equation*}
1-\frac{2\alpha_n(1-\gamma)\lambda_{\min }}{1+\alpha_n\phi_{\max}^2} \le \frac{1}{1+\eta\alpha_n}
\end{equation*}
where $\eta$ is given by
$$
\eta = \frac{2(1-\gamma)\lambda_{\min}}{1+\mathbb{I}\{\phi_{\max}^2 > 2(1-\gamma)\lambda_{\min}\}(\phi_{\max}^2 - 2(1-\gamma)\lambda_{\min})\alpha_1}.
$$
Since $\frac{1}{1+x} \le \exp\left(\frac{-x}{1+x}\right)$, we have
$
\frac{1}{1+\eta\alpha_n} \le \exp\left(\frac{-\eta\alpha_n}{1+\eta\alpha_n}\right).
$
Continuing from \eqref{FIN_TD0_RECUR}, 
we have
\begin{align}
    \mathbb{E}_{\mu_{\pi_\star}}\left\|\boldsymbol{w}_{\star}-\boldsymbol{w}^{\text{im}}_{n+1}\right\|^2
    \le  \exp\left(\frac{-\eta\alpha_n}{1+\eta\alpha_n}\right)\mathbb{E}_{\mu_{\pi_\star}}\left\|\boldsymbol{w}_{\star}-\boldsymbol{w}^{\text{im}}_{n}\right\|^2 + 2\alpha_n\mathbb{E}_{\mu_{\pi_\star}}\left\{ \xi_n(\boldsymbol{w}^{\text{im}}_{n})\right\}+  \mathcal{O}(\alpha_n^2). \label{FIN_UP_TD0_RECUR}
\end{align}
Repeated applications of \eqref{FIN_UP_TD0_RECUR} with Lemma \ref{APPEND_LEMMA:RANDOM_STOCHASTIC_BOUND} give us
\begin{align}
\mathbb{E}_{\mu_{\pi_\star}}\left\|\boldsymbol{w}_{\star}-\boldsymbol{w}^{\text{im}}_{n+1}\right\|^2 & \lesssim \left\{\prod_{i=1}^n  \exp\left(\frac{-\eta\alpha_i}{1+\eta\alpha_i}\right)\right\}\left\|\boldsymbol{w}_{\star}-\boldsymbol{w}^{\text{im}}_{1}\right\|^2 + \sum_{i=1}^n\left\{\prod_{k=i+1}^n \exp\left(\frac{-\eta\alpha_k}{1+\eta\alpha_k}\right)\right\} \alpha_i^2 \nonumber \\
& \quad +  \tau_{\alpha_n}\sum_{i=1}^{\tau_{\alpha_n}}\left\{\prod_{k=i+1}^n \exp\left(\frac{-\eta\alpha_k}{1+\eta\alpha_k}\right)\right\} \alpha_i \nonumber \\
&\quad + \tau_{\alpha_n}\sum_{i=\tau_{\alpha_n}+1}^{n}\left\{\prod_{k=i+1}^n \exp\left(\frac{-\eta\alpha_k}{1+\eta\alpha_k}\right)\right\} \alpha_{i-\tau_{\alpha_n}}\alpha_i. \nonumber
\end{align}
Since
$
\sum \frac{\alpha_k}{1+\eta\alpha_k} = \sum \alpha_k - \sum \frac{\eta\alpha_k^2}{1+\eta\alpha_k} = \sum \alpha_k + \mathcal{O}(1)
$
for $\alpha_k = \alpha_1/k^s, s \in (0.5, 1]$, we get
\begin{align}
\mathbb{E}_{\mu_{\pi_\star}}\left\|\boldsymbol{w}_{\star}-\boldsymbol{w}^{\text{im}}_{n+1}\right\|^2
& \lesssim \exp\left(-\eta\sum_{i=1}^n \alpha_i\right)\left\|\boldsymbol{w}_{\star}-\boldsymbol{w}^{\text{im}}_{1}\right\|^2 \label{IMP_DEC_TD0_FIR} \\ 
&
\quad + \sum_{i=1}^n\exp\left(-\eta\sum_{k=i+1}^n \alpha_k\right) \alpha_i^2 \label{IMP_DEC_TD0_SEC}  \\
& \quad +  \tau_{\alpha_n}\sum_{i=1}^{\tau_{\alpha_n}}\exp\left(-\eta\sum_{k=i+1}^n \alpha_k \right) \alpha_i \label{IMP_DEC_TD0_THR}  \\
&\quad +  \tau_{\alpha_n}\sum_{i=\tau_{\alpha_n}+1}^{n}\exp\left(-\eta\sum_{k=i+1}^n \alpha_k\right)  \alpha_{i-\tau_{\alpha_n}}\alpha_i. \label{IMP_DEC_TD0_FOUR} 
\end{align}
We now bound each term separately. For the first term, we have
\begin{align*}
\eqref{IMP_DEC_TD0_FIR} = \exp\left(-\eta\alpha_1\sum_{i=1}^n \frac{1}{i^s}\right)  \left\|\boldsymbol{w}_{\star}-\boldsymbol{w}^{\text{im}}_{1}\right\|^2 \leq \exp\left\{-\eta\alpha_1
\psi(n)\right\} \left\|\boldsymbol{w}_{\star}-\boldsymbol{w}^{\text{im}}_{1}\right\|^2
\end{align*}
where 
$$
\psi(n) := \begin{cases}
\frac{n^{1-s}-1}{1-s} \quad &\text{for} ~s \in (0.5, 1)\\
\log n \quad &\text{for} ~s =1.
\end{cases}
$$
For the remaining terms, by Lemma \ref{LEMMA:prelim_TD_exp_bound}, we get
\begin{align*}
\eqref{IMP_DEC_TD0_SEC} 
= \mathcal{O}(\alpha_n), \quad \eqref{IMP_DEC_TD0_THR} 
=\mathcal{O}(\tau_{\alpha_n}\alpha_n), \quad \eqref{IMP_DEC_TD0_FOUR} 
= \mathcal{O}(\tau_{\alpha_n} \alpha_{(n-\tau_{\alpha_n})\vee 1})  
\end{align*}
for $s \in (0.5, 1)$ or $s = 1$ and $\eta \alpha_1 > 1$. Combining everything, we get
\begin{align*}
\mathbb{E}_{\mu_{\pi_\star}}\left\|\boldsymbol{w}_{\star}-\boldsymbol{w}^{\text{im}}_{n+1}\right\|^2 &\lesssim  \exp\left\{-\eta\alpha_1
\psi(n)\right\} \left\|\boldsymbol{w}_{\star}-\boldsymbol{w}^{\text{im}}_{1}\right\|^2 + \mathcal{O}\left(\tau_{\alpha_n} \alpha_{(n-\tau_{\alpha_n})\vee1} \right). 
\end{align*}
\end{proof}

\subsubsection{Finite-time analysis for implicit TD($\lambda$)}
\noindent Recall that, in TD($\lambda$) algorithm, we defined
\begin{align*}
    S_{\boldsymbol{z}_n}(\boldsymbol{w}) &:= r_n \boldsymbol{e}_{n} + \gamma \boldsymbol{e}_{n} \boldsymbol{\phi}_{n+1}^{\top} \boldsymbol{w} - \boldsymbol{e}_{n} \boldsymbol{\phi}_{n}^{\top} \boldsymbol{w} \\
    S(\boldsymbol{w}) &:= \mathbb{E}\left(r_n \boldsymbol{e}_{-\infty:n}\right) + \mathbb{E}\left(\gamma \boldsymbol{e}_{-\infty:n} \boldsymbol{\phi}_{n+1}^{\top}\right) \boldsymbol{w} - \mathbb{E}\left(\boldsymbol{e}_{-\infty:n} \boldsymbol{\phi}_{n}^{\top}\right) \boldsymbol{w}
\end{align*}
where $\boldsymbol{e}_{-\infty:n} := \sum_{k=0}^\infty (\lambda\gamma)^k \boldsymbol{\phi}_{n-k}$.
In addition to these notations, we also define
\begin{align*}    
    S_{\ell:n}(\boldsymbol{w}) &:= r_n \boldsymbol{e}_{\ell:n} + \gamma \boldsymbol{e}_{\ell:n} \boldsymbol{\phi}_{n+1}^{\top} \boldsymbol{w} - \boldsymbol{e}_{\ell:n} \boldsymbol{\phi}_{n}^{\top} \boldsymbol{w}\\
    \xi_{n}(\boldsymbol{w}) &:=\left\{S_{\boldsymbol{z}_n}(\boldsymbol{w})-S(\boldsymbol{w})\right\}^{\top}\left(\boldsymbol{w}-\boldsymbol{w}_{\star}\right), ~~ \forall \boldsymbol{w}\in \mathbb{R}^{d}\\
    \xi_{\ell:n}(\boldsymbol{w}) &:=\left\{S_{\ell:n}(\boldsymbol{w})-S(\boldsymbol{w})\right\}^{\top}\left(\boldsymbol{w}-\boldsymbol{w}_{\star}\right), ~~ \forall \boldsymbol{w}\in \mathbb{R}^{d}
\end{align*}
where $\boldsymbol{e}_{\ell:n} := \sum_{k=0}^{n-\ell} (\lambda\gamma)^k \boldsymbol{\phi}_{n-k}$. The following results from \citep{bhandari2018finite} will be used to establish the finite-time MSE bound.

\begin{lemma}[Lemma 16 of \citet{bhandari2018finite}]\label{APPEND_LEMMA:TSIT_TD_LAMB_LOWER}
For all $\boldsymbol{w} \in \mathbb{R}^d$, $\kappa = \frac{\gamma(1-\lambda)}{1-\lambda\gamma}$.
\[
(\boldsymbol{w}_{\star} - w)^{\top} S(\boldsymbol{w}) \geq (1 - \kappa) \|V_{\boldsymbol{w}_{\star}} - V_{\boldsymbol{W}}\|_D^2.
\]
\end{lemma}

\begin{lemma}[Lemma 17 of \citet{bhandari2018finite}]\label{APPEND_LEMMA:EL_TR_BOUND}
With probability one, for all $\boldsymbol{w} \in \{\tilde{\boldsymbol{w}}: \|\tilde{\boldsymbol{w}}\| \le R\}$ and $i \in \mathbb{N}$, $\left\|S_{\boldsymbol{z}_i}(\boldsymbol{w})\right\| \le B$, $\left\|S(\boldsymbol{w})\right\| \le B$, where $B:= r_{\max}e_{\max}+2e_{\max}\phi_{\max}R$.
\end{lemma}

\begin{lemma}
\label{LEMMA:TDLAMB_FIN_PROJ_ITER} With probability one, let $\kappa = \frac{\gamma(1-\lambda)}{1-\lambda\gamma}$. Then, for all $n \in \mathbb{N}$,
$$
\left\|\boldsymbol{w}_{\star}-\boldsymbol{w}^{\text{im}}_{n+1}\right\|^2 
\le \|\boldsymbol{w}_{\star} - \boldsymbol{w}^{\text{im}}_{n}\|^2 - \frac{2\alpha_n(1-\lambda\gamma)^2(1 - \kappa)}{1+\alpha_n e_{\max}^2}\left\|V_{\boldsymbol{w}_{\star}}-V_{\boldsymbol{w}^{\text{im}}_{n}}\right\|_{\boldsymbol{D}}^2 + 2\tilde \alpha_n \xi_n(\boldsymbol{w}_{n}) + \alpha_n^2 B^2.
$$
\end{lemma}
\begin{proof}
\begin{align}
\|\boldsymbol{w}_{\star} - \boldsymbol{w}^{\text{im}}_{n+1}\|^2 
&= \left\|\Pi_{R}(\boldsymbol{w}_{\star}) - \Pi_{R}\{\boldsymbol{w}^{\text{im}}_n + \tilde \alpha_n S_{\boldsymbol{z}_n}(\boldsymbol{w}^{\text{im}}_{n})\}\right\|^2 \label{LEMMA_RECUR_TDL_2} \\
&= \left\|\boldsymbol{w}_{\star} - \boldsymbol{w}^{\text{im}}_n - \tilde \alpha_n S_{\boldsymbol{z}_n}(\boldsymbol{w}^{\text{im}}_{n})\right\|^2 \label{LEMMA_RECUR_TDL_3}\\
&= \|\boldsymbol{w}_{\star} - \boldsymbol{w}^{\text{im}}_{n}\|^2 - 2\tilde \alpha_n S_{\boldsymbol{z}_n}(\boldsymbol{w}^{\text{im}}_{n})^{\top} (\boldsymbol{w}_{\star} - \boldsymbol{w}^{\text{im}}_{n}) + \left\|\tilde \alpha_n S_{\boldsymbol{z}_n}(\boldsymbol{w}^{\text{im}}_{n})\right\|^2 \nonumber \\
&\le \|\boldsymbol{w}_{\star} - \boldsymbol{w}^{\text{im}}_{n}\|^2 - 2\tilde \alpha_n S_{\boldsymbol{z}_n}(\boldsymbol{w}^{\text{im}}_{n})^{\top} (\boldsymbol{w}_{\star} - \boldsymbol{w}^{\text{im}}_{n}) + \alpha_n^2 B^2 \label{LEMMA_RECUR_TDL_5} \\
&= \|\boldsymbol{w}_{\star} - \boldsymbol{w}^{\text{im}}_{n}\|^2 - 2\tilde \alpha_n S(\boldsymbol{w}^{\text{im}}_{n})^{\top} (\boldsymbol{w}_{\star} - \boldsymbol{w}^{\text{im}}_{n}) + 2\tilde \alpha_n \xi_n(\boldsymbol{w}^{\text{im}}_{n}) +  \alpha_n^2 B^2 \nonumber \\
&\le \|\boldsymbol{w}_{\star} - \boldsymbol{w}^{\text{im}}_{n}\|^2 - 2\tilde \alpha_n(1 - \kappa)\left\|V_{\boldsymbol{w}_{\star}}-V_{\boldsymbol{w}^{\text{im}}_{n}}\right\|_{\boldsymbol{D}}^2 + 2\tilde \alpha_n \xi_n(\boldsymbol{w}^{\text{im}}_{n}) + \alpha_n^2 B^2 \label{LEMMA_RECUR_TDL_7} \\
&\le \|\boldsymbol{w}_{\star} - \boldsymbol{w}^{\text{im}}_{n}\|^2 - \frac{2\alpha_n(1-\lambda\gamma)^2(1 - \kappa)}{(1-\lambda\gamma)^2+\alpha_n e_{\max}^2}\left\|V_{\boldsymbol{w}_{\star}}-V_{\boldsymbol{w}^{\text{im}}_{n}}\right\|_{\boldsymbol{D}}^2 + 2\tilde \alpha_n \xi_n(\boldsymbol{w}^{\text{im}}_{n}) + \alpha_n^2 B^2
\label{LEMMA_RECUR_TDL_8} \\
&\le \|\boldsymbol{w}_{\star} - \boldsymbol{w}^{\text{im}}_{n}\|^2 - \frac{2\alpha_n(1-\lambda\gamma)^2(1 - \kappa)}{1+\alpha_n 
e_{\max}^2}\left\|V_{\boldsymbol{w}_{\star}}-V_{\boldsymbol{w}^{\text{im}}_{n}}\right\|_{\boldsymbol{D}}^2 + 2\tilde \alpha_n \xi_n(\boldsymbol{w}^{\text{im}}_{n}) + \alpha_n^2 B^2
\label{LEMMA_RECUR_TDL_9} 
\end{align}
where \eqref{LEMMA_RECUR_TDL_2} is due to the fact that $\boldsymbol{w}_{\star} = \Pi_{R}(\boldsymbol{w}_{\star})$, 
\eqref{LEMMA_RECUR_TDL_3} is thanks to non-expansiveness of the projection operator on the convex set, 
\eqref{LEMMA_RECUR_TDL_5} comes from Lemma \ref{APPEND_LEMMA:EL_TR_BOUND} with $\tilde\alpha_n \le \alpha_n$, and \eqref{LEMMA_RECUR_TDL_7} is from Lemma \ref{APPEND_LEMMA:TSIT_TD_LAMB_LOWER}. Finally, \eqref{LEMMA_RECUR_TDL_8} is the direct consequence of Lemma \ref{APPEND_LEMMA:IMP_STEP} and \eqref{LEMMA_RECUR_TDL_9} is due to $(1-\lambda\gamma)^2 < 1$.
\end{proof}

\begin{lemma}[Lemma 19 of \citet{bhandari2018finite}]\label{APPEND_LEMMA:BHANDARI_19}
With probability one, given $\ell \le i$, 
\begin{enumerate}
    \item $|\xi_{\ell:i}(\boldsymbol{w})| \le 2B^2$
    \item $|\xi_{\ell:i}(\boldsymbol{w})- \xi_{\ell:i}(\boldsymbol{w}')| \lesssim \|\boldsymbol{w}-\boldsymbol{w}'\|$
    \item $|\xi_{i}(\boldsymbol{w})- \xi_{i-\tau:i}(\boldsymbol{w})| \lesssim (\lambda\gamma)^\tau$~ \text{for all} ~$\tau \le i$
    \item $|\xi_{i}(\boldsymbol{w})- \xi_{-\infty:i}(\boldsymbol{w})| \lesssim (\lambda\gamma)^i$ 
\end{enumerate}
for all $\boldsymbol{w}, \boldsymbol{w}' \in  \{\tilde{\boldsymbol{w}}: \|\tilde{\boldsymbol{w}}\| \le R\}$. 
    
\end{lemma}

\begin{definition}\label{APPEND_DEF:DOUBLE_MIX_LAMB} 
Given $\epsilon > 0$, we define a $\lambda$-mixing time $\tau_{\lambda, \alpha_n}$ to be
\begin{align*}
  \tau_{\lambda, \alpha_n} = \max\{\tau_{\alpha_n}, \tau_{\alpha_n}^{\lambda}\} ~\text{where}~
  \tau^{\lambda}_{\epsilon} = \min\left\{n\in \mathbb{N} \mid (\lambda\gamma)^n \le \epsilon\right\}.
\end{align*}
\end{definition}

\begin{lemma}\label{APPEND_LEMMA:TD_LAMB_RANDOM_STOCHASTIC_BOUND}
Given a non-increasing sequence $(\alpha_{n})_{n \in \mathbb{N}}$, for any fixed $i < n$,
\begin{enumerate}
    \item if $2\tau_{\lambda, \alpha_n} < i $,
        $$\mathbb{E}_{\mu_{\pi_\star}}\left\{ \xi_{i}\left(\boldsymbol{w}^{\text{im}}_{i}\right)\right\} \lesssim \tau_{\lambda, \alpha_n} \alpha_{i-2\tau_{\lambda, \alpha_n}}.
        $$
    \item if $i \le 2\tau_{\lambda, \alpha_n}$, 
        $$
        \mathbb{E}_{\mu_{\pi_\star}}\left\{ \xi_{i}\left(\boldsymbol{w}^{\text{im}}_{i}\right)\right\} \lesssim \tau_{\lambda, \alpha_n}.
        $$
\end{enumerate}
\end{lemma}
\begin{proof} 
\textbf{Claim 1:} We first consider the case where $i > 2\tau_{\lambda, \alpha_n}$ and obtain a bound for $\mathbb{E}_{\mu_{\pi_\star}}\left\{\xi_{i}(\boldsymbol{w}^\text{im}_i)\right\}$. Notice that
\begin{align}
\mathbb{E}_{\mu_{\pi_\star}}\left\{\xi_{i}(\boldsymbol{w}^\text{im}_i)\right\} &\le \left|\mathbb{E}_{\mu_{\pi_\star}}\left\{\xi_{i}(\boldsymbol{w}^\text{im}_i)\right\} - \mathbb{E}_{\mu_{\pi_\star}}\left\{\xi_{i}\left(\boldsymbol{w}^\text{im}_{i-2\tau_{\lambda, \alpha_n}}\right)\right\} \right| \label{TDL_XI_N_FIRST_BD} \\ 
& \quad + \left|\mathbb{E}_{\mu_{\pi_\star}}\left\{\xi_{i}\left(\boldsymbol{w}^\text{im}_{i-2\tau_{\lambda, \alpha_n}}\right)\right\} - \mathbb{E}_{\mu_{\pi_\star}}\left\{\xi_{i-\tau_{\lambda, \alpha_n}:i}\left(\boldsymbol{w}^\text{im}_{i-2\tau_{\lambda, \alpha_n}}\right)\right\}  \right| \label{TDL_XI_N_SECOND_BD}\\
& \quad + \left| \mathbb{E}_{\mu_{\pi_\star}}\left\{\xi_{i-\tau_{\lambda, \alpha_n}:i}\left(\boldsymbol{w}^\text{im}_{i-2\tau_{\lambda, \alpha_n}}\right)\right\}  \right|. \label{TDL_XI_N_THIRD_BD}
\end{align}

To get an upper bound of the term in \eqref{TDL_XI_N_FIRST_BD}, notice that
\begin{align*}
\left|\xi_{i}(\boldsymbol{w}^\text{im}_i) - \xi_{i}\left(\boldsymbol{w}^\text{im}_{i-2\tau_{\lambda, \alpha_n}}\right) \right| \lesssim  \left\|\boldsymbol{w}^\text{im}_i - \boldsymbol{w}^\text{im}_{i-2\tau_{\lambda, \alpha_n}}\right\| \lesssim  \sum_{k=i-2\tau_{\lambda, \alpha_n}}^{i-1} \|\boldsymbol{w}^\text{im}_{k+1} - \boldsymbol{w}^\text{im}_k\| 
\end{align*}
where the second inequality comes from Lemma \ref{APPEND_LEMMA:BHANDARI_19} and the third inequality is thanks to the triangle inequality. Note that 
$$ 
\left\|\boldsymbol{w}^{\text{im}}_{k+1}-\boldsymbol{w}^{\text{im}}_{k}\right\| = \left\|\Pi_{R}(\boldsymbol{w}^{\text{im}}_{k}+ \tilde \alpha_k S_{\boldsymbol{z}_k}(\boldsymbol{w}^{\text{im}}_{k}))-\Pi_{R}(\boldsymbol{w}^{\text{im}}_{k})\right\| \le 
\left\|\boldsymbol{w}^{\text{im}}_{k} + \tilde \alpha_k S_{\boldsymbol{z}_k}(\boldsymbol{w}^{\text{im}}_{k})-\boldsymbol{w}^{\text{im}}_{k}\right\| \le \alpha_k B
$$
where in the first inequality, we have used the non-expansiveness of the projection operator, and 
in the second inequality, both Lemma \ref{APPEND_LEMMA:IMP_STEP} and \ref{APPEND_LEMMA:EL_TR_BOUND} were used. Therefore, we have
\begin{align}\label{TDL_XI_N_FIRST_BD_FIRST_PRELIM}
\left|\xi_{i}(\boldsymbol{w}^\text{im}_i) - \xi_{i}\left(\boldsymbol{w}^\text{im}_{i-2\tau_{\lambda, \alpha_n}}\right) \right| \lesssim \sum_{k=i-2\tau_{\lambda, \alpha_n}}^{i-1} \alpha_k
\end{align}
which leads to
\begin{align}
\left|\mathbb{E}_{\mu_{\pi_\star}}\left\{\xi_{i}(\boldsymbol{w}^\text{im}_i)\right\} - \mathbb{E}_{\mu_{\pi_\star}}\left\{\xi_{i}\left(\boldsymbol{w}^\text{im}_{i-2\tau_{\lambda, \alpha_n}}\right)\right\} \right| &\le  \mathbb{E}_{\mu_{\pi_\star}}\left\{\left| \xi_{i}(\boldsymbol{w}^\text{im}_i) - \xi_{i}\left(\boldsymbol{w}^\text{im}_{i-2\tau_{\lambda, \alpha_n}}\right)\right| \right\} \nonumber \\ &\lesssim \sum_{k=i-2\tau_{\lambda, \alpha_n}}^{i-1} \alpha_k \label{TDL_XI_N_FIRST_BD_FIRST} 
\end{align}
where the first inequality is due to Jensen's inequality and the second inequality is thanks to \eqref{TDL_XI_N_FIRST_BD_FIRST_PRELIM}. 

To obtain an upper bound of \eqref{TDL_XI_N_SECOND_BD}, note that the third claim of Lemma \ref{APPEND_LEMMA:BHANDARI_19}, we have
\begin{equation}\label{TDL_XI_N_FIRST_BD_SECOND}
\left|\mathbb{E}_{\mu_{\pi_\star}}\left\{\xi_{i}\left(\boldsymbol{w}^\text{im}_{i-2\tau_{\lambda, \alpha_n}}\right)\right\} - \mathbb{E}_{\mu_{\pi_\star}}\left\{\xi_{i-\tau_{\lambda, \alpha_n}:i}\left(\boldsymbol{w}^\text{im}_{i-2\tau_{\lambda, \alpha_n}}\right)\right\}  \right| \lesssim (\lambda\gamma)^{\tau_{\lambda, \alpha_n}} \le \alpha_n
\end{equation}
where the last inequality is due to the definition of the $\lambda$-mixing time $\tau_{\lambda, \alpha_n}$.

Next, we derive an upper bound of \eqref{TDL_XI_N_THIRD_BD}. For a fixed $w$, one can treat $\xi_{i-\tau_{\lambda, \alpha_n}:i}(\boldsymbol{w})$ as a function of $\{\boldsymbol{z}_{i-\tau_{\lambda, \alpha_n}}, \cdots, \boldsymbol{z}_{i}\}$, where $\boldsymbol{z}_k= (\boldsymbol{x}_{k}, \boldsymbol{x}_{k+1})$ for $k \in \{i-\tau_{\lambda, \alpha_n}, \cdots, i\}$. Furthermore, one can view $\boldsymbol{w}^\text{im}_{i-2\tau_{\lambda, \alpha_n}}$ as a function of $\{\boldsymbol{z}_1, \cdots, \boldsymbol{z}_{i-2\tau_{\lambda, \alpha_n}-1}\}$. Now consider $\xi_{i-\tau_{\lambda, \alpha_n}:i}\left(\boldsymbol{w}^\text{im}_{i-2\tau_{\lambda, \alpha_n}}\right)$, which is a function of both 
$$
\boldsymbol{u} = \{\boldsymbol{z}_1, \cdots, \boldsymbol{z}_{i-2\tau_{\lambda, \alpha_n}-1}\} ~~\text{as well as}~~ \tilde{\boldsymbol{u}} = \{\boldsymbol{z}_{i-\tau_{\lambda, \alpha_n}}, \cdots, \boldsymbol{z}_{i}\}.
$$ 
We set $h(\boldsymbol{u}, \tilde{\boldsymbol{u}}) = \xi_{i-\tau_{\lambda, \alpha_n}:i} \left(\boldsymbol{w}^\text{im}_{i-2\tau_{\lambda, \alpha_n}}\right)$ to invoke Lemma \ref{APPEND_LEMMA:BHAND9}. The condition for Lemma \ref{APPEND_LEMMA:BHAND9} is met since 
$$
\boldsymbol{u} =\{\boldsymbol{z}_1, \cdots, \boldsymbol{z}_{i-2\tau_{\lambda, \alpha_n}-1}\} \to \boldsymbol{x}_{i-2\tau_{\lambda, \alpha_n}} \to \boldsymbol{x}_{i-\tau_{\lambda, \alpha_n}} \to \{\boldsymbol{z}_{i-\tau_{\lambda, \alpha_n}}, \cdots, \boldsymbol{z}_{i}\} = \tilde{\boldsymbol{u}}
$$ 
forms a Markov chain. Therefore, we get
\begin{equation}\label{TDL_XI_MIX_BOUND}
\mathbb{E}_{\mu_{\pi_\star}} \left\{h(\boldsymbol{u}, \tilde{\boldsymbol{u}}) \right\} - \mathbb{E} \left\{h(\boldsymbol{u}', \tilde{\boldsymbol{u}}') \right\} \le 2\|h\|_\infty m\rho ^{\tau_{\lambda, \alpha_n}}
\end{equation}
where $\boldsymbol{u}' = \{\boldsymbol{z}'_1, \ldots, \boldsymbol{z}'_{i-2\tau_{\lambda,\alpha_n}-1}\}$ and $\tilde{\boldsymbol{u}}' = \{\boldsymbol{z}'_{i-\tau_{\lambda,\alpha_n}}, \ldots, \boldsymbol{z}'_{i}\}$ are independent copies with the same marginal distributions as $\boldsymbol{u}$ and $\tilde{\boldsymbol{u}}$, respectively. Let us denote the $(i-2\tau_{\lambda, \alpha_n})^{\text{th}}$ implicit TD($\lambda$) iterate computed using $\boldsymbol{u}'$ as $\boldsymbol{w}_{i-2\tau_{\lambda, \alpha_n}}'$. From the law of iterated expectations, we have
$$
\mathbb{E} \left\{h(\boldsymbol{u}', \tilde{\boldsymbol{u}}') \right\} = \mathbb{E} \left [\mathbb{E} \left\{ \xi_{i-\tau_{\lambda, \alpha_n}:i}\left(\boldsymbol{w}_{i-2\tau_{\lambda, \alpha_n}}'\right)\mid \boldsymbol{u}' \right\} \right].
$$
Now, for any fixed $w$, by the definition of $\xi_{i-\tau_{\lambda, \alpha_n}:i}(\boldsymbol{w})$, we know
\begin{align*}\label{TDL_xi_bound}
\mathbb{E}_{\mu_{\pi_\star}}\left\{\xi_{i-\tau_{\lambda, \alpha_n}:i}(\boldsymbol{w})\right\} &= \left[\mathbb{E}_{\mu_{\pi_\star}}\left\{S_{i-\tau_{\lambda, \alpha_n}:i}(\boldsymbol{w})\right\}-S(\boldsymbol{w})\right]^{\top}\left(\boldsymbol{w}-\boldsymbol{w}_{\star}\right) \nonumber \\
&=  \mathbb{E}_{\mu_{\pi_\star}}\left\{S_{i-\tau_{\lambda, \alpha_n}:i}(\boldsymbol{w})-S_{-\infty:i}(\boldsymbol{w})\right\}^{\top}\left(\boldsymbol{w}-\boldsymbol{w}_{\star}\right).
\end{align*}
The second equality follows from
$$
\mathbb{E}_{\mu_{\pi_\star}}\left\{S_{i-\tau_{\lambda, \alpha_n}:i}(\boldsymbol{w})\right\}-S(\boldsymbol{w}) = \mathbb{E}_{\mu_{\pi_\star}}\left\{S_{i-\tau_{\lambda, \alpha_n}:i}(\boldsymbol{w})\right\}-\mathbb{E}_{\mu_{\pi_\star}}\left\{S_{-\infty:i}(\boldsymbol{w})\right\} 
=  \mathbb{E}_{\mu_{\pi_\star}}\left\{S_{i-\tau_{\lambda, \alpha_n}:i}(\boldsymbol{w})-S_{-\infty:i}(\boldsymbol{w})\right\}.
$$
Notice that
\begin{align*}
   \left | \left\{S_{i-\tau_{\lambda, \alpha_n}:i}(\boldsymbol{w}) - S_{-\infty:i}(\boldsymbol{w})\right\}^{\top}(\boldsymbol{w}-\boldsymbol{w}_{\star}) \right | &= \left|\xi_{i-\tau_{\lambda, \alpha_n}:i}(\boldsymbol{w}) - \xi_{-\infty:i}(\boldsymbol{w})\right| \\ &\le \left|\xi_{i-\tau_{\lambda, \alpha_n}:i}(\boldsymbol{w}) - \xi_{i}(\boldsymbol{w})\right| + \left|\xi_{i}(\boldsymbol{w}) - \xi_{-\infty:i}(\boldsymbol{w})\right|\\ &\lesssim (\lambda\gamma)^{\tau_{\lambda, \alpha_n}} 
\end{align*}
where the first inequality is due to the triangle inequality and the last inequality follows from combining claims 3 and 4 of Lemma \ref{APPEND_LEMMA:BHANDARI_19} with $\tau_{\lambda, \alpha_n} \le n.$ This yields
\begin{equation}\label{TDL_INDEP_TERM_BOUND}
\mathbb{E} \left\{h(\boldsymbol{u}', \tilde{\boldsymbol{u}}') \right\} 
\lesssim  (\lambda\gamma)^{\tau_{\lambda, \alpha_n}}.
\end{equation}
Combining \eqref{TDL_XI_MIX_BOUND} and \eqref{TDL_INDEP_TERM_BOUND}, we arrive at 
\begin{align}\label{TDL_XI_N_FIRST_BD_THIRD}
\mathbb{E}_{\mu_{\pi_\star}} \left\{\xi_{i-\tau_{\lambda, \alpha_n}:i} \left(\boldsymbol{w}^\text{im}_{i-2\tau_{\lambda, \alpha_n}}\right)\right\} 
\lesssim m\rho ^{\tau_{\lambda, \alpha_n}} + (\lambda\gamma)^{\tau_{\lambda, \alpha_n}} \le \alpha_n
\end{align}
where the last inequality is due to the definition of $\lambda$-mixing time $\tau_{\lambda, \alpha_n}$. Combining \eqref{TDL_XI_N_FIRST_BD_FIRST}, \eqref{TDL_XI_N_FIRST_BD_SECOND} and \eqref{TDL_XI_N_FIRST_BD_THIRD}, we get
\begin{align*}
    \mathbb{E}_{\mu_{\pi_\star}}\{\xi_{i}\left(\boldsymbol{w}^{\text{im}}_{i}\right)\} \lesssim  \sum_{k=i-2\tau_{\lambda, \alpha_n}}^{i-1} \alpha_k + \alpha_n \lesssim \tau_{\lambda, \alpha_n} \alpha_{i-2\tau_{\lambda, \alpha_n}} + \alpha_n \lesssim \tau_{\lambda, \alpha_n} \alpha_{i-2\tau_{\lambda, \alpha_n}}.
\end{align*}
Combined with Lemma \ref{APPEND_LEMMA:IMP_STEP}, we get the first claim. \\

\noindent \textbf{Claim 2:} We next consider the case where $i \le 2\tau_{\lambda, \alpha_n}$. Using the triangle inequality, we get that
\begin{align}
\mathbb{E}_{\mu_{\pi_\star}}\left\{\xi_{i}(\boldsymbol{w}^\text{im}_i)\right\} &\le \left|\mathbb{E}_{\mu_{\pi_\star}}\left\{\xi_{i}(\boldsymbol{w}^\text{im}_i)\right\} - \mathbb{E}_{\mu_{\pi_\star}}\left\{\xi_{i}\left(\boldsymbol{w}^\text{im}_{1}\right)\right\} \right| \label{TDL_XI_N_SECOND_FIRST_BD} \\ 
& \quad + \left|\mathbb{E}_{\mu_{\pi_\star}}\left\{\xi_{i}\left(\boldsymbol{w}^\text{im}_{1}\right)\right\} - \mathbb{E}_{\mu_{\pi_\star}}\left\{\xi_{-\infty:i}\left(\boldsymbol{w}^\text{im}_{1}\right)\right\}  \right| \label{TDL_XI_N_SECOND_SECOND_BD}\\
& \quad + \left| \mathbb{E}_{\mu_{\pi_\star}}\left\{\xi_{-\infty:i}\left(\boldsymbol{w}^\text{im}_{1}\right)\right\}  \right|. \label{TDL_XI_N_SECOND_THIRD_BD}
\end{align}
Similar argument as the above can be applied to obtain a bound for \eqref{TDL_XI_N_SECOND_FIRST_BD}. Specifically, we have 
\begin{align*}
\left|\xi_{i}(\boldsymbol{w}^\text{im}_i) - \xi_{i}\left(\boldsymbol{w}^\text{im}_{1}\right) \right| \lesssim \left\|\boldsymbol{w}^\text{im}_i - \boldsymbol{w}^\text{im}_{1}\right\| \lesssim \sum_{k=1}^{i-1} \|\boldsymbol{w}^\text{im}_{k+1} - \boldsymbol{w}^\text{im}_{k}\|
\end{align*}
where the first inequality comes from Lemma \ref{APPEND_LEMMA:BHANDARI_19} and the second inequality is thanks to the triangle inequality. Recall that 
$$ 
\left\|\boldsymbol{w}^{\text{im}}_{k+1}-\boldsymbol{w}^{\text{im}}_{k}\right\| = \left\|\Pi_{R}\{\boldsymbol{w}^{\text{im}}_{k}+ \tilde \alpha_k S_{\boldsymbol{z}_k}(\boldsymbol{w}^{\text{im}}_{k})\}-\Pi_{R}(\boldsymbol{w}^{\text{im}}_{k})\right\| \le 
\left\|\boldsymbol{w}^{\text{im}}_{k} + \tilde \alpha_k S_{\boldsymbol{z}_k}(\boldsymbol{w}^{\text{im}}_{k})-\boldsymbol{w}^{\text{im}}_{k}\right\| \le \alpha_k B
$$
where in the first inequality, we have used the non-expansiveness of the projection operator, and 
in the second inequality, both Lemma \ref{APPEND_LEMMA:IMP_STEP} and \ref{APPEND_LEMMA:EL_TR_BOUND} were used. Therefore, we have
\begin{align}\label{TDL_XI_N_SECOND_BD_FIRST_PRELIM}
\left|\xi_{i}(\boldsymbol{w}^\text{im}_i) - \xi_{i}\left(\boldsymbol{w}^\text{im}_{1}\right) \right| \lesssim \sum_{k=1}^{i-1} \alpha_k
\end{align}
which leads to
\begin{equation}\label{TDL_XI_N_SECOND_BD_FIRST}
\left|\mathbb{E}_{\mu_{\pi_\star}}\left\{\xi_{i}(\boldsymbol{w}^\text{im}_i)\right\} - \mathbb{E}_{\mu_{\pi_\star}}\left\{\xi_{i}\left(\boldsymbol{w}^\text{im}_{1}\right)\right\} \right| \le  \mathbb{E}_{\mu_{\pi_\star}}\left\{\left| \xi_{i}(\boldsymbol{w}^\text{im}_i) - \xi_{i}\left(\boldsymbol{w}^\text{im}_{1}\right)\right| \right\} \lesssim \sum_{k=1}^{i-1} \alpha_k 
\end{equation}
where the first inequality is by Jensen's inequality and the second inequality is thanks to \eqref{TDL_XI_N_SECOND_BD_FIRST_PRELIM}. Furthermore, from the fourth claim of Lemma \ref{APPEND_LEMMA:BHANDARI_19}, we can obtain an upper bound of \eqref{TDL_XI_N_SECOND_SECOND_BD} as follows
\begin{equation}\label{TDL_XI_N_SECOND_BD_SECOND}
\left|\mathbb{E}_{\mu_{\pi_\star}}\left\{\xi_{i}\left(\boldsymbol{w}^\text{im}_{1}\right)\right\} - \mathbb{E}_{\mu_{\pi_\star}}\left\{\xi_{-\infty:i}\left(\boldsymbol{w}^\text{im}_{1}\right)\right\}  \right| \le (\lambda\gamma)^i.
\end{equation}
Lastly, by definition, since $\boldsymbol{w}^{\text{im}}_1$ is fixed, we have $\mathbb{E}_{\mu_{\pi_\star}}\left\{\xi_{-\infty:i}\left(\boldsymbol{w}^\text{im}_{1}\right)\right\} = 0$. Combining \eqref{TDL_XI_N_SECOND_BD_FIRST} and \eqref{TDL_XI_N_SECOND_BD_SECOND}, we have
\begin{align*}
\mathbb{E}_{\mu_{\pi_\star}}\left\{\xi_{i}(\boldsymbol{w}^\text{im}_i)\right\} \lesssim
\sum_{k=1}^{i-1} \alpha_k + (\lambda\gamma)^i \lesssim \tau_{\lambda, \alpha_n} + (\lambda\gamma)^i \lesssim \tau_{\lambda, \alpha_n}.
\end{align*}
\end{proof}

\noindent We now establish a finite-time MSE bound of implicit TD($\lambda$).

\begin{proof}[Proof of Theorem \ref{THM:FIN_PROJ_TDL}]
Starting from Lemma \ref{LEMMA:TDLAMB_FIN_PROJ_ITER}, we have
\begin{align}
\mathbb{E}_{\mu_{\pi_\star}}\left\|\boldsymbol{w}_{\star}-\boldsymbol{w}^{\text{im}}_{n+1}\right\|^2 &\le \mathbb{E}_{\mu_{\pi_\star}}\|\boldsymbol{w}_{\star} - \boldsymbol{w}^{\text{im}}_{n}\|^2 - \frac{2\alpha_n(1-\lambda\gamma)^2(1 - \kappa)}{1+\alpha_n e_{\max}^2}\mathbb{E}_{\mu_{\pi_\star}}\left\|V_{\boldsymbol{w}_{\star}}-V_{\boldsymbol{w}^{\text{im}}_{n}}\right\|_{\boldsymbol{D}}^2 \nonumber \\ 
&\quad + 2\mathbb{E}_{\mu_{\pi_\star}}\left\{\tilde \alpha_n \xi_n(\boldsymbol{w}^{\text{im}}_{n})\right\} + \alpha_n^2 B^2 \nonumber  \\
&\le \mathbb{E}_{\mu_{\pi_\star}}\|\boldsymbol{w}_{\star} - \boldsymbol{w}^{\text{im}}_{n}\|^2 - \frac{2\alpha_n(1-\lambda\gamma)^2(1 - \kappa)\lambda_{\min }}{1+\alpha_n  e_{\max}^2}\mathbb{E}_{\mu_{\pi_\star}}\left\|\boldsymbol{w}_{\star}-\boldsymbol{w}^{\text{im}}_{n}\right\|^2 \nonumber \\
&\quad + 2\alpha_n\mathbb{E}_{\mu_{\pi_\star}}\left\{ \xi_n(\boldsymbol{w}^{\text{im}}_{n})\right\} + 2\mathbb{E}_{\mu_{\pi_\star}}\left\{(\tilde \alpha_n - \alpha_n) \xi_n(\boldsymbol{w}^{\text{im}}_{n})\right\}  + \alpha_n^2 B^2 \nonumber \\
&= \mathbb{E}_{\mu_{\pi_\star}}\|\boldsymbol{w}_{\star} - \boldsymbol{w}^{\text{im}}_{n}\|^2 - \frac{2\alpha_n(1-\lambda\gamma)(1 - \gamma)\lambda_{\min }}{1+\alpha_n e_{\max}^2}\mathbb{E}_{\mu_{\pi_\star}}\left\|\boldsymbol{w}_{\star}-\boldsymbol{w}^{\text{im}}_{n}\right\|^2 \nonumber \\
&\quad + 2\alpha_n\mathbb{E}_{\mu_{\pi_\star}}\left\{ \xi_n(\boldsymbol{w}^{\text{im}}_{n})\right\} + \mathcal{O}(\alpha_n^2) \label{FIN_TDL_RECUR}
\end{align}
where the second inequality is due to Lemma \ref{APPEND_LEMMA:BHAND1} and the third one is thanks to Lemma \ref{APPEND_LEMMA:BHANDARI_19} and $|\tilde \alpha_n - \alpha_n| \le \alpha_n^2 e_{\max}^2$. One can show that
\begin{align*}
&1-\frac{2\alpha_n(1-\lambda\gamma)(1-\gamma)\lambda_{\min }}{1+\alpha_n e_{\max}^2} \le \frac{1}{1+\eta_{\lambda}\alpha_n} \quad \text{where}\\
&\eta_{\lambda} = \frac{2(1-\lambda\gamma)(1-\gamma)\lambda_{\min}}{1+\mathbb{I}\{e_{\max}^2 > 2(1-\lambda\gamma)(1-\gamma)\lambda_{\min}\}(e_{\max}^2 - 2(1-\lambda\gamma)(1-\gamma)\lambda_{\min})\alpha_1}.
\end{align*}
Furthermore, since $\frac{1}{1+x} \le \exp\left(\frac{-x}{1+x}\right)$, we have
$
\frac{1}{1+\eta_{\lambda}\alpha_n} \le \exp\left(\frac{-\eta_{\lambda}\alpha_n}{1+\eta_{\lambda}\alpha_n}\right).
$
Continuing from \eqref{FIN_TDL_RECUR}, we have
\begin{align}
    \mathbb{E}_{\mu_{\pi_\star}}\left\|\boldsymbol{w}_{\star}-\boldsymbol{w}^{\text{im}}_{n+1}\right\|^2 
    &\le  \exp\left(\frac{-\eta_{\lambda}\alpha_n}{1+\eta_{\lambda}\alpha_n }\right)\mathbb{E}_{\mu_{\pi_\star}}\left\|\boldsymbol{w}_{\star}-\boldsymbol{w}^{\text{im}}_{n}\right\|^2 + 2\alpha_n\mathbb{E}_{\mu_{\pi_\star}}\left\{ \xi_n(\boldsymbol{w}^{\text{im}}_{n})\right\} + \mathcal{O}(\alpha_n^2) \nonumber \\
    & \lesssim \left\{\prod_{i=1}^n  \exp\left(\frac{-\eta_{\lambda}\alpha_i}{1+\eta_{\lambda}\alpha_i}\right)\right\}\left\|\boldsymbol{w}_{\star}-\boldsymbol{w}^{\text{im}}_{1}\right\|^2 
    + \sum_{i=1}^n\left\{\prod_{k=i+1}^n \exp\left(\frac{-\eta_{\lambda}\alpha_k}{1+\eta_{\lambda}\alpha_k}\right)\right\} \alpha_i^2 \nonumber \\
    & \quad +\tau_{\lambda, \alpha_n}\sum_{i=1}^{2\tau_{\lambda, \alpha_n}}\left\{\prod_{k=i+1}^n \exp\left(\frac{-\eta_{\lambda}\alpha_k}{1+\eta_{\lambda}\alpha_k}\right)\right\} \alpha_i \nonumber \\
    &\quad + \tau_{\lambda, \alpha_n}\sum_{i=2\tau_{\lambda, \alpha_n}+1}^{n}\left\{\prod_{k=i+1}^n \exp\left(\frac{-\eta_{\lambda}\alpha_k}{1+\eta_{\lambda}\alpha_k}\right)\right\}  \alpha_{i-2\tau_{\lambda, \alpha_n}}\alpha_i \nonumber
\end{align}
where the second inequality is due to Lemma \ref{APPEND_LEMMA:TD_LAMB_RANDOM_STOCHASTIC_BOUND}. Notice that 
$
\sum \frac{\alpha_k}{1+\eta_{\lambda}\alpha_k} = \sum \alpha_k + \mathcal{O}(1)
$
for $\alpha_k = \alpha_1/k^s, s \in (0.5, 1]$. Therefore, we get
\begin{align*}
\mathbb{E}_{\mu_{\pi_\star}}\left\|\boldsymbol{w}_{\star}-\boldsymbol{w}^{\text{im}}_{n+1}\right\|^2
& \lesssim \exp\left(-\eta_{\lambda}\sum_{i=1}^n \alpha_i\right)\left\|\boldsymbol{w}_{\star}-\boldsymbol{w}^{\text{im}}_{1}\right\|^2 
+ \sum_{i=1}^n\exp\left(-\eta_{\lambda}\sum_{k=i+1}^n \alpha_k\right)\alpha_i^2 \\
& \quad + \tau_{\lambda, \alpha_n}\sum_{i=1}^{2\tau_{\lambda, \alpha_n}}\exp\left(-\eta_{\lambda}\sum_{k=i+1}^n \alpha_k \right) \alpha_i 
+ \tau_{\lambda, \alpha_n}\sum_{i=2\tau_{\lambda, \alpha_n}+1}^{n}\exp\left(-\eta_{\lambda}\sum_{k=i+1}^n \alpha_k\right)  \alpha_{i-2\tau_{\lambda, \alpha_n}}\alpha_i.
\end{align*}
We now bound each term separately. For the first term, we have
\begin{align*}
\exp\left(-\eta_{\lambda}\sum_{i=1}^n \alpha_i\right)\left\|\boldsymbol{w}_{\star}-\boldsymbol{w}^{\text{im}}_{1}\right\|^2 &= \exp\left(-\eta_{\lambda}\alpha_1\sum_{i=1}^n \frac{1}{i^s}\right)  \left\|\boldsymbol{w}_{\star}-\boldsymbol{w}^{\text{im}}_{1}\right\|^2 \leq \exp\left\{-\eta_{\lambda}\alpha_1
\psi(n)\right\} \left\|\boldsymbol{w}_{\star}-\boldsymbol{w}^{\text{im}}_{1}\right\|^2
\end{align*}
where 
$$
\psi(n) := \begin{cases}
\frac{n^{1-s}-1}{1-s} \quad &\text{for} ~s \in (0.5, 1)\\
\log n \quad &\text{for} ~s =1.
\end{cases}
$$
For the remaining terms, by Lemma \ref{LEMMA:prelim_TD_exp_bound}, we get
\begin{align*}
\sum_{i=1}^n\exp\left(-\eta_{\lambda}\sum_{k=i+1}^n \alpha_k\right)\alpha_i^2 
&= \mathcal{O}(\alpha_n)\\
\tau_{\lambda, \alpha_n}\sum_{i=1}^{2\tau_{\lambda,\alpha_n}}\exp\left(-\eta_{\lambda}\sum_{k=i+1}^n \alpha_k \right) \alpha_i 
&=\mathcal{O}(\tau_{\lambda,\alpha_n}\alpha_n)\\
\tau_{\lambda, \alpha_n}\sum_{i=2\tau_{\lambda,\alpha_n}+1}^{n}\exp\left(-\eta_{\lambda}\sum_{k=i+1}^n \alpha_k\right)  \alpha_{i-2\tau_{\lambda,\alpha_n}}\alpha_i 
&=\mathcal{O}(\tau_{\lambda, \alpha_n} \alpha_{(n-2\tau_{\lambda,\alpha_n})\vee 1})  
\end{align*}
for $s \in (0.5, 1)$ or $s = 1$ and 
$\eta_\lambda \alpha_1 > 1$. 
Combining everything, we get
\begin{align*}
\mathbb{E}_{\mu_{\pi_\star}}\left\|\boldsymbol{w}_{\star}-\boldsymbol{w}^{\text{im}}_{n+1}\right\|^2 &\lesssim \exp\left\{-\eta_{\lambda}\alpha_1
\psi(n)\right\} \left\|\boldsymbol{w}_{\star}-\boldsymbol{w}^{\text{im}}_{1}\right\|^2 + \mathcal{O}\left(\tau_{\lambda,\alpha_n} \alpha_{(n-2\tau_{\lambda,\alpha_n})\vee 1} \right). 
\end{align*}
\end{proof}

\section{Theoretical analysis for implicit TDC}
\noindent For the ease of presentation, we abbreviate the superscript for the implicit update and consider the following implicit TDC updates given by
\begin{align}
\boldsymbol{w}_{n+1} & = \boldsymbol{w}_{n} + \alpha'_n \rho_n \left(r_n + \gamma \boldsymbol{\phi}_{n+1}^{\top} \boldsymbol{w}_{n} -\boldsymbol{\phi}_{n}^{\top} \boldsymbol{w}_{n}\right)\boldsymbol{\phi}_{n} - \alpha_n \rho_n \gamma \left(\boldsymbol{\phi}_{n}^{\top} \boldsymbol{u}_{n}\right) \left\{\boldsymbol{\phi}_{n+1}- \alpha'_n\rho_n \left(\boldsymbol{\phi}_{n}^{\top} \boldsymbol{\phi}_{n+1}\right) \boldsymbol{\phi}_{n} \right\} \label{imp_W_matrix} \\ 
 \boldsymbol{u}_{n+1} & =\boldsymbol{u}_{n}+ \beta'_n \rho_n \left(r_n + \gamma \boldsymbol{\phi}_{n+1}^{\top} \boldsymbol{w}_{n} - \boldsymbol{\phi}_{n}^{\top} \boldsymbol{w}_{n}\right)\boldsymbol{\phi}_{n} - \beta'_n \rho_n \boldsymbol{\phi}_{n} \boldsymbol{\phi}_{n}^{\top} \boldsymbol{u}_{n} \label{imp_U_matrix} 
\end{align}

\noindent where $\alpha'_n = \frac{\alpha_n}{1+\alpha_n\rho_n \|\boldsymbol{\phi}_{n}\|^2}$ and $\beta'_n = \frac{\beta_n}{1+\beta_n\rho_n \|\boldsymbol{\phi}_{n}\|^2}$. We first list notations and establish a linear stochastic approximation form of the implicit TDC update.

\begin{itemize}
\item $\pi_b$: behavior policy / $\pi_\star$: target policy
\item $\mu_{\pi_b}$ = stationary distribution of the Markov chain $\{(\boldsymbol{x}_{n}, a_n, \boldsymbol{x}_{n+1})\}_{n \ge 0}$ under behavior policy 
\item $\mu_{\pi_\star}$ = stationary distribution of the Markov chain $\{(\boldsymbol{x}_{n}, a_n, \boldsymbol{x}_{n+1})\}_{n \ge 0}$ under target policy 
\item $\rho(\boldsymbol{x},a)=\pi_\star(a \mid \boldsymbol{x}) / \pi_{b}(a \mid \boldsymbol{x}), \quad \rho_{\max} = \max_{\boldsymbol{x} \in \mathcal{X}, a \in \mathcal{A}} \rho(\boldsymbol{x}, a)$
\item $\boldsymbol{O}_n=\left(\boldsymbol{x}_{n}, a_n, \boldsymbol{x}_{n+1}\right)$ denotes the observation at time $n \in \mathbb{N}$
\item $\boldsymbol{A}=\mathbb{E}_{\mu_{\pi_{b}}}\left[\rho(\boldsymbol{x},a) \phi(\boldsymbol{x})\left\{\gamma \phi\left(x'\right)-\phi(\boldsymbol{x})\right\}^{\top}\right],$ \quad $\boldsymbol{A}_n = \rho_n \boldsymbol{\phi}_{n}\left(\gamma \boldsymbol{\phi}_{n+1}-\boldsymbol{\phi}_{n}\right)^{\top}$ 
\item $\boldsymbol{B}=-\gamma \mathbb{E}_{\mu_{\pi_{b}}}\left[\rho(\boldsymbol{x},a)\phi(x')\phi(\boldsymbol{x})^{\top}\right]$
\item $\boldsymbol{B}^{s}_n=-\gamma \rho_n \boldsymbol{\phi}_{n+1}\boldsymbol{\phi}_{n}^{\top}, \quad \boldsymbol{B}_{n}=-\gamma \rho_n \left\{\boldsymbol{\phi}_{n+1}- \alpha'_n \rho_n \left(\boldsymbol{\phi}_{n}^{\top} \boldsymbol{\phi}_{n+1}\right) \boldsymbol{\phi}_{n} \right\}\boldsymbol{\phi}_{n}^{\top}$
\item $\boldsymbol{C}=-\mathbb{E}_{\mu_{\pi_{b}}}\left[\rho(\boldsymbol{x}, a)\phi(\boldsymbol{x}) \phi(\boldsymbol{x})^{\top}\right], \quad \boldsymbol{C}_n = -\rho_n\boldsymbol{\phi}_{n} \boldsymbol{\phi}_{n}^{\top}$ 
\item $\boldsymbol{b}=\mathbb{E}_{\mu_{\pi_{b}}}\left[\rho(\boldsymbol{x}, a)r(\boldsymbol{x}, a) \phi(\boldsymbol{x})\right], \quad \boldsymbol{b}_{n}=\rho_n r_n \boldsymbol{\phi}_{n}$
\item $f_{1}\left(\boldsymbol{w}_{n}, \boldsymbol{O}_n\right)=\left(\boldsymbol{A}_n-\boldsymbol{B}^{s}_n \boldsymbol{C}^{-1} \boldsymbol{A}\right) \boldsymbol{w}_{n}+\left(\boldsymbol{b}_{n}-\boldsymbol{B}^{s}_n \boldsymbol{C}^{-1} \boldsymbol{b}\right)$, 
\item $\bar{f}_{1}\left(\boldsymbol{w}_{n}\right) = \left(\boldsymbol{A}-\boldsymbol{B} \boldsymbol{C}^{-1} \boldsymbol{A}\right) \boldsymbol{w}_{n}+\left(\boldsymbol{b}-\boldsymbol{B}\boldsymbol{C}^{-1}  \boldsymbol{b}\right)$
\item $\boldsymbol{v}_n = \boldsymbol{u}_{n} + \boldsymbol{C}^{-1}(\boldsymbol{b}+\boldsymbol{A}\boldsymbol{w}_{n})$ denotes a tracking error vector at time $n \in \mathbb{N}$ 
\item $g_{1}\left(\boldsymbol{v}_n, \boldsymbol{O}_n\right)=\boldsymbol{B}^{s}_n \boldsymbol{v}_n$, \quad $\bar{g}_{1}\left(\boldsymbol{v}_n\right)=\boldsymbol{B}\boldsymbol{v}_n$
\item $f_{2}\left(\boldsymbol{w}_{n}, \boldsymbol{O}_n\right)=\left(\boldsymbol{A}_n-\boldsymbol{C}_n \boldsymbol{C}^{-1} \boldsymbol{A}\right) \boldsymbol{w}_{n}+\left(\boldsymbol{b}_{n}-\boldsymbol{C}_n \boldsymbol{C}^{-1} \boldsymbol{b}\right)$
\item $g_{2}\left(\boldsymbol{v}_n, \boldsymbol{O}_n\right)=\boldsymbol{C}_n \boldsymbol{v}_n$, \quad $\bar{g}_{2}\left(\boldsymbol{v}_n\right)=\boldsymbol{C}\boldsymbol{v}_n$
\item $\lambda_{c}$ = minimum absolute eigenvalue of the matrix $\boldsymbol{C}$
\item $\tau_{\alpha_n} = \min\{i \ge 0: m\rho^i \le \alpha_n \}, \quad \tau_{\beta_n} = \min\{i \ge 0: m\rho^i \le \beta_n \}$
\end{itemize}

\noindent Based on the introduced notations,  we can rewrite \eqref{imp_W_matrix} and \eqref{imp_U_matrix} as
\begin{align*}
    \boldsymbol{w}_{n+1} &= \boldsymbol{w}_{n} + \alpha'_n \left(\boldsymbol{b}_{n} + \boldsymbol{A}_n \boldsymbol{w}_{n}\right) + \alpha_n \boldsymbol{B}_{n} \boldsymbol{u}_{n} \\
    \boldsymbol{u}_{n+1} &= \boldsymbol{u}_{n} + \beta'_n\left(\boldsymbol{b}_{n} + \boldsymbol{A}_n \boldsymbol{w}_{n} + \boldsymbol{C}_n \boldsymbol{u}_{n}\right).
\end{align*}
Corresponding projected implicit TDC algorithms are provided below
\begin{align}
 \boldsymbol{w}_{n+1} &= \prod_{R_{\boldsymbol{w}}}\left\{\boldsymbol{w}_{n} + \alpha'_n \left(\boldsymbol{b}_{n} + \boldsymbol{A}_n \boldsymbol{w}_{n}\right) + \alpha_n \boldsymbol{B}_{n} \boldsymbol{u}_{n}\right\} \label{proj_imp_W_matrix} \\
 \boldsymbol{u}_{n+1} &= \prod_{R_{\boldsymbol{u}}}\left\{\boldsymbol{u}_{n} + \beta'_n\left(\boldsymbol{b}_{n} + \boldsymbol{A}_n \boldsymbol{w}_{n} + \boldsymbol{C}_n \boldsymbol{u}_{n}\right)\right\} \label{proj_imp_U_matrix}.
\end{align}

\noindent To facilitate theoretical analysis, we rewrite the above projected linear stochastic approximation form into the following expressions:
\begin{align*}
    \boldsymbol{w}_{n+1} &= \Pi_{R_{\boldsymbol{w}}} \left\{\boldsymbol{w}_{n} + \alpha'_n \left(\boldsymbol{b}_{n} + \boldsymbol{A}_n \boldsymbol{w}_{n}\right) + \alpha_n \boldsymbol{B}_{n} \boldsymbol{u}_{n}\right\} \\
    &= \Pi_{R_{\boldsymbol{w}}} \left\{\boldsymbol{w}_{n} + \alpha'_n \left(\boldsymbol{b}_{n} + \boldsymbol{A}_n \boldsymbol{w}_{n} + \boldsymbol{B}^{s}_n \boldsymbol{u}_{n} \right) + (\alpha_n \boldsymbol{B}_{n} - \alpha'_n \boldsymbol{B}_{n}^s) \boldsymbol{u}_{n}\right\} \\
    & = \Pi_{R_{\boldsymbol{w}}} \left[\boldsymbol{w}_{n} + \alpha'_n \left\{f_1(\boldsymbol{w}_{n}, \boldsymbol{O}_n) + g_1(\boldsymbol{v}_n, \boldsymbol{O}_n) \right\} + (\alpha_n \boldsymbol{B}_{n} - \alpha'_n \boldsymbol{B}_{n}^s) \boldsymbol{u}_{n}\right].
\end{align*}
Plugging $\boldsymbol{u}_{n} = \boldsymbol{v}_n - \boldsymbol{C}^{-1}(\boldsymbol{b} +  \boldsymbol{A}\boldsymbol{w}_{n})$ to \eqref{proj_imp_U_matrix}, we also have
\begin{align*}
     \boldsymbol{v}_{n+1} &= \Pi_{R_{\boldsymbol{u}}} \left\{\boldsymbol{v}_n - \boldsymbol{C}^{-1}(\boldsymbol{b}+\boldsymbol{A}\boldsymbol{w}_{n}) + \beta'_n\left(\boldsymbol{b}_{n} + \boldsymbol{A}_n \boldsymbol{w}_{n} + \boldsymbol{C}_n \boldsymbol{u}_{n}\right)\right\} + \boldsymbol{C}^{-1}(\boldsymbol{b}+\boldsymbol{A}\boldsymbol{w}_{n+1})\\
     &= \Pi_{R_{\boldsymbol{u}}} \left[\boldsymbol{v}_n - \boldsymbol{C}^{-1}(\boldsymbol{b}+\boldsymbol{A}\boldsymbol{w}_{n}) + \beta'_n\left\{\boldsymbol{b}_{n} - \boldsymbol{C}_n\boldsymbol{C}^{-1}\boldsymbol{b} +  \boldsymbol{A}_n \boldsymbol{w}_{n} - \boldsymbol{C}_n\boldsymbol{C}^{-1}\boldsymbol{A}\boldsymbol{w}_{n} + \boldsymbol{C}_n \boldsymbol{u}_{n} + \boldsymbol{C}_n\boldsymbol{C}^{-1}(\boldsymbol{b}+\boldsymbol{A}\boldsymbol{w}_{n})\right\}\right] \\
     &\quad + \boldsymbol{C}^{-1}(\boldsymbol{b}+\boldsymbol{A}\boldsymbol{w}_{n+1}) \\
     & = \Pi_{R_{\boldsymbol{u}}} \left[\boldsymbol{v}_n + \beta'_n\left\{f_2(\boldsymbol{w}_{n}, \boldsymbol{O}_n) + g_2(\boldsymbol{v}_n, \boldsymbol{O}_n)\right\} - \boldsymbol{C}^{-1}(\boldsymbol{b}+\boldsymbol{A}\boldsymbol{w}_{n})\right] + \boldsymbol{C}^{-1}(\boldsymbol{b}+\boldsymbol{A}\boldsymbol{w}_{n+1}).
\end{align*}

\subsection{Technical lemmas for finite-time analysis for projected implicit TDC}

\noindent In this section, we establish preliminary lemmas used in the proof of projected implicit TDC's finite-time MSE bounds. 

\begin{lemma}\label{LEMMA:B_bound}
For all $n \in \mathbb{N}$,
\begin{align*}
    &(a) ~\|\boldsymbol{B}^{s}_n\| \le \gamma\rho_{\max}\phi_{\max}^2 \\
    &(b) ~\|\boldsymbol{B}_{n}\| \le \gamma\rho_{\max}\phi_{\max}^2 (1 + \alpha_1 \rho_{\max}\phi_{\max}^2) \\
    &(c) ~\|\boldsymbol{B}_{n} - \boldsymbol{B}^{s}_n\| \lesssim  \alpha_n \\
    &(d) ~ \|\alpha_n \boldsymbol{B}_{n} - \alpha'_n \boldsymbol{B}_{n}^s\| \lesssim \alpha_n^2.
\end{align*}
\end{lemma}
\begin{proof}
Recall the definition $\boldsymbol{B}^{s}_n=-\gamma \rho_n \boldsymbol{\phi}_{n+1}\boldsymbol{\phi}_{n}^{\top},~ \boldsymbol{B}_{n}=-\gamma \rho_n \left\{\boldsymbol{\phi}_{n+1}- \alpha'_n \rho_n \left(\boldsymbol{\phi}_{n}^{\top} \boldsymbol{\phi}_{n+1}\right) \boldsymbol{\phi}_{n} \right\}\boldsymbol{\phi}_{n}^{\top}$. Part (a) follows from the Cauchy-Schwarz inequality. For part (b), 
\begin{align*}
\left\|-\gamma \rho_n \left\{\boldsymbol{\phi}_{n+1}- \alpha'_n \rho_n \left(\boldsymbol{\phi}_{n}^{\top} \boldsymbol{\phi}_{n+1}\right) \boldsymbol{\phi}_{n} \right\}\boldsymbol{\phi}_{n}^{\top}\right\| &\le \gamma \rho_{\max}\left(\phi_{\max} + \alpha'_n \rho_{\max}\phi_{\max}^3 \right)\phi_{\max}\\ &\le \gamma \rho_{\max}\left(1 + \alpha_1 \rho_{\max}\phi_{\max}^2 \right)\phi_{\max}^2.
\end{align*}
For part (c),
$$
\|\boldsymbol{B}_{n} - \boldsymbol{B}^{s}_n\| = \left\|\gamma \rho^2_n \alpha'_n \left(\boldsymbol{\phi}_{n}^{\top} \boldsymbol{\phi}_{n+1}\right) \boldsymbol{\phi}_{n} \boldsymbol{\phi}_{n}^{\top} \right\| \le \gamma \rho_{\max}^2 \phi_{\max}^4 \alpha_n .
$$
For part (d),
\begin{align*}
\|\alpha_n \boldsymbol{B}_{n} - \alpha'_n \boldsymbol{B}_{n}^s\| &= \alpha_n \left\|\boldsymbol{B}_{n} - \frac{1}{1+\alpha_n \rho_n \|\boldsymbol{\phi}_{n}\|^2} \boldsymbol{B}_{n}^s \right\| \\ &\le \alpha_n \left\|\boldsymbol{B}_{n} - \boldsymbol{B}_{n}^s\right\| + \alpha_n\left(1-\frac{1}{1+\alpha_n \rho_n \|\boldsymbol{\phi}_{n}\|^2}\right) \left\|\boldsymbol{B}_{n}^s \right\| \\
&\le \gamma \rho_{\max}^2 \phi_{\max}^4 \alpha_n^2 + \frac{\alpha_n^2 \rho_n \|\boldsymbol{\phi}_{n}\|^2}{1+\alpha_n\rho_n\|\boldsymbol{\phi}_{n}\|^2} \gamma \rho_{\max} \phi_{\max}^2.
\end{align*}
where the second inequality is due to (a) and (c).
\end{proof}

\begin{lemma}\label{LEMMA:f1_bound} For any $\boldsymbol{w} \in \mathbb{R}^{d}$ such that $\|\boldsymbol{w}\| \leq R_{\boldsymbol{w}}$,
$$
\left\|f_{1}(\boldsymbol{w}, \boldsymbol{O}_{n})\right\| \leq K_{f_{1}}, \quad \left\|\bar f_{1}(\boldsymbol{w})\right\| \leq K_{f_{1}}, \quad \forall n \in \mathbb{N}.
$$ 
Here, $K_{f_{1}}$ is a positive constant independent of $w$.
\end{lemma}
\begin{proof} By the definition of $f_{1}(\boldsymbol{w}, \boldsymbol{O}_{n})$, and  $\lambda_{c}=\min |\lambda(C)|$, we obtain
$$
\begin{aligned}
\left\|f_{1}(\boldsymbol{w}, \boldsymbol{O}_{n})\right\| & =\left\|\left(\boldsymbol{A}_{n}-\boldsymbol{B}^{s}_{n} \boldsymbol{C}^{-1} \boldsymbol{A}\right) \boldsymbol{w}+\left(\boldsymbol{b}_{n}-\boldsymbol{B}^{s}_{n} \boldsymbol{C}^{-1} \boldsymbol{b}\right)\right\| \\
& \leq\left\|\left(\boldsymbol{A}_{n}-\boldsymbol{B}^{s}_{n} \boldsymbol{C}^{-1} \boldsymbol{A}\right) \boldsymbol{w}\right\|+\left\|\left(\boldsymbol{b}_{n}-\boldsymbol{B}^{s}_{n} \boldsymbol{C}^{-1} \boldsymbol{b}\right)\right\| \\
& \leq\left(\left\|\boldsymbol{A}_{n}\right\|+\left\|\boldsymbol{B}^{s}_{n}\right\|\left\|\boldsymbol{C}^{-1}\right\|\|\boldsymbol{A}\|\right)\|\boldsymbol{w}\|+\left\|\boldsymbol{b}_{n}\right\|+\left\|\boldsymbol{B}^{s}_{n}\right\|\left\|\boldsymbol{C}^{-1}\right\|\|\boldsymbol{b}\| \\
& \leq\left\{(1+\gamma) \rho_{\max} \phi_{\max}^2 +\frac{1}{\lambda_c} \gamma(1+\gamma) \rho_{\max}^2\phi_{\max}^4\right\}R_{\boldsymbol{w}}+\rho_{\max} r_{\max}\phi_{\max}+\frac{1}{\lambda_c} \gamma \rho_{\max}^2 r_{\max}\phi_{\max}^3 =: K_{f_{1}} .
\end{aligned}
$$
The same argument holds for $\bar f_{1}$.
\end{proof}

\begin{lemma}\label{LEMMA:f2_bound} For any $\boldsymbol{w} \in \mathbb{R}^{d}$ such that $\|\boldsymbol{w}\| \leq R_{\boldsymbol{w}},$
$$
\left\|f_{2}(\boldsymbol{w}, \boldsymbol{O}_{n})\right\| \leq K_{f_{2}}, \quad \forall n \in \mathbb{N}.
$$
Here, $K_{f_{2}}$ is a positive constant independent of $w$.
\end{lemma}
\begin{proof} By the definition of $f_{2}(\boldsymbol{w}, \boldsymbol{O}_{n})$, and $\lambda_{c}=\min |\lambda(\boldsymbol{C})|$, we obtain
$$
\begin{aligned}
\left\|f_{2}(\boldsymbol{w}, \boldsymbol{O}_{n})\right\| & =\left\|\left(\boldsymbol{A}_{n}-\boldsymbol{C}_{n} \boldsymbol{C}^{-1} \boldsymbol{A}\right) \boldsymbol{w}+\left(\boldsymbol{b}_{n}-\boldsymbol{C}_{n} \boldsymbol{C}^{-1} \boldsymbol{b}\right)\right\| \\
& \leq\left\|\left(\boldsymbol{A}_{n}-\boldsymbol{C}_{n} \boldsymbol{C}^{-1} \boldsymbol{A}\right) \boldsymbol{w}\right\|+\left\|\left(\boldsymbol{b}_{n}-\boldsymbol{C}_{n} \boldsymbol{C}^{-1} \boldsymbol{b}\right)\right\| \\
& \leq\left(\left\|\boldsymbol{A}_{n}\right\|+\left\|\boldsymbol{C}_{n}\right\|\left\|\boldsymbol{C}^{-1}\right\|\|\boldsymbol{A}\|\right)\|\boldsymbol{w}\|+\left\|\boldsymbol{b}_{n}\right\|+\left\|\boldsymbol{C}_{n}\right\|\left\|\boldsymbol{C}^{-1}\right\|\|\boldsymbol{b}\| \\
& \leq\left\{(1+\gamma) \rho_{\max}\phi_{\max}^2+\frac{1}{\lambda_c}(1+\gamma) \rho^2_{\max}\phi_{\max}^4\right\} R_{\boldsymbol{w}}+\rho_{\max} r_{\max}\phi_{\max}+\frac{1}{\lambda_c} \rho^2_{\max} r_{\max}\phi_{\max}^3 =: K_{f_{2}}
\end{aligned}
$$
\end{proof}

\begin{lemma}\label{LEMMA:g_bound}
For any $\boldsymbol{u}, \boldsymbol{w} \in \mathbb{R}^d$ such that $\|\boldsymbol{u}\| \le R_{\boldsymbol{u}}$ and $\|\boldsymbol{w}\| \le R_{\boldsymbol{w}}$, 
\begin{align*}
&(a) ~\|\boldsymbol{v}\| \leq R_{\boldsymbol{v}}:=R_{\boldsymbol{u}} + 2\|\boldsymbol{C}^{-1}\| \|\boldsymbol{A}\|R_{\boldsymbol{w}}\\
&(b) ~\left\|g_{1}(\boldsymbol{v}, \boldsymbol{O}_{n})\right\| \leq K_{g_{1}}, ~\left\|\bar g_{1}{\boldsymbol{v}}\right\| \leq K_{g_{1}}, \quad \forall n \in \mathbb{N}\\
&(c) ~ \left\|g_{2}(\boldsymbol{v}, \boldsymbol{O}_{n})\right\| \leq K_{g_{2}}, ~ \left\|\bar g_{2}{\boldsymbol{v}}\right\| \leq K_{g_{2}},  \quad \forall n \in \mathbb{N}.
\end{align*}
Here, $R_{\boldsymbol{v}}$, $K_{g_{1}}$ and $K_{g_{2}}$ are some positive constants independent of $w$ and $u$. 
\end{lemma}
\begin{proof}
For (a),
\begin{align*}
\|\boldsymbol{v}\| &= \left\|\boldsymbol{u} + \boldsymbol{C}^{-1}(\boldsymbol{b}+\boldsymbol{A}\boldsymbol{w})\right\| \\&\le \|\boldsymbol{u}\| + \|\boldsymbol{C}^{-1}(\boldsymbol{A}\boldsymbol{w}_\star+\boldsymbol{A}\boldsymbol{w})\| \\
&\le R_{\boldsymbol{u}} + 2\|\boldsymbol{C}^{-1}\| \|\boldsymbol{A}\|R_{\boldsymbol{w}}.
\end{align*}
For (b), by the definition of $g_{1}(\boldsymbol{v}, \boldsymbol{O}_{n})$, we obtain 
$$
\left\|g_{1}\left(\boldsymbol{v}, \boldsymbol{O}_n\right)\right\|=\left\|\boldsymbol{B}^{s}_n \boldsymbol{v}\right\| \leq\left\|\boldsymbol{B}^{s}_{n}\right\|\|\boldsymbol{v}\| \leq \gamma \rho_{\max}\phi_{\max}^2 R_{\boldsymbol{v}} = : K_{g_1}.
$$
The same argument holds for $\bar g_1$. For (c), by the definition of $g_{2}(\boldsymbol{v}, \boldsymbol{O}_{n})$, we obtain
$$
\left\|g_{2}(\boldsymbol{v}, \boldsymbol{O}_{n})\right\|=\left\|\boldsymbol{C}_{n} \boldsymbol{v}\right\| \leq\left\|\boldsymbol{C}_{n}\right\|\|\boldsymbol{v}\| \leq \rho_{\max}\phi_{\max}^2  R_{\boldsymbol{v}} =: K_{g_{2}}.
$$
The same argument holds for $\bar g_2$.
\end{proof}

\begin{lemma}\label{LEMMA:f1_zeta_bound}
Let $\zeta_{f_{1}}(\boldsymbol{w}, \boldsymbol{O}_{n}):= \langle f_1(\boldsymbol{w}, \boldsymbol{O}_n) - \bar f_1(\boldsymbol{w}), \boldsymbol{w} - \boldsymbol{w}_{\star} \rangle$. For any $\boldsymbol{w}, \boldsymbol{w}' \in \mathbb{R}^{d}$ such that $\|\boldsymbol{w}\| \leq R_{\boldsymbol{w}}$ and $\|\boldsymbol{w}'\| \leq R_{\boldsymbol{w}}$, we have 
\begin{align*}
&(a) ~\left|\zeta_{f_{1}}(\boldsymbol{w}, \boldsymbol{O}_{n})\right| \leq 4 R_{\boldsymbol{w}} K_{f_{1}}, \quad \forall n \in \mathbb{N}\\
&(b)~\left|\zeta_{f_{1}}(\boldsymbol{w}, \boldsymbol{O}_{n})-\zeta_{f_{1}}\left(\boldsymbol{w}', \boldsymbol{O}_{n}\right)\right| \lesssim \left\|\boldsymbol{w}-\boldsymbol{w}'\right\|, \quad \forall n \in \mathbb{N}.
\end{align*}
\end{lemma}

\begin{proof} For (a), we get
$$
\left|\zeta_{f_{1}}(\boldsymbol{w}, \boldsymbol{O}_{n})\right| \leq\left(\left\|f_{1}(\boldsymbol{w}, \boldsymbol{O}_{n})\right\|+\left\|\bar{f}_{1}(\boldsymbol{w})\right\|\right)\left(\|\boldsymbol{w}\|+\left\|\boldsymbol{w}_{\star}\right\|\right) \leq 4 R_{\boldsymbol{w}} K_{f_{1}}.
$$
For (b), we derive the bound as follows
\begin{align*}
& \left|\zeta_{f_{1}}(\boldsymbol{w}, \boldsymbol{O}_{n})-\zeta_{f_{1}}\left(\boldsymbol{w}', \boldsymbol{O}_{n}\right)\right| \\
& =\left|\left\langle f_{1}(\boldsymbol{w}, \boldsymbol{O}_{n})-\bar{f}_1(\boldsymbol{w}), \boldsymbol{w}-\boldsymbol{w}_{\star}\right\rangle-\left\langle f_{1}\left(\boldsymbol{w}', \boldsymbol{O}_{n}\right)-\bar{f}_{1}\left(\boldsymbol{w}'\right), \boldsymbol{w}'-\boldsymbol{w}_{\star}\right\rangle\right| \\
& \leq\left\|\boldsymbol{w}-\boldsymbol{w}_{\star}\right\|\left\|f_{1}(\boldsymbol{w}, \boldsymbol{O}_{n})-\bar{f}_{1}(\boldsymbol{w})-f_{1}\left(\boldsymbol{w}', \boldsymbol{O}_{n}\right)+\bar{f}_{1}\left(\boldsymbol{w}'\right)\right\|+\left\|f_{1}\left(\boldsymbol{w}', \boldsymbol{O}_{n}\right)-\bar{f}_{1}\left(\boldsymbol{w}'\right)\right\|\left\|\boldsymbol{w}-\boldsymbol{w}'\right\| \\
& \leq\left\|\boldsymbol{w}-\boldsymbol{w}_{\star}\right\|\left(\left\|f_{1}(\boldsymbol{w}, \boldsymbol{O}_{n})-f_{1}\left(\boldsymbol{w}', \boldsymbol{O}_{n}\right)\right\|+\left\|\bar{f}_{1}\left(\boldsymbol{w}'\right)-\bar{f}_{1}(\boldsymbol{w})\right\|\right)+\left\|f_{1}\left(\boldsymbol{w}', \boldsymbol{O}_{n}\right)-\bar{f}_{1}\left(\boldsymbol{w}'\right)\right\|\left\|\boldsymbol{w}-\boldsymbol{w}'\right\| \\
& \leq 2 R_{\boldsymbol{w}}\left(\left\|\left(\boldsymbol{A}_{n}-\boldsymbol{B}^{s}_{n} \boldsymbol{C}^{-1} \boldsymbol{A}\right)\left(\boldsymbol{w}-\boldsymbol{w}'\right)\right\|+\left\|\left(\boldsymbol{A}-\boldsymbol{B} \boldsymbol{C}^{-1} \boldsymbol{A}\right)\left(\boldsymbol{w}'-\boldsymbol{w}\right)\right\|\right)+2 K_{f_{1}}\left\|\boldsymbol{w}-\boldsymbol{w}'\right\| \\
& \leq 4 R_{\boldsymbol{w}}(1 + \gamma) \rho_{\max} \phi_{\max}^2 \left(1 + \gamma \rho_{\max}\phi_{\max}^2/\lambda_{c}\right)\left\|\boldsymbol{w}-\boldsymbol{w}'\right\|+2 K_{f_{1}}\left\|\boldsymbol{w}-\boldsymbol{w}'\right\|.
\end{align*}
\end{proof}

\begin{lemma}\label{LEMMA:f2_zeta_bound}
Let $\zeta_{f_{2}}\left(\boldsymbol{w}, \boldsymbol{v}, \boldsymbol{O}_{n}\right):= \langle f_2(\boldsymbol{w},\boldsymbol{O}_n), \boldsymbol{v} \rangle$. For any $\boldsymbol{w}, \boldsymbol{w}', \boldsymbol{v},  \boldsymbol{v}' \in \mathbb{R}^{d}$ such that $\|\boldsymbol{w}\| \leq R_{\boldsymbol{w}}$, $\| \boldsymbol{w}'\| \leq R_{\boldsymbol{w}}$, $\|\boldsymbol{v}\| \leq R_{\boldsymbol{v}}$ and $\| \boldsymbol{v}'\| \leq R_{\boldsymbol{v}}$, 
\begin{align*}
&(a)~\left|\zeta_{f_{2}}\left(\boldsymbol{w}, \boldsymbol{v}, \boldsymbol{O}_{n}\right)\right| \leq  K_{f_{2}}R_{\boldsymbol{v}}, \quad \forall n \in \mathbb{N}\\
&(b)~\left|\zeta_{f_{2}}\left(\boldsymbol{w}, \boldsymbol{v}, \boldsymbol{O}_{n}\right)-\zeta_{f_{2}}\left(\boldsymbol{w}', \boldsymbol{v}', \boldsymbol{O}_{n}\right)\right| \lesssim \left\|\boldsymbol{w}-\boldsymbol{w}'\right\| + \left\|\boldsymbol{v}-\boldsymbol{v}'\right\|, \quad \forall n \in \mathbb{N}.
\end{align*}
\end{lemma}
\begin{proof}
For (a), by the definition, we have 
$$
\left|\zeta_{f_{2}}\left(\boldsymbol{w}, \boldsymbol{v}, \boldsymbol{O}_{n}\right)\right|=\left|\left\langle f_{2}(\boldsymbol{w}, \boldsymbol{O}_{n}), \boldsymbol{v}\right\rangle\right| \leq \left\|f_{2}(\boldsymbol{w}, \boldsymbol{O}_{n})\right\|\|\boldsymbol{v}\| \leq K_{f_{2}}R_{\boldsymbol{v}}.
$$
For (b), we derive the bound as follows
\begin{align*}
\left|\zeta_{f_{2}}\left(\boldsymbol{w}, \boldsymbol{v}, \boldsymbol{O}_{n}\right)-\zeta_{f_{2}}\left(\boldsymbol{w}', \boldsymbol{v}', \boldsymbol{O}_{n}\right)\right| & =\left|\left\langle f_{2}(\boldsymbol{w}, \boldsymbol{O}_{n}), \boldsymbol{v}\right\rangle-\left\langle f_{2}\left(\boldsymbol{w}', \boldsymbol{O}_{n}\right), \boldsymbol{v}'\right\rangle\right| \\
& \leq\|\boldsymbol{v}\|\left\|f_{2}(\boldsymbol{w}, \boldsymbol{O}_{n})-f_{2}\left(\boldsymbol{w}', \boldsymbol{O}_{n}\right)\right\|+\left\|f_{2}\left(\boldsymbol{w}', \boldsymbol{O}_{n}\right)\right\|\left\|\boldsymbol{v}-\boldsymbol{v}'\right\| \\
& \leq R_{\boldsymbol{v}}\left\|\left(\boldsymbol{A}_{n}-\boldsymbol{C}_{n} \boldsymbol{C}^{-1} \boldsymbol{A}\right)\left(\boldsymbol{w}-\boldsymbol{w}'\right)\right\|+ K_{f_{2}}\left\|\boldsymbol{v}-\boldsymbol{v}'\right\| \\
& \leq R_{\boldsymbol{v}}(1+\gamma)\left(\rho_{\max}\phi_{\max}^2+\rho_{\max}^2\phi_{\max}^4/\lambda_c\right)\left\|\boldsymbol{w}-\boldsymbol{w}'\right\|+ K_{f_{2}}\left\|\boldsymbol{v}-\boldsymbol{v}'\right\|. 
\end{align*}
\end{proof}

\begin{lemma}\label{LEMMA:g2_zeta_bound} Let $\zeta_{g_{2}}(\boldsymbol{v}, \boldsymbol{O}_{n}):= \langle g_2(\boldsymbol{v}, \boldsymbol{O}_n) - \bar g_2(\boldsymbol{v}), \boldsymbol{v} \rangle$, where $\boldsymbol{v} = \boldsymbol{u} + \boldsymbol{C}^{-1}(\boldsymbol{b}+\boldsymbol{A}\boldsymbol{w})$. For all $\boldsymbol{v}, \boldsymbol{v}' \in \mathbb{R}^{d}$ such that $\|\boldsymbol{v}\| \leq R_{\boldsymbol{v}}$ and $\|\boldsymbol{v}'\| \leq R_{\boldsymbol{v}}$, we have 
\begin{align*}
&(a) ~\left|\zeta_{g_{2}}(\boldsymbol{v}, \boldsymbol{O}_{n})\right| \leq 2 K_{g_{2}}R_{\boldsymbol{v}}, \quad \forall n \in \mathbb{N}\\
& (b) ~\left|\zeta_{g_{2}}(\boldsymbol{v}, \boldsymbol{O}_{n})-\zeta_{g_{2}}\left(\boldsymbol{v}', \boldsymbol{O}_{n}\right)\right| \lesssim \left\|\boldsymbol{v}-\boldsymbol{v}'\right\|, \quad \forall n \in \mathbb{N}.
\end{align*}
\end{lemma}
\begin{proof} 
For part (a), by the definition of $\zeta_{g_{2}}(\boldsymbol{v}, \boldsymbol{O}_{n})$, we have $$
\left|\zeta_{g_{2}}(\boldsymbol{v}, \boldsymbol{O}_{n})\right|= \left|\left\langle g_{2}(\boldsymbol{v}, \boldsymbol{O}_{n})-\bar{g}_{2}(\boldsymbol{v}), \boldsymbol{v}\right\rangle\right| \leq\left(\left\|g_{2}(\boldsymbol{v}, \boldsymbol{O}_{n})\right\|+\left\|\bar{g}_{2}(\boldsymbol{v})\right\|\right)\|\boldsymbol{v}\| \leq 2 K_{g_{2}}R_{\boldsymbol{v}}.
$$
For part (b), we derive the bound as follows.
\begin{align*}
& \left|\zeta_{g_{2}}(\boldsymbol{v}, \boldsymbol{O}_{n})-\zeta_{g_{2}}\left(\boldsymbol{v}', \boldsymbol{O}_{n}\right)\right| \\
& =\left|\left\langle g_{2}(\boldsymbol{v}, \boldsymbol{O}_{n})-\bar{g}_{2}(\boldsymbol{v}), \boldsymbol{v}\right\rangle-\left\langle g_{2}\left(\boldsymbol{v}', \boldsymbol{O}_{n}\right)-\bar{g}_{2}\left(\boldsymbol{v}'\right), \boldsymbol{v}'\right\rangle\right| \\
& \leq\|\boldsymbol{v}\|\left\|g_{2}(\boldsymbol{v}, \boldsymbol{O}_{n})-\bar{g}_{2}(\boldsymbol{v})-g_{2}\left(\boldsymbol{v}', \boldsymbol{O}_{n}\right)+\bar{g}_{2}\left(\boldsymbol{v}'\right)\right\|+\left\|g_{2}\left(\boldsymbol{v}', \boldsymbol{O}_{n}\right)-\bar{g}_{2}\left(\boldsymbol{v}'\right)\right\|\left\|\boldsymbol{v}-\boldsymbol{v}'\right\| \\
& =\|\boldsymbol{v}\|\left\|\left(\boldsymbol{C}_{n}-\boldsymbol{C}\right)\left(\boldsymbol{v}-\boldsymbol{v}'\right)\right\|+\left\|g_{2}\left(\boldsymbol{v}', \boldsymbol{O}_{n}\right)-\bar{g}_{2}\left(\boldsymbol{v}'\right)\right\|\left\|\boldsymbol{v}-\boldsymbol{v}'\right\| \\
& \leq  R_{\boldsymbol{v}}\left(\left\|\boldsymbol{C}_{n}\right\|+\|\boldsymbol{C}\|\right)\left\|\boldsymbol{v}-\boldsymbol{v}'\right\|+2 K_{g_{2}}\left\|\boldsymbol{v}-\boldsymbol{v}'\right\| \\
& \leq 2 R_{\boldsymbol{v}} \rho_{\max}\phi_{\max}^2\left\|\boldsymbol{v}-\boldsymbol{v}'\right\|+2 K_{g_{2}}\left\|\boldsymbol{v}-\boldsymbol{v}'\right\|. 
\end{align*}
\end{proof}

\begin{lemma}\label{LEMMA:expect_zeta_f1_bound} For $n \in \mathbb{N}$, suppose $i \le n$ and $(\alpha_n)_{n \in \mathbb{N}}$ is a non-increasing sequence. If $i \leq \tau_{\alpha_n},$
$$
\mathbb{E}_{\mu_{\pi_b}}\left\{\zeta_{f_{1}}\left(\boldsymbol{w}_i, \boldsymbol{O}_{i}\right)\right\} \lesssim \tau_{\alpha_n}.
$$ 
Otherwise,
$$
\mathbb{E}_{\mu_{\pi_b}}\left\{\zeta_{f_{1}}\left(\boldsymbol{w}_{i}, \boldsymbol{O}_{i}\right)\right\} \lesssim \alpha_{n} + \left(\alpha_{i-\tau_{\alpha_n}} + \alpha_{i-\tau_{\alpha_n}}^2\right)\tau_{\alpha_n}.
$$
\end{lemma}
\begin{proof} Note that for any $i \in \mathbb{N}$,
\begin{align}
\left\|\boldsymbol{w}_{i+1}-\boldsymbol{w}_{i}\right\| & =\left\|\Pi_{R_{\boldsymbol{w}}} \left[\boldsymbol{w}_i + \alpha'_i \left\{f_1(\boldsymbol{w}_i, \boldsymbol{O}_{i}) + g_1(\boldsymbol{v}_{i}, \boldsymbol{O}_{i}) \right\} + \alpha_i \boldsymbol{B}_i \boldsymbol{u}_i - \alpha'_i \boldsymbol{B}_i^s \boldsymbol{u}_i\right]-\Pi_{R_{\boldsymbol{w}}} \boldsymbol{w}_{i}\right\| \nonumber \\
& \leq\left\|\boldsymbol{w}_i + \alpha'_i \left\{f_1(\boldsymbol{w}_i, \boldsymbol{O}_{i}) + g_1(\boldsymbol{v}_{i}, \boldsymbol{O}_{i}) \right\} + \alpha_i \boldsymbol{B}_i \boldsymbol{u}_i - \alpha'_i \boldsymbol{B}_i^s \boldsymbol{u}_i-\boldsymbol{w}_{i}\right\| \nonumber \\
& \leq \alpha_{i}\left\|f_{1}\left(\boldsymbol{w}_{i}, \boldsymbol{O}_{i}\right)+g_{1}\left(\boldsymbol{v}_{i}, \boldsymbol{O}_{i}\right)\right\| + \|\alpha_i \boldsymbol{B}_i \boldsymbol{u}_i - \alpha'_i \boldsymbol{B}_i^s \boldsymbol{u}_i \| \nonumber \\
& \lesssim \alpha_i + \alpha_i^2 \label{TDC_w_recursion}
\end{align}
where the last inequality follows from Lemma \ref{LEMMA:B_bound}, Lemma \ref{LEMMA:f1_bound}, and Lemma \ref{LEMMA:g_bound}. Applying the Lipschitz continuity of $\zeta_{f_{1}}\left(\boldsymbol{w}, \boldsymbol{O}_{i}\right)$, obtained in part (b) of Lemma \ref{LEMMA:f1_zeta_bound}, for $i > \tau_{\alpha_n}$, it follows that
$$
\left|\zeta_{f_{1}}\left(\boldsymbol{w}_{i}, \boldsymbol{O}_{i}\right)-\zeta_{f_{1}}\left(\boldsymbol{w}_{i-\tau_{\alpha_n}}, \boldsymbol{O}_{i}\right)\right| \lesssim \left\|\boldsymbol{w}_i-\boldsymbol{w}_{i-\tau_{\alpha_n}}\right\| \lesssim \sum_{k=i-\tau_{\alpha_n}}^{i-1} \alpha_{k} + \sum_{k=i-\tau_{\alpha_n}}^{i-1} \alpha_{k}^2.
$$
We now provide an upper bound for $\mathbb{E}_{\mu_{\pi_b}}\left\{\zeta_{f_{1}}\left(\boldsymbol{w}_{i-\tau_{\alpha_n}}, \boldsymbol{O}_{i}\right)\right\}$. To this end, we define $\boldsymbol{w}_{i-\tau_{\alpha_n}}'$ and $\boldsymbol{O}_{i}'=\left(\boldsymbol{x}_{i}', a_{i}', \boldsymbol{x}_{i+1}'\right)$, which are drawn independently from the marginal distributions of $\boldsymbol{w}_{i-\tau_{\alpha_n}}$ and $\boldsymbol{O}_{i}$. From part (a) of Lemma \ref{LEMMA:f1_zeta_bound} and Lemma \ref{APPEND_LEMMA:BHAND9}, we have
$$
\mathbb{E}_{\mu_{\pi_b}}\left\{\zeta_{f_{1}}\left(\boldsymbol{w}_{i-\tau_{\alpha_n}}, \boldsymbol{O}_{i}\right)\right\} \leq\left|\mathbb{E}_{\mu_{\pi_b}}\left\{\zeta_{f_{1}}\left(\boldsymbol{w}_{i-\tau_{\alpha_n}}, \boldsymbol{O}_{i}\right)\right\}-\mathbb{E}\left\{\zeta_{f_{1}}\left(\boldsymbol{w}_{i-\tau_{\alpha_n}}', \boldsymbol{O}_{i}'\right)\right\}\right| \leq 8 R_{\boldsymbol{w}} K_{f_{1}} m \rho^{\tau_{\alpha_n}}.
$$
It follows that
\begin{align*}
\mathbb{E}_{\mu_{\pi_b}}\left\{\zeta_{f_{1}}\left(\boldsymbol{w}_{i}, \boldsymbol{O}_{i}\right)\right\} & \lesssim \mathbb{E}_{\mu_{\pi_b}}\left\{\zeta_{f_{1}}\left(\boldsymbol{w}_{i-\tau_{\alpha_n}}, \boldsymbol{O}_{i}\right)\right\} + \sum_{k=i-\tau_{\alpha_n}}^{i-1} \alpha_{k} + \sum_{k=i-\tau_{\alpha_n}}^{i-1} \alpha_{k}^2 \\
& \lesssim 8 R_{\boldsymbol{w}} K_{f_{1}} m \rho^{\tau_{\alpha_n}} + \left(\alpha_{i-\tau_{\alpha_n}} +  \alpha_{i-\tau_{\alpha_n}}^2\right) \tau_{\alpha_n} \\
& \lesssim  \alpha_n + \left(\alpha_{i-\tau_{\alpha_n}} +  \alpha_{i-\tau_{\alpha_n}}^2\right) \tau_{\alpha_n}.
\end{align*}
On the other hand, if $i \leq \tau_{\alpha_n}$, 
\begin{align*}
\mathbb{E}_{\mu_{\pi_b}}\left\{\zeta_{f_{1}}\left(\boldsymbol{w}_{i}, \boldsymbol{O}_{i}\right)\right\} \lesssim \mathbb{E}_{\mu_{\pi_b}}\left\{\zeta_{f_{1}}\left(\boldsymbol{w}_{1}, \boldsymbol{O}_{i}\right)\right\} + \sum_{k=1}^{i-1} \alpha_{k} + \sum_{k=1}^{i-1} \alpha_{k}^2 \lesssim \tau_{\alpha_n}.
\end{align*}
\end{proof}

\begin{lemma}\label{LEMMA:expect_zeta_f2_bound}
For $n \in \mathbb{N}$, suppose $i \le n$. Furthermore, $(\alpha_n)_{n \in \mathbb{N}}$ and $(\beta_n)_{n \in \mathbb{N}}$ are non-increasing sequences. If $\alpha_n/\beta_n$ is a non-increasing sequence, for $i \leq \tau_{\beta_n},$
$$
\mathbb{E}_{\mu_{\pi_b}}\left\{\zeta_{f_{2}}\left(\boldsymbol{w}_{i}, \boldsymbol{v}_{i}, \boldsymbol{O}_{i}\right)\right\} \lesssim \tau_{\beta_n}. 
$$ 
Otherwise,
$$
\mathbb{E}_{\mu_{\pi_b}}\left\{\zeta_{f_{2}}\left(\boldsymbol{w}_{i}, \boldsymbol{v}_{i}, \boldsymbol{O}_{i}\right)\right\} \lesssim  \beta_{n} + \tau_{\beta_n} \beta_{i-\tau_{\beta_n}}.
$$
\end{lemma}

\begin{proof}
Notice that
\begin{align}
& \left\|\boldsymbol{v}_{i+1}-\boldsymbol{v}_{i}\right\| =\left\|\Pi_{R_{\boldsymbol{u}}}\left[\boldsymbol{v}_{i}+\beta'_{i}\left\{f_{2}\left(\boldsymbol{w}_{i}, \boldsymbol{O}_{i}\right)+g_{2}\left(\boldsymbol{v}_{i}, \boldsymbol{O}_{i}\right)\right\}-\boldsymbol{C}^{-1}\left(\boldsymbol{b}+\boldsymbol{A} \boldsymbol{w}_{i}\right)\right]+\boldsymbol{C}^{-1}\left(\boldsymbol{b}+\boldsymbol{A} \boldsymbol{w}_{i+1}\right)-\boldsymbol{v}_{i}\right\| \nonumber \\
& =\left\|\Pi_{R_{\boldsymbol{u}}}\left[\boldsymbol{v}_{i}+\beta'_{i}\left\{f_{2}\left(\boldsymbol{w}_{i}, \boldsymbol{O}_{i}\right)+g_{2}\left(\boldsymbol{v}_{i}, \boldsymbol{O}_{i}\right)\right\}-\boldsymbol{C}^{-1}\left(\boldsymbol{b}+\boldsymbol{A} \boldsymbol{w}_{i}\right)\right]+\boldsymbol{C}^{-1}\left(\boldsymbol{b}+\boldsymbol{A} \boldsymbol{w}_{i}\right)-\boldsymbol{v}_{i}+\boldsymbol{C}^{-1} \boldsymbol{A}\left(\boldsymbol{w}_{i+1}-\boldsymbol{w}_{i}\right)\right\|\nonumber \\
& \leq\left\|\Pi_{R_{\boldsymbol{u}}}\left[\boldsymbol{v}_{i}+\beta'_{i}\left\{f_{2}\left(\boldsymbol{w}_{i}, \boldsymbol{O}_{i}\right)+g_{2}\left(\boldsymbol{v}_{i}, \boldsymbol{O}_{i}\right)\right\}-\boldsymbol{C}^{-1}\left(\boldsymbol{b}+\boldsymbol{A} \boldsymbol{w}_{i}\right)\right]-\Pi_{R_{\boldsymbol{u}}}\left\{\boldsymbol{v}_{i}-\boldsymbol{C}^{-1}\left(\boldsymbol{b}+\boldsymbol{A} \boldsymbol{w}_{i}\right)\right\}\right\| \nonumber \\
& \quad+\left\|\boldsymbol{C}^{-1} \boldsymbol{A}\left(\boldsymbol{w}_{i+1}-\boldsymbol{w}_{i}\right)\right\| \nonumber \\
& \leq\left\|\boldsymbol{v}_{i}+\beta'_{i}\left\{f_{2}\left(\boldsymbol{w}_{i}, \boldsymbol{O}_{i}\right)+g_{2}\left(\boldsymbol{v}_{i}, \boldsymbol{O}_{i}\right)\right\}-\boldsymbol{C}^{-1}\left(\boldsymbol{b}+\boldsymbol{A} \boldsymbol{w}_{i}\right)-\left\{\boldsymbol{v}_{i}-\boldsymbol{C}^{-1}\left(\boldsymbol{b}+\boldsymbol{A} \boldsymbol{w}_{i}\right)\right\}\right\|+\left\|\boldsymbol{C}^{-1} \boldsymbol{A}\left(\boldsymbol{w}_{i+1}-\boldsymbol{w}_{i}\right)\right\| \nonumber \\
& \leq \beta_{i}\left\|f_{2}\left(\boldsymbol{w}_{i}, \boldsymbol{O}_{i}\right)+g_{2}\left(\boldsymbol{v}_{i}, \boldsymbol{O}_{i}\right)\right\|+\left\|\boldsymbol{C}^{-1} \boldsymbol{A}\left(\boldsymbol{w}_{i+1}-\boldsymbol{w}_{i}\right)\right\| \nonumber \\
&\lesssim \beta_i + \alpha_i + \alpha_i^2 \nonumber \\
&\le \beta_i\left(1 + \frac{\alpha_1}{\beta_1} + \frac{\alpha_1^2}{\beta_1}\right) 
\label{TDC_v_recursion}
\end{align}
where the first inequality follows from the fact that $\|\boldsymbol{u}_i\| = \|\boldsymbol{v}_{i} - \boldsymbol{C}^{-1}(\boldsymbol{b}+\boldsymbol{A}\boldsymbol{w}_i)\| \le R_{\boldsymbol{u}}$, the fourth inequality is thanks to \eqref{TDC_w_recursion} and the last inequality is due to 
$
\frac{\alpha_i}{\beta_i} \le \frac{\alpha_1}{\beta_1} ~\text{and}~\frac{\alpha^2_i}{\beta_i} \le \frac{\alpha_1^2}{\beta_1}.
$
Applying the Lipschitz continuous property of $\zeta_{f_{2}}\left(\boldsymbol{w}, \boldsymbol{v}, \boldsymbol{O}_{i}\right)$ in part (b) of Lemma \ref{LEMMA:f2_zeta_bound}, for $i > \tau_{\beta_n}$ it follows that
$$
\begin{aligned}
\left|\zeta_{f_{2}}\left(\boldsymbol{w}_{i}, \boldsymbol{v}_{i}, \boldsymbol{O}_{i}\right)-\zeta_{f_{2}}\left(\boldsymbol{w}_{i-\tau_{\beta_n}}, \boldsymbol{v}_{i-\tau_{\beta_n}}, \boldsymbol{O}_{i}\right)\right| &\lesssim \left\|\boldsymbol{w}_{i}-\boldsymbol{w}_{i-\tau_{\beta_n}}\right\| + \left\|\boldsymbol{v}_{i}-\boldsymbol{v}_{i-\tau_{\beta_n}}\right\| \\
& \quad \lesssim  \sum_{k=i-\tau_{\beta_n}}^{i-1} \alpha_{k} +  \sum_{k=i-\tau_{\beta_n}}^{i-1} \alpha^2_{k} + \sum_{k=i-\tau_{\beta_n}}^{i-1} \beta_{k}
\end{aligned}
$$
where the second inequality follows from \eqref{TDC_w_recursion} and \eqref{TDC_v_recursion}. The next step is to provide an upper bound for $\mathbb{E}_{\mu_{\pi_b}}\left\{\zeta_{f_{2}}\left(\boldsymbol{w}_{i-\tau_{\beta_n}}, \boldsymbol{v}_{i-\tau_{\beta_n}}, \boldsymbol{O}_{i}\right)\right\}$. To this end, we define $(\boldsymbol{w}_{i-\tau_{\beta_n}}', \boldsymbol{v}_{i-\tau_{\beta_n}}')$ and $\boldsymbol{O}_{i}'=\left(\boldsymbol{x}_{i}', a_{i}', \boldsymbol{x}_{i+1}'\right)$ which are independently drawn from the marginal distributions of $\left(\boldsymbol{w}_{i-\tau_{\beta_n}}, \boldsymbol{v}_{i-\tau_{\beta_n}}\right)$ and $\boldsymbol{O}_{i}$. It can be shown that $\mathbb{E}\left\{\zeta_{f_{2}}\left(\boldsymbol{w}_{i-\tau_{\beta_n}}', \boldsymbol{v}_{i-\tau_{\beta_n}}', \boldsymbol{O}_{i}'\right)\right\}=0$. By Lemma \ref{LEMMA:f2_zeta_bound} and Lemma \ref{APPEND_LEMMA:BHAND9}, we get
\begin{align*}
   &\mathbb{E}_{\mu_{\pi_b}}\left\{\zeta_{f_{2}}\left(\boldsymbol{w}_{i-\tau_{\beta_n}}, \boldsymbol{v}_{i-\tau_{\beta_n}}, \boldsymbol{O}_{i}\right)\right\} \\&\leq\left|\mathbb{E}_{\mu_{\pi_b}}\left\{\zeta_{f_{2}}\left(\boldsymbol{w}_{i-\tau_{\beta_n}}, \boldsymbol{v}_{i-\tau_{\beta_n}}, \boldsymbol{O}_{i}\right)\right\}-\mathbb{E}\left\{\zeta_{f_{2}}\left(\boldsymbol{w}_{i-\tau_{\beta_n}}', \boldsymbol{v}_{i-\tau_{\beta_n}}', \boldsymbol{O}_{i}'\right)\right\}\right|\\
   &\lesssim m \rho^{\tau_{\beta_n}}. 
\end{align*}
It follows that
\begin{align*}
&\mathbb{E}_{\mu_{\pi_b}}\left\{\zeta_{f_{2}}\left(\boldsymbol{w}_{i}, \boldsymbol{v}_{i}, \boldsymbol{O}_{i}\right)\right\}\\ & \lesssim \mathbb{E}_{\mu_{\pi_b}}\left\{\zeta_{f_{2}}\left(\boldsymbol{w}_{i-\tau_{\beta_n}}, \boldsymbol{v}_{i-\tau_{\beta_n}}, \boldsymbol{O}_{i}\right)\right\} + \sum_{k=i-\tau_{\beta_n}}^{i-1} \alpha_{k} + \sum_{k=i-\tau_{\beta_n}}^{i-1} \alpha^2_{k} + \sum_{k=i-\tau_{\beta_n}}^{i-1} \beta_{k} \\
& \lesssim  m \rho^{\tau_{\beta_n}} +  \tau_{\beta_n} \alpha_{i-\tau_{\beta_n}} + \tau_{\beta_n} \alpha_{i-\tau_{\beta_n}}^2 + \tau_{\beta_n} \beta_{i-\tau_{\beta_n}} \\
& \lesssim  \beta_{n}+\left[\max \left\{1, \frac{\alpha_1}{\beta_1}\right\} +\max \left\{1, \frac{\alpha_1^2}{\beta_1}\right\} + 1\right] \tau_{\beta_n} \beta_{i-\tau_{\beta_n}} 
\end{align*}
where the last inequality is thanks to $\frac{\alpha^2_{i-\tau_{\beta_n}}}{\beta_{i-\tau_{\beta_n}}} \le \frac{\alpha_1^2}{\beta_1}.$
Similarly, for $i \leq \tau_{\beta_n}$, it follows that
\begin{align*}
\mathbb{E}_{\mu_{\pi_b}}\left\{\zeta_{f_{2}}\left(\boldsymbol{w}_{i}, \boldsymbol{v}_{i}, \boldsymbol{O}_{i}\right)\right\} & \lesssim \mathbb{E}_{\mu_{\pi_b}}\left\{\zeta_{f_{2}}\left(\boldsymbol{w}_{1}, \boldsymbol{v}_1, \boldsymbol{O}_{i}\right)\right\} + \sum_{k=1}^{i-1} \alpha_{k} + \sum_{k=1}^{i-1} \alpha^2_{k} + \sum_{k=1}^{i-1} \beta_{k} \lesssim \tau_{\beta_n}. 
\end{align*}
\end{proof}

\begin{lemma}\label{LEMMA:expect_zeta_g2_bound} For $n \in \mathbb{N}$, suppose $i \le n$ and $(\beta_n)_{n \in \mathbb{N}}$ is a non-increasing sequence. If $i \leq \tau_{\beta_n}$, 
$$
\mathbb{E}_{\mu_{\pi_b}}\left\{\zeta_{g_{2}}\left(\boldsymbol{v}_{i}, \boldsymbol{O}_{i}\right)\right\}\lesssim \tau_{\beta_n}. 
$$ 
Otherwise,
$$
\mathbb{E}_{\mu_{\pi_b}}\left\{\zeta_{g_{2}}\left(\boldsymbol{v}_{i}, \boldsymbol{O}_{i}\right)\right\} \lesssim  \beta_{n} +  \tau_{\beta_n} \beta_{i-\tau_{\beta_n}}.
$$
\end{lemma}
\begin{proof}
Applying the Lipschitz continuous property of $\zeta_{g_{2}}\left(\boldsymbol{v}, \boldsymbol{O}_{i}\right)$ established in part (b) of Lemma \ref{LEMMA:g2_zeta_bound} with \eqref{TDC_v_recursion}, for $i > \tau_{\beta_n}$, we have
$$
\left|\zeta_{g_{2}}\left(\boldsymbol{v}_{i}, \boldsymbol{O}_{i}\right)-\zeta_{g_{2}}\left(\boldsymbol{v}_{i-\tau_{\beta_n}}, \boldsymbol{O}_{i}\right)\right| \lesssim \left\|\boldsymbol{v}_{i}-\boldsymbol{v}_{i-\tau_{\beta_n}}\right\| \lesssim  \sum_{k=i-\tau_{\beta_n}}^{i-1} \beta_{k}.
$$
Like the previous two Lemmas, we provide an upper bound for $\mathbb{E}_{\mu_{\pi_b}}\{\zeta_{g_{2}}\left(\boldsymbol{v}_{i-\tau_{\beta_n}}, \boldsymbol{O}_{i}\right)\}$. To this end, we define an independent $\boldsymbol{v}_{i-\tau_{\beta_n}}'$ and $\boldsymbol{O}_{i}'=\left(
\boldsymbol{x}_{i}', a_{i}', \boldsymbol{x}_{i+1}'\right)$ which are independently drawn from marginal distributions of $\boldsymbol{v}_{i-\tau_{\beta_n}}$ and $\boldsymbol{O}_{i}$. Using part (a) of Lemma \ref{LEMMA:g2_zeta_bound} and Lemma \ref{APPEND_LEMMA:BHAND9}, we obtain
$$
\mathbb{E}_{\mu_{\pi_b}}\left\{\zeta_{g_{2}}\left(\boldsymbol{v}_{i-\tau_{\beta_n}}, \boldsymbol{O}_{i}\right)\right\} \leq\left|\mathbb{E}_{\mu_{\pi_b}}\left\{\zeta_{g_{2}}\left(\boldsymbol{v}_{i-\tau_{\beta_n}}, \boldsymbol{O}_{i}\right)\right\}-\mathbb{E}\left\{\zeta_{g_{2}}\left(\boldsymbol{v}_{i-\tau_{\beta_n}}', \boldsymbol{O}_{i}'\right)\right\}\right| \lesssim m \rho^{\tau_{\beta_n}}.
$$
Therefore, it follows that
\begin{align*}
\mathbb{E}_{\mu_{\pi_b}}\left\{\zeta_{g_{2}}\left(\boldsymbol{v}_{i}, \boldsymbol{O}_{i}\right)\right\} &\lesssim \mathbb{E}_{\mu_{\pi_b}}\left\{\zeta_{g_{2}}\left(\boldsymbol{v}_{i-\tau_{\beta_n}}, \boldsymbol{O}_{i}\right)\right\} + \sum_{k=i-\tau_{\beta_n}}^{i-1} \beta_{k}\lesssim  m \rho^{\tau_{\beta_n}} + \tau_{\beta_n} \beta_{i-\tau_{\beta_n}} \lesssim  \beta_{n} + \tau_{\beta_n} \beta_{i-\tau_{\beta_n}}.
\end{align*}
For $i \leq \tau_{\beta_n}$, we have that
$$
\mathbb{E}_{\mu_{\pi_b}}\left\{\zeta_{g_{2}}\left(\boldsymbol{v}_{i}, \boldsymbol{O}_{i}\right)\right\} \lesssim \mathbb{E}_{\mu_{\pi_b}}\left\{\zeta_{g_{2}}\left(\boldsymbol{v}_{1}, \boldsymbol{O}_{i}\right)\right\}+ \sum_{k=1}^{i-1} \beta_{k} \lesssim \tau_{\beta_n}. 
$$
\end{proof}

\begin{lemma}\label{LEMMA:f1_zeta_exp_bound}
For $\sigma\in(0,1)$, $\lambda_w < 0$, let $\alpha_{n}=\frac{\alpha_1}{n^{\sigma}}$, 
Then
\begin{align*}
\sum_{i=1}^{n} e^{\lambda_{w} \sum_{k=i+1}^{n} \alpha_k} \alpha_{i}\mathbb{E}_{\mu_{\pi_b}}\left\{ \zeta_{f_{1}}\left(\boldsymbol{w}_{i}, \boldsymbol{O}_{i}\right)\right\} = \mathcal{O}\left(\log n / n^{\sigma} \right). 
\end{align*}
\end{lemma}
\begin{proof} 
Applying Lemma \ref{LEMMA:expect_zeta_f1_bound}, 
it follows that
\begin{align*}
& \sum_{i=1}^{n} e^{\lambda_{w} \sum_{k=i+1}^{n} \alpha_k}  \alpha_{i}\mathbb{E}_{\mu_{\pi_b}}\left[\zeta_{f_{1}}\left(\boldsymbol{w}_{i}, \boldsymbol{O}_{i}\right)\right] \\
& \lesssim \tau_{\alpha_n} \sum_{i=1}^{\tau_{\alpha_n}} e^{\lambda_{w} \sum_{k=i+1}^{n} \alpha_k} \alpha_{i} + \alpha_{n} \sum_{i=\tau_{\alpha_n}+1}^{n} e^{\lambda_{w} \sum_{k=i+1}^{n} \alpha_k} \alpha_{i} \\ &\quad +  \tau_{\alpha_n} \sum_{i=\tau_{\alpha_n}+1}^{n} e^{\lambda_{w} \sum_{k=i+1}^{n} \alpha_k} \alpha_{i-\tau_{\alpha_n}}\alpha_{i}  + \tau_{\alpha_n} \sum_{i=\tau_{\alpha_n}+1}^{n} e^{\lambda_{w} \sum_{k=i+1}^{n} \alpha_k} \alpha^2_{i-\tau_{\alpha_n}} \alpha_{i} \nonumber \\
& \lesssim \tau_{\alpha_n} \sum_{i=1}^{\tau_{\alpha_n}} e^{\lambda_{w} \sum_{k=i+1}^{n} \alpha_k} \alpha_{i} + \alpha_{n} \sum_{i=\tau_{\alpha_n}+1}^{n} e^{\lambda_{w} \sum_{k=i+1}^{n} \alpha_k} \alpha_{i} +\tau_{\alpha_n} \sum_{i=\tau_{\alpha_n}+1}^{n} e^{\lambda_{w} \sum_{k=i+1}^{n} \alpha_k} \alpha_{i-\tau_{\alpha_n}} \alpha_{i}. \nonumber
\end{align*}
Applying Lemma \ref{LEMMA:prelim_TD_exp_bound}, we obtain:
\begin{align*}
\sum_{i=1}^{\tau_{\alpha_n}} e^{\lambda_{w} \sum_{k=i+1}^{n} \alpha_k} \alpha_{i} \lesssim \alpha_n, \quad
\alpha_{n} \sum_{i=\tau_{\alpha_n}+1}^{n} e^{\lambda_{w} \sum_{k=i+1}^{n} \alpha_k} \alpha_{i} \lesssim \alpha_n, \quad \sum_{i=\tau_{\alpha_n}+1}^{n} e^{\lambda_{w} \sum_{k=i+1}^{n} \alpha_k} \alpha_{i-\tau_{\alpha_n}} \alpha_{i} \lesssim \alpha_{(n-\tau_{\alpha_n})\vee 1}.
\end{align*}
Combined with the fact that $\tau_{\alpha_n} = \mathcal{O}(\log n)$, we obtain the desired result.
\end{proof}

\begin{lemma}\label{LEMMA:f2_zeta_exp_bound} For $\nu \in (0,1)$, $\lambda_u < 0$, let $\beta_{n}=\frac{\beta_1}{n^{\nu}}$. Then
$$
\sum_{i=1}^{n} e^{\lambda_{u} \sum_{k=i+1}^{n}\beta_k}  \beta_i\mathbb{E}_{\mu_{\pi_b}}\left\{\zeta_{f_{2}}\left(\boldsymbol{w}_{i}, \boldsymbol{v}_{i}, \boldsymbol{O}_{i}\right)\right\} 
= \mathcal{O}\left(\log n /n^{\nu} \right).
$$
\end{lemma}
\begin{proof} Applying Lemma \ref{LEMMA:expect_zeta_f2_bound}, 
it follows that
\begin{align*}
\sum_{i=1}^{n} e^{\lambda_{u} \sum_{k=i+1}^{n} \beta_k}  \beta_{i}\mathbb{E}_{\mu_{\pi_b}}\left\{\zeta_{f_{2}}\left(\boldsymbol{w}_{i}, \boldsymbol{v}_{i}, \boldsymbol{O}_{i}\right)\right\} &\lesssim  \tau_{\beta_n} \sum_{i=1}^{\tau_{\beta_n}} e^{\lambda_{u} \sum_{k=i+1}^{n} \beta_k} \beta_{i} +  \beta_{n} \sum_{i=\tau_{\beta_n}+1}^{n} e^{\lambda_{u} \sum_{k=i+1}^{n} \beta_k} \beta_{i} \\
& \quad + \tau_{\beta_n} \sum_{i=\tau_{\beta_n}+1}^{n} e^{\lambda_{u} \sum_{k=i+1}^{n} \beta_k} \beta_{i-\tau_{\beta_n}} \beta_{i}.
\end{align*}
Applying Lemma \ref{LEMMA:prelim_TD_exp_bound}, we have
\begin{align*}
&\tau_{\beta_n} \sum_{i=1}^{\tau_{\beta_n}} e^{\lambda_{u} \sum_{k=i+1}^{n} \beta_k} \beta_{i} \lesssim \tau_{\beta_n} \beta_n \\ & \beta_{n} \sum_{i=\tau_{\beta_n}+1}^{n} e^{\lambda_{u} \sum_{k=i+1}^{n} \beta_k} \beta_{i} \lesssim \beta_n \\ & \tau_{\beta_n} \sum_{i=\tau_{\beta_n}+1}^{n} e^{\lambda_{u} \sum_{k=i+1}^{n} \beta_k} \beta_{i-\tau_{\beta_n}} \beta_{i} \lesssim \tau_{\beta_n}\beta_{(n-\tau_{\beta_n})\vee 1}.
\end{align*}
Combined with the fact that $\tau_{\beta_n} = \mathcal{O}(\log n)$, we obtain the desired result.
\end{proof}

\begin{lemma}\label{LEMMA:g2_zeta_exp_bound}
For $\nu \in (0,1), \lambda_u < 0$, let $\beta_{n}=\frac{\beta_1}{n^{\nu}}$. 
Then
$$
\sum_{i=1}^{n} e^{\lambda_{u} \sum_{k=i+1}^{n}\beta_k}  \beta_{i}\mathbb{E}_{\mu_{\pi_b}}\left\{\zeta_{g_{2}}\left(\boldsymbol{v}_{i}, \boldsymbol{O}_{i}\right)\right\} = \mathcal{O}\left(\log n / n^{\nu} \right).
$$
\end{lemma}
\begin{proof}
Applying Lemma \ref{LEMMA:expect_zeta_g2_bound}, it follows that
$$
\begin{aligned}
\sum_{i=1}^{n} e^{\lambda_{u} \sum_{k=i+1}^{n} \beta_k}  \beta_{i}\mathbb{E}_{\mu_{\pi_b}}\left\{\zeta_{g_{2}}\left(\boldsymbol{v}_{i}, \boldsymbol{O}_{i}\right)\right\} 
& \lesssim \tau_{\beta_n} \sum_{i=1}^{\tau_{\beta_n}} e^{\lambda_{u} \sum_{k=i+1}^{n} \beta_k} \beta_{i} + \beta_{n} \sum_{i=\tau_{\beta_n}+1}^{n} e^{\lambda_{u} \sum_{k=i+1}^{n}\beta_k} \beta_{i} \\
& \quad +  \tau_{\beta_n} \sum_{i=\tau_{\beta_n}+1}^{n} e^{\lambda_{u} \sum_{k=i+1}^{n} \beta_k} \beta_{i-\tau_{\beta_n}} \beta_{i}.
\end{aligned}
$$
Now we invoke Lemma \ref{LEMMA:prelim_TD_exp_bound}, and get
$$
\begin{aligned}
& \tau_{\beta_n} \sum_{i=1}^{\tau_{\beta_n}} e^{\lambda_{u} \sum_{k=i+1}^{n} \beta_k} \beta_{i} \lesssim \tau_{\beta_n} \beta_n\\
& \beta_{n} \sum_{i=\tau_{\beta_n}+1}^{n} e^{\lambda_{u} \sum_{k=i+1}^{n} \beta_k} \beta_{i} \lesssim \beta_n \\
& \tau_{\beta_n} \sum_{i=\tau_{\beta_n}+1}^{n} e^{\lambda_{u} \sum_{k=i+1}^{n} \beta_k} \beta_{i-\tau_{\beta_n}} \beta_{i} \lesssim \tau_{\beta_n} \beta_{(n-\tau_{\beta_n})\vee 1}.
\end{aligned}
$$
Combined with the fact that $\tau_{\beta_n} = \mathcal{O}(\log n)$, we obtain the desired result.
\end{proof}

\begin{lemma}\label{LEMMA:MATRIX_C_BOUND} For $0<\nu<\sigma<1, \lambda_u < 0$, let $\alpha_{n}=\frac{\alpha_1}{n^{\sigma}}$ and $\beta_{n}=\frac{\beta_1}{n^{\nu}}$. 
Then
$$
\begin{aligned}
& \sum_{i=1}^{n} e^{\lambda_{u} \sum_{k=i+1}^{n} \beta_k} \mathbb{E}_{\mu_{\pi_b}}\left\langle \boldsymbol{C}^{-1} \boldsymbol{A}\left(\boldsymbol{w}_{i+1}-\boldsymbol{w}_{i}\right), \boldsymbol{v}_{i}\right\rangle = \mathcal{O}\left(\frac{1}{n^{\sigma-\nu}} \right).
\end{aligned}
$$
\end{lemma}
\begin{proof} Using the same arguments as in Lemma \ref{LEMMA:expect_zeta_f1_bound} and \eqref{TDC_w_recursion}, it follows that
\begin{align*}
&\sum_{i=1}^{n} e^{\lambda_{u} \sum_{k=i+1}^{n} \beta_k} \mathbb{E}_{\mu_{\pi_b}}\left\langle \boldsymbol{C}^{-1} \boldsymbol{A}\left(\boldsymbol{w}_{i+1}-\boldsymbol{w}_{i}\right), \boldsymbol{v}_{i}\right\rangle \\ &\leq \sum_{i=1}^{n} e^{\lambda_{u} \sum_{k=i+1}^{n} \beta_k} \mathbb{E}_{\mu_{\pi_b}}\left(\left\|\boldsymbol{C}^{-1}\right\|\|\boldsymbol{A}\|\left\|\boldsymbol{w}_{i+1}-\boldsymbol{w}_{i}\right\|\left\|\boldsymbol{v}_{i}\right\|\right) \\
 &\lesssim \sum_{i=1}^{n} e^{\lambda_{u} \sum_{k=i+1}^{n} \beta_k}\alpha_i. 
\end{align*}
Following the arguments used in Lemma \ref{LEMMA:prelim_TD_exp_bound}, we get
\begin{equation*}
\sum_{i=1}^{n} e^{\left(\lambda_{u} / 2\right) \sum_{k=i+1}^{n} \beta_k} \beta_{i} \lesssim \beta_1
\end{equation*}
as well as
\begin{equation*}
\max _{i \in[1, n]}\left\{e^{\left(\lambda_{u} / 2\right) \sum_{k=i+1}^{n} \beta_k} \frac{1}{i^{\sigma-\nu}}\right\} \lesssim \frac{1}{n^{\sigma-\nu}}
\end{equation*}
which yields the desired result.
\end{proof}

\begin{lemma}\label{LEMMA:MATRIX_C_BOUND_REPEATED} 
For $\alpha_{n}=\frac{\alpha_1}{n^{\sigma}}$ and $\beta_{n}=\frac{\beta_1}{n^{\nu}}$, suppose $\mathbb{E}_{\mu_{\pi_b}}\|\boldsymbol{v}_{i}\|^2 = \mathcal{O}\left(\frac{\log i}{i^\nu} \right) + \mathcal{O}\left( \frac{1}{i^{\sigma-\nu}}\right)$. If $\sigma>\frac{3}{2} \nu$, 
$$
\sum_{i=1}^{n} e^{\lambda_{u} \sum_{k=i+1}^{n} \beta_k} \mathbb{E}_{\mu_{\pi_b}}\left\langle \boldsymbol{C}^{-1} \boldsymbol{A}\left(\boldsymbol{w}_{i+1}-\boldsymbol{w}_{i}\right), \boldsymbol{v}_{i}\right\rangle=\mathcal{O}\left(n^{-\nu}\right)
$$
and if $\nu<\sigma \leq \frac{3}{2} \nu$, 
$$
\sum_{i=1}^{n} e^{\lambda_{u} \sum_{k=i+1}^{n} \beta_k} \mathbb{E}_{\mu_{\pi_b}}\left\langle \boldsymbol{C}^{-1} \boldsymbol{A}\left(\boldsymbol{w}_{i+1}-\boldsymbol{w}_{i}\right), \boldsymbol{v}_{i}\right\rangle = \mathcal{O}\left(\frac{1}{n^{2(\sigma-\nu)-\epsilon}}\right)
$$
where $\epsilon$ is any constant in $(0, \sigma-\nu]$.
\end{lemma}
\begin{proof}
If $\sigma \geq 2 \nu$, Lemma \ref{LEMMA:MATRIX_C_BOUND} implies that
$$
\sum_{i=1}^{n} e^{\lambda_{u} \sum_{k=i+1}^{n}\beta_k} \mathbb{E}_{\mu_{\pi_b}}\left\langle \boldsymbol{C}^{-1} \boldsymbol{A}\left(\boldsymbol{w}_{i+1}-\boldsymbol{w}_{i}\right), \boldsymbol{v}_{i}\right\rangle=\mathcal{O}\left(n^{-\nu}\right).
$$
If $\sigma < 2 \nu$, it follows that $\mathbb{E}_{\mu_{\pi_b}}\left\|\boldsymbol{v}_{i}\right\|^2=\mathcal{O}\left(\frac{1}{i^{\sigma-\nu}}\right)$. Hence there exists $T \in \mathbb{N}$ such that
\begin{align}
& \mathbb{E}_{\mu_{\pi_b}}\left\|\boldsymbol{v}_{i}\right\|^2 \leq R_{\boldsymbol{v}}^2, \quad \forall  i \leq T  \label{v_norm_short_bound}\\
& \mathbb{E}_{\mu_{\pi_b}}\left\|\boldsymbol{v}_{i}\right\|^2 \lesssim \frac{1}{i^{(\sigma-\nu)}}, \quad \forall i>T. \label{v_norm_large_bound}
\end{align}
Note that
\begin{align*}
& \sum_{i=1}^{n} e^{\lambda_{u} \sum_{k=i+1}^{n} \beta_k} \mathbb{E}_{\mu_{\pi_b}}\left\langle \boldsymbol{C}^{-1} \boldsymbol{A}\left(\boldsymbol{w}_{i+1}-\boldsymbol{w}_{i}\right), \boldsymbol{v}_{i}\right\rangle \\
& \lesssim \sum_{i=1}^{n} e^{\lambda_{u} \sum_{k=i+1}^{n} \beta_k} \alpha_{i} \sqrt{\mathbb{E}_{\mu_{\pi_b}}\left\|\boldsymbol{v}_{i}\right\|^2}  \\
& \lesssim \left(\sum_{i=1}^{T} e^{\lambda_{u} \sum_{k=i+1}^{n} \beta_k} \alpha_{i} \sqrt{\mathbb{E}_{\mu_{\pi_b}}\left\|\boldsymbol{v}_{i}\right\|^2}+\sum_{i=T+1}^{n} e^{\lambda_{u} \sum_{k=i+1}^{n} \beta_k} \alpha_{i} \sqrt{\mathbb{E}_{\mu_{\pi_b}}\left\|\boldsymbol{v}_{i}\right\|^2}\right)\\
&\lesssim \sum_{i=1}^{T} e^{\lambda_{u} \sum_{k=i+1}^{n} \beta_k} \beta_{i} \frac{1}{i^{(\sigma-\nu)}}+ \sum_{i=T+1}^{n} e^{\lambda_{u} \sum_{k=i+1}^{n} \beta_k} \beta_{i} \frac{1}{i^{1.5(\sigma-\nu)}}
\end{align*}
where the last inequality follows from \eqref{v_norm_short_bound} and \eqref{v_norm_large_bound}. Following arguments of Lemma \ref{LEMMA:prelim_TD_exp_bound} with \eqref{alpha_bound_cbc}, we get
$$
\sum_{i=1}^{T} e^{\lambda_{u} \sum_{k=i+1}^{n} \beta_k} \beta_{i} \frac{1}{i^{(\sigma-\nu)}} \leq \sum_{i=1}^{T} e^{\lambda_{u} \sum_{k=i+1}^{n } \beta_k} \beta_{i} \le e^{\frac{\lambda_u\beta_1}{1-\nu}\left\{(n+1)^{1-\nu}-(T+1)^{1-\nu}\right\}} 
$$
and
$$
\sum_{i=T+1}^{n} e^{\lambda_{u} \sum_{k=i+1}^{n} \beta_k} \beta_{i} \frac{1}{i^{1.5(\sigma-\nu)}} \lesssim \frac{1}{n^{1.5(\sigma-\nu)}}.
$$
It follows that
$$
\sum_{i=1}^{n} e^{\lambda_{u} \sum_{k=i+1}^{n} \beta_k}  \mathbb{E}_{\mu_{\pi_b}}\left\langle \boldsymbol{C}^{-1} \boldsymbol{A}\left(\boldsymbol{w}_{i+1}-\boldsymbol{w}_{i}\right), \boldsymbol{v}_{i}\right\rangle=\mathcal{O}\left(\frac{1}{n^{1.5(\sigma-\nu)}}\right).
$$
If $\frac{3}{2} \nu<\sigma \leq 2 \nu$, we have $\mathbb{E}_{\mu_{\pi_b}}\left\|\boldsymbol{v}_{i}\right\|^2=\mathcal{O}\left(\frac{1}{i^{1.5(\sigma-\nu)}}\right)$. Following the same steps as above, we have
$$
\sum_{i=1}^{n} e^{\lambda_{u} \sum_{k=i+1}^{n} \beta_k} \mathbb{E}_{\mu_{\pi_b}}\left\langle \boldsymbol{C}^{-1} \boldsymbol{A}\left(\boldsymbol{w}_{i+1}-\boldsymbol{w}_{i}\right), \boldsymbol{v}_{i}\right\rangle=\mathcal{O}\left(\frac{1}{n^{1.75(\sigma-\nu)}}\right)
$$
and $\mathbb{E}_{\mu_{\pi_b}}\left\|\boldsymbol{v}_{i}\right\|^2=\mathcal{O}\left(\frac{1}{i^{1.75(\sigma-\nu)}}\right)$. Repeating the same arguments for a total of $N=$ $\left\lceil-\log _{2}\left(2-\frac{\nu}{\sigma-\nu}\right)\right\rceil$ times, we have
$$
\sum_{i=1}^{n} e^{\lambda_{u} \sum_{k=i+1}^{n} \beta_k} \mathbb{E}_{\mu_{\pi_b}}\left\langle \boldsymbol{C}^{-1} \boldsymbol{A}\left(\boldsymbol{w}_{i+1}-\boldsymbol{w}_{i}\right), \boldsymbol{v}_{i}\right\rangle=\mathcal{O}\left(\frac{1}{n^{\left(2-2^{-N}\right)(\sigma-\nu)}}\right)=\mathcal{O}\left(\frac{1}{n^{\nu}}\right).
$$
If $\nu<\sigma \leq \frac{3}{2} \nu$, we repeat previous steps for a total number $N=\left\lceil\log _{2}\left(\frac{\sigma-\nu}{\epsilon}\right)\right\rceil$ of times to arrive
$$
\sum_{i=1}^{n} e^{\lambda_{u} \sum_{k=i+1}^{n}\beta_k} \mathbb{E}_{\mu_{\pi_b}}\left\langle \boldsymbol{C}^{-1} \boldsymbol{A}\left(\boldsymbol{w}_{i+1}-\boldsymbol{w}_{i}\right), \boldsymbol{v}_{i}\right\rangle=\mathcal{O}\left(\frac{1}{n^{2(\sigma-\nu)-\epsilon}}\right).
$$
\end{proof}

\begin{lemma}\label{LEMMA:MATRIX_B_BOUND} For $x, y \in [0,1]$ and $\sigma\in(0,1)$, suppose $\alpha_1>0, \alpha_{n}=\frac{\alpha_1}{n^{\sigma}}$. If $\mathbb{E}_{\mu_{\pi_b}}\left\|\boldsymbol{v}_{i}\right\|^2= \mathcal{O}\left(\frac{\log i}{i^{\nu}} + \frac{1}{i^\nu}\right)^{x}$ and $\mathbb{E}_{\mu_{\pi_b}}\left\|\boldsymbol{w}_{i}-\boldsymbol{w}_{\star}\right\|^2=\mathcal{O}\left(\frac{\log i}{i^{\nu}} + \frac{1}{i^\nu}\right)^{y}$, we have
$$
\sum_{i=1}^{n} e^{\lambda_{w} \sum_{k=i+1}^{n} \alpha_k}  \alpha_{i}\mathbb{E}_{\mu_{\pi_b}}\left\langle \boldsymbol{B}^s_{i} \boldsymbol{v}_{i}, \boldsymbol{w}_{i}-\boldsymbol{w}_{\star}\right\rangle=\mathcal{O}\left(\frac{\log n}{n^{\nu}}+\frac{1}{n^{\nu}}\right)^{0.5(x+y)}.
$$
If $\mathbb{E}_{\mu_{\pi_b}}\left\|\boldsymbol{v}_{i}\right\|^2=\mathcal{O}\left(\frac{\log i}{i^{\nu}}+\frac{1}{i^{2(\sigma-\nu)-\epsilon}}\right)^{x}$ and $\mathbb{E}_{\mu_{\pi_b}}\left\|\boldsymbol{w}_{i}-\boldsymbol{w}_{\star}\right\|^2=\mathcal{O}\left(\frac{\log i}{i^{\nu}}+\frac{1}{i^{2(\sigma-\nu)-\epsilon}}\right)^{y}$, we have
$$
\sum_{i=1}^{n} e^{\lambda_{w} \sum_{k=i+1}^{n}\alpha_k}  \alpha_{i}\mathbb{E}_{\mu_{\pi_b}}\left\langle \boldsymbol{B}^s_{i} \boldsymbol{v}_{i}, \boldsymbol{w}_{i}-\boldsymbol{w}_{\star}\right\rangle=\mathcal{O}\left(\frac{\log n}{n^{\nu}}+\frac{1}{n^{2(\sigma-\nu)-\epsilon}}\right)^{0.5(x+y)}.
$$
\end{lemma}

\begin{proof} Consider the first case. Without loss of generality, we assume there exist $T>0$ such that
$$
\begin{aligned}
& \mathbb{E}_{\mu_{\pi_b}}\left\|\boldsymbol{v}_{i}\right\|^2 \leq R_{\boldsymbol{v}}^2, \quad \mathbb{E}_{\mu_{\pi_b}}\left\|\boldsymbol{w}_{i}-\boldsymbol{w}_{\star}\right\|^2 \leq R_{\boldsymbol{w}}^2, \quad \forall i \leq T \\
& \mathbb{E}_{\mu_{\pi_b}}\left\|\boldsymbol{v}_{i}\right\|^2 \lesssim \left(\frac{\log i +1}{i^{\nu}}\right)^{x}, \quad \mathbb{E}_{\mu_{\pi_b}}\left\|\boldsymbol{w}_{i}-\boldsymbol{w}_{\star}\right\|^2 \leq \left(\frac{\log i + 1}{i^{\nu}}\right)^{y},  \quad \forall i>T.
\end{aligned}
$$
Therefore, we get 
$$
\begin{aligned}
&\sum_{i=1}^{n} e^{\lambda_{w} \sum_{k=i+1}^{n}\alpha_k}  \alpha_{i}\mathbb{E}_{\mu_{\pi_b}}\left\langle \boldsymbol{B}^s_{i} \boldsymbol{v}_{i}, \boldsymbol{w}_{i}-\boldsymbol{w}_{\star}\right\rangle \leq \sum_{i=1}^{n} e^{\lambda_{w} \sum_{k=i+1}^{n} \alpha_k} \alpha_{i} \sqrt{\mathbb{E}_{\mu_{\pi_b}}\left\|\boldsymbol{B}^s_{i} \boldsymbol{v}_{i}\right\|^2} \sqrt{\mathbb{E}_{\mu_{\pi_b}}\left\|\boldsymbol{w}_{i}-\boldsymbol{w}_{\star}\right\|^2} \\
& \lesssim \sum_{i=1}^{n} e^{\lambda_{w} \sum_{k=i+1}^{n} \alpha_k} \alpha_{i} \sqrt{\mathbb{E}_{\mu_{\pi_b}}\left\|\boldsymbol{v}_{i}\right\|^2} \sqrt{\mathbb{E}_{\mu_{\pi_b}}\left\|\boldsymbol{w}_{i}-\boldsymbol{w}_{\star}\right\|^2} \\
&\lesssim \sum_{i=1}^{T} e^{\lambda_{w} \sum_{k=i+1}^{n} \alpha_k} \alpha_{i} \sqrt{\mathbb{E}_{\mu_{\pi_b}}\left\|\boldsymbol{v}_{i}\right\|^2} \sqrt{\mathbb{E}_{\mu_{\pi_b}}\left\|\boldsymbol{w}_{i}-\boldsymbol{w}_{\star}\right\|^2}+\sum_{i=T+1}^{n} e^{\lambda_{w} \sum_{k=i+1}^{n} \alpha_k} \alpha_{i} \sqrt{\mathbb{E}_{\mu_{\pi_b}}\left\|\boldsymbol{v}_{i}\right\|^2} \sqrt{\mathbb{E}_{\mu_{\pi_b}}\left\|\boldsymbol{w}_{i}-\boldsymbol{w}_{\star}\right\|^2} \\
&\lesssim  \sum_{i=1}^{T} e^{\lambda_{w} \sum_{k=i+1}^{n} \alpha_k} \alpha_{i} + \sum_{i=T+1}^{n} e^{\lambda_{w} \sum_{k=i+1}^{n}\alpha_k} \alpha_{i}\left(\frac{\log i + 1}{i^{\nu}}\right)^{0.5(x+y)}.
\end{aligned}
$$
 Following arguments of Lemma \ref{LEMMA:prelim_TD_exp_bound} with \eqref{alpha_bound_cbc}, we get
$$
\sum_{i=1}^{T} e^{\lambda_{w} \sum_{k=i+1}^{n} \alpha_k} \alpha_{i} \lesssim  e^{\frac{\lambda_{w} \alpha_1}{1-\sigma}\left\{(1+n)^{1-\sigma}-(1+T)^{1-\sigma}\right\}}
$$
and
\begin{align*}
&\sum_{i=T+1}^{n} e^{\lambda_{w} \sum_{k=i+1}^{n} \alpha_k} \alpha_{i}\left(\frac{\log i + 1}{i^{\nu}}\right)^{0.5(x+y)} \lesssim \left(\frac{\log n + 1}{n^{\nu}}\right)^{0.5(x+y)}.
\end{align*}
The proof for the second case can be done with analogous arguments.
\end{proof}

\begin{lemma}\label{LEMMA:MATRIX_B_BOUND_REPEATED} Suppose $\mathbb{E}_{\mu_{\pi_b}}\|\boldsymbol{w}_{i+1}-\boldsymbol{w}_{\star}\|^2 = \mathcal{O}\left(\frac{\log i}{i^\nu} + \frac{1}{i^\nu} \right)^{0.5}$ and $\epsilon' \in (0, 0.5]$. For $0<\frac{3}{2} \nu<\sigma<1$, if $\mathbb{E}_{\mu_{\pi_b}}\left\|\boldsymbol{v}_{i}\right\|^2 = \mathcal{O}\left(\frac{\log i}{i^{\nu}}\right) + \mathcal{O}\left(\frac{1}{i^\nu}\right)$, we have
$$
\sum_{i=1}^{n} e^{\lambda_{w} \sum_{k=i+1}^{n} \alpha_k}\alpha_{i} \mathbb{E}_{\mu_{\pi_b}} \left\langle \boldsymbol{B}^s_{i} \boldsymbol{v}_{i}, \boldsymbol{w}_{i}-\boldsymbol{w}_{\star}\right\rangle = \mathcal{O}\left(\frac{\log n}{n^{\nu}} + \frac{1}{n^\nu}\right)^{1-\epsilon'}.
$$
For $0<\nu<\sigma \leq \frac{3}{2} \nu < 1$, if $\mathbb{E}_{\mu_{\pi_b}}\left\|\boldsymbol{v}_{i}\right\|^2 = \mathcal{O}\left(\frac{\log i}{i^{\nu}}\right) + \mathcal{O}\left(\frac{1}{i^{2(\sigma-\nu)-\epsilon}}\right)$, we have
$$
\sum_{i=1}^{n} e^{\lambda_{w} \sum_{k=i+1}^{n}\alpha_k}\alpha_{i} \mathbb{E}_{\mu_{\pi_b}} \left\langle \boldsymbol{B}^s_{i} \boldsymbol{v}_{i}, \boldsymbol{w}_{i}-\boldsymbol{w}_{\star}\right\rangle = \mathcal{O}\left(\frac{\log n}{n^{\nu}}+\frac{1}{n^{2(\sigma-\nu)-\epsilon}}\right)^{1-\epsilon'}.
$$
\end{lemma}
\begin{proof} Suppose $\mathbb{E}_{\mu_{\pi_b}}\left\|\boldsymbol{w}_{i}-\boldsymbol{w}_{\star}\right\|^2 \leq 4 R_{\boldsymbol{w}}^2 = \mathcal{O}(1)$, applying Lemma \ref{LEMMA:MATRIX_B_BOUND}, we have
\begin{equation*}
\sum_{i=1}^{n} e^{\lambda_{w} \sum_{k=i+1}^{n} \alpha_k}\alpha_{i} \mathbb{E}_{\mu_{\pi_b}}\left\langle \boldsymbol{B}^s_{i} \boldsymbol{v}_{i}, \boldsymbol{w}_{i}-\boldsymbol{w}_{\star}\right\rangle = \mathcal{O}\left(\frac{\log n}{n^{\nu}}+\frac{1}{n^{\nu}}\right)^{0.5}.
\end{equation*}
From the assumption $\mathbb{E}_{\mu_{\pi_b}}\left\|\boldsymbol{w}_{i}-\boldsymbol{w}_{\star}\right\|^2 = \mathcal{O}\left(\frac{\log i}{i^{\nu}} + \frac{1}{i^\nu}\right)^{0.5}$ with Lemma \ref{LEMMA:MATRIX_B_BOUND}, we obtain
\begin{equation*}
\sum_{i=1}^{n} e^{\lambda_{w} \sum_{k=i+1}^{n}\alpha_k} \alpha_{i}\mathbb{E}_{\mu_{\pi_b}}\left\langle \boldsymbol{B}^s_{i} \boldsymbol{v}_{i}, \boldsymbol{w}_{i}-\boldsymbol{w}_{\star}\right\rangle=\mathcal{O}\left(\frac{\log n}{n^{\nu}}+\frac{1}{n^{\nu}}\right)^{0.75}.
\end{equation*}
Repeating the above steps for a total number of $N=\left\lceil\log _{2}\left(\frac{1}{\epsilon'}\right)\right\rceil$ times, we have
$$
\begin{aligned}
\sum_{i=1}^{n} e^{\lambda_{w} \sum_{k=i+1}^{n} \alpha_k}\alpha_{i} \mathbb{E}_{\mu_{\pi_b}}\left\langle \boldsymbol{B}^s_{i} \boldsymbol{v}_{i}, \boldsymbol{w}_{i}-\boldsymbol{w}_{\star}\right\rangle & = \mathcal{O}\left(\frac{\log n}{n^{\nu}}+\frac{1}{n^{\nu}}\right)^{1-\frac{1}{2^{N}}} = \mathcal{O}\left(\frac{\log n}{n^{\nu}}+\frac{1}{n^{\nu}}\right)^{1-\epsilon'}.
\end{aligned}
$$
The proof for the second case is analogous.
\end{proof}

\subsection{Finite-time analysis for projected implicit TDC}
\noindent We finally provide proofs for the finite-time MSE bounds of the implicit TDC algorithm under a decreasing step size sequence.

\begin{proof}[Proof of Theorem \ref{thm:tdc_decr}] For $n \in \mathbb{N}$, we know
\begin{align*}
\left\|\boldsymbol{w}_{n+1}-\boldsymbol{w}_{\star}\right\|^2 &=  \left\|\Pi_{R_{\boldsymbol{w}}}\left[\boldsymbol{w}_{n}+\alpha'_{n}\left\{f_{1}\left(\boldsymbol{w}_{n}, \boldsymbol{O}_{n}\right)+g_{1}\left(\boldsymbol{v}_{n}, \boldsymbol{O}_{n}\right)\right\} + (\alpha_n \boldsymbol{B}_{n} - \alpha'_n \boldsymbol{B}_{n}^s) \boldsymbol{u}_{n}\right]-\boldsymbol{w}_{\star}\right\|^2 \\
&= \left\|\Pi_{R_{\boldsymbol{w}}}\left[\boldsymbol{w}_{n}+\alpha'_{n}\left\{f_{1}\left(\boldsymbol{w}_{n}, \boldsymbol{O}_{n}\right)+g_{1}\left(\boldsymbol{v}_{n}, \boldsymbol{O}_{n}\right)\right\} + (\alpha_n \boldsymbol{B}_{n} - \alpha'_n \boldsymbol{B}_{n}^s) \boldsymbol{u}_{n}\right]-\Pi_{R_{\boldsymbol{w}}}\boldsymbol{w}_{\star}\right\|^2 \\
&\leq \left\|\boldsymbol{w}_{n}-\boldsymbol{w}_{\star}+\alpha'_{n}\left\{f_{1}\left(\boldsymbol{w}_{n}, \boldsymbol{O}_{n}\right)+g_{1}\left(\boldsymbol{v}_{n}, \boldsymbol{O}_{n}\right)\right\} + (\alpha_n \boldsymbol{B}_{n} - \alpha'_n \boldsymbol{B}_{n}^s) \boldsymbol{u}_{n} \right\|^2 \\
&= \left\|\boldsymbol{w}_{n}-\boldsymbol{w}_{\star}+\alpha'_{n}\left\{f_{1}\left(\boldsymbol{w}_{n}, \boldsymbol{O}_{n}\right)+g_{1}\left(\boldsymbol{v}_{n}, \boldsymbol{O}_{n}\right)\right\}\right\|^2 + \left\|(\alpha_n \boldsymbol{B}_{n} - \alpha'_n \boldsymbol{B}_{n}^s) \boldsymbol{u}_{n} \right\|^2 \\
&\quad + 2\left\langle \boldsymbol{w}_{n}-\boldsymbol{w}_{\star}+\alpha'_{n}\left\{f_{1}\left(\boldsymbol{w}_{n}, \boldsymbol{O}_{n}\right)+g_{1}\left(\boldsymbol{v}_{n}, \boldsymbol{O}_{n}\right)\right\}, (\alpha_n \boldsymbol{B}_{n} - \alpha'_n \boldsymbol{B}_{n}^s) \boldsymbol{u}_{n} \right\rangle \\
&\le \left\|\boldsymbol{w}_{n}-\boldsymbol{w}_{\star}\right\|^2 + 2 \alpha'_{n}\left\langle f_{1}\left(\boldsymbol{w}_{n}, \boldsymbol{O}_{n}\right), \boldsymbol{w}_{n}-\boldsymbol{w}_{\star}\right\rangle + 2 \alpha'_{n}\left\langle g_{1}\left(\boldsymbol{v}_{n}, \boldsymbol{O}_{n}\right), \boldsymbol{w}_{n}-\boldsymbol{w}_{\star}\right\rangle \\
& \quad + \alpha_{n}^2\left\|f_{1}\left(\boldsymbol{w}_{n}, \boldsymbol{O}_{n}\right)+g_{1}\left(\boldsymbol{v}_{n}, \boldsymbol{O}_{n}\right)\right\|^2 + \left\|(\alpha_n \boldsymbol{B}_{n} - \alpha'_n \boldsymbol{B}_{n}^s) \boldsymbol{u}_{n} \right\|^2 \\
&\quad + 2 \left\| \boldsymbol{w}_{n}-\boldsymbol{w}_{\star}+\alpha'_{n}\left\{f_{1}\left(\boldsymbol{w}_{n}, \boldsymbol{O}_{n}\right)+g_{1}\left(\boldsymbol{v}_{n}, \boldsymbol{O}_{n}\right)\right\}\right\|\left\| (\alpha_n \boldsymbol{B}_{n} - \alpha'_n \boldsymbol{B}_{n}^s) \boldsymbol{u}_{n} \right \| \\
& \leq  \left\|\boldsymbol{w}_{n}-\boldsymbol{w}_{\star}\right\|^2 + 2 \alpha'_{n}\left\langle\bar{f}_{1}\left(\boldsymbol{w}_{n}\right), \boldsymbol{w}_{n}-\boldsymbol{w}_{\star}\right\rangle + 2 \alpha'_{n}\left\langle f_{1}\left(\boldsymbol{w}_{n}, \boldsymbol{O}_{n}\right)-\bar{f}_{1}\left(\boldsymbol{w}_{n}\right), \boldsymbol{w}_{n}-\boldsymbol{w}_{\star}\right\rangle \\
&\quad + 2 \alpha'_{n}\left\langle g_{1}\left(\boldsymbol{v}_{n}, \boldsymbol{O}_{n}\right), \boldsymbol{w}_{n}-\boldsymbol{w}_{\star}\right\rangle + 2 \alpha_{n}^2K_{f_1}^2 + 2 \alpha_{n}^2K_{g_1}^2 + \left\|(\alpha_n \boldsymbol{B}_{n} - \alpha'_n \boldsymbol{B}_{n}^s) \boldsymbol{u}_{n} \right\|^2 \\
&\quad + 2 \left \{2R_{\boldsymbol{w}}+\alpha_{n}\left(K_{f_{1}}+K_{g_{1}}\right)\right\} \left\|(\alpha_n \boldsymbol{B}_{n} - \alpha'_n \boldsymbol{B}_{n}^s) \boldsymbol{u}_{n} \right \|.
\end{align*}
Applying Lemma \ref{LEMMA:B_bound} and combining terms, we have
\begin{align*}
\left\|\boldsymbol{w}_{n+1}-\boldsymbol{w}_{\star}\right\|^2 &\leq  \left\|\boldsymbol{w}_{n}-\boldsymbol{w}_{\star}\right\|^2 + 2 \alpha'_{n}\left\langle\left(A^{\top} \boldsymbol{C}^{-1} \boldsymbol{A}\right)\left(\boldsymbol{w}_{n}-\boldsymbol{w}_{\star}\right), \boldsymbol{w}_{n}-\boldsymbol{w}_{\star}\right\rangle \\ &\quad + 2\alpha_{n}'\left\langle f_{1}\left(\boldsymbol{w}_{n}, \boldsymbol{O}_{n}\right)-\bar{f}_{1}\left(\boldsymbol{w}_{n}\right), \boldsymbol{w}_{n}-\boldsymbol{w}_{\star}\right\rangle + 2 \alpha'_{n}\left\langle g_{1}\left(\boldsymbol{v}_{n}, \boldsymbol{O}_{n}\right), \boldsymbol{w}_{n}-\boldsymbol{w}_{\star}\right\rangle + \mathcal{O}(\alpha_n^2) \\
&\leq \left(1-\alpha'_{n}\left|\lambda_{w}\right|\right)\left\|\boldsymbol{w}_{n}-\boldsymbol{w}_{\star}\right\|^2 + 2 \alpha'_{n} \zeta_{f_{1}}\left(\boldsymbol{w}_{n}, \boldsymbol{O}_{n}\right) + 2 \alpha'_{n}\left\langle \boldsymbol{B}^{s}_{n} \boldsymbol{v}_{n}, \boldsymbol{w}_{n}-\boldsymbol{w}_{\star}\right\rangle + \mathcal{O}(\alpha_n^2) 
\end{align*}
where in the second inequality we used facts $2 \lambda_{\max }\left(A^{\top} \boldsymbol{C}^{-1} \boldsymbol{A}\right) \leq \lambda_{w}<0$ and $\zeta_{f_{1}}\left(\boldsymbol{w}_{n}, \boldsymbol{O}_{n}\right)=\left\langle f_{1}\left(\boldsymbol{w}_{n}, \boldsymbol{O}_{n}\right)-\bar{f}_{1}\left(\boldsymbol{w}_{n}\right), \boldsymbol{w}_{n}-\boldsymbol{w}_{\star}\right\rangle$.
Now note that $\alpha'_n \ge \alpha_n/(1+\alpha_n \rho_{\max}\phi_{\max}^2) =: \underline{\alpha_n}$ holds almost surely. Furthermore, we know from Lemma \ref{LEMMA:B_bound} and Lemma \ref{LEMMA:f1_zeta_bound} that 
\begin{align*}
    \alpha'_{n} \zeta_{f_{1}}\left(\boldsymbol{w}_{n}, \boldsymbol{O}_{n}\right) &= \alpha_{n} \zeta_{f_{1}}\left(\boldsymbol{w}_{n}, \boldsymbol{O}_{n}\right) + (\alpha'_{n} - \alpha_n) \zeta_{f_{1}}\left(\boldsymbol{w}_{n}, \boldsymbol{O}_{n}\right) = \alpha_{n} \zeta_{f_{1}}\left(\boldsymbol{w}_{n}, \boldsymbol{O}_{n}\right) + \mathcal{O}(\alpha_{n}^2) \\
     \alpha'_{n}\left\langle \boldsymbol{B}^{s}_{n} \boldsymbol{v}_{n}, \boldsymbol{w}_{n}-\boldsymbol{w}_{\star}\right\rangle &=  \alpha_{n}\left\langle \boldsymbol{B}^{s}_{n} \boldsymbol{v}_{n}, \boldsymbol{w}_{n}-\boldsymbol{w}_{\star}\right\rangle +  (\alpha'_{n}-\alpha_n)\left\langle \boldsymbol{B}^{s}_{n} \boldsymbol{v}_{n}, \boldsymbol{w}_{n}-\boldsymbol{w}_{\star}\right\rangle = \alpha_{n}\left\langle \boldsymbol{B}^{s}_{n} \boldsymbol{v}_{n}, \boldsymbol{w}_{n}-\boldsymbol{w}_{\star}\right\rangle + \mathcal{O}(\alpha_{n}^2)
\end{align*}
which leads to 
\begin{align*}
\left\|\boldsymbol{w}_{n+1}-\boldsymbol{w}_{\star}\right\|^2 
\leq \left(1-\underline{\alpha_n}\left|\lambda_{w}\right|\right)\left\|\boldsymbol{w}_{n}-\boldsymbol{w}_{\star}\right\|^2 + 2 \alpha_{n} \zeta_{f_{1}}\left(\boldsymbol{w}_{n}, \boldsymbol{O}_{n}\right) + 2 \alpha_{n}\left\langle \boldsymbol{B}^{s}_{n} \boldsymbol{v}_{n}, \boldsymbol{w}_{n}-\boldsymbol{w}_{\star}\right\rangle + \mathcal{O}(\alpha_n^2).
\end{align*}
Note that
\begin{align*}
 1-\underline{\alpha_n}\left|\lambda_{w}\right| &= 1-\frac{\alpha_n\left|\lambda_{w}\right|}{1+\alpha_n \rho_{\max} \phi_{\max}^2} \le \frac{1}{1+\eta_w \alpha_n} \in (0,1) \quad \text{where} \\  \eta_w &= \frac{\left|\lambda_{w}\right|}{1 + \mathbb{I}\{\rho_{\max}\phi_{\max}^2 > |\lambda_w|\}(\rho_{\max}\phi_{\max}^2 - |\lambda_w|)\alpha_1}.
\end{align*}
With $\frac{1}{1+x} \le \exp(\frac{-x}{1+x})$, we get
\begin{align*}
\left\|\boldsymbol{w}_{n+1}-\boldsymbol{w}_{\star}\right\|^2 
\leq \exp\left(\frac{-\eta_w \alpha_n}{1+\eta_w \alpha_n}\right)\left\|\boldsymbol{w}_{n}-\boldsymbol{w}_{\star}\right\|^2 + 2 \alpha_{n} \zeta_{f_{1}}\left(\boldsymbol{w}_{n}, \boldsymbol{O}_{n}\right) + 2 \alpha_{n}\left\langle \boldsymbol{B}^{s}_{n} \boldsymbol{v}_{n}, \boldsymbol{w}_{n}-\boldsymbol{w}_{\star}\right\rangle + \mathcal{O}(\alpha_n^2).
\end{align*}
Telescoping the above inequality and taking the expectation on both sides yields that
\begin{align}
\mathbb{E}_{\mu_{\pi_b}}\left\|\boldsymbol{w}_{n+1}-\boldsymbol{w}_{\star}\right\|^2 &\lesssim  {\left\{\prod_{i=1}^{n}\exp\left(\frac{-\eta_w \alpha_i}{1+\eta_w \alpha_i}\right)\right\}\left\|\boldsymbol{w}_{1}-\boldsymbol{w}_{\star}\right\|^2 } \nonumber \\
&\quad + \sum_{i=1}^{n}\left\{\prod_{k=i+1}^{n}\exp\left(\frac{-\eta_w \alpha_k}{1+\eta_w \alpha_k}\right)\right\} \alpha_i\mathbb{E}_{\mu_{\pi_b}}\left\{\zeta_{f_{1}}\left(\boldsymbol{w}_{i}, \boldsymbol{O}_{i}\right)\right\}\nonumber \\
& \quad + \sum_{i=1}^{n}\left\{\prod_{k=i+1}^{n}\exp\left(\frac{-\eta_w \alpha_k}{1+\eta_w \alpha_k}\right)\right\} \alpha_i\mathbb{E}_{\mu_{\pi_b}}\left\langle \boldsymbol{B}^s_{i} \boldsymbol{v}_{i}, \boldsymbol{w}_{i}-\boldsymbol{w}_{\star}\right\rangle\nonumber \\
&\quad + \sum_{i=1}^{n}\left\{\prod_{k=i+1}^{n}\exp\left(\frac{-\eta_w \alpha_k}{1+\eta_w \alpha_k}\right)\right\} \alpha_i^2.
\nonumber
\end{align}
Since $\sum \frac{\alpha_k}{1+\eta_w \alpha_k} = \sum \alpha_k + \mathcal{O}(1)$, we know 
\begin{align*}
    \prod_{i=1}^{n}\exp\left(\frac{-\eta_w \alpha_i}{1+\eta_w \alpha_i}\right) \lesssim \exp\left(-\eta_w \sum_{i=1}^n \alpha_i \right), \quad \prod_{k=i+1}^{n}\exp\left(\frac{-\eta_w \alpha_k}{1+\eta_w \alpha_k}\right) \lesssim \exp\left(-\eta_w \sum_{k=i+1}^n \alpha_k \right)
\end{align*}
which gives us
\begin{align}
\mathbb{E}_{\mu_{\pi_b}}\left\|\boldsymbol{w}_{n+1}-\boldsymbol{w}_{\star}\right\|^2 &\lesssim  \exp\left(-\eta_w \sum_{i=1}^n \alpha_i \right)\left\|\boldsymbol{w}_{1}-\boldsymbol{w}_{\star}\right\|^2 + \sum_{i=1}^{n}\exp\left(-\eta_w \sum_{k=i+1}^n \alpha_k \right) \alpha_i\mathbb{E}_{\mu_{\pi_b}}\left\{\zeta_{f_{1}}\left(\boldsymbol{w}_{i}, \boldsymbol{O}_{i}\right)\right\}\nonumber \\
& \quad + \sum_{i=1}^{n}\exp\left(-\eta_w \sum_{k=i+1}^n \alpha_k \right)\alpha_i\mathbb{E}_{\mu_{\pi_b}}\left\langle \boldsymbol{B}^s_{i} \boldsymbol{v}_{i}, \boldsymbol{w}_{i}-\boldsymbol{w}_{\star}\right\rangle  + \sum_{i=1}^{n}\exp\left(-\eta_w \sum_{k=i+1}^n \alpha_k \right)\alpha_i^2.
\nonumber
\end{align}
From Lemma \ref{LEMMA:f1_zeta_exp_bound}, Lemma \ref{LEMMA:MATRIX_B_BOUND_REPEATED} and Lemma \ref{LEMMA:prelim_TD_exp_bound}, we obtain
\begin{align*}
&\mathbb{E}_{\mu_{\pi_b}}\left\|\boldsymbol{w}_{n+1}-\boldsymbol{w}_{\star}\right\|^2\\ &\lesssim  \exp\left[-\eta_w \alpha_1\left\{(1+n)^{1-\sigma}-1\right\}\right]\left\|\boldsymbol{w}_{1}-\boldsymbol{w}_{\star}\right\|^2 + \mathcal{O}\left(\frac{\log n}{n^{\sigma}} \right) + \mathcal{O}\left(\frac{\log n}{n^{\nu}}+h(\sigma, \nu)\right)^{1-\epsilon'} + \mathcal{O}\left(\frac{1}{n^\sigma}\right)\\
&\lesssim  \exp\left[-\eta_w \alpha_1\left\{(1+n)^{1-\sigma}-1\right\}\right]\left\|\boldsymbol{w}_{1}-\boldsymbol{w}_{\star}\right\|^2 + \mathcal{O}\left(\frac{\log n}{n^{\sigma}} \right) + \mathcal{O}\left(\frac{\log n}{n^{\nu}}+h(\sigma, \nu)\right)^{1-\epsilon'}
\end{align*}
where
$$
h(\sigma, \nu)=\begin{cases}
\frac{1}{n^{\nu}}, & \sigma>1.5 \nu \\
\frac{1}{n^{2(\sigma-\nu)-\epsilon}} & \nu<\sigma \leq 1.5 \nu
\end{cases}
$$
for an arbitrarily small constant $\epsilon \in(0, \sigma-\nu]$. We now justify the condition of Lemma \ref{LEMMA:MATRIX_B_BOUND_REPEATED}.\\

\noindent We next bound the recursion of the tracking error vector $\boldsymbol{v}_{n}$ as follows. For $n \in \mathbb{N}$, note that
\begin{align*}
\left\|\boldsymbol{v}_{n+1}\right\|^2= & \left\|\Pi_{R_{\boldsymbol{u}}}\left[\boldsymbol{v}_{n}+\beta'_{n}\left\{f_{2}\left(\boldsymbol{w}_{n}, \boldsymbol{O}_{n}\right)+g_{2}\left(\boldsymbol{v}_{n}, \boldsymbol{O}_{n}\right)\right\}-\boldsymbol{C}^{-1}\left(\boldsymbol{b}+\boldsymbol{A} \boldsymbol{w}_{n}\right)\right]+\boldsymbol{C}^{-1}\left(\boldsymbol{b}+\boldsymbol{A}\boldsymbol{w}_{n+1}\right)\right\|^2 \\
= & \left\|\Pi_{R_{\boldsymbol{u}}}\left[\boldsymbol{v}_{n}+\beta'_{n}\left\{f_{2}\left(\boldsymbol{w}_{n}, 
\boldsymbol{O}_{n}\right)+g_{2}\left(\boldsymbol{v}_{n}, \boldsymbol{O}_{n}\right)\right\}-\boldsymbol{C}^{-1}\left(\boldsymbol{b}+\boldsymbol{A}\boldsymbol{w}_{n}\right)\right]-\Pi_{R_{\boldsymbol{u}}}\left\{-\boldsymbol{C}^{-1}\left(\boldsymbol{b}+\boldsymbol{A}\boldsymbol{w}_{n+1}\right)\right\}\right\|^2 \\
\leq & \left\|\boldsymbol{v}_n+\beta'_{n}\left\{f_{2}\left(\boldsymbol{w}_{n}, \boldsymbol{O}_{n}\right)+g_{2}\left(\boldsymbol{v}_{n}, \boldsymbol{O}_{n}\right)\right\}+\boldsymbol{C}^{-1} \boldsymbol{A}\left(\boldsymbol{w}_{n+1}-\boldsymbol{w}_{n}\right)\right\|^2 \\
= & \left\|\boldsymbol{v}_{n}\right\|^2+2 \beta'_{n}\left\langle f_{2}\left(\boldsymbol{w}_{n}, \boldsymbol{O}_{n}\right), \boldsymbol{v}_{n}\right\rangle+2 \beta'_{n}\left\langle g_{2}\left(\boldsymbol{v}_{n}, \boldsymbol{O}_{n}\right), \boldsymbol{v}_{n}\right\rangle+2\left\langle \boldsymbol{C}^{-1} \boldsymbol{A}\left(\boldsymbol{w}_{n+1}-\boldsymbol{w}_{n}\right), \boldsymbol{v}_{n}\right\rangle \\
& +\left\|\beta'_{n} f_{2}\left(\boldsymbol{w}_{n}, \boldsymbol{O}_{n}\right)+\beta'_{n} g_{2}\left(\boldsymbol{v}_{n}, \boldsymbol{O}_{n}\right)+\boldsymbol{C}^{-1} \boldsymbol{A}\left(\boldsymbol{w}_{n+1}-\boldsymbol{w}_{n}\right)\right\|^2 \\
\leq & \left\|\boldsymbol{v}_{n}\right\|^2+2 \beta'_{n}\left\langle\bar{g}_{2}\left(\boldsymbol{v}_{n}\right), \boldsymbol{v}_{n}\right\rangle+2 \beta'_{n}\left\langle f_{2}\left(\boldsymbol{w}_{n}, \boldsymbol{O}_{n}\right), \boldsymbol{v}_{n}\right\rangle+2 \beta'_{n}\left\langle g_{2}\left(\boldsymbol{v}_{n}, \boldsymbol{O}_{n}\right)-\bar{g}_{2}\left(\boldsymbol{v}_{n}\right), \boldsymbol{v}_{n}\right\rangle \\
& +2\left\langle \boldsymbol{C}^{-1} \boldsymbol{A}\left(\boldsymbol{w}_{n+1}-\boldsymbol{w}_{n}\right), \boldsymbol{v}_{n}\right\rangle + 3 \beta_{n}^2\left\|f_{2}\left(\boldsymbol{w}_{n}, \boldsymbol{O}_{n}\right)\right\|^2 + 3 \beta_{n}^2\left\|g_{2}\left(\boldsymbol{v}_{n}, \boldsymbol{O}_{n}\right)\right\|^2 \\
& + 3\left\|\boldsymbol{C}^{-1} \boldsymbol{A}\left(\boldsymbol{w}_{n+1}-\boldsymbol{w}_{n}\right)\right\|^2 \\
\leq & \left\|\boldsymbol{v}_{n}\right\|^2 + 2 \beta'_{n}\left\langle \boldsymbol{C} \boldsymbol{v}_{n}, \boldsymbol{v}_{n}\right\rangle + 2 \beta'_{n}\left\langle f_{2}\left(\boldsymbol{w}_{n}, \boldsymbol{O}_{n}\right), \boldsymbol{v}_{n}\right\rangle + 2 \beta'_{n}\left\langle g_{2}\left(\boldsymbol{v}_{n}, \boldsymbol{O}_{n}\right)-\bar{g}_{2}\left(\boldsymbol{v}_{n}\right), \boldsymbol{v}_{n}\right\rangle \\
& + 2\left\langle \boldsymbol{C}^{-1} \boldsymbol{A}\left(\boldsymbol{w}_{n+1}-\boldsymbol{w}_{n}\right), \boldsymbol{v}_{n}\right\rangle + 3 \beta_{n}^2\left\|f_{2}\left(\boldsymbol{w}_{n}, \boldsymbol{O}_{n}\right)\right\|^2 + 3 \beta_{n}^2\left\|g_{2}\left(\boldsymbol{v}_{n}, \boldsymbol{O}_{n}\right)\right\|^2 \\
& + 3\left\|\boldsymbol{C}^{-1}\right\|^2\|\boldsymbol{A}\|^2\left\|\alpha'_n\left\{f_{1}\left(\boldsymbol{w}_{n}, \boldsymbol{O}_{n}\right)+g_{1}\left(\boldsymbol{v}_{n}, \boldsymbol{O}_{n}\right)\right\} + (\alpha_n \boldsymbol{B}_{n} - \alpha'_n \boldsymbol{B}_{n}^s) \boldsymbol{u}_{n} \right\|^2 \\
\leq & \left\|\boldsymbol{v}_{n}\right\|^2 + 2 \beta'_{n}\left\langle \boldsymbol{C} \boldsymbol{v}_{n}, \boldsymbol{v}_{n}\right\rangle + 2 \beta'_{n}\left\langle f_{2}\left(\boldsymbol{w}_{n}, \boldsymbol{O}_{n}\right), \boldsymbol{v}_{n}\right\rangle + 2 \beta'_{n}\left\langle g_{2}\left(\boldsymbol{v}_{n}, \boldsymbol{O}_{n}\right)-\bar{g}_{2}\left(\boldsymbol{v}_{n}\right), \boldsymbol{v}_{n}\right\rangle \\
& + 2\left\langle \boldsymbol{C}^{-1} \boldsymbol{A}\left(\boldsymbol{w}_{n+1}-\boldsymbol{w}_{n}\right), \boldsymbol{v}_{n}\right\rangle +  \mathcal{O}(\beta_n^2) + \mathcal{O}(\alpha_n^2)\\  
\leq & \left(1-\beta'_{n}\left|\lambda_{u}\right|\right)\left\|\boldsymbol{v}_{n}\right\|^2 + 2 \beta'_{n} \zeta_{f_{2}}\left(\boldsymbol{w}_{n}, \boldsymbol{v}_{n}, \boldsymbol{O}_{n}\right) + 2 \beta'_{n} \zeta_{g_{2}}\left(\boldsymbol{v}_{n}, \boldsymbol{O}_{n}\right) + 2\left\langle \boldsymbol{C}^{-1} \boldsymbol{A}\left(\boldsymbol{w}_{n+1}-\boldsymbol{w}_{n}\right), \boldsymbol{v}_{n}\right\rangle \\
& + \mathcal{O}(\beta_n^2) + \mathcal{O}(\alpha_n^2)
\end{align*}
where $\lambda_{\max }(2 \boldsymbol{C}) \leq \lambda_{u}<0, \zeta_{f_{2}}\left(\boldsymbol{w}_{n}, \boldsymbol{v}_{n}, \boldsymbol{O}_{n}\right)=\left\langle f_{2}\left(\boldsymbol{w}_{n}, \boldsymbol{O}_{n}\right), \boldsymbol{v}_{n}\right\rangle, \zeta_{g_{2}}\left(\boldsymbol{v}_{n}, \boldsymbol{O}_{n}\right)=\left\langle g_{2}\left(\boldsymbol{v}_{n}, \boldsymbol{O}_{n}\right)-\bar{g}_{2}\left(\boldsymbol{v}_{n}\right), \boldsymbol{v}_{n}\right\rangle$. 
From Lemma \ref{LEMMA:f2_zeta_bound} and Lemma \ref{LEMMA:g2_zeta_bound}, we know that
\begin{align*}
    \beta'_{n} \zeta_{f_{2}}\left(\boldsymbol{w}_{n}, \boldsymbol{v}_{n}, \boldsymbol{O}_{n}\right) &=  \beta_{n} \zeta_{f_{2}}\left(\boldsymbol{w}_{n}, \boldsymbol{v}_{n}, \boldsymbol{O}_{n}\right) + \mathcal{O}(\beta_n^2) \\
    \beta'_{n} \zeta_{g_{2}}\left(\boldsymbol{v}_{n}, \boldsymbol{O}_{n}\right) &= \beta_{n} \zeta_{g_{2}}\left(\boldsymbol{v}_{n}, \boldsymbol{O}_{n}\right) + \mathcal{O}(\beta_n^2).
\end{align*}
Furthermore, $\beta'_n \ge \beta_n/(1+\beta_n \rho_{\max}\phi_{\max}^2) = : \underline{\beta_n}$ holds almost surely. With these identities, taking the expectation on both sides, we have
\begin{align*}
\mathbb{E}\left\|\boldsymbol{v}_{n+1}\right\|^2 \leq & \left(1-\underline{\beta_{n}}\left|\lambda_{u}\right|\right) \mathbb{E}\left\|\boldsymbol{v}_{n}\right\|^2 + 2\beta_n\mathbb{E}\left\{\zeta_{f_{2}}\left(\boldsymbol{w}_{n}, \boldsymbol{v}_{n}, \boldsymbol{O}_{n}\right)\right\}+2  \beta_{n}\mathbb{E}\left\{\zeta_{g_{2}}\left(\boldsymbol{v}_{n}, \boldsymbol{O}_{n}\right)\right\} \nonumber \\
& + 2 \mathbb{E}\left\langle \boldsymbol{C}^{-1} \boldsymbol{A}\left(\boldsymbol{w}_{n+1}-\boldsymbol{w}_{n}\right), \boldsymbol{v}_{n}\right\rangle + \mathcal{O}(\beta_{n}^2)  + \mathcal{O}(\alpha_{n}^2).
\end{align*}
Note that
\begin{align*}
    1-\underline{\beta_{n}}\left|\lambda_{u}\right| &= 1-\frac{\beta_{n}\left|\lambda_{u}\right|}{1 + \beta_n \rho_{\max}\phi_{\max}^2} \le \frac{1}{1 + \eta_v \beta_n}\in (0,1) \quad \text{where} \\
    \eta_v &= \frac{|\lambda_u|}{1 + \mathbb{I}\{\rho_{\max}\phi_{\max}^2 > |\lambda_u|\} (\rho_{\max} \phi_{\max}^2 - |\lambda_u|)\beta_1}.
\end{align*}
With $\frac{1}{1+x} \le \exp(\frac{-x}{1+x})$ and $(1+i)^{-\nu} \geq(1+i)^{-\sigma}$ for all $i \in \mathbb{N}$, we get
\begin{align}
\mathbb{E}_{\mu_{\pi_b}}\left\|\boldsymbol{v}_{n+1}\right\|^2 \lesssim & \exp\left(\frac{-\eta_v \beta_n}{1+\eta_v \beta_n}\right) \mathbb{E}_{\mu_{\pi_b}}\left\|\boldsymbol{v}_{n}\right\|^2 + \beta_n\mathbb{E}_{\mu_{\pi_b}}\left\{\zeta_{f_{2}}\left(\boldsymbol{w}_{n}, \boldsymbol{v}_{n}, \boldsymbol{O}_{n}\right)\right\} + \beta_{n}\mathbb{E}_{\mu_{\pi_b}}\left\{\zeta_{g_{2}}\left(\boldsymbol{v}_{n}, \boldsymbol{O}_{n}\right)\right\} \nonumber \\
& + \mathbb{E}_{\mu_{\pi_b}}\left\langle \boldsymbol{C}^{-1} \boldsymbol{A}\left(\boldsymbol{w}_{n+1}-\boldsymbol{w}_{n}\right), \boldsymbol{v}_{n}\right\rangle + \mathcal{O}(\beta_{n}^2). \label{v_recursion_decreasing} 
\end{align}
Telescoping the above inequality, we obtain
\begin{align*}
\mathbb{E}_{\mu_{\pi_b}}\left\|\boldsymbol{v}_{n+1}\right\|^2 &\lesssim  {\left\{\prod_{i=1}^{n}\exp\left(\frac{-\eta_v \beta_i}{1+\eta_v \beta_i}\right)\right\}\left\|\boldsymbol{v}_1\right\|^2 } + \sum_{i=1}^{n}\left\{\prod_{k=i+1}^{n}\exp\left(\frac{-\eta_v \beta_k}{1+\eta_v \beta_k}\right)\right\} \beta_{i} \mathbb{E}_{\mu_{\pi_b}}\left\{\zeta_{f_{2}}\left(\boldsymbol{w}_{i}, \boldsymbol{v}_{i}, \boldsymbol{O}_{i}\right)\right\} \\
&\quad  + \sum_{i=1}^{n}\left\{\prod_{k=i+1}^{n}\exp\left(\frac{-\eta_v \beta_k}{1+\eta_v \beta_k}\right)\right\} \beta_{i} \mathbb{E}_{\mu_{\pi_b}}\left\{\zeta_{g_{2}}\left(\boldsymbol{v}_{i}, \boldsymbol{O}_{i}\right)\right\} + \sum_{i=1}^{n}\left\{\prod_{k=i+1}^{n}\exp\left(\frac{-\eta_v \beta_k}{1+\eta_v \beta_k}\right)\right\} \beta_{i}^2 \\
& \quad + \sum_{i=1}^{n}\left\{\prod_{k=i+1}^{n}\exp\left(\frac{-\eta_v \beta_k}{1+\eta_v \beta_k}\right)\right\} \mathbb{E}_{\mu_{\pi_b}}\left\langle \boldsymbol{C}^{-1} \boldsymbol{A}\left(\boldsymbol{w}_{i+1}-\boldsymbol{w}_{i}\right), \boldsymbol{v}_{i}\right\rangle. 
\end{align*}
Since $\sum \frac{\beta_k}{1+\eta_v \beta_k} = \sum \beta_k + \mathcal{O}(1)$, we know 
\begin{align*}
    \prod_{i=1}^{n}\exp\left(\frac{-\eta_v \beta_i}{1+\eta_v \beta_i}\right) \lesssim \exp\left(-\eta_v \sum_{i=1}^n \beta_i \right), \quad \prod_{k=i+1}^{n}\exp\left(\frac{-\eta_v \beta_k}{1+\eta_v \beta_k}\right) \lesssim \exp\left(-\eta_v \sum_{k=i+1}^n \beta_k \right),
\end{align*}
which gives us 
\begin{align*}
\mathbb{E}_{\mu_{\pi_b}}\left\|\boldsymbol{v}_{n+1}\right\|^2 &\lesssim  \exp\left(-\eta_v \sum_{i=1}^n \beta_i \right)\left\|\boldsymbol{v}_1\right\|^2 + \sum_{i=1}^{n}\exp\left(-\eta_v \sum_{k=i+1}^n \beta_k \right) \beta_{i} \mathbb{E}_{\mu_{\pi_b}}\left\{\zeta_{f_{2}}\left(\boldsymbol{w}_{i}, \boldsymbol{v}_{i}, \boldsymbol{O}_{i}\right)\right\} \\
&\quad  + \sum_{i=1}^{n}\exp\left(-\eta_v \sum_{k=i+1}^n \beta_k \right) \beta_{i} \mathbb{E}_{\mu_{\pi_b}}\left\{\zeta_{g_{2}}\left(\boldsymbol{v}_{i}, \boldsymbol{O}_{i}\right)\right\} + \sum_{i=1}^{n}\exp\left(-\eta_v \sum_{k=i+1}^n \beta_k \right) \beta_{i}^2  \\
& \quad + \sum_{i=1}^{n}\exp\left(-\eta_v \sum_{k=i+1}^n \beta_k \right) \mathbb{E}_{\mu_{\pi_b}}\left\langle \boldsymbol{C}^{-1} \boldsymbol{A}\left(\boldsymbol{w}_{i+1}-\boldsymbol{w}_{i}\right), \boldsymbol{v}_{i}\right\rangle. 
\end{align*}
Combining Lemma \ref{LEMMA:f2_zeta_exp_bound}, Lemma \ref{LEMMA:g2_zeta_exp_bound}, and Lemma \ref{LEMMA:MATRIX_C_BOUND}, we obtain
\begin{align}
\mathbb{E}_{\mu_{\pi_b}}\left\|\boldsymbol{v}_{n+1}\right\|^2 &\lesssim \exp\left[-\eta_v \beta_1\left\{(1+n)^{1-\nu}-1\right\}\right]\left\|\boldsymbol{v}_1\right\|^2 + \mathcal{O}\left(\frac{\log n}{n^{\nu}} \right) + \mathcal{O}\left(\frac{1}{n^{\nu}} \right) + \mathcal{O}\left(\frac{1}{n^{\sigma-\nu}} \right)\nonumber\\
&= \mathcal{O}\left(\frac{\log n}{n^{\nu}} \right)+ \mathcal{O}\left(\frac{1}{n^{\sigma-\nu}} \right). \label{norm_v_repeat_condition}
\end{align}
Applying Lemma \ref{LEMMA:MATRIX_C_BOUND_REPEATED}, we can further refine to yield
$$
\mathbb{E}_{\mu_{\pi_b}}\left\|\boldsymbol{v}_{n}\right\|^2 = \mathcal{O}\left(\frac{\log n}{n^{\nu}}\right)+\mathcal{O}\left(h(\sigma, \nu)\right)
$$
with
$$
h(\sigma, \nu)=\left\{\begin{array}{lr}
\frac{1}{n^{\nu}}, & \sigma>1.5 \nu \\
\frac{1}{n^{2(\sigma-\nu)-\epsilon}}, & \nu<\sigma \leq 1.5 \nu
\end{array}\right.
$$
where $\epsilon \in(0, \sigma-\nu]$ can be an arbitrarily small constant.
\end{proof}

\section{Additional numerical experiments}\label{APPEND_SEC_LAMBDA}
In this section, we provide numerical experimental results of implicit TD(0.5) along with standard TD(0.5) methods with and without projection on the 11-state random walk environments, considered in Subsection \ref{SUBSEC:RW_ABS}. Similar to what we have observed for TD(0) algorithms, both implicit TD(0.5) and projected implicit TD(0.5) proved considerably more robust to the choice of step size than their standard TD(0.5) counterparts.
\begin{figure}[H]
 \centering
    \includegraphics[height=.25\textwidth]{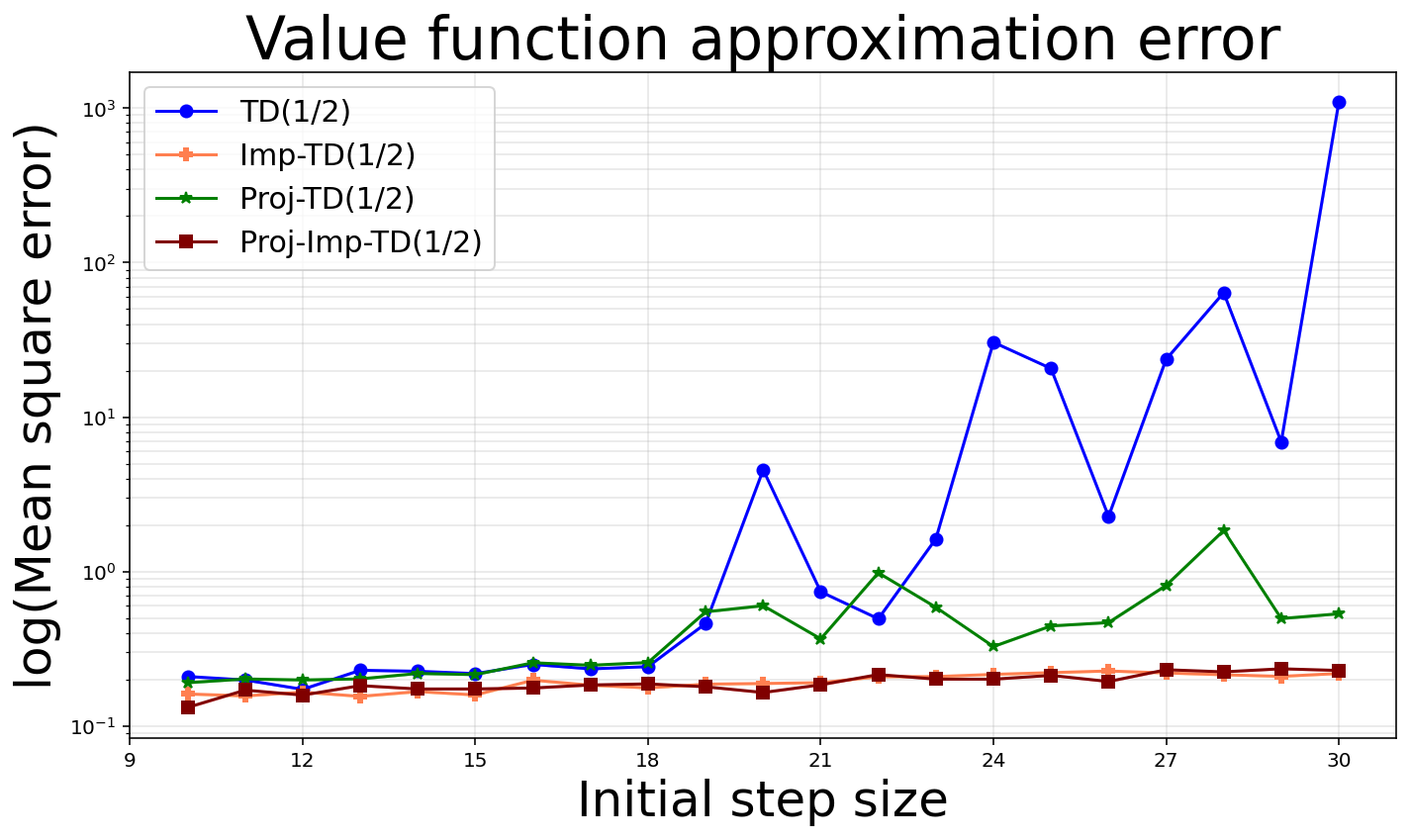}
    \includegraphics[height=.25\textwidth]{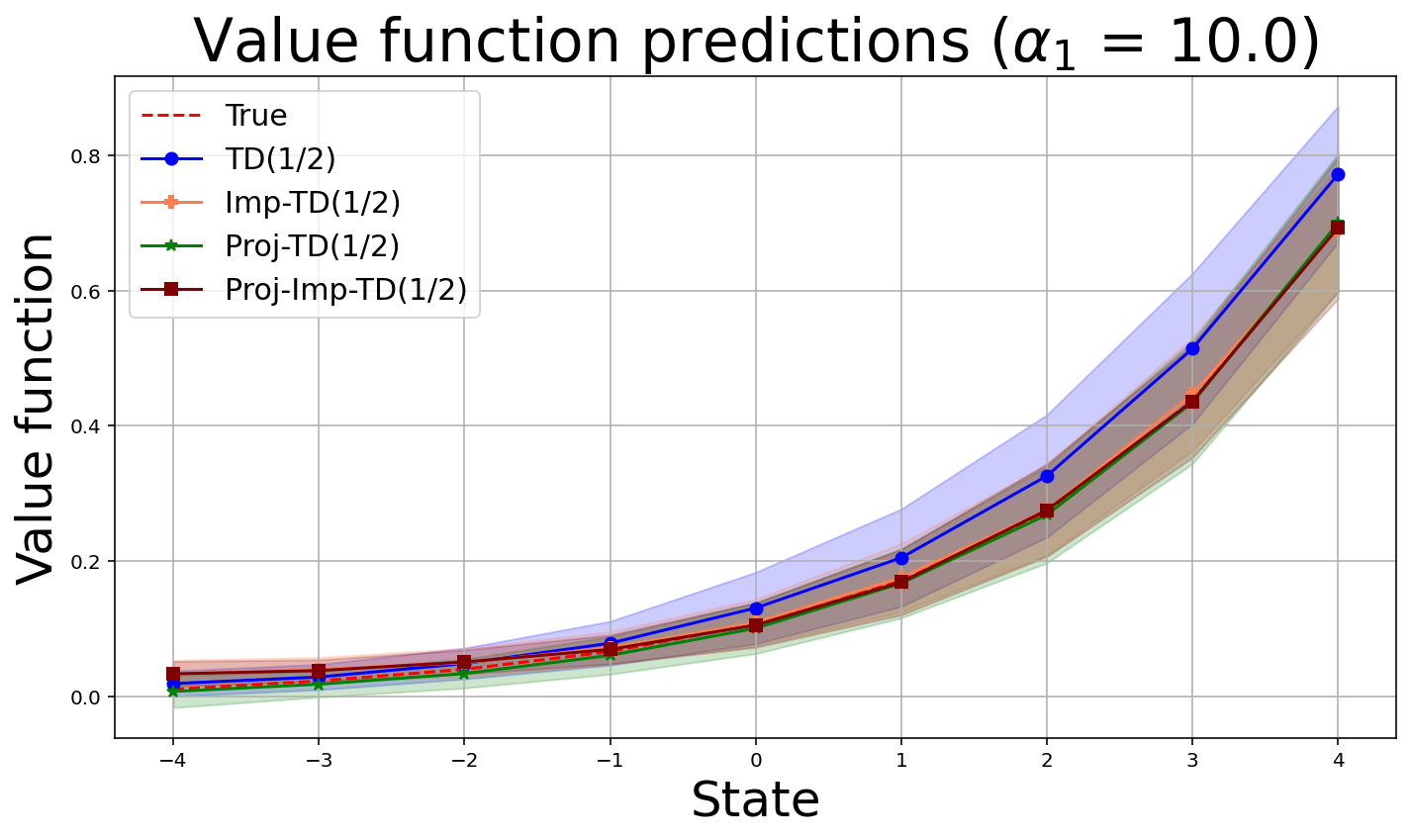}
    \includegraphics[height=.25\textwidth]{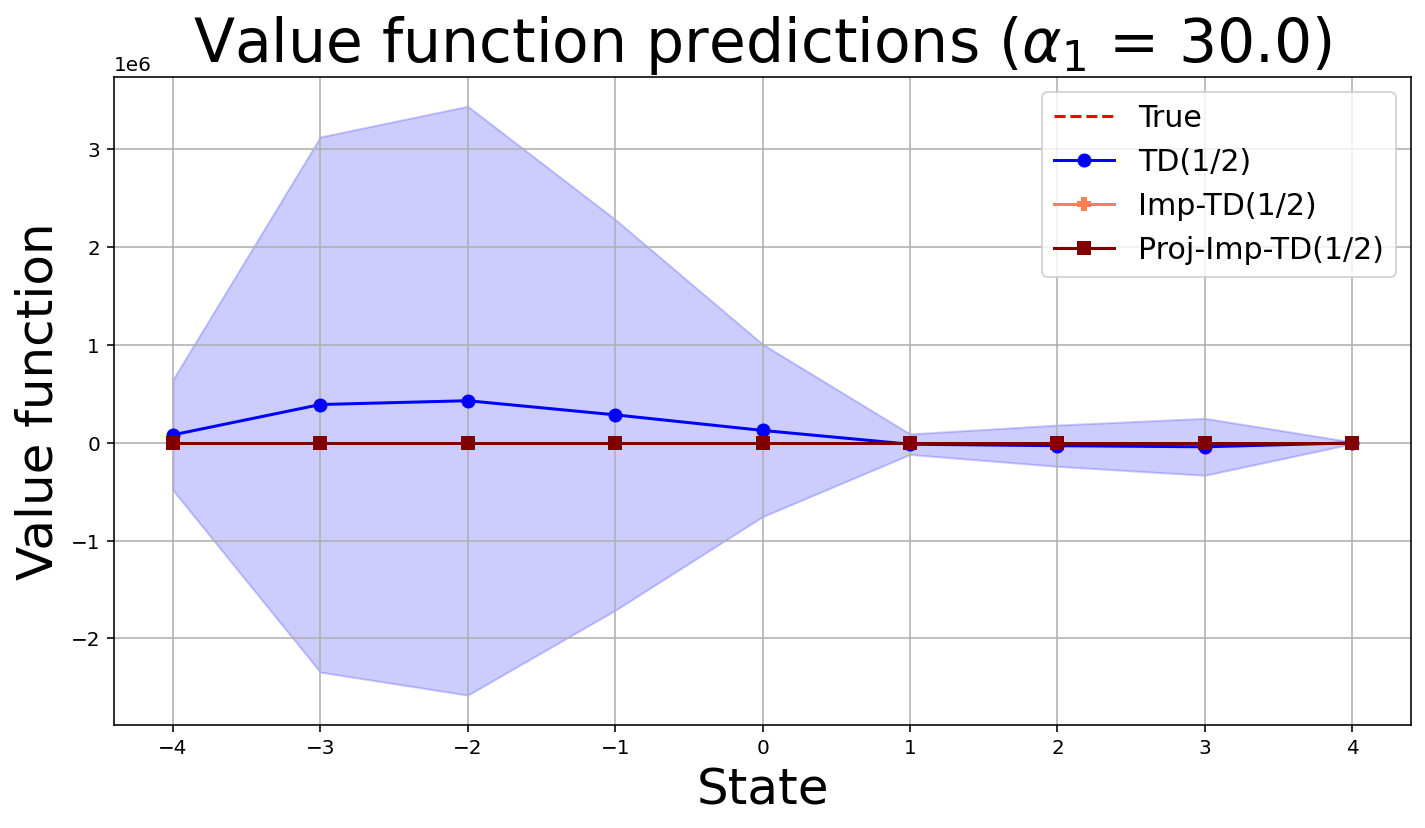}
    \includegraphics[height=.25\textwidth]{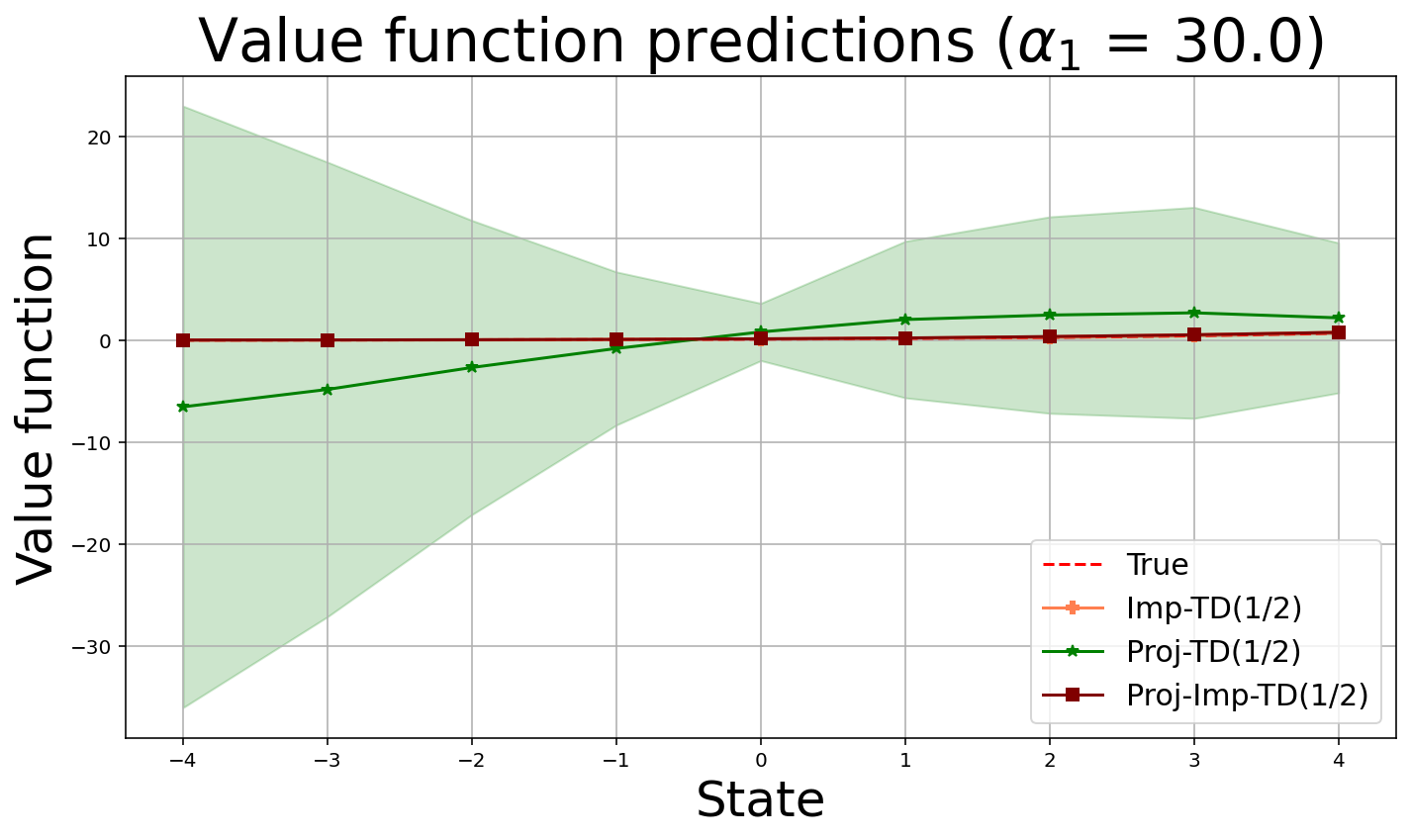}
    \caption{\textbf{Top left:} Value function approximation error (log-MSE) versus initial step size over the interval $[10.0, 30.0]$. TD(0.5) exhibits the sharpest increase in error as the initial step size grows, projected TD(0.5) increases more moderately, while implicit TD(0.5) and projected implicit TD(0.5) remain nearly flat across the entire range, reflecting their robustness to large step sizes. \textbf{Top right:} Value function approximation with a small initial step size ($\alpha_1 = 10.0$). All four algorithms accurately recover the true value function, with tight standard deviation bands. \textbf{Bottom left:} Value function approximation with a large initial step size ($\alpha_1 = 30.0$), comparing TD(0.5) against implicit TD(0.5) and projected implicit TD(0.5). TD(0.5) exhibits severe numerical instability, with estimates and standard deviation bands growing to extreme magnitudes, while both implicit variants remain stable and closely track the true value function. \textbf{Bottom right:} Value function approximation with a large initial step size ($\alpha_1 = 30.0$), comparing projected TD(0.5) against implicit TD(0.5) and projected implicit TD(0.5). Projected TD(0.5) exhibits instability, with a wider confidence band, while both implicit variants remain stable with tight bands. }
\label{fig:VALUE_FUNC_TDL}
\end{figure}
\vskip 0.2in

\bibliographystyle{plainnat}
\bibliography{sample}

\end{document}